%% file: arxiv.tex
\documentclass[10pt,twocolumn,letterpaper]{article}

\usepackage{iccv}
\usepackage{times}
\usepackage{epsfig}
\usepackage{graphicx}
\usepackage{amsmath}
\usepackage{amssymb}

\usepackage{multirow}
\usepackage{arydshln}
\usepackage{xcolor}  
\usepackage[export]{adjustbox}
\usepackage{bbm}
\usepackage[accsupp]{axessibility}
\usepackage{dblfloatfix}

\usepackage{pifont}
\newcommand{\cmark}{\color{teal}{\ding{51}}\color{black}}%
\newcommand{\xmark}{\color{violet}{\ding{55}}\color{black}}%
\def\draft{false}

\usepackage[breaklinks=true,bookmarks=false]{hyperref} 
\iccvfinalcopy 

\def\iccvPaperID{9008} 

\begin{document}

\title{Smoothness Similarity Regularization for Few-Shot GAN Adaptation}




\author{
     Vadim Sushko$^{1,2}$  \qquad
     Ruyu Wang$^1$ \qquad
     Juergen Gall$^{2,3}$ \\
     {\normalsize
     $^1$Bosch Center for Artificial Intelligence \qquad
     $^2$University of Bonn}\\
     {\normalsize$^3$Lamarr Institute for Machine Learning and Artificial Intelligence}\\
     {\tt\small vad221@gmail.com} \qquad {\tt\small ruyu.wang@de.bosch.com} \qquad {\tt\small gall@iai.uni-bonn.de}
}

\maketitle


\begin{abstract}
   The task of few-shot GAN adaptation aims to adapt a pre-trained GAN model to a small dataset with very few training images. While existing methods perform well when the dataset for pre-training is structurally similar to the target dataset, the approaches suffer from training instabilities or memorization issues when the objects in the two domains have a very different structure.  
   To mitigate this limitation, we propose a new smoothness similarity regularization that transfers the inherently learned smoothness of the pre-trained GAN to the few-shot target domain even if the two domains are very different. We evaluate our approach by adapting an unconditional and a class-conditional GAN to diverse few-shot target domains. Our proposed method significantly outperforms prior few-shot GAN adaptation methods in the challenging case of structurally dissimilar source-target domains, while performing on par with the state of the art for similar source-target domains. 

\end{abstract}
\vspace{-2ex}

\input{tex/1_intro.tex}

\input{tex/2_related_work.tex}

\input{tex/3_method.tex}
\input{tex/4_experiments.tex}
\input{tex/5_conclusion.tex}

{\small
\bibliographystyle{ieee_fullname}
\bibliography{references}
}

\begin{table*}[h!]
	\begin{center}
		\textbf{\Large{Smoothness Similarity Regularization for Few-Shot GAN Adaptation \vspace{0.5ex} \\ \textit{Supplementary material}}} \vspace{1.5ex}
	\end{center}
\end{table*}

\newpage
\newpage

\renewcommand{\thesection}{\Alph{section}}
\renewcommand{\thetable}{\Alph{table}}
\renewcommand{\thefigure}{\Alph{figure}}

\setcounter{section}{0}
\setcounter{figure}{0}   
\setcounter{table}{0}

\input{tex/7_supplementary.tex}

\end{document}

%% file: tex/1_intro.tex
\section{Introduction}
\label{sec:intro}


Generative adversarial networks (GANs) have been shown to be powerful at various image synthesis tasks \cite{choi2020stargan,sushko2020you,chan2022efficient,NEURIPS2021_076ccd93,sauer2022stylegan,sauer2023stylegan}. The success of these models is in large part enabled by the availability of large datasets for training, typically consisting of thousands of images. However, there are many applications and computer vision tasks such as one-shot or few-shot learning \cite{boudiaf2021few,tian2020prior}, out-of-distribution detection \cite{ren2019likelihood}, or long-tailed recognition tasks \cite{gupta2019lvis} where the number of available training images is very low. 

\input{figures/teaser.tex}

Since training a GAN from scratch on very few samples does not perform well as shown in Fig.~\ref{fig:teaser}, a common strategy is to fine-tune a pre-trained GAN model on the few-shot dataset, typically employing additional regularization losses to penalize the degradation of the diversity \cite{ojha2021few,xiao2022few}. This approach, referred to as few-shot GAN adaptation, performs well when the target domain is structurally very similar to the dataset that has been used for pre-training, e.g., photographs vs.\ sketches of human faces. However, the performance drastically degrades in case of large dissimilarities between the source and target domain as shown in Fig.~\ref{fig:teaser}. Such dissimilarities are a major bottleneck of using GANs in other disciplines like medicine, production, or crop science, where there is a lack of large datasets due to privacy, confidentiality, or simply lack of data. Motivated by this fact, we extend the protocol for few-shot GAN adaptation by investigating also pairs of datasets that are very different like churches and shells as shown in Fig.~\ref{fig:teaser}.   

To improve few-shot GAN adaptation in the case of structurally dissimilar pairs, we propose a new GAN adaptation strategy. Firstly, we propose a new smoothness similarity regularization for the generator. Our key observation is that pre-trained GAN generators, regardless of the exact structure of objects in the pre-training dataset, learn well-structured and smooth latent spaces. For example, prior works demonstrated that various local shifts in the latent space can lead to interpretable and smooth transitions of output images, such as translation of objects in the scene or changing their size \cite{voynov2020unsupervised,harkonen2020ganspace,shen2021closed}. As we show in our experiments, the proposed smoothness similarity regularization enables the transfer of this desirable property to other few-shot image domains without compromising the synthesis quality. Secondly, to overcome overfitting issues, we revisit the adversarial loss function of the discriminator and propose a simple yet efficient modification by computing the loss at different layers of the discriminator. This leads to the mitigation of overfitting and a more stabilized adaptation of the model to diverse target domains.


We evaluate our approach by adapting an unconditional \cite{Karras2019stylegan2} and a class-conditional GAN \cite{Brock2019} to diverse few-shot target domains. Our model significantly outperforms previous state-of-the-art methods in image quality and diversity in the challenging case of dissimilar source and target domains, while performing on par with the state of the art on structurally similar dataset pairs.
In summary, our contributions are as follows: (i) We extend the evaluation protocol for few-shot GAN adaptation by including new dataset pairs that are structurally much less similar than was considered in prior work. (ii) We propose a new smoothness similarity regularization, which enables diverse synthesis in the target domain by transferring the learned smoothness of a pre-trained GAN. (iii) We revisit the adversarial loss function of the discriminator to stabilize few-shot GAN adaptation across diverse target domains. (iv) Our proposed model enables high-quality synthesis in the challenging case of dissimilar source and target domains, significantly outperforming prior methods. In addition, we show that our method can be applied to different classes of GAN architectures, including unconditional and class-conditional GAN models.

%% file: figures/teaser.tex
\begin{figure}[t]
	\includegraphics[draft=\draft,width=0.99\linewidth]{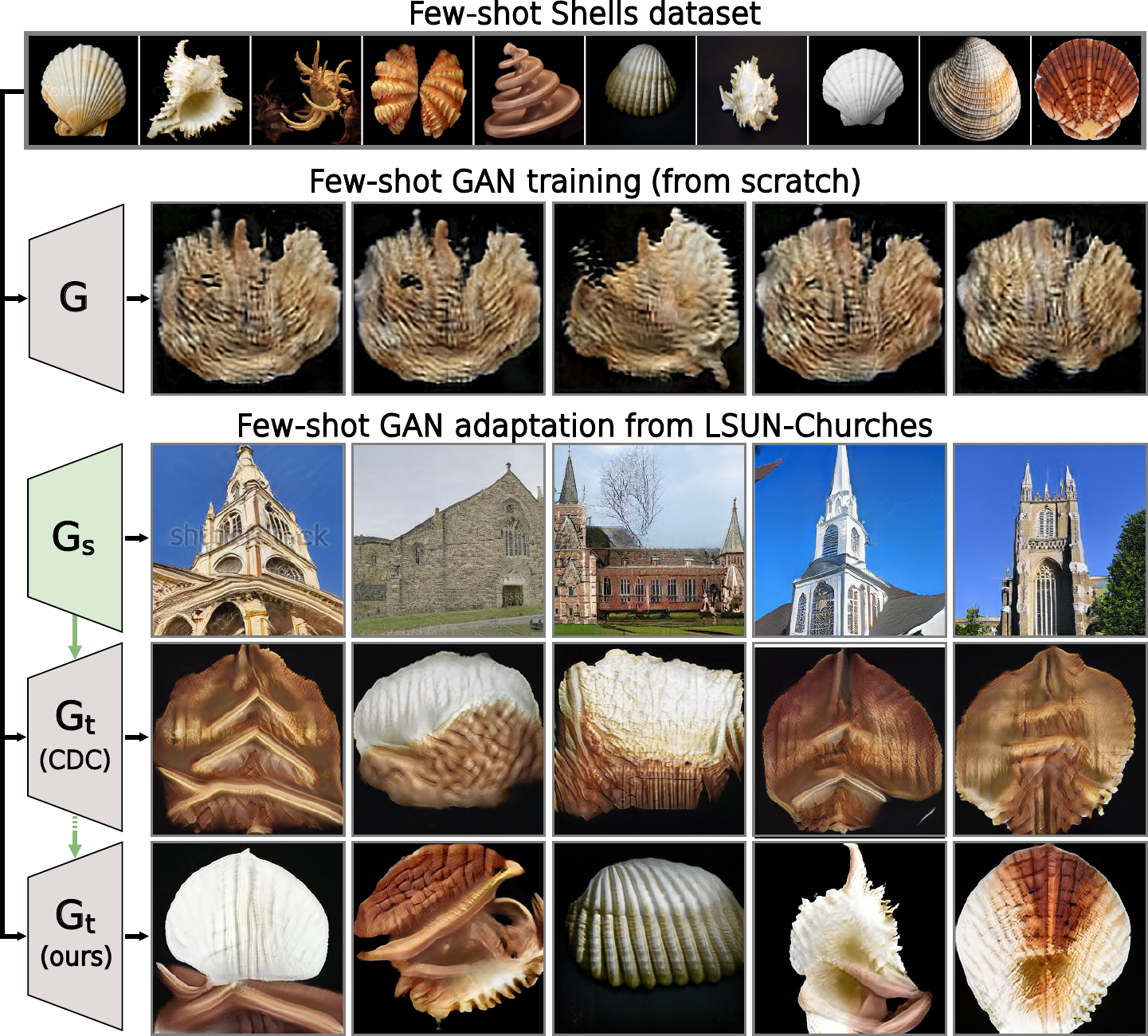}
	\caption{Training a GAN model $G$ on a few-shot dataset (row 1) from scratch fails due to training instabilities (row 2). We thus aim to adapt a GAN $G_s$ that has been pre-trained on a large dataset like LSUN-Church (row 3) to the target few-shot dataset ($G_t$). While fine-tuning \cite{ojha2021few} does not perform well either if source and target are dissimilar (row 4), our approach generates diverse and realistic images (row 5) by transferring the smoothness properties of $G_s$.
 }
	\label{fig:teaser}
 \vspace{-3ex}
\end{figure}

%% file: tex/2_related_work.tex
\section{Related Work}
\label{sec:related}

To address the image generation problem in the low data regime, existing works mainly follow three research lines -- one-shot, low-shot, and few-shot learning. One-shot generation methods \cite{Shaham2019SinGANLA, sushko2021one} focus on leveraging the internal patch distribution within a single image, however, their extension to capture the distribution of a small collection of images is non-trivial. In low-shot learning \cite{Zhao2020DifferentiableAF}, several works  \cite{Zhao2020DifferentiableAF,Karras2020TrainingGA} proposed to mitigate the limited-data-induced overfitting issue by adapting data augmentations to the generative networks. Others \cite{anonymous2021towards, Cui2021GenCoGC} stabilized the training process and reduced overfitting by revising the network design. Despite the promising performance in many low data regimes (typically having 100+ images), these low-shot methods fail in the extremely few-shot setting (e.g., 10 images). Our work lies in the scope of few-shot learning.

\textbf{Few-shot image synthesis.} 
Conventional few-shot learning aims at learning a discriminative classifier under limited data scenarios. In the context of image synthesis with GANs, the goal instead is to produce diverse new images from the learned distribution while preventing overfitting to the few training samples. A straightforward way is to treat it as a domain adaptation problem and incorporate the commonly used transfer learning technique, i.e., fine-tuning, to ease the need for data. However, naive fine-tuning (TGAN) \cite{Wang2018TransferringGG} often suffers from overfitting and results in poor performance. Researchers proposed remedies such as mining suitable parts of the latent space before fine-tuning \cite{Wang2020MineGANEK} or restricting weight updates, for example, updating only the BatchNorm parameters of the generator \cite{Noguchi2019ImageGF}, penalizing drastic changes in important weights \cite{li2020few}, or freezing the earliest layers of the discriminator (FreezeD) \cite{Mo2020FreezeDA}. 
More recent works focused on introducing different regularizations to preserve specific knowledge from the pre-trained model and prevent diversity degradation \cite{zhao2022closer}.  For example, CDC \cite{ojha2021few} proposed to preserve the pair-wise perceptual similarity between samples from the source domain and to transfer it to the target domain, while RSSA \cite{xiao2022few} designed a novel consistency term to align the structural information between source and target domains. Although the two aforementioned methods constitute the current state of the art in few-shot generative learning, their assumptions impose strong constraints on the structure of the few-shot target domain. As we show in experiments, they fail in the more challenging regime when the source and target domains are not restrictively similar. Most recently, \cite{yunqingfew} proposed to replace prior knowledge preservation criteria with adaptation-aware kernel modulation (AdAM), which relaxed the source-target proximity requirement of previous methods to some extent.
In this work, we take a step further and introduce a new regularization term to preserve the generator's smoothness properties that are not limited to a specific domain, enabling successful adaptation between image domains of unprecedented structural dissimilarity.

\input{figures/method.tex}

\textbf{Smoothness of image generators.}
Smooth transitions in the latent space are an important property for generative models, where it is believed to be a sign of a well-conditioned generator. Models trained on large datasets naturally possess this property with or without explicit regularization \cite{Brock2019, Karras2019stylegan2}. 
For example, StyleGANv2 \cite{Karras2019stylegan2} introduced a regularization based on the perceptual path length measure (PPL) \cite{Karras2018ASG}, which encourages that a fixed-size step in the latent space results in a fixed-magnitude change in the image space. However, achieving a smooth mapping of the generator is difficult for few-shot image synthesis since there are not enough training samples. 
Thus, MixDL \cite{kong2022few} sought to alleviate the ``staircase'' latent space interpolations, i.e., jumps  between training samples, by introducing a continuous coefficient vector and enforcing smooth interpolations between training images. Although the two above regularizers aim to encourage smoother interpolations between training samples and thus mitigate mode collapse, they are not designed to take advantage of the available pre-training knowledge. In contrast, in this work we develop a new smoothness similarity regularization that leverages the well-structured latent space of a pre-trained GAN generator. In effect, our approach enables high-quality few-shot image synthesis by transferring smooth and realistic image transitions of pre-trained GANs to diverse few-shot domains.


%% file: figures/method.tex
\begin{figure*}[t]
    \vspace{-1ex}
	\includegraphics[width=0.98\linewidth]{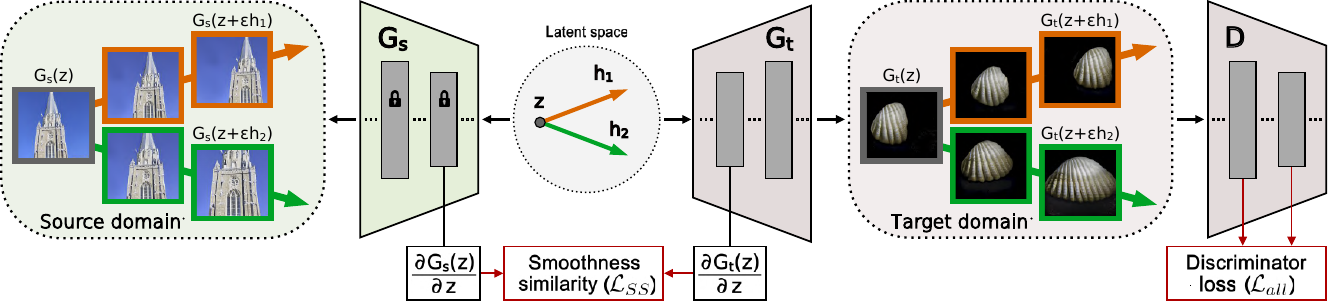}
	\caption{
 Given a pre-trained generator $G_s$, the proposed smoothness similarity regularization preserves the learned smoothness of $G_s$ while adapting it to a target domain with very few images.   
 To mitigate overfitting to the target domain, the discriminator loss utilizes features at various layers and automatically adjusts the impact of different semantic scales to the similarity of the source and target domain. 
 }
	\label{fig:method}
    \vspace{-2ex}
\end{figure*}

%% file: tex/3_method.tex
\section{Method}
\label{Sec:method}

In the task of few-shot GAN adaptation, we are given a small target dataset $T$ and a pre-trained GAN model, consisting of a discriminator $D$ and a generator $G_s$, which produces an image $x =  G_s(z)$ from a continuous input variable $z$, such as a random noise vector or a continuous class embedding. The goal is to adapt the generator to the target dataset such that it generates diverse and realistic images in the domain of the target dataset as shown in Fig.~\ref{fig:teaser}. We denote the adapted target generator by $G_t$.


To achieve few-shot synthesis with a high image quality and diversity, our model should adhere to the following two properties. 
Firstly, the generator $G_t$ should not only memorize and generate the target images, which will be addressed by the smoothness similarity regularization (Sec.~\ref{Sec:method:G}).
Secondly, the discriminator $D$ must avoid overfitting to the few target images in order to provide useful supervision for $G_t$ (Sec.~\ref{Sec:method:D}). The overview of our method is shown in Fig.~\ref{fig:method}.

\subsection{Smoothness similarity regularization for $G_t$}
\label{Sec:method:G}

In a low data regime like ours, $G_t$ can easily overfit to the target dataset $T$ and collapse to reproducing only the few modes represented in the training data. When walking in the latent space of such a generator, one would observe ``staircase'' patterns, where minor shifts in the latent space cause discontinuous transitions in the output image space (as shown in row 4 of Fig.~\ref{fig:ablation_interp}).  
Naturally, to achieve a synthesis of high diversity, it is desirable for $G_t$ to avoid such discontinuities, as having smoother image transitions allows to generate intermediate samples that can exhibit novel features. Therefore, in our model we aim to encourage $G_t$ to produce smooth latent space interpolations, in which all the intermediate images are realistic.

Our approach is based on the observation that GANs trained on large datasets tend to have a well-structured latent space~\cite{voynov2020unsupervised,harkonen2020ganspace,shen2021closed}, in which different latent space directions can lead to smooth and interpretable image transitions. For example, in a generator pre-trained on a large dataset of churches, latent directions can emerge causing smooth zooming or translation of churches (see Fig.~\ref{fig:method}). Our observation is that the nature of such image transitions (e.g., zooming or translation) is remarkably general. Thus, we propose a regularizer that utilizes this smoothness property of the source generator $G_s$ as a cue  while adapting it to another image domain, which can be very different from the domain that was used for pre-training.
For example, as shown in Fig.~\ref{fig:method}, the same latent directions of churches can cause similar zooming or translation effects on shells.




Mathematically, the smoothness of the generator can be represented via a Jacobian matrix $J_{G^l}(z) = || \partial G^l(z) / \partial z ||$, quantifying how the generator's intermediate features after the $l$-th block change under local shifts in the latent space. As we want the same latent shift to cause perceptually similar image transitions in the source and target domains, we design a regularization term that brings the Jacobian matrices of $G^l_s$ and $G^l_t$ closer together. As the computation of full Jacobian matrices is expensive, we use an unbiased estimator of their products with a Gaussian vector \cite{dauphin2015equilibrated,Karras2019stylegan2}, which can be computed with standard back-propagation:
\begin{equation}
    J_{G^l}^T(z) \cdot y = \mathbbm{E}_{(y) \sim N(0, 1)} \nabla_z \langle G^l(z) , y \rangle,
    \label{eq:helper}
\end{equation}
where $y$ is a Gaussian tensor of the same shape as $G^l$. Our smoothness similarity regularization is then expressed as:
\begin{small}
\begin{equation}
    \mathcal{L}_{SS} = \lambda_{SS} \cdot \mathbbm{E}_{(z, y) \sim N(0, 1)} || \nabla_z \langle G_s^l(z) , y \rangle - \nabla_z \langle G_t^l(z) , y \rangle ||_2,
    \label{eq:g_loss}
\end{equation}
\end{small}
where $\lambda_{SS}$ steers the impact of the regularizer. As shown in Fig.~\ref{fig:method}, the smoothness similarity regularization depends on both generators, but only $G_t$ is updated.  
It is interesting to note that the Jacobian matrix is also used for the path length regularization \cite{Karras2019stylegan2}, which forces $J_{G}(z)$ to be orthogonal up to a global scale at any $z$. While this alternative regularizer also induces some form of smoothness, it does not transfer the inherently learned smoothness of a pre-trained GAN. We show in Sec.~\ref{sec:experiments:stylegan2} that it struggles to enforce the realism of intermediate images. 
Furthermore, our approach shares the motivation with some prior regularization approaches that use noise perturbations to enforce diversity \cite{ojha2021few,xiao2022few}. In contrast to Eq.~\ref{eq:g_loss}, these approaches incorporate non-gradient components, e.g., assuming similarity of images $G_s(z)$$\leftrightarrow$$G_t(z)$ or distributions $d(G_t(z_1),G_t(z_2))$$\leftrightarrow$$d(G_s(z_1),G_s(z_2))$. As such assumptions are violated when source and target domains are dissimilar, they perform poorly compared to our smoothness similarity regularization $\mathcal{L}_{SS}$ as shown in the experiments.

\subsection{Revisiting the $D$ adversarial loss}
\label{Sec:method:D}

To identify what kind of image transitions look realistic for the target domain, $G_t$ requires strong supervision from the discriminator on image realism at different semantic scales. This includes the colors and textures of objects, as well as object shapes, especially if their distribution is different from the shapes of objects in the source domain. Learning the concept of image realism in low data regimes is, however, challenging due to the problem of overfitting. 

Typically, a GAN discriminator consists of several consecutive blocks $\{D^i\}_{i=1}^{N}$ and computes for each given image $x$ a real/fake logit after the last block $l = s^N \circ D^N (x)$, where $s^N$ is a final processing layer such as a convolution. When adapting such a discriminator to a very small dataset, it is prone to memorizing the training set \cite{sushko2021learning}, leading to mode collapse and poor diversity of synthesized images \cite{ojha2021few}. 
A possible solution \cite{ojha2021few, xiao2022few} to overcome memorization is to use variants of the PatchGAN discriminator \cite{isola2017image}, discarding the latest discriminator layers: $l = s^k \circ D^k (x), k < N$. This solution allows to adapt colors and textures of generated images to the target domain while avoiding the memorization problem. However, it naturally has a limited capacity to learn more high-level semantic scene properties such as the shapes of objects, which we show in the experiments.

\input{figures/qual_comparison_distant1.tex}

In order to avoid memorization, and yet to balance the adaptation of colors, textures, and shapes of generated objects to a new domain, we hypothesize that a more flexible attention to different levels of image realism is required by the discriminator.
To this end, we perform a simple yet efficient modification to the loss function of the discriminator. Given a discriminator $\{D^i\}_{i=1}^{N}$ and its adversarial loss function $\mathcal{L_D}(l)$ used for pre-training (e.g., cross-entropy or hinge loss), we design the discriminator to produce real/fake logits after \textit{each} discriminator's block, and correspondingly compute the loss as the average across all blocks:
\vspace{-0ex}
\begin{equation}
    \mathcal{L}_{all}(x) = \frac{1}{N} \sum\limits_{i=1}^N \mathcal{L_D}  [ l^i (x) ], ~~~l^i (x) =  s^i \circ D^i(x). 
    \label{eq:d_loss}
\end{equation}
\vspace{-1ex}

With the new objective, $D$ is given more freedom to utilize the features extracted at different scales to compute the loss. Our finding is that $D$ dynamically adapts the magnitude of the loss at each scale to the target domain, without explicit supervision (see Fig.~\ref{fig:plots_losses}). 
Consequently, we observe a strong overall stabilization effect on the adaptation performance across diverse source-target dataset pairs.

%% file: figures/qual_comparison_distant1.tex
\begin{figure*}[t]
\begin{centering}
\setlength{\tabcolsep}{0.53in}
\renewcommand{\arraystretch}{1}
\par\end{centering}
\begin{centering}

\vspace{-0.5ex}
\begin{tabular}{@{\hskip -0.02in}c@{\hskip 0.04in}:c@{\hskip 0.01in}c@{\hskip 0.01in}c@{\hskip 0.01in}c@{\hskip 0.01in}c@{\hskip 0.01in}c@{\hskip 0.01in}c@{\hskip 0.01in}c@{\hskip 0.01in}c@{\hskip 0.01in}c@{\hskip 0.01in}c@{\hskip 0.01in}c@{\hskip 0.01in}c}

  \multirow{-1}{*}{\begin{tabular}{c@{\hskip 0.01in}c@{\hskip 0.03in}} 
  \frame{\includegraphics[draft=\draft,width=0.075\linewidth, height=0.070\linewidth]{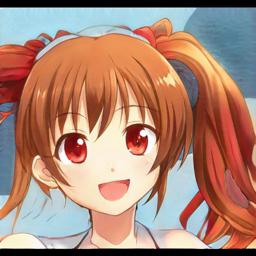}} &
  \frame{\includegraphics[draft=\draft,width=0.075\linewidth, height=0.070\linewidth]{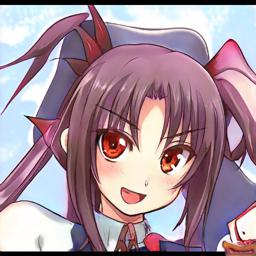}} \tabularnewline[-3pt]	
  \frame{\includegraphics[draft=\draft,width=0.075\linewidth, height=0.070\linewidth]{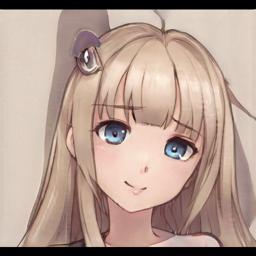}} &
  \frame{\includegraphics[draft=\draft,width=0.075\linewidth, height=0.070\linewidth]{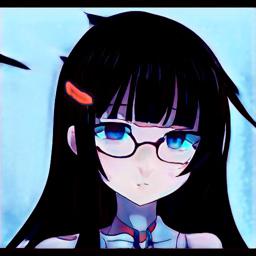}} \tabularnewline[-3pt]
  \frame{\includegraphics[draft=\draft,width=0.075\linewidth, height=0.070\linewidth]{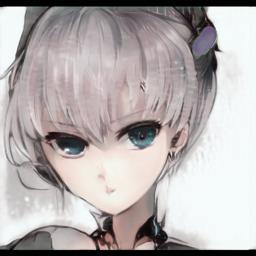}} &
  \frame{\includegraphics[draft=\draft,width=0.075\linewidth, height=0.070\linewidth]{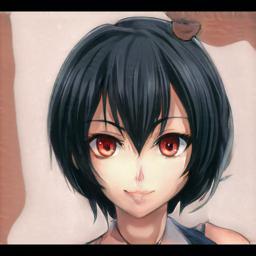}} \tabularnewline[-3pt] 	
  \frame{\includegraphics[draft=\draft,width=0.075\linewidth, height=0.070\linewidth]{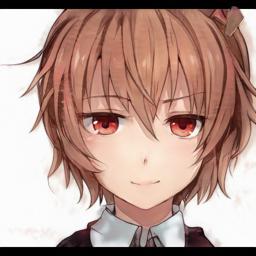}} &
  \frame{\includegraphics[draft=\draft,width=0.075\linewidth, height=0.070\linewidth]{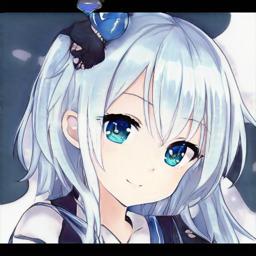}} \tabularnewline[-3pt]
  \frame{\includegraphics[draft=\draft,width=0.075\linewidth, height=0.070\linewidth]{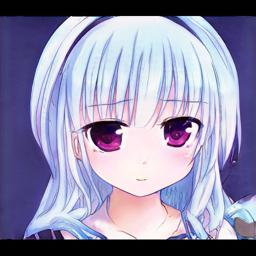}} &
  \frame{\includegraphics[draft=\draft,width=0.075\linewidth, height=0.070\linewidth]{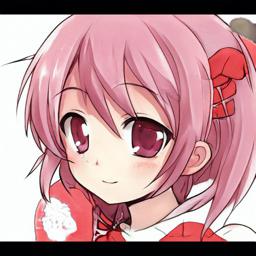}} \tabularnewline[+1pt] 
  \multicolumn{2}{@{\hskip -0.02in}c@{\hskip 0.03in}}{\small 10-shot} \tabularnewline[-2pt] 
  \multicolumn{2}{@{\hskip -0.02in}c@{\hskip 0.03in}}{\small Anime-Faces} \tabularnewline[-3pt]
  \end{tabular}} &
  
  \frame{\includegraphics[draft=\draft,width=0.075\linewidth, height=0.070\linewidth]{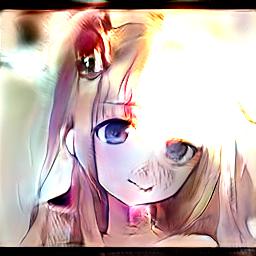}} & 
  \frame{\includegraphics[draft=\draft,width=0.075\linewidth, height=0.070\linewidth]{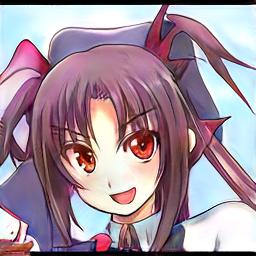}} & 
  \frame{\includegraphics[draft=\draft,width=0.075\linewidth, height=0.070\linewidth]{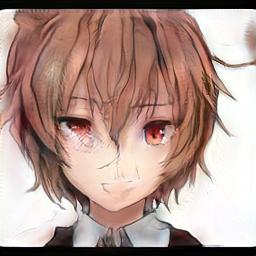}} & 
  \frame{\includegraphics[draft=\draft,width=0.075\linewidth, height=0.070\linewidth]{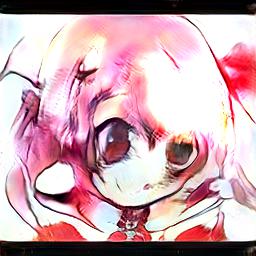}} & 
  \frame{\includegraphics[draft=\draft,width=0.075\linewidth, height=0.070\linewidth]{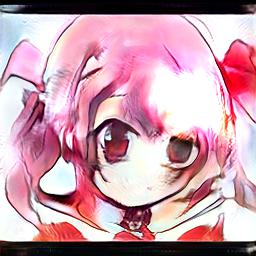}} & 
  \frame{\includegraphics[draft=\draft,width=0.075\linewidth, height=0.070\linewidth]{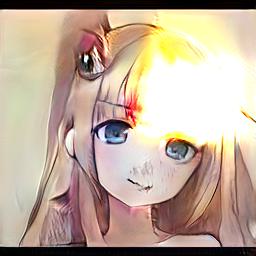}} & 
  \frame{\includegraphics[draft=\draft,width=0.075\linewidth, height=0.070\linewidth]{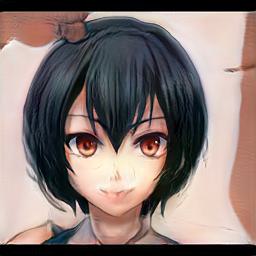}} & 
  \frame{\includegraphics[draft=\draft,width=0.075\linewidth, height=0.070\linewidth]{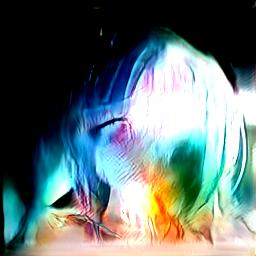}} & 
  \frame{\includegraphics[draft=\draft,width=0.075\linewidth, height=0.070\linewidth]{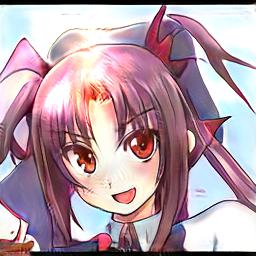}} & 
  \frame{\includegraphics[draft=\draft,width=0.075\linewidth, height=0.070\linewidth]{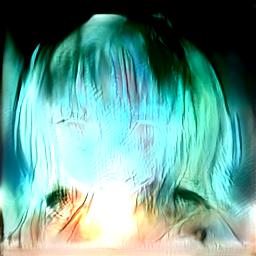}} &~~\rotatebox{270}{{\hspace{-3.6em} TGAN \hspace{-3.0em} } }
  \tabularnewline[-3pt]
  &
  \frame{\includegraphics[draft=\draft,width=0.075\linewidth, height=0.070\linewidth]{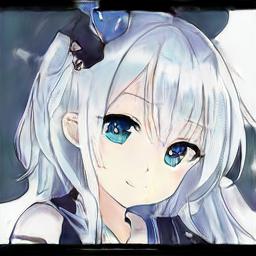}} & 
  \frame{\includegraphics[draft=\draft,width=0.075\linewidth, height=0.070\linewidth]{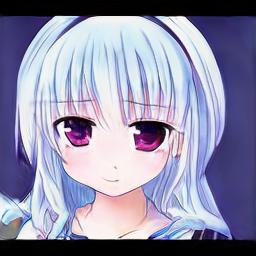}} & 
  \frame{\includegraphics[draft=\draft,width=0.075\linewidth, height=0.070\linewidth]{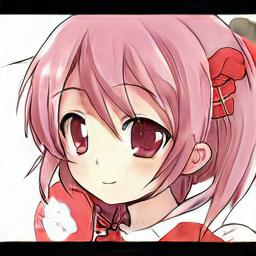}} & 
  \frame{\includegraphics[draft=\draft,width=0.075\linewidth, height=0.070\linewidth]{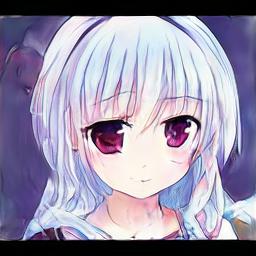}} & 
  \frame{\includegraphics[draft=\draft,width=0.075\linewidth, height=0.070\linewidth]{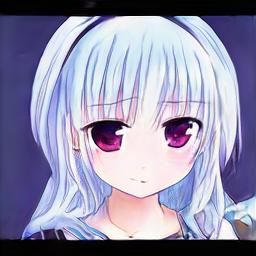}} & 
  \frame{\includegraphics[draft=\draft,width=0.075\linewidth, height=0.070\linewidth]{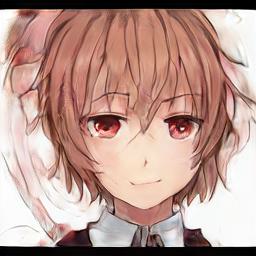}} & 
  \frame{\includegraphics[draft=\draft,width=0.075\linewidth, height=0.070\linewidth]{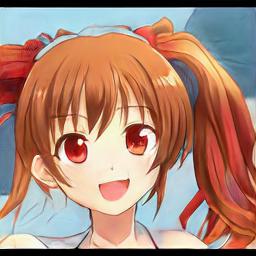}} & 
  \frame{\includegraphics[draft=\draft,width=0.075\linewidth, height=0.070\linewidth]{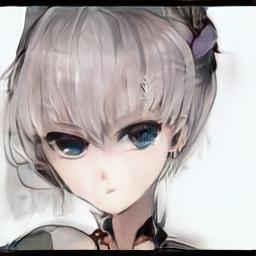}} & 
  \frame{\includegraphics[draft=\draft,width=0.075\linewidth, height=0.070\linewidth]{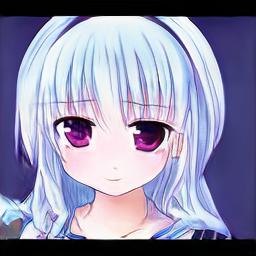}} & 
  \frame{\includegraphics[draft=\draft,width=0.075\linewidth, height=0.070\linewidth]{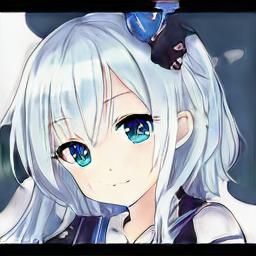}} &~~\rotatebox{270}{{\hspace{-3.7em} FreezeD \hspace{-3.0em} } }
  \tabularnewline[-3pt]
  &
  \frame{\includegraphics[draft=\draft,width=0.075\linewidth, height=0.070\linewidth]{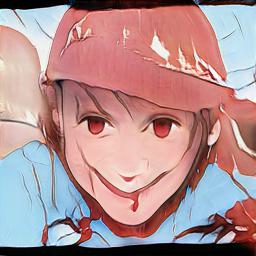}} & 
  \frame{\includegraphics[draft=\draft,width=0.075\linewidth, height=0.070\linewidth]{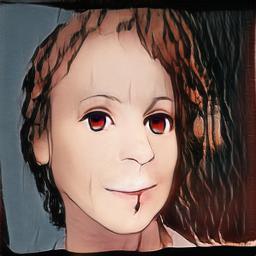}} & 
  \frame{\includegraphics[draft=\draft,width=0.075\linewidth, height=0.070\linewidth]{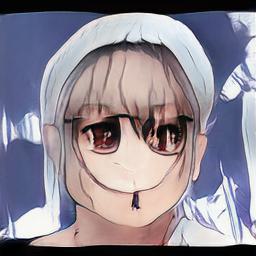}} & 
  \frame{\includegraphics[draft=\draft,width=0.075\linewidth, height=0.070\linewidth]{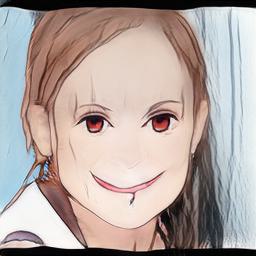}} & 
  \frame{\includegraphics[draft=\draft,width=0.075\linewidth, height=0.070\linewidth]{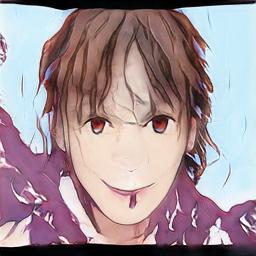}} & 
  \frame{\includegraphics[draft=\draft,width=0.075\linewidth, height=0.070\linewidth]{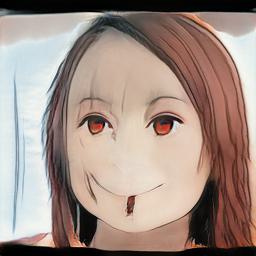}} & 
  \frame{\includegraphics[draft=\draft,width=0.075\linewidth, height=0.070\linewidth]{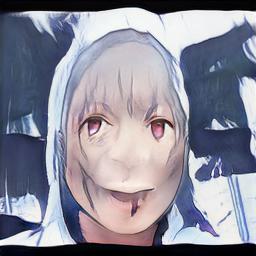}} & 
  \frame{\includegraphics[draft=\draft,width=0.075\linewidth, height=0.070\linewidth]{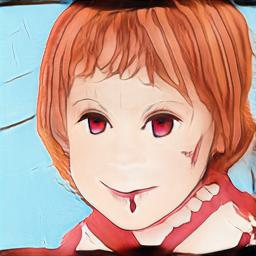}} & 
  \frame{\includegraphics[draft=\draft,width=0.075\linewidth, height=0.070\linewidth]{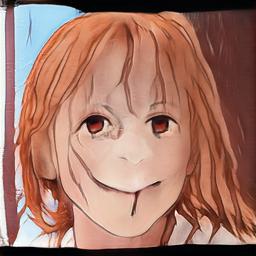}} & 
  \frame{\includegraphics[draft=\draft,width=0.075\linewidth, height=0.070\linewidth]{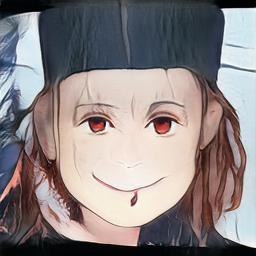}} &  ~~\rotatebox{270}{{\hspace{-3.0em} CDC \hspace{-3.0em} } }
  \tabularnewline[-3pt]
  &
  \frame{\includegraphics[draft=\draft,width=0.075\linewidth, height=0.070\linewidth]{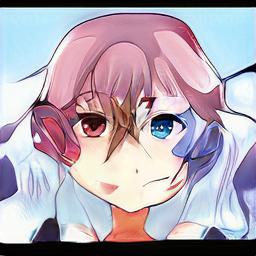}} & 
  \frame{\includegraphics[draft=\draft,width=0.075\linewidth, height=0.070\linewidth]{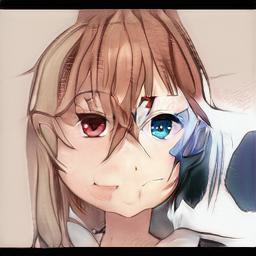}} & 
  \frame{\includegraphics[draft=\draft,width=0.075\linewidth, height=0.070\linewidth]{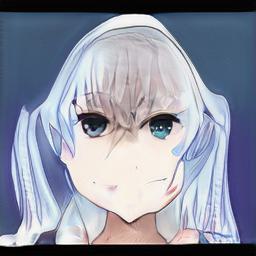}} & 
  \frame{\includegraphics[draft=\draft,width=0.075\linewidth, height=0.070\linewidth]{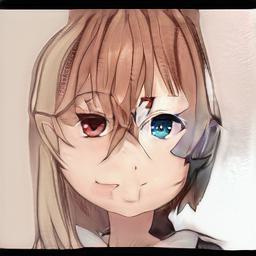}} & 
  \frame{\includegraphics[draft=\draft,width=0.075\linewidth, height=0.070\linewidth]{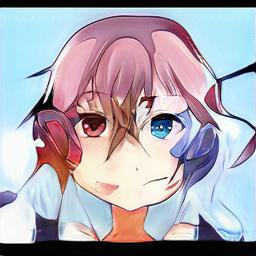}} & 
  \frame{\includegraphics[draft=\draft,width=0.075\linewidth, height=0.070\linewidth]{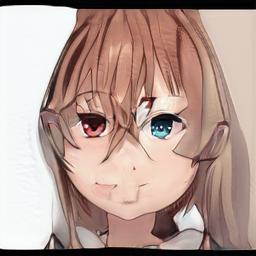}} & 
  \frame{\includegraphics[draft=\draft,width=0.075\linewidth, height=0.070\linewidth]{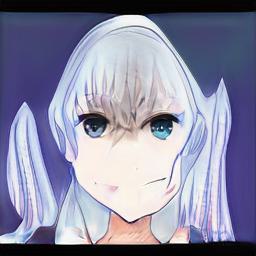}} & 
  \frame{\includegraphics[draft=\draft,width=0.075\linewidth, height=0.070\linewidth]{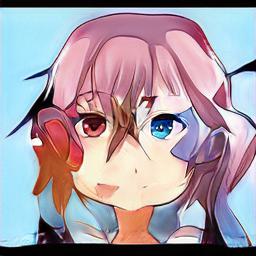}} & 
  \frame{\includegraphics[draft=\draft,width=0.075\linewidth, height=0.070\linewidth]{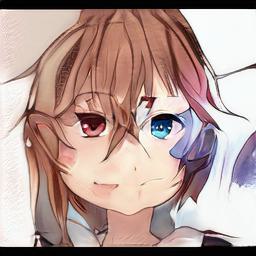}} & 
  \frame{\includegraphics[draft=\draft,width=0.075\linewidth, height=0.070\linewidth]{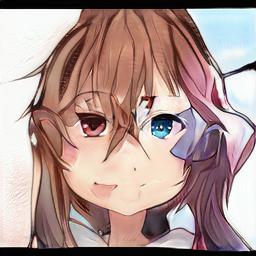}} &  ~~\rotatebox{270}{{\hspace{-3.2em} RSSA \hspace{-3.0em} } }
  \tabularnewline[-3pt]
  &
  \frame{\includegraphics[draft=\draft,width=0.075\linewidth, height=0.070\linewidth]{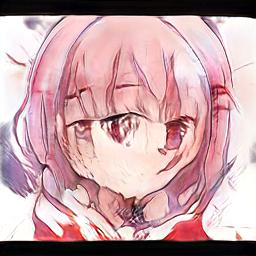}} & 
  \frame{\includegraphics[draft=\draft,width=0.075\linewidth, height=0.070\linewidth]{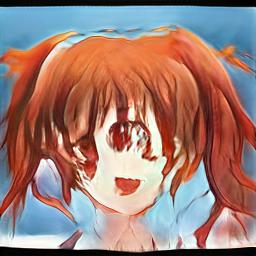}} & 
  \frame{\includegraphics[draft=\draft,width=0.075\linewidth, height=0.070\linewidth]{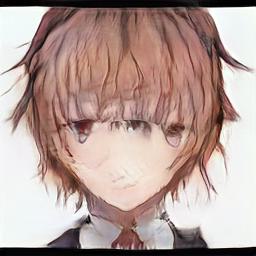}} & 
  \frame{\includegraphics[draft=\draft,width=0.075\linewidth, height=0.070\linewidth]{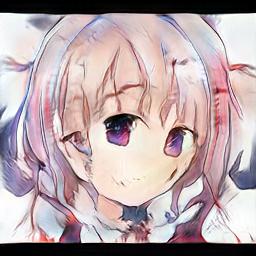}} & 
  \frame{\includegraphics[draft=\draft,width=0.075\linewidth, height=0.070\linewidth]{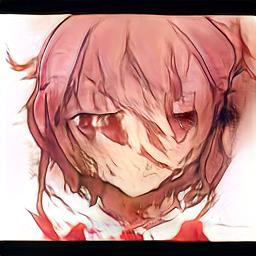}} & 
  \frame{\includegraphics[draft=\draft,width=0.075\linewidth, height=0.070\linewidth]{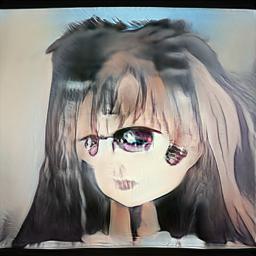}} & 
  \frame{\includegraphics[draft=\draft,width=0.075\linewidth, height=0.070\linewidth]{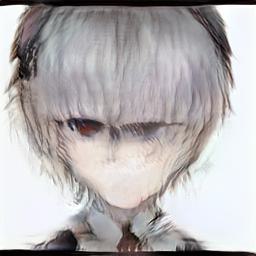}} & 
  \frame{\includegraphics[draft=\draft,width=0.075\linewidth, height=0.070\linewidth]{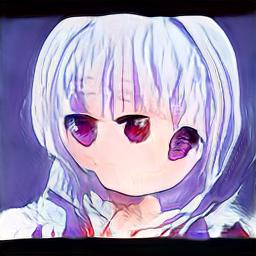}} & 
  \frame{\includegraphics[draft=\draft,width=0.075\linewidth, height=0.070\linewidth]{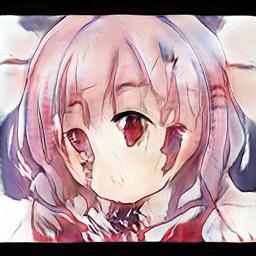}} & 
  \frame{\includegraphics[draft=\draft,width=0.075\linewidth, height=0.070\linewidth]{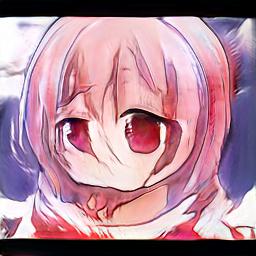}} &  ~~\rotatebox{270}{{\hspace{-3.4em} AdAM \hspace{-3.0em} } }
  \tabularnewline[-3pt]
  &
  \frame{\includegraphics[draft=\draft,width=0.075\linewidth, height=0.070\linewidth]{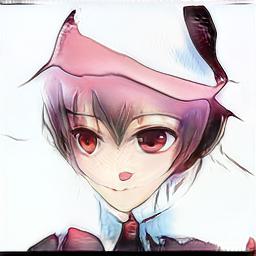}} & 
  \frame{\includegraphics[draft=\draft,width=0.075\linewidth, height=0.070\linewidth]{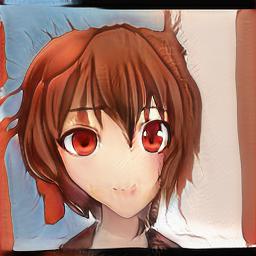}} & 
  \frame{\includegraphics[draft=\draft,width=0.075\linewidth, height=0.070\linewidth]{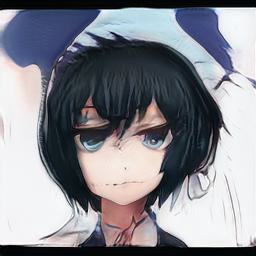}} & 
  \frame{\includegraphics[draft=\draft,width=0.075\linewidth, height=0.070\linewidth]{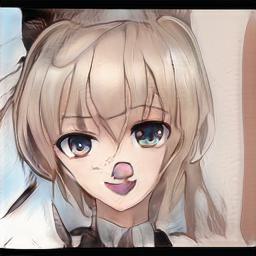}} & 
  \frame{\includegraphics[draft=\draft,width=0.075\linewidth, height=0.070\linewidth]{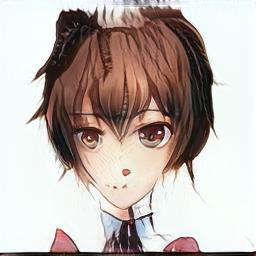}} & 
  \frame{\includegraphics[draft=\draft,width=0.075\linewidth, height=0.070\linewidth]{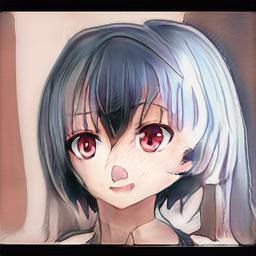}} & 
  \frame{\includegraphics[draft=\draft,width=0.075\linewidth, height=0.070\linewidth]{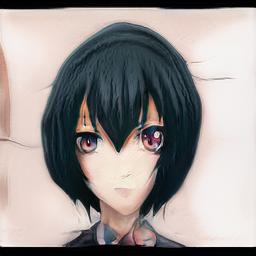}} & 
  \frame{\includegraphics[draft=\draft,width=0.075\linewidth, height=0.070\linewidth]{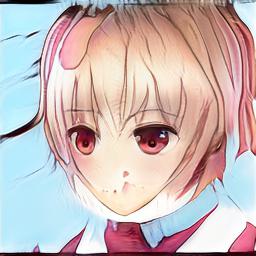}} & 
  \frame{\includegraphics[draft=\draft,width=0.075\linewidth, height=0.070\linewidth]{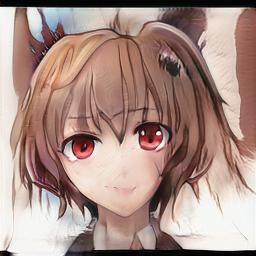}} & 
  \frame{\includegraphics[draft=\draft,width=0.075\linewidth, height=0.070\linewidth]{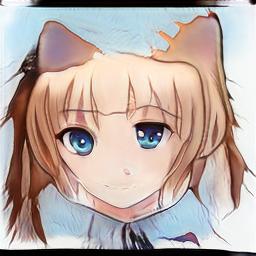}} & ~~\rotatebox{270}{{\hspace{-3.1em} \textbf{Ours} \hspace{-3.0em} } }
  \tabularnewline[-3pt]
  &
  \frame{\includegraphics[draft=\draft,width=0.075\linewidth, height=0.070\linewidth]{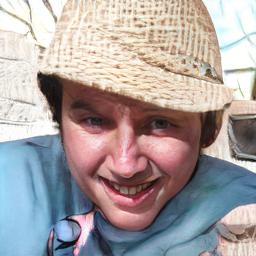}} & 
  \frame{\includegraphics[draft=\draft,width=0.075\linewidth, height=0.070\linewidth]{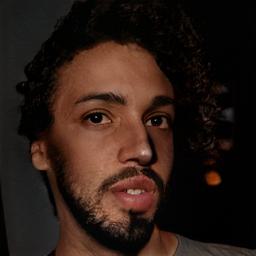}} & 
  \frame{\includegraphics[draft=\draft,width=0.075\linewidth, height=0.070\linewidth]{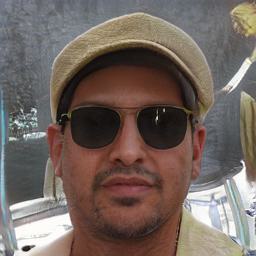}} & 
  \frame{\includegraphics[draft=\draft,width=0.075\linewidth, height=0.070\linewidth]{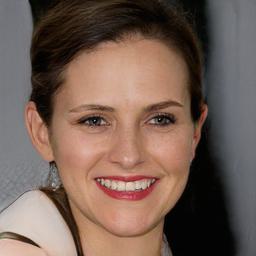}} & 
  \frame{\includegraphics[draft=\draft,width=0.075\linewidth, height=0.070\linewidth]{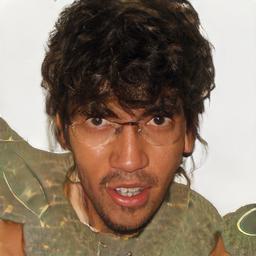}} & 
  \frame{\includegraphics[draft=\draft,width=0.075\linewidth, height=0.070\linewidth]{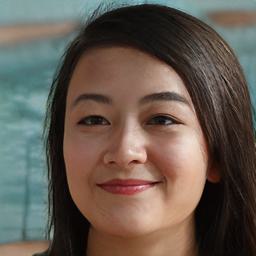}} & 
  \frame{\includegraphics[draft=\draft,width=0.075\linewidth, height=0.070\linewidth]{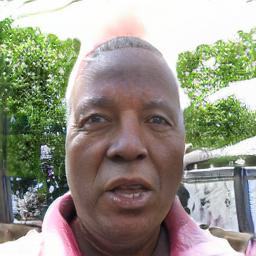}} & 
  \frame{\includegraphics[draft=\draft,width=0.075\linewidth, height=0.070\linewidth]{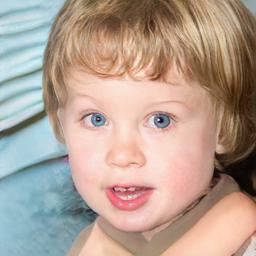}} & 
  \frame{\includegraphics[draft=\draft,width=0.075\linewidth, height=0.070\linewidth]{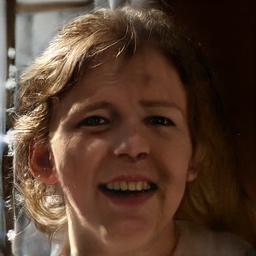}} & 
  \frame{\includegraphics[draft=\draft,width=0.075\linewidth, height=0.070\linewidth]{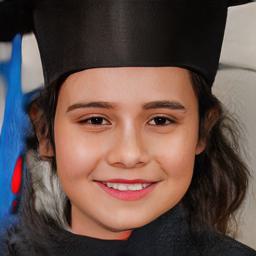}} & ~~\rotatebox{270}{{\hspace{-3.5em} Source 
  \hspace{-3.0em} } }
  \tabularnewline[-4pt]
\end{tabular}

\vspace{1.5ex}

\begin{tabular}{@{\hskip -0.02in}c@{\hskip 0.04in}:c@{\hskip 0.01in}c@{\hskip 0.01in}c@{\hskip 0.01in}c@{\hskip 0.01in}c@{\hskip 0.01in}c@{\hskip 0.01in}c@{\hskip 0.01in}c@{\hskip 0.01in}c@{\hskip 0.01in}c@{\hskip 0.01in}c@{\hskip 0.01in}c@{\hskip 0.01in}c}

  \multirow{-1}{*}{\begin{tabular}{c@{\hskip 0.01in}c@{\hskip 0.03in}} 
  \frame{\includegraphics[draft=\draft,width=0.075\linewidth, height=0.070\linewidth]{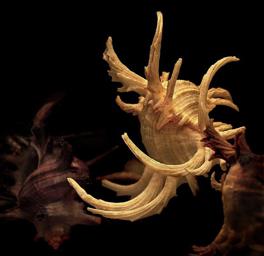}} &
  \frame{\includegraphics[draft=\draft,width=0.075\linewidth, height=0.070\linewidth]{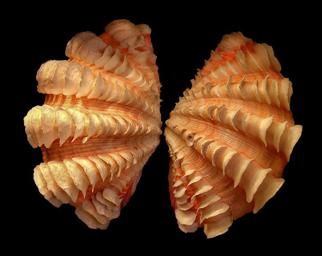}} \tabularnewline[-3pt]	
  \frame{\includegraphics[draft=\draft,width=0.075\linewidth, height=0.070\linewidth]{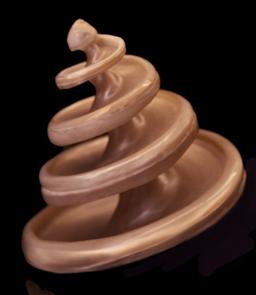}} &
  \frame{\includegraphics[draft=\draft,width=0.075\linewidth, height=0.070\linewidth]{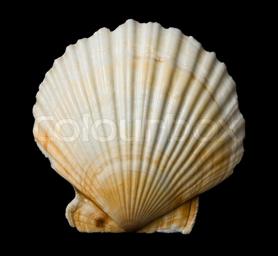}} \tabularnewline[-3pt]
  \frame{\includegraphics[draft=\draft,width=0.075\linewidth, height=0.070\linewidth]{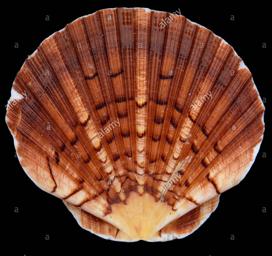}} &
  \frame{\includegraphics[draft=\draft,width=0.075\linewidth, height=0.070\linewidth]{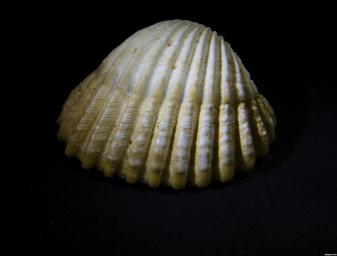}} \tabularnewline[-3pt] 	
  \frame{\includegraphics[draft=\draft,width=0.075\linewidth, height=0.070\linewidth]{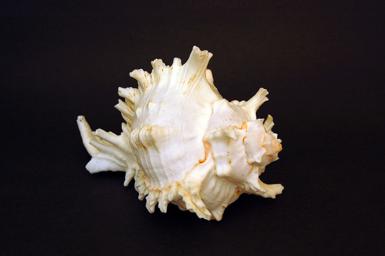}} &
  \frame{\includegraphics[draft=\draft,width=0.075\linewidth, height=0.070\linewidth]{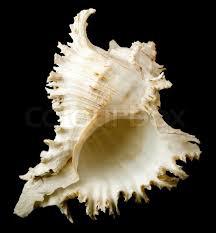}} \tabularnewline[-3pt]
  \frame{\includegraphics[draft=\draft,width=0.075\linewidth, height=0.070\linewidth]{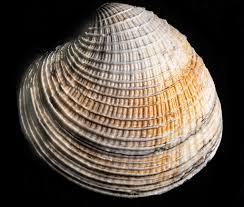}} &
  \frame{\includegraphics[draft=\draft,width=0.075\linewidth, height=0.070\linewidth]{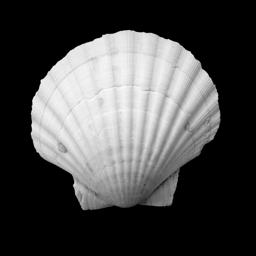}}\tabularnewline[+1pt] 
  \multicolumn{2}{@{\hskip -0.02in}c@{\hskip 0.03in}}{\small 10-shot} \tabularnewline[-2pt] 
  \multicolumn{2}{@{\hskip -0.02in}c@{\hskip 0.03in}}{\small Shells} \tabularnewline[-3pt]
  \end{tabular}} &
  
  \frame{\includegraphics[draft=\draft,width=0.075\linewidth, height=0.070\linewidth]{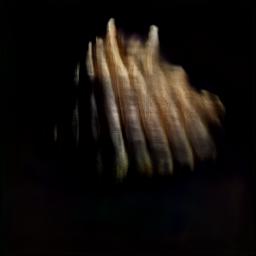}} & 
  \frame{\includegraphics[draft=\draft,width=0.075\linewidth, height=0.070\linewidth]{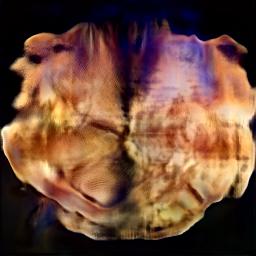}} & 
  \frame{\includegraphics[draft=\draft,width=0.075\linewidth, height=0.070\linewidth]{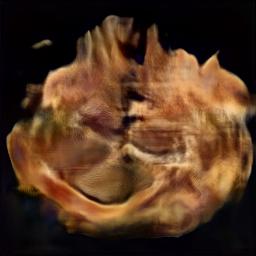}} & 
  \frame{\includegraphics[draft=\draft,width=0.075\linewidth, height=0.070\linewidth]{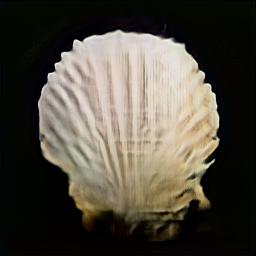}} & 
  \frame{\includegraphics[draft=\draft,width=0.075\linewidth, height=0.070\linewidth]{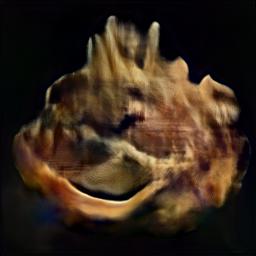}} & 
  \frame{\includegraphics[draft=\draft,width=0.075\linewidth, height=0.070\linewidth]{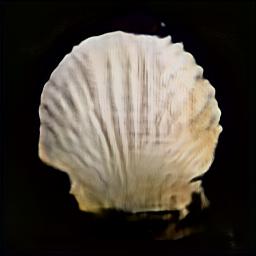}} & 
  \frame{\includegraphics[draft=\draft,width=0.075\linewidth, height=0.070\linewidth]{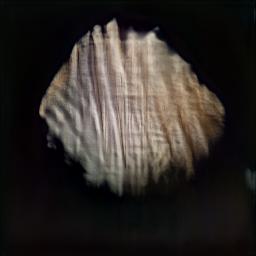}} & 
  \frame{\includegraphics[draft=\draft,width=0.075\linewidth, height=0.070\linewidth]{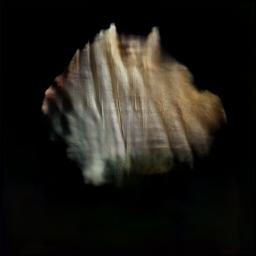}} & 
  \frame{\includegraphics[draft=\draft,width=0.075\linewidth, height=0.070\linewidth]{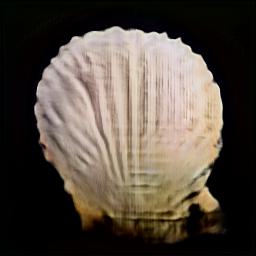}} & 
  \frame{\includegraphics[draft=\draft,width=0.075\linewidth, height=0.070\linewidth]{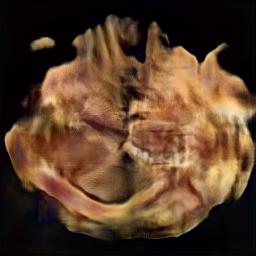}} & ~~\rotatebox{270}{{\hspace{-3.6em} TGAN \hspace{-3.0em} } }
  \tabularnewline[-3pt]
  &
  \frame{\includegraphics[draft=\draft,width=0.075\linewidth, height=0.070\linewidth]{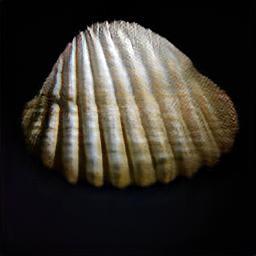}} & 
  \frame{\includegraphics[draft=\draft,width=0.075\linewidth, height=0.070\linewidth]{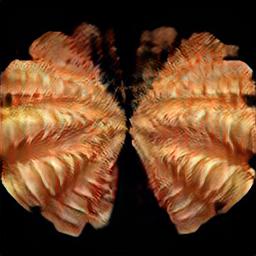}} & 
  \frame{\includegraphics[draft=\draft,width=0.075\linewidth, height=0.070\linewidth]{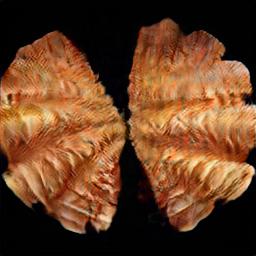}} & 
  \frame{\includegraphics[draft=\draft,width=0.075\linewidth, height=0.070\linewidth]{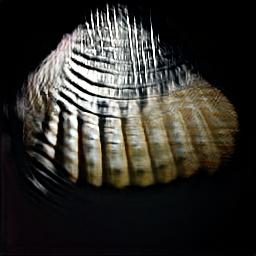}} & 
  \frame{\includegraphics[draft=\draft,width=0.075\linewidth, height=0.070\linewidth]{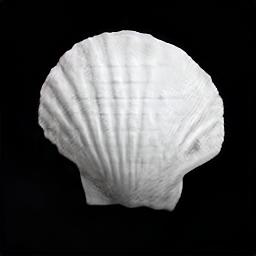}} & 
  \frame{\includegraphics[draft=\draft,width=0.075\linewidth, height=0.070\linewidth]{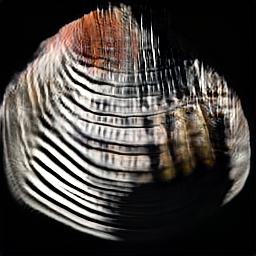}} & 
  \frame{\includegraphics[draft=\draft,width=0.075\linewidth, height=0.070\linewidth]{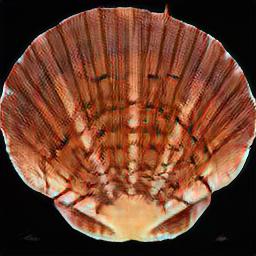}} & 
  \frame{\includegraphics[draft=\draft,width=0.075\linewidth, height=0.070\linewidth]{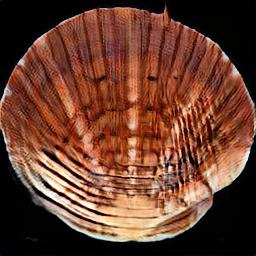}} & 
  \frame{\includegraphics[draft=\draft,width=0.075\linewidth, height=0.070\linewidth]{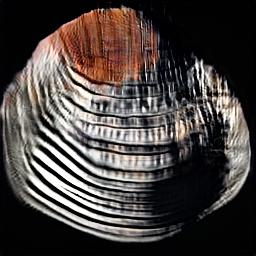}} & 
  \frame{\includegraphics[draft=\draft,width=0.075\linewidth, height=0.070\linewidth]{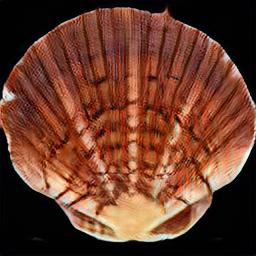}} & ~~\rotatebox{270}{{\hspace{-3.7em} FreezeD \hspace{-3.0em} } }
  \tabularnewline[-3pt]
  &
  \frame{\includegraphics[draft=\draft,width=0.075\linewidth, height=0.070\linewidth]{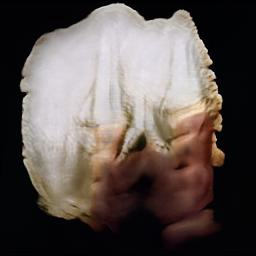}} & 
  \frame{\includegraphics[draft=\draft,width=0.075\linewidth, height=0.070\linewidth]{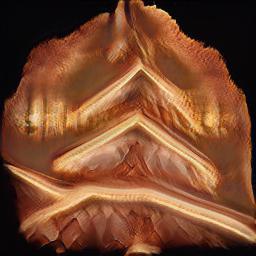}} & 
  \frame{\includegraphics[draft=\draft,width=0.075\linewidth, height=0.070\linewidth]{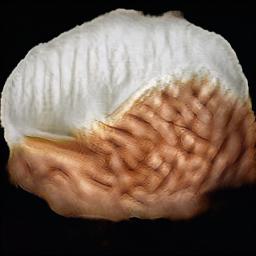}} & 
  \frame{\includegraphics[draft=\draft,width=0.075\linewidth, height=0.070\linewidth]{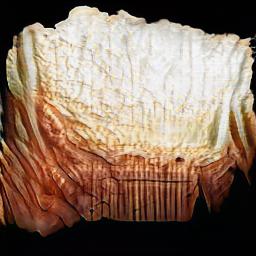}} & 
  \frame{\includegraphics[draft=\draft,width=0.075\linewidth, height=0.070\linewidth]{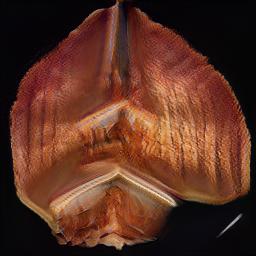}} & 
  \frame{\includegraphics[draft=\draft,width=0.075\linewidth, height=0.070\linewidth]{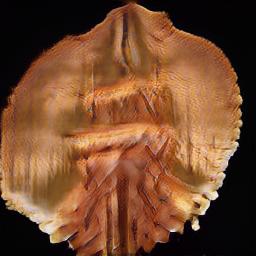}} & 
  \frame{\includegraphics[draft=\draft,width=0.075\linewidth, height=0.070\linewidth]{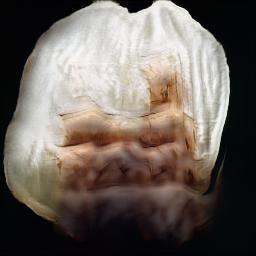}} & 
  \frame{\includegraphics[draft=\draft,width=0.075\linewidth, height=0.070\linewidth]{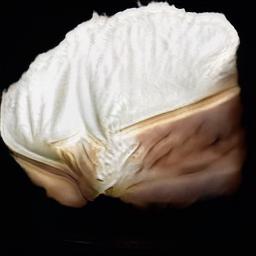}} & 
  \frame{\includegraphics[draft=\draft,width=0.075\linewidth, height=0.070\linewidth]{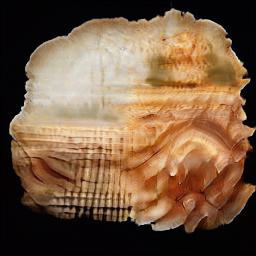}} & 
  \frame{\includegraphics[draft=\draft,width=0.075\linewidth, height=0.070\linewidth]{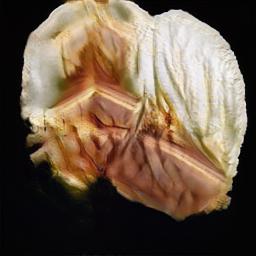}} & ~~\rotatebox{270}{{\hspace{-3.0em} CDC \hspace{-3.0em} } }
  \tabularnewline[-3pt]
  &
  \frame{\includegraphics[draft=\draft,width=0.075\linewidth, height=0.070\linewidth]{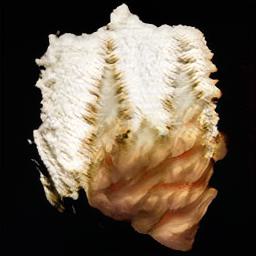}} & 
  \frame{\includegraphics[draft=\draft,width=0.075\linewidth, height=0.070\linewidth]{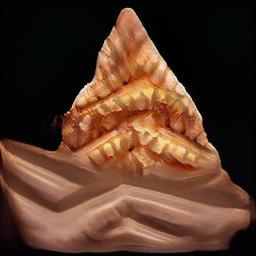}} & 
  \frame{\includegraphics[draft=\draft,width=0.075\linewidth, height=0.070\linewidth]{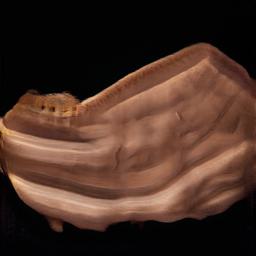}} & 
  \frame{\includegraphics[draft=\draft,width=0.075\linewidth, height=0.070\linewidth]{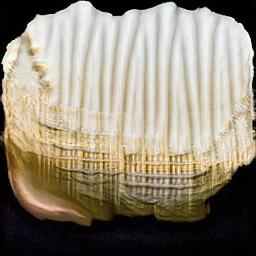}} & 
  \frame{\includegraphics[draft=\draft,width=0.075\linewidth, height=0.070\linewidth]{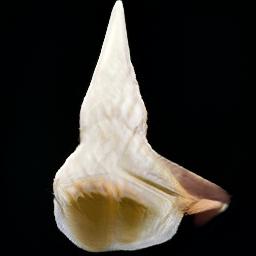}} & 
  \frame{\includegraphics[draft=\draft,width=0.075\linewidth, height=0.070\linewidth]{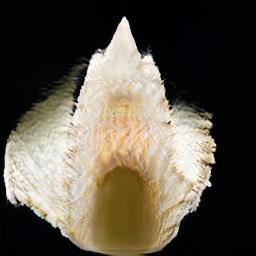}} & 
  \frame{\includegraphics[draft=\draft,width=0.075\linewidth, height=0.070\linewidth]{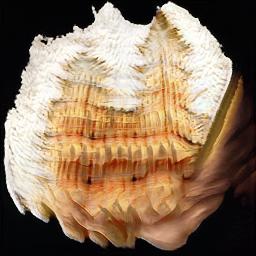}} & 
  \frame{\includegraphics[draft=\draft,width=0.075\linewidth, height=0.070\linewidth]{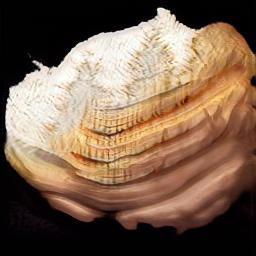}} & 
  \frame{\includegraphics[draft=\draft,width=0.075\linewidth, height=0.070\linewidth]{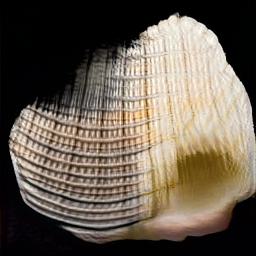}} & 
  \frame{\includegraphics[draft=\draft,width=0.075\linewidth, height=0.070\linewidth]{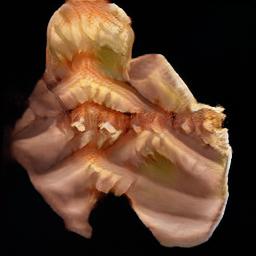}} & ~~\rotatebox{270}{{\hspace{-3.2em} RSSA \hspace{-3.0em} } }
  \tabularnewline[-3pt]
  &
  \frame{\includegraphics[draft=\draft,width=0.075\linewidth, height=0.070\linewidth]{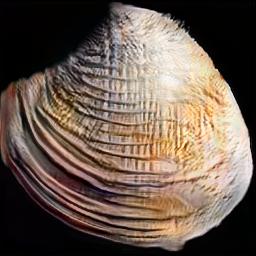}} & 
  \frame{\includegraphics[draft=\draft,width=0.075\linewidth, height=0.070\linewidth]{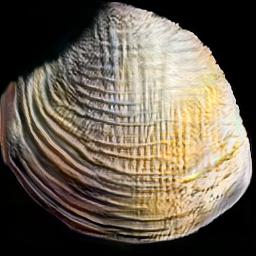}} & 
  \frame{\includegraphics[draft=\draft,width=0.075\linewidth, height=0.070\linewidth]{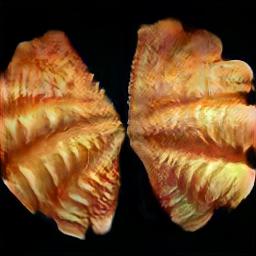}} & 
  \frame{\includegraphics[draft=\draft,width=0.075\linewidth, height=0.070\linewidth]{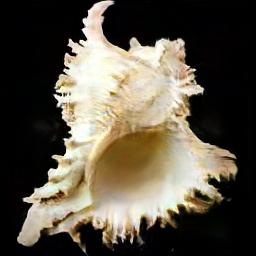}} & 
  \frame{\includegraphics[draft=\draft,width=0.075\linewidth, height=0.070\linewidth]{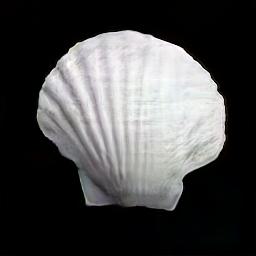}} & 
  \frame{\includegraphics[draft=\draft,width=0.075\linewidth, height=0.070\linewidth]{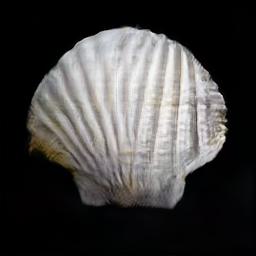}} & 
  \frame{\includegraphics[draft=\draft,width=0.075\linewidth, height=0.070\linewidth]{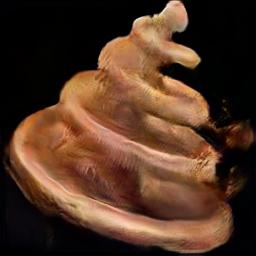}} & 
  \frame{\includegraphics[draft=\draft,width=0.075\linewidth, height=0.070\linewidth]{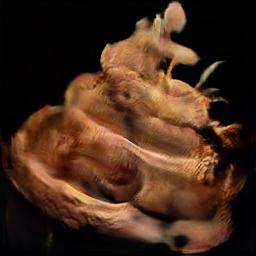}} & 
  \frame{\includegraphics[draft=\draft,width=0.075\linewidth, height=0.070\linewidth]{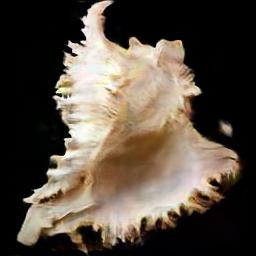}} & 
  \frame{\includegraphics[draft=\draft,width=0.075\linewidth, height=0.070\linewidth]{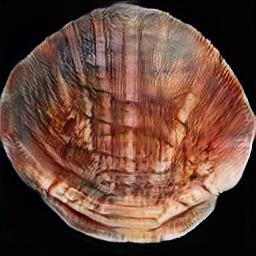}} & ~~\rotatebox{270}{{\hspace{-3.4em} AdAM \hspace{-3.0em} } }
  \tabularnewline[-3pt]
  &
  \frame{\includegraphics[draft=\draft,width=0.075\linewidth, height=0.070\linewidth]{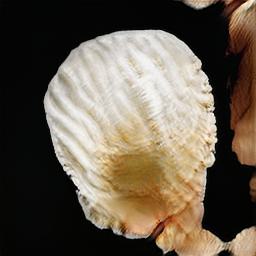}} & 
  \frame{\includegraphics[draft=\draft,width=0.075\linewidth, height=0.070\linewidth]{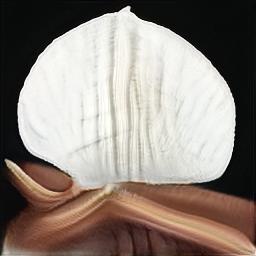}} & 
  \frame{\includegraphics[draft=\draft,width=0.075\linewidth, height=0.070\linewidth]{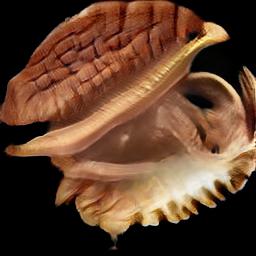}} & 
  \frame{\includegraphics[draft=\draft,width=0.075\linewidth, height=0.070\linewidth]{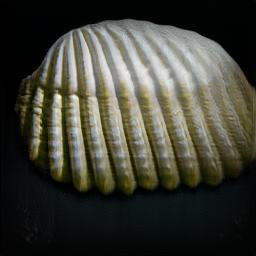}} & 
  \frame{\includegraphics[draft=\draft,width=0.075\linewidth, height=0.070\linewidth]{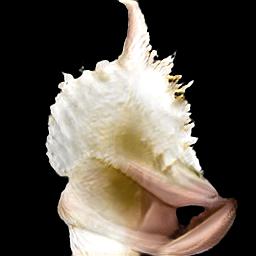}} & 
  \frame{\includegraphics[draft=\draft,width=0.075\linewidth, height=0.070\linewidth]{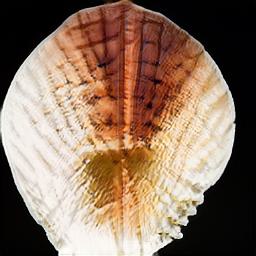}} & 
  \frame{\includegraphics[draft=\draft,width=0.075\linewidth, height=0.070\linewidth]{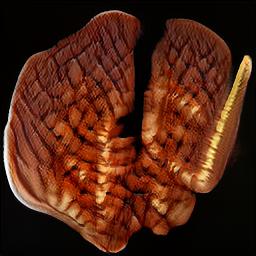}} & 
  \frame{\includegraphics[draft=\draft,width=0.075\linewidth, height=0.070\linewidth]{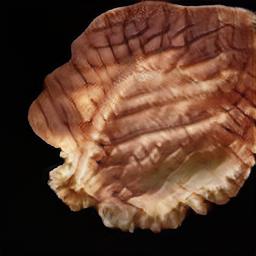}} & 
  \frame{\includegraphics[draft=\draft,width=0.075\linewidth, height=0.070\linewidth]{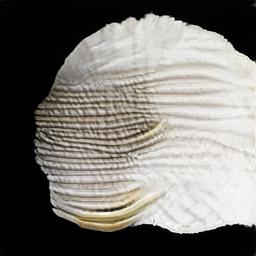}} & 
  \frame{\includegraphics[draft=\draft,width=0.075\linewidth, height=0.070\linewidth]{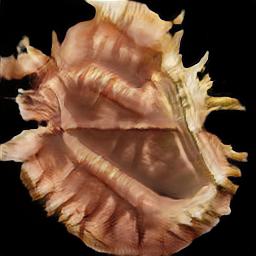}} & ~~\rotatebox{270}{{\hspace{-3.1em} \textbf{Ours} \hspace{-3.0em} } }
  \tabularnewline[-3pt]
  &
  \frame{\includegraphics[draft=\draft,width=0.075\linewidth, height=0.070\linewidth]{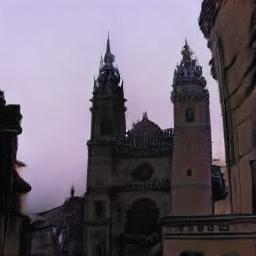}} & 
  \frame{\includegraphics[draft=\draft,width=0.075\linewidth, height=0.070\linewidth]{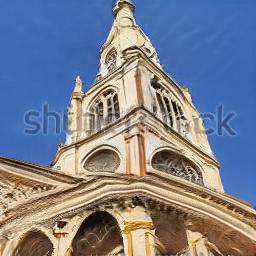}} & 
  \frame{\includegraphics[draft=\draft,width=0.075\linewidth, height=0.070\linewidth]{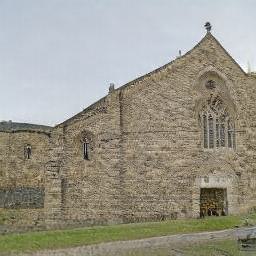}} & 
  \frame{\includegraphics[draft=\draft,width=0.075\linewidth, height=0.070\linewidth]{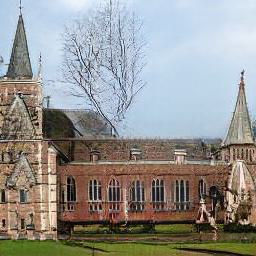}} & 
  \frame{\includegraphics[draft=\draft,width=0.075\linewidth, height=0.070\linewidth]{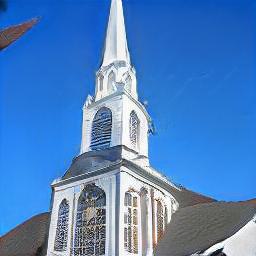}} & 
  \frame{\includegraphics[draft=\draft,width=0.075\linewidth, height=0.070\linewidth]{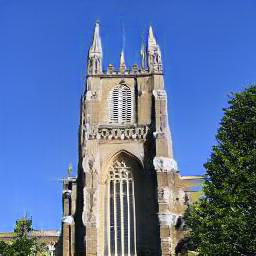}} & 
  \frame{\includegraphics[draft=\draft,width=0.075\linewidth, height=0.070\linewidth]{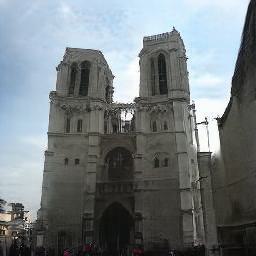}} & 
  \frame{\includegraphics[draft=\draft,width=0.075\linewidth, height=0.070\linewidth]{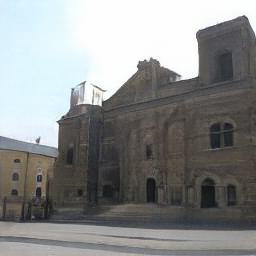}} & 
  \frame{\includegraphics[draft=\draft,width=0.075\linewidth, height=0.070\linewidth]{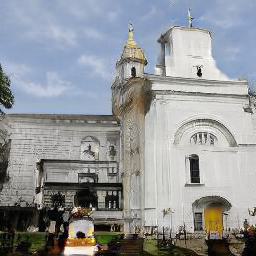}} & 
  \frame{\includegraphics[draft=\draft,width=0.075\linewidth, height=0.070\linewidth]{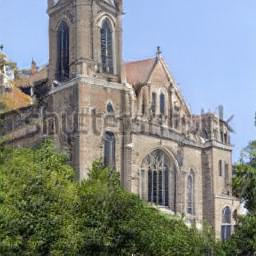}} & ~~\rotatebox{270}{{\hspace{-3.5em} Source 
  \hspace{-3.0em} } }
  \tabularnewline[-4pt]
\end{tabular}

\par\end{centering}
\vspace{5pt}
\caption{Visual comparison to prior methods on \textit{Face$\rightarrow$Anime} and \textit{Church$\rightarrow$Shells}, the source-target dataset pairs with a dissimilar structure (e.g., shapes of objects). In this challenging regime, we observe that prior methods suffer from training instabilities, memorization issues, or inability to adapt the shapes of objects to the new domain. In contrast, our method generates images that look realistic, flexibly combine features of different target images, and transfer the variation of images from the source domain to the target domain. 
}
\label{fig:qualitative_comparison_distant1}
\vspace{-2.5ex}
\end{figure*}

%% file: tex/4_experiments.tex
\section{Experiments}
\label{sec:experiments}

To demonstrate that our approach for few-shot GAN adaptation can be applied to unconditional and class-conditional GANs, we selected for each category a popular GAN architecture: unconditional StyleGANv2 \cite{Karras2019stylegan2} and class-conditional BigGAN \cite{Brock2019}. 
For both models, we test our approach on a variety of source-target domain pairs. We focus on $10$-shot target adaptation in the main paper, but we provide results for $1$-shot and $5$-shot adaptation in the supplementary material. For fair comparisons with prior works, most of our ablations and comparisons are conducted with StyleGANv2. 

\subsection{Adaptation of unconditional GAN}
\label{sec:experiments:stylegan2}

\textbf{Datasets.} In contrast to previous works that mostly considered pairs of similar datasets like \textit{Face$\rightarrow$Sketch} and \textit{Face$\rightarrow$Sunglasses}, we extend the protocol by including structurally dissimilar pairs of source and target domains, which is a more challenging task and is our primary interest. As source generators, we use StyleGANv2 checkpoints pre-trained on FFHQ \cite{Karras2018ASG}, LSUN-Church, and LSUN-Horse \cite{yu15lsun}. For the target datasets, we selected 10-shot subsets of various commonly used few-shot datasets, such as Anime-Face, Shells, or Pokemons \cite{Zhao2020DifferentiableAF,anonymous2021towards}. Results on more datasets are shown in the supplementary material.


\textbf{Training details.} We fine-tune StyleGANv2 using the  $\mathcal{L}_{SS}$ and $\mathcal{L}_{all}$ loss terms as presented in Sec.~\ref{Sec:method}. For the smoothness similarity regularization, we use the intermediate features $G^l$ at resolution (32$\times$32) and set $\lambda_{SS} = 5.0$. 
We follow \cite{ojha2021few} in choosing all the other hyperparameters, such as image resolution (256$\times$256), learning rates, and batch size. Our experiments across all datasets use the same model configuration and set of hyperparameters.

\input{figures/qual_comparison_close1.tex}

\textbf{Baselines.} We compare our method to most recent few-shot GAN adaptation approaches: TGAN \cite{Wang2018TransferringGG},  FreezeD \cite{Mo2020FreezeDA}, CDC \cite{ojha2021few}, RSSA \cite{xiao2022few}, and AdAM \cite{yunqingfew}. In addition, we compare our proposed smoothness similarity regularizer $\mathcal{L}_{SS}$ to other regularization techniques: 
path length regularization (PPL) \cite{Karras2019stylegan2} and MixDL \cite{kong2022few}.


\textbf{Evaluation.} 
In low data regimes, it is necessary to judge results both in quality and diversity aspects, as there is a trade-off between them \cite{robb2021fewshot,sushko2021learning}. We measure the quality with FID \cite{heuselttur2017} between a held-out validation set and a generated set of the same size. Following \cite{ojha2021few}, we evaluate diversity with the intra-LPIPS, clustering generated images according to their nearest training samples and computing the average LPIPS \cite{zhang2018unreasonable} of all the clusters. 
We train all models for 30k epochs in case of dissimilar domain pairs and for 5k on closer domains, evaluating metrics every 1k epochs. Final checkpoints in all experiments correspond to best FID.

\input{tables/comparison_distant.tex}

\textbf{Results with dissimilar source-target domains.}
We first present our results on the source-target domain pairs with dissimilar structure: \textit{Face$\rightarrow$Anime}, \textit{Church$\rightarrow$Shells}, and \textit{Horse$\rightarrow$Pokemon} (see Fig.~\ref{fig:qualitative_comparison_distant1} and supplementary material). Our general observation from Fig.~\ref{fig:qualitative_comparison_distant1} is that in this challenging regime prior methods suffer either from training instabilities, memorization issues, or inability to adapt the shape of objects to the new domain. For example, for \textit{Face$\rightarrow$Anime}, despite an apparent correspondence between the two domains, none of the prior methods successfully transfers the distribution of head poses to the anime style, e.g., 
overfitting too strongly to the 10 provided samples (FreezeD),
failing to adapt the shape of faces to the style of anime (CDC),
or not generating high-quality anime-faces due to instabilities (TGAN, RSSA, AdAM).
Similarly, for \textit{Church$\rightarrow$Shells}, we observe that prior methods 
produce only copies of the example shells (FreezeD, AdAM), 
generate shells of unrealistic church-like shapes (CDC, RSSA),
or suffer from instabilities (TGAN).
In contrast, our method achieves high-quality synthesis, in which the generated images (i) look like realistic anime-faces and shells; (ii) flexibly combine features observed in different target images (e.g., anime hair color can be combined with various eye colors or background styles); and (iii) meaningfully transfer the variation of images from the source domain (e.g., generated shells adjust to the positions and shapes of churches). 

The quantitative comparison in Table~\ref{table:comparison_unrelated} confirms our analysis, where our method achieves the best quality and diversity scores across all datasets. We note a high average relative improvement of more than 18\% and 11\% in FID and LPIPS compared to the highest scores achieved by prior methods. Overall, we conclude that our method significantly improves over prior works on few-shot GAN adaptation with dissimilar source and target domains.

\input{tables/comparison_close.tex}

\textbf{Results with close source-target domains.}
Next, we follow the evaluation of prior works and compare the models on similar source and target domains, such as adaptation of human faces to a different style. The visual results for \textit{Face$\rightarrow$Sketch} and \textit{Face$\rightarrow$Sunglasses} are shown in Fig.~\ref{fig:qualitative_comparison_close1}. Our method successfully performs the few-shot adaptation in this setting, adapting the colors and textures of faces to the gray-scale sketch domain, or adding a novel attribute (sunglasses). 
We note that our method is not explicitly designed to transfer all the details of a face from the source domain, thus changes in the generated images like facial hair are expected. 
Yet, we observe that our method generally does not lose distinctive features of faces in source images, performing on par with previous state-of-the-art methods. The quantitative comparison is provided in Table~\ref{table:comparison_close}\footnote{FID evaluation differs from prior works (discussed in suppl.\ material).}: on both datasets our method achieves the best FID scores and performs on par with the best performer in LPIPS. 


\input{figures/ablation.tex}

\textbf{Ablations.} 
We demonstrate the importance of our proposed loss terms in Fig.~\ref{fig:ablation_interp}, which shows latent space interpolations of trained models and their similarity to the pre-trained source model $G_s$ (row 1). 
Firstly, we note that the plain StyleGANv2 model (row 2) suffers from instabilities in our low data regime, achieving poor image quality and diversity and having ``staircase''-like latent space interpolations. Applying $\mathcal{L}_{SS}$ without $\mathcal{L}_{all}$ (row 3) helps to achieve diverse synthesis with smooth interpolations, but it is not enough to achieve good image quality. On the other hand, using $\mathcal{L}_{all}$ (row 4) helps to overcome instabilities and improve image quality, but it cannot maintain smooth interpolations and high diversity. Finally, our full model (row 5) allows a higher-quality, diverse synthesis with smooth and realistic latent space interpolations. Note how the image transitions mimic the behaviour of the source model (churches and shells change shapes and positions similarly), allowing to achieve diverse and realistic synthesis.

\input{tables/ablation.tex}


The effect of $\mathcal{L}_{all}$ is further demonstrated in Fig.~\ref{fig:plots_losses}, where we show the contribution of different $D$ blocks to the adversarial loss at different epochs. We note the ability of the discriminator to identify correct loss contributions adaptively for different source-target domain pairs. For example for \textit{Face$\rightarrow$Anime}, the network concentrates mostly on the earliest $D$ blocks to adapt the colors and textures of faces to a new style. In contrast, for the more distant domains \textit{Church$\rightarrow$Shells}, the network learns to attribute a higher weight to the later blocks to also adapt higher-level features, such as shapes of objects. In effect, we observe a stabilized adaptation of colors, textures, and shapes of objects across diverse source-target pairs. Using PatchGAN \cite{ojha2021few} as discriminator loss does not achieve such a balance as it focuses mostly on lower-scale features (row 6 in Fig.~\ref{fig:ablation_interp}).

\input{figures/plots_losses.tex}

\input{tables/ablation_SS.tex}

\input{figures/biggan.tex}

Our observations are confirmed by the quantitative study in Table~\ref{table:ablation}: without $\mathcal{L}_{SS}$ the model does not achieve high diversity (high LPIPS), while $\mathcal{L}_{all}$ is necessary for high image quality (low FID). We conclude that both our proposed loss terms are important to achieve high-quality synthesis. More ablations on $\mathcal{L}_{SS}$ and $\mathcal{L}_{all}$ can be found in the supplementary material.

Lastly, Table~\ref{table:ablation_SS} provides a comparison of our proposed $\mathcal{L}_{SS}$ loss term to other regularizers: path length regularization (PPL) \cite{Karras2018ASG} and MixDL \cite{kong2022few}. While all regularizers help to achieve smoother latent space interpolations and thus improve the quality and diversity metrics, our smoothness similarity regularization enables the highest performance in both FID and LPIPS. While our approach transfers the learned smoothness of the source generator to the target domain, PPL and MixDL resort to gradually interpolating between the provided training samples, which leads to latent space interpolations that either look unrealistic or lack diversity (rows 7-8 in Fig.~\ref{fig:ablation_interp}). This demonstrates that transferring smoothness from a pre-trained generator is beneficial to enforce image transitions that are realistic and diverse.

\subsection{Adaptation of class-conditional GAN}
\label{sec:experiments:biggan}

Our approach is not limited to unconditional GANs, but it can also be applied to a class-conditional GAN model. We selected BigGAN \cite{Brock2019} for our experiments as it is a popular backbone architecture for class-conditional image synthesis on ImageNet~\cite{imagenet_cvpr09}. We make two modifications to enable the adaptation of the model to unconditional few-shot datasets. Firstly, we remove the conditioning of the discriminator via the projection layer \cite{miyato2018cgans}. Secondly, we treat the generator's learned continuous class embedding as part of the latent space, thus sampling a Gaussian vector in the joint noise-class space at each fine-tuning epoch. This way, the generator produces an image based on a single input vector in an unconditional fashion. We then fine-tune the pre-trained model using our loss terms $\mathcal{L}_{SS}$ and $\mathcal{L}_{all}$  as presented in Sec.~\ref{Sec:method}. We use image resolution 256$\times$256 and batch size of 32. The hyperparameters for $\mathcal{L}_{SS}$ are the same as for StyleGANv2:  intermediate features $G^l$ at resolution (32$\times$32) and $\lambda_{SS} = 5.0$. We train for 30k epochs and select checkpoints by best FID.

\input{tables/biggan.tex}

\textbf{Datasets.}
As the source generator, we use the BigGAN checkpoint pre-trained on class-conditional ImageNet at resolution 256$\times$256. We demonstrate 10-shot adaptation results with two commonly used few-shot generation datasets: Oxford-Flowers \cite{nilsback2006visual} and Pokemons \cite{anonymous2021towards}. We use the same model configuration for both datasets. 

\textbf{Results.} 
%
Fig.~\ref{fig:qualitative_biggan} demonstrates latent space interpolations of the source and target generators. We note that a simple fine-tuning of BigGAN suffers from training instabilities and mode collapse. In contrast, our method successfully adapts BigGAN to generate diverse images in the target domains. We highlight that our method transfers smooth and realistic image transitions from the well-learned BigGAN's noise-class space, 
despite significant dissimilarities between ImageNet and the few-shot datasets, in particular Pokemons. For example, it can be noticed how the latent space interpolations in the target domains mimic the source domain, e.g., the generated flowers and pokemons change their position and size similarly to dogs and wolves (5$^{\text{th}}$-10$^{\text{th}}$ columns in Fig.~\ref{fig:qualitative_biggan}) or stretch their shape to mimic the proportions of busses (11$^{\text{th}}$-14$^{\text{th}}$ columns).



Table~\ref{table:biggan} shows the importance of our proposed loss terms. Our observations are consistent with the ablations with StylGANv2: $\mathcal{L}_{all}$ is necessary to avoid instabilities and achieve a good image quality (low FID), while $\mathcal{L}_{SS}$ is required to achieve smooth latent space interpolations and good diversity (high LPIPS). We conclude that our method successfully extends to the adaptation of class-conditional models, where target domains benefit from the rich noise-class space learned on a multi-class dataset such as ImageNet. 
More details and results are provided in the supplementary material.

%% file: figures/qual_comparison_close1.tex
\begin{figure*}[t]
\begin{centering}
\setlength{\tabcolsep}{0.01in}
\renewcommand{\arraystretch}{1}
\par\end{centering}
\begin{centering}

\begin{tabular}{@{\hskip -0.13in}c@{\hskip 0.10in}c@{\hskip 0.01in}c@{\hskip 0.01in}c@{\hskip 0.01in}c@{\hskip 0.01in}c@{\hskip 0.01in}c@{\hskip 0.01in}c:c@{\hskip 0.01in}c@{\hskip 0.01in}c@{\hskip 0.01in}c@{\hskip 0.01in}c@{\hskip 0.01in}c@{\hskip 0.01in}c}

~~\rotatebox{90}{{\hspace{0.0em} Real \hspace{0.0em} }} &

\multicolumn{7}{@{\hskip -0.04in}c:}{
\frame{\includegraphics[draft=\draft,width=0.0465\linewidth, height=0.0465\linewidth]{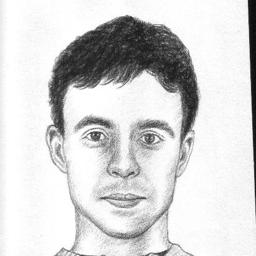}} \hspace{-1.0ex}
\frame{\includegraphics[draft=\draft,width=0.0465\linewidth, height=0.0465\linewidth]{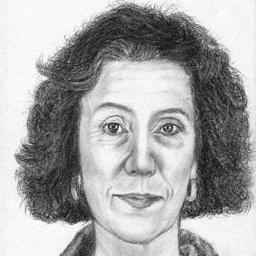}} \hspace{-1.0ex}
\frame{\includegraphics[draft=\draft,width=0.0465\linewidth, height=0.0465\linewidth]{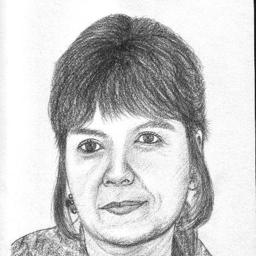}} \hspace{-1.0ex}
\frame{\includegraphics[draft=\draft,width=0.0465\linewidth, height=0.0465\linewidth]{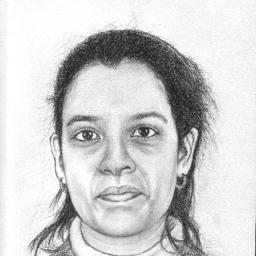}} \hspace{-1.0ex}
\frame{\includegraphics[draft=\draft,width=0.0465\linewidth, height=0.0465\linewidth]{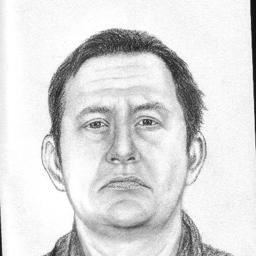}} \hspace{-1.0ex}
\frame{\includegraphics[draft=\draft,width=0.0465\linewidth, height=0.0465\linewidth]{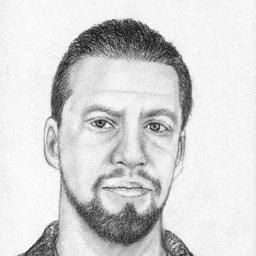}} \hspace{-1.0ex}
\frame{\includegraphics[draft=\draft,width=0.0465\linewidth, height=0.0465\linewidth]{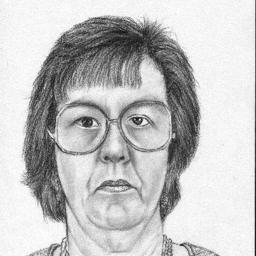}} \hspace{-1.0ex}
\frame{\includegraphics[draft=\draft,width=0.0465\linewidth, height=0.0465\linewidth]{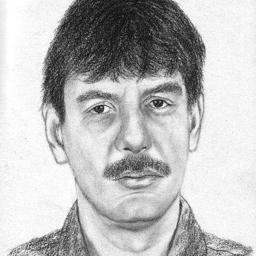}} \hspace{-1.0ex}
\frame{\includegraphics[draft=\draft,width=0.0465\linewidth, height=0.0465\linewidth]{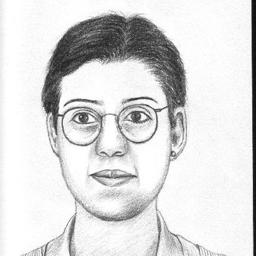}} \hspace{-1.0ex}
\frame{\includegraphics[draft=\draft,width=0.0465\linewidth, height=0.0465\linewidth]{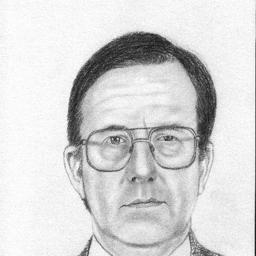}}  
} &
\multicolumn{7}{@{\hskip 0.04in}c}{
\frame{\includegraphics[draft=\draft,width=0.0465\linewidth, height=0.0465\linewidth]{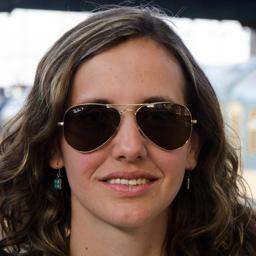}} \hspace{-1.0ex}
\frame{\includegraphics[draft=\draft,width=0.0465\linewidth, height=0.0465\linewidth]{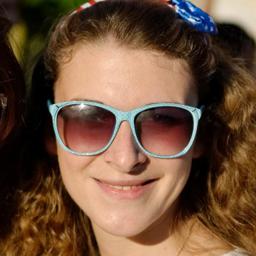}} \hspace{-1.0ex}
\frame{\includegraphics[draft=\draft,width=0.0465\linewidth, height=0.0465\linewidth]{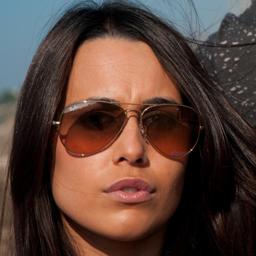}} \hspace{-1.0ex}
\frame{\includegraphics[draft=\draft,width=0.0465\linewidth, height=0.0465\linewidth]{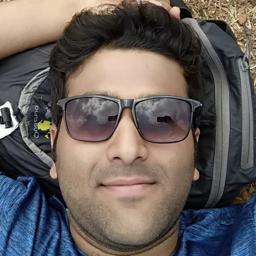}} \hspace{-1.0ex}
\frame{\includegraphics[draft=\draft,width=0.0465\linewidth, height=0.0465\linewidth]{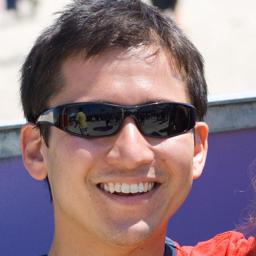}} \hspace{-1.0ex}
\frame{\includegraphics[draft=\draft,width=0.0465\linewidth, height=0.0465\linewidth]{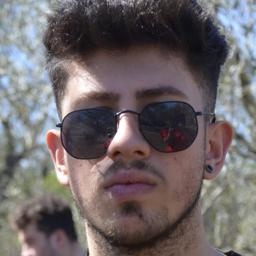}} \hspace{-1.0ex}
\frame{\includegraphics[draft=\draft,width=0.0465\linewidth, height=0.0465\linewidth]{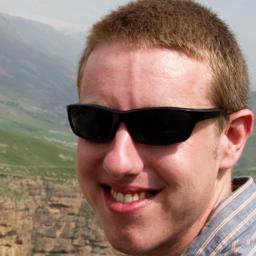}} \hspace{-1.0ex}
\frame{\includegraphics[draft=\draft,width=0.0465\linewidth, height=0.0465\linewidth]{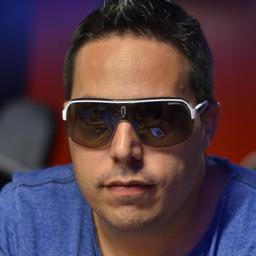}} \hspace{-1.0ex}
\frame{\includegraphics[draft=\draft,width=0.0465\linewidth, height=0.0465\linewidth]{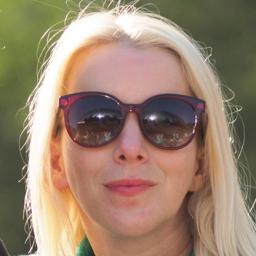}} \hspace{-1.0ex}
\frame{\includegraphics[draft=\draft,width=0.0465\linewidth, height=0.0465\linewidth]{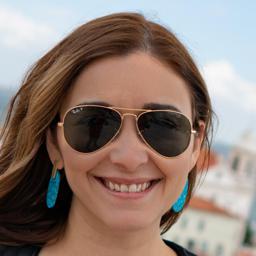}}  
}

\tabularnewline[2pt]

~~\rotatebox{90}{{\hspace{0.0em} Source \hspace{-3.0em} }} &
\frame{\includegraphics[draft=\draft,width=0.067\linewidth, height=0.067\linewidth]{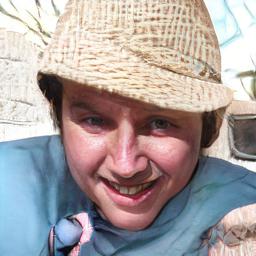}} & 
\frame{\includegraphics[draft=\draft,width=0.067\linewidth, height=0.067\linewidth]{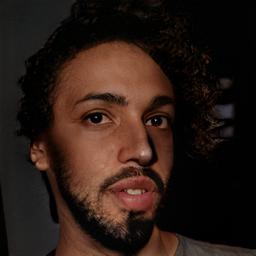}} &
\frame{\includegraphics[draft=\draft,width=0.067\linewidth, height=0.067\linewidth]{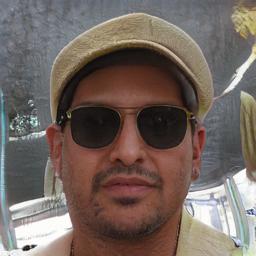}} &
\frame{\includegraphics[draft=\draft,width=0.067\linewidth, height=0.067\linewidth]{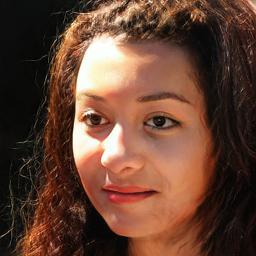}} &
\frame{\includegraphics[draft=\draft,width=0.067\linewidth, height=0.067\linewidth]{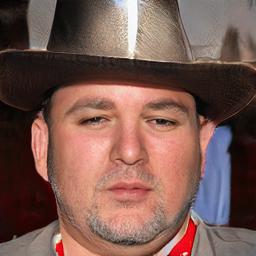}} &
\frame{\includegraphics[draft=\draft,width=0.067\linewidth, height=0.067\linewidth]{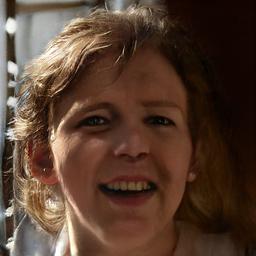}} &
\frame{\includegraphics[draft=\draft,width=0.067\linewidth, height=0.067\linewidth]{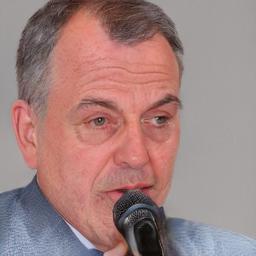}} &

\frame{\includegraphics[draft=\draft,width=0.067\linewidth, height=0.067\linewidth]{figures/qualitative_results/ffhq-sketch/source/p_2_000000.jpg}} & 
\frame{\includegraphics[draft=\draft,width=0.067\linewidth, height=0.067\linewidth]{figures/qualitative_results/ffhq-sketch/source/p_6_000000.jpg}} &
\frame{\includegraphics[draft=\draft,width=0.067\linewidth, height=0.067\linewidth]{figures/qualitative_results/ffhq-sketch/source/p_7_000000.jpg}} &
\frame{\includegraphics[draft=\draft,width=0.067\linewidth, height=0.067\linewidth]{figures/qualitative_results/ffhq-sketch/source/p_8_000000.jpg}} &
\frame{\includegraphics[draft=\draft,width=0.067\linewidth, height=0.067\linewidth]{figures/qualitative_results/ffhq-sketch/source/p_17_000000.jpg}} &
\frame{\includegraphics[draft=\draft,width=0.067\linewidth, height=0.067\linewidth]{figures/qualitative_results/ffhq-sketch/source/p_22_000000.jpg}} &
\frame{\includegraphics[draft=\draft,width=0.067\linewidth, height=0.067\linewidth]{figures/qualitative_results/ffhq-sketch/source/p_23_000000.jpg}}
\tabularnewline[-3pt]

~~\rotatebox{90}{{\hspace{0.4em} CDC \hspace{-3.0em} }} &
\frame{\includegraphics[draft=\draft,width=0.067\linewidth, height=0.067\linewidth]{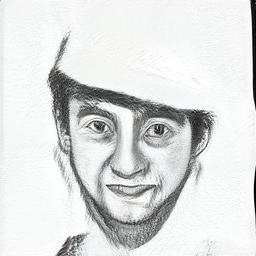}} & 
\frame{\includegraphics[draft=\draft,width=0.067\linewidth, height=0.067\linewidth]{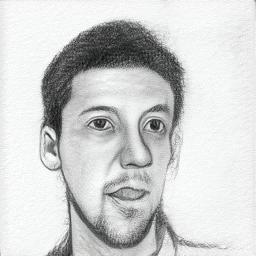}} &
\frame{\includegraphics[draft=\draft,width=0.067\linewidth, height=0.067\linewidth]{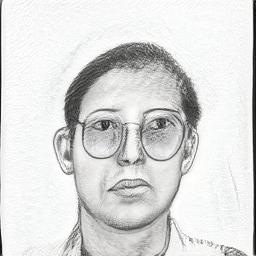}} &
\frame{\includegraphics[draft=\draft,width=0.067\linewidth, height=0.067\linewidth]{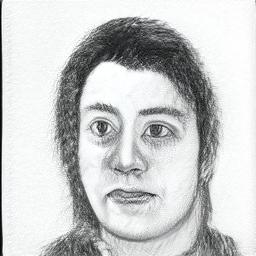}} &
\frame{\includegraphics[draft=\draft,width=0.067\linewidth, height=0.067\linewidth]{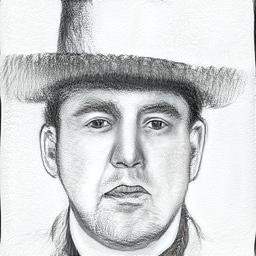}} &
\frame{\includegraphics[draft=\draft,width=0.067\linewidth, height=0.067\linewidth]{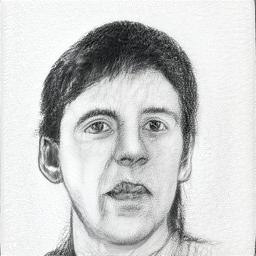}} &
\frame{\includegraphics[draft=\draft,width=0.067\linewidth, height=0.067\linewidth]{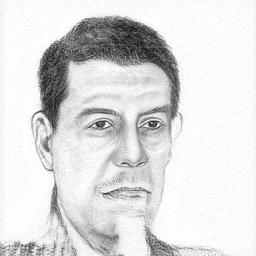}} &

\frame{\includegraphics[draft=\draft,width=0.067\linewidth, height=0.067\linewidth]{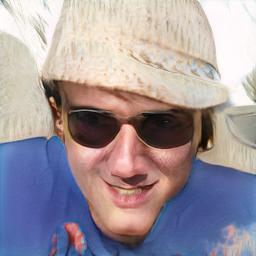}} & 
\frame{\includegraphics[draft=\draft,width=0.067\linewidth, height=0.067\linewidth]{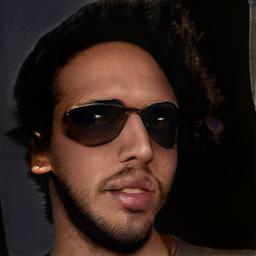}} &
\frame{\includegraphics[draft=\draft,width=0.067\linewidth, height=0.067\linewidth]{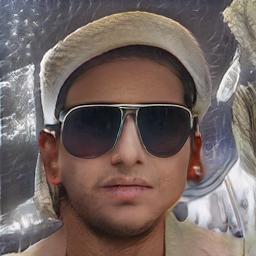}} &
\frame{\includegraphics[draft=\draft,width=0.067\linewidth, height=0.067\linewidth]{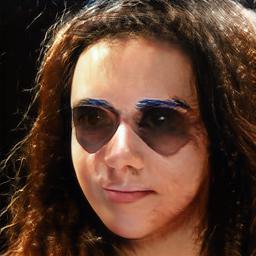}} &
\frame{\includegraphics[draft=\draft,width=0.067\linewidth, height=0.067\linewidth]{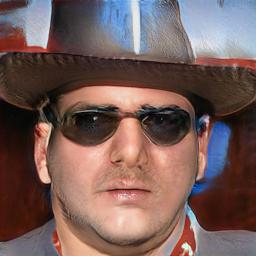}} &
\frame{\includegraphics[draft=\draft,width=0.067\linewidth, height=0.067\linewidth]{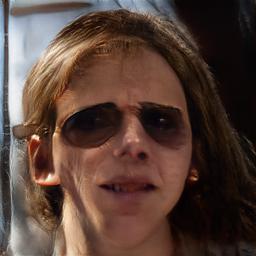}} &
\frame{\includegraphics[draft=\draft,width=0.067\linewidth, height=0.067\linewidth]{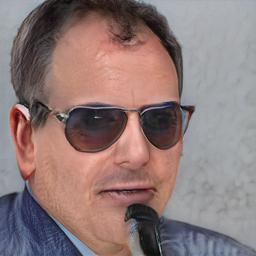}}
\tabularnewline[-3pt]

~~\rotatebox{90}{{\hspace{0.3em} RSSA \hspace{-3.0em} }} &
\frame{\includegraphics[draft=\draft,width=0.067\linewidth, height=0.067\linewidth]{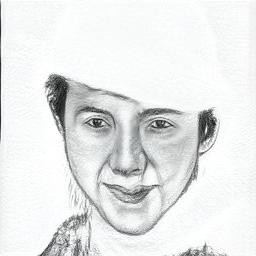}} & 
\frame{\includegraphics[draft=\draft,width=0.067\linewidth, height=0.067\linewidth]{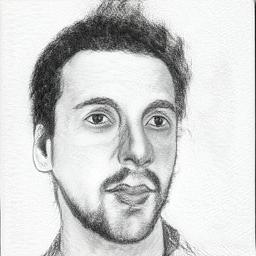}} &
\frame{\includegraphics[draft=\draft,width=0.067\linewidth, height=0.067\linewidth]{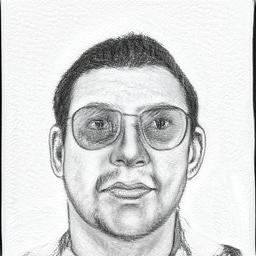}} &
\frame{\includegraphics[draft=\draft,width=0.067\linewidth, height=0.067\linewidth]{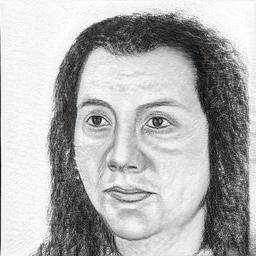}} &
\frame{\includegraphics[draft=\draft,width=0.067\linewidth, height=0.067\linewidth]{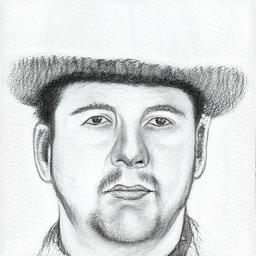}} &
\frame{\includegraphics[draft=\draft,width=0.067\linewidth, height=0.067\linewidth]{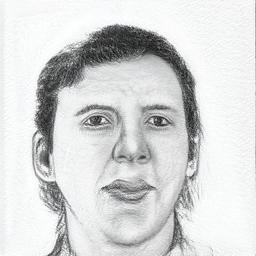}} &
\frame{\includegraphics[draft=\draft,width=0.067\linewidth, height=0.067\linewidth]{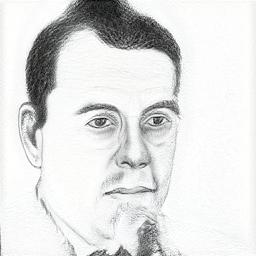}} &

\frame{\includegraphics[draft=\draft,width=0.067\linewidth, height=0.067\linewidth]{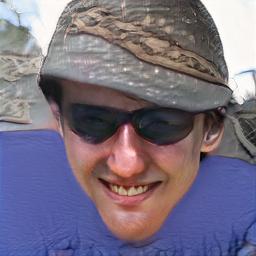}} & 
\frame{\includegraphics[draft=\draft,width=0.067\linewidth, height=0.067\linewidth]{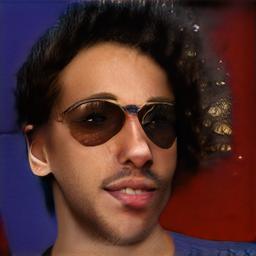}} &
\frame{\includegraphics[draft=\draft,width=0.067\linewidth, height=0.067\linewidth]{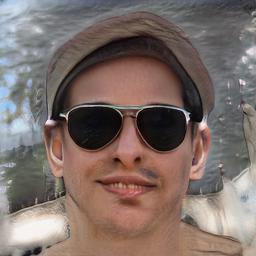}} &
\frame{\includegraphics[draft=\draft,width=0.067\linewidth, height=0.067\linewidth]{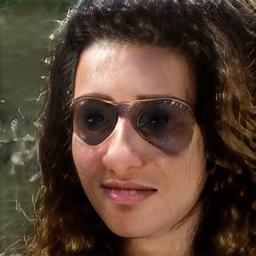}} &
\frame{\includegraphics[draft=\draft,width=0.067\linewidth, height=0.067\linewidth]{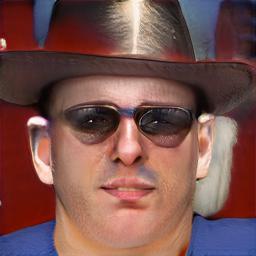}} &
\frame{\includegraphics[draft=\draft,width=0.067\linewidth, height=0.067\linewidth]{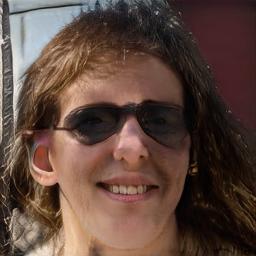}} &
\frame{\includegraphics[draft=\draft,width=0.067\linewidth, height=0.067\linewidth]{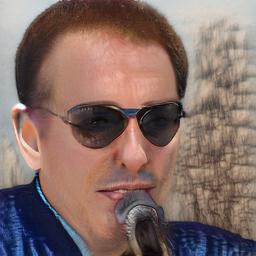}}
\tabularnewline[-3pt]

~~\rotatebox{90}{{\hspace{0.1em} AdAM \hspace{-3.0em} }} &
\frame{\includegraphics[draft=\draft,width=0.067\linewidth, height=0.067\linewidth]{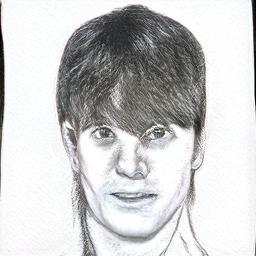}} & 
\frame{\includegraphics[draft=\draft,width=0.067\linewidth, height=0.067\linewidth]{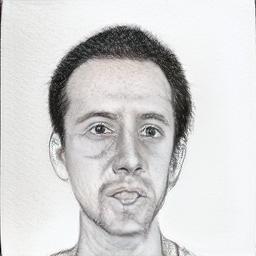}} &
\frame{\includegraphics[draft=\draft,width=0.067\linewidth, height=0.067\linewidth]{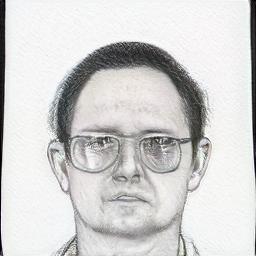}} &
\frame{\includegraphics[draft=\draft,width=0.067\linewidth, height=0.067\linewidth]{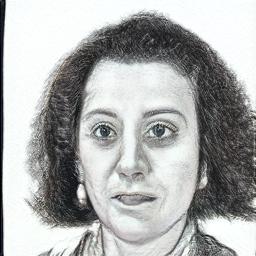}} &
\frame{\includegraphics[draft=\draft,width=0.067\linewidth, height=0.067\linewidth]{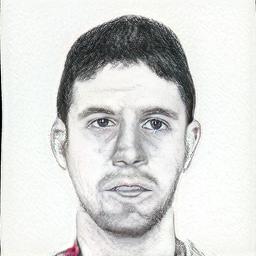}} &
\frame{\includegraphics[draft=\draft,width=0.067\linewidth, height=0.067\linewidth]{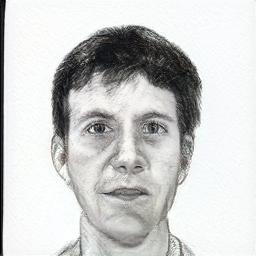}} &
\frame{\includegraphics[draft=\draft,width=0.067\linewidth, height=0.067\linewidth]{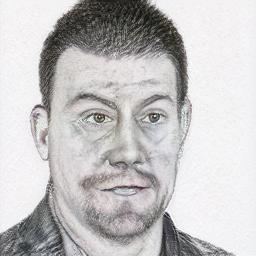}} &

\frame{\includegraphics[draft=\draft,width=0.067\linewidth, height=0.067\linewidth]{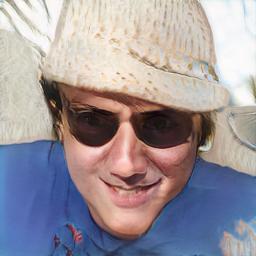}} & 
\frame{\includegraphics[draft=\draft,width=0.067\linewidth, height=0.067\linewidth]{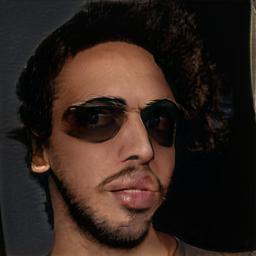}} &
\frame{\includegraphics[draft=\draft,width=0.067\linewidth, height=0.067\linewidth]{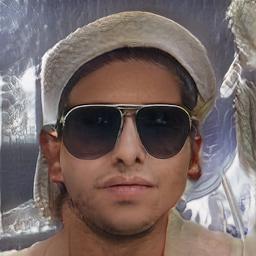}} &
\frame{\includegraphics[draft=\draft,width=0.067\linewidth, height=0.067\linewidth]{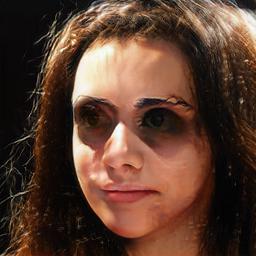}} &
\frame{\includegraphics[draft=\draft,width=0.067\linewidth, height=0.067\linewidth]{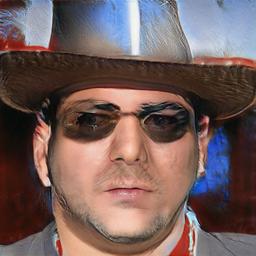}} &
\frame{\includegraphics[draft=\draft,width=0.067\linewidth, height=0.067\linewidth]{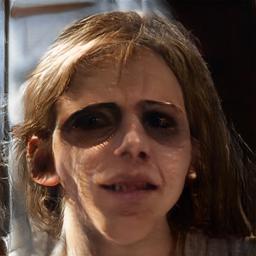}} &
\frame{\includegraphics[draft=\draft,width=0.067\linewidth, height=0.067\linewidth]{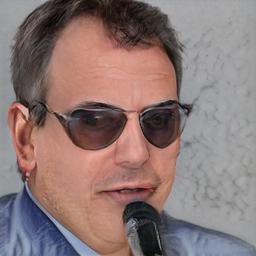}}
\tabularnewline[-3pt]

~~\rotatebox{90}{{\hspace{0.4em} Ours \hspace{-3.0em} }} &
\frame{\includegraphics[draft=\draft,width=0.067\linewidth, height=0.067\linewidth]{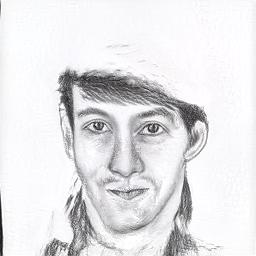}} & 
\frame{\includegraphics[draft=\draft,width=0.067\linewidth, height=0.067\linewidth]{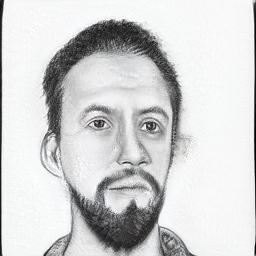}} &
\frame{\includegraphics[draft=\draft,width=0.067\linewidth, height=0.067\linewidth]{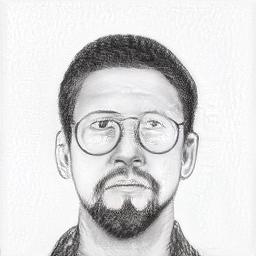}} &
\frame{\includegraphics[draft=\draft,width=0.067\linewidth, height=0.067\linewidth]{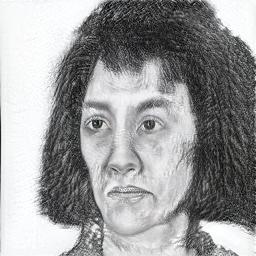}} &
\frame{\includegraphics[draft=\draft,width=0.067\linewidth, height=0.067\linewidth]{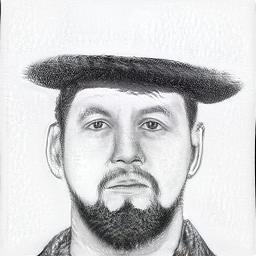}} &
\frame{\includegraphics[draft=\draft,width=0.067\linewidth, height=0.067\linewidth]{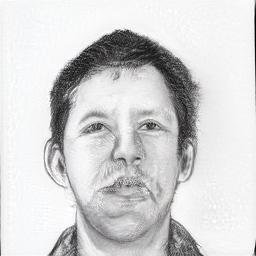}} &
\frame{\includegraphics[draft=\draft,width=0.067\linewidth, height=0.067\linewidth]{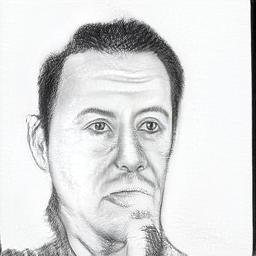}} &

\frame{\includegraphics[draft=\draft,width=0.067\linewidth, height=0.067\linewidth]{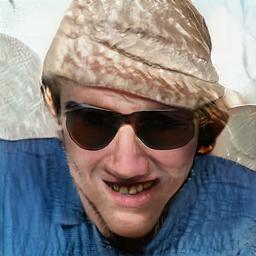}} & 
\frame{\includegraphics[draft=\draft,width=0.067\linewidth, height=0.067\linewidth]{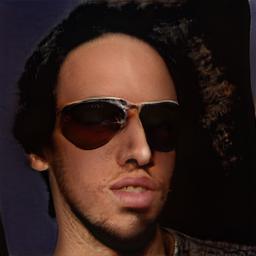}} &
\frame{\includegraphics[draft=\draft,width=0.067\linewidth, height=0.067\linewidth]{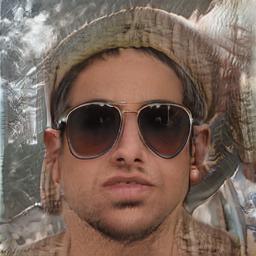}} &
\frame{\includegraphics[draft=\draft,width=0.067\linewidth, height=0.067\linewidth]{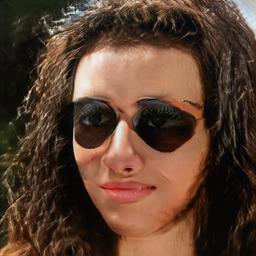}} &
\frame{\includegraphics[draft=\draft,width=0.067\linewidth, height=0.067\linewidth]{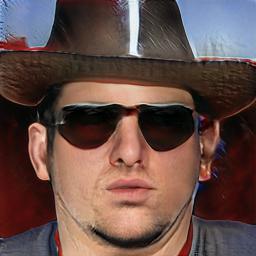}} &
\frame{\includegraphics[draft=\draft,width=0.067\linewidth, height=0.067\linewidth]{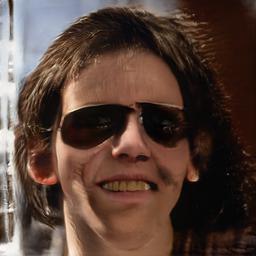}} &
\frame{\includegraphics[draft=\draft,width=0.067\linewidth, height=0.067\linewidth]{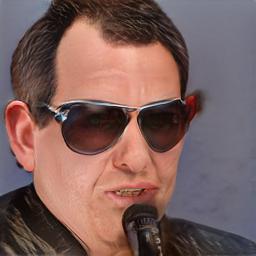}}
\tabularnewline[-3pt]

\end{tabular}
\par\end{centering}
\vspace{5pt}
\caption{Visual comparison to most recent prior methods on \textit{Face$\rightarrow$Sketch} and \textit{Church$\rightarrow$Sunglasses}, the dataset pairs depicting similar image domains. In this regime, our method performs on par with previous state of the art. See Table~\ref{table:comparison_close} for a quantitative comparison.}
\label{fig:qualitative_comparison_close1}
\vspace{-2ex}
\end{figure*}

%% file: tables/comparison_distant.tex
\begin{table}[t]

	\setlength{\tabcolsep}{0.55em}
	\renewcommand{\arraystretch}{0.95}
	\centering
		\begin{tabular}{@{\hskip -0.04in}l@{\hskip 0.02in}|c@{\hskip 0.01in}c@{\hskip 0.02in}|c@{\hskip -0.05in}c@{\hskip -0.06in}|c@{\hskip -0.05in}c@{\hskip -0.10in}}
			\multirow{2}{*}{\footnotesize{} Method} & \multicolumn{2}{c|}{\footnotesize{} \hspace{-0.7ex} \textbf{Face$\rightarrow$Anime}} & \multicolumn{2}{c|}{\footnotesize{} \hspace{-1.5ex} \textbf{Church$\rightarrow$Shells}} & \multicolumn{2}{c}{\footnotesize{} \hspace{-1.0ex}\textbf{Horse$\rightarrow$Pokemons}}
            \tabularnewline 
            & \footnotesize{} FID$\downarrow$  & \footnotesize{} LPIPS$\uparrow$  & \footnotesize{} FID$\downarrow$  & \footnotesize{} LPIPS$\uparrow$  & \footnotesize{} FID$\downarrow$  & \footnotesize{} LPIPS$\uparrow$ 
            \tabularnewline
            
			\hline 	

			{\footnotesize{} \textbf{TGAN} \cite{Wang2018TransferringGG} } & \footnotesize{153.2} & \footnotesize{0.29}  & \footnotesize{205.3} & \footnotesize{0.22}  & \footnotesize{115.0} & \footnotesize{0.52} \tabularnewline

			{\footnotesize{} \textbf{FreezeD} \cite{Mo2020FreezeDA} } & \footnotesize{112.4} & \footnotesize{0.22}  & \footnotesize{180.8} & \footnotesize{0.27}  & \footnotesize{123.3} & \footnotesize{0.49}   \tabularnewline

			{\footnotesize{} \textbf{CDC} \cite{ojha2021few} } & \footnotesize{140.2} & \footnotesize{0.50}  & \footnotesize{187.9} & \footnotesize{0.48}  & \footnotesize{109.5} & \footnotesize{0.55}  \tabularnewline

			{\footnotesize{} \textbf{RSSA} \cite{xiao2022few} } & \footnotesize{133.2} & \footnotesize{0.37}  & \footnotesize{182.4} & \footnotesize{0.44}   & \footnotesize{117.3} & \footnotesize{0.54}   \tabularnewline

			{\footnotesize{} \textbf{AdAM} \cite{yunqingfew} } & \footnotesize{116.4} & \footnotesize{0.42}   & \footnotesize{152.4} & \footnotesize{0.28}  & \footnotesize{106.5} & \footnotesize{0.55}  \tabularnewline

			{\footnotesize{} \textbf{Ours}} & \textbf{\footnotesize{97.3}} & \textbf{\footnotesize{0.57}}   & \textbf{\footnotesize{140.5}} & \textbf{\footnotesize{0.53}}  & \textbf{\footnotesize{84.1}} & \textbf{\footnotesize{0.61}}  \tabularnewline

\end{tabular}
\vspace{0.5ex}
\caption{Comparison of the adaptation performance in case of dissimilar source-target domains. Bold denotes best performance.}
\label{table:comparison_unrelated} %
\vspace{-2ex}
\end{table}

%% file: tables/comparison_close.tex
\begin{table}[t]

	\setlength{\tabcolsep}{0.55em}
	\renewcommand{\arraystretch}{0.95}
	\centering
		\begin{tabular}{@{\hskip -0.04in}l@{\hskip 0.02in}|c@{\hskip 0.01in}c@{\hskip 0.02in}|c@{\hskip -0.05in}c@{\hskip -0.06in}}
			\multirow{2}{*}{\footnotesize{} Method} & \multicolumn{2}{c|}{\footnotesize{} \hspace{-0.7ex} \textbf{Face$\rightarrow$Sketch}} & \multicolumn{2}{c}{\footnotesize{} \hspace{-1.5ex} \textbf{Face$\rightarrow$Sunglasses}} 
            \tabularnewline 
            & \footnotesize{} FID$\downarrow$  & \footnotesize{} LPIPS$\uparrow$  & \footnotesize{} FID$\downarrow$  & \footnotesize{} LPIPS$\uparrow$ 
            \tabularnewline
            
			\hline 	

			{\footnotesize{} \textbf{TGAN} \cite{Wang2018TransferringGG} } & \footnotesize{54.2} & \footnotesize{0.38}  & \footnotesize{36.8} & \footnotesize{0.56}   \tabularnewline

			{\footnotesize{} \textbf{FreezeD} \cite{Mo2020FreezeDA} } & \footnotesize{48.8} & \footnotesize{0.32}  & \footnotesize{32.0} & \footnotesize{0.59}    \tabularnewline

			{\footnotesize{} \textbf{CDC} \cite{ojha2021few} } & \footnotesize{54.2} & \footnotesize{0.40}  & \footnotesize{30.5} & \footnotesize{0.59}   \tabularnewline

			{\footnotesize{} \textbf{RSSA} \cite{xiao2022few} } & \footnotesize{61.4} & \textbf{\footnotesize{0.45}}  & \footnotesize{36.3} & \footnotesize{0.58}     \tabularnewline

			{\footnotesize{} \textbf{AdAM} \cite{yunqingfew} } & \footnotesize{56.3} & \footnotesize{0.37}   & \footnotesize{31.1} & \textbf{\footnotesize{0.60}}   \tabularnewline

			{\footnotesize{} \textbf{Ours}} & \textbf{\footnotesize{45.2}} & \footnotesize{0.44}   & \textbf{\footnotesize{27.5}} & \textbf{\footnotesize{0.60}}   \tabularnewline

\end{tabular}
\vspace{1.5ex}
\caption{Comparison in case of structurally close source-target domains. Bold denotes best performance.}
\label{table:comparison_close} %
\vspace{-2.3ex}
\end{table}

%% file: figures/ablation.tex
\begin{figure}[t]
\begin{centering}
\setlength{\tabcolsep}{0.01in}
\renewcommand{\arraystretch}{1}
\par\end{centering}
\begin{centering}

\vspace{-0.5ex}
\begin{tabular}{@{\hskip -0.08in}c@{\hskip 0.05in}c@{\hskip 0.01in}c@{\hskip 0.01in}c@{\hskip 0.01in}c@{\hskip 0.01in}c@{\hskip 0.01in}c@{\hskip 0.01in}c@{\hskip -0.01in}c}

\multicolumn{8}{c}{\small $\leftarrow$ Interpolation in the latent space $\rightarrow$}
\tabularnewline[0.3pt]

~~\rotatebox{90}{{\hspace{-0.2em} \small (Source) \hspace{-3.0em} }} &
\frame{\includegraphics[draft=\draft,width=0.125\linewidth, height=0.125\linewidth]{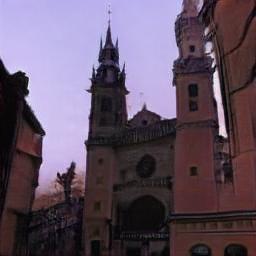}} & 
\frame{\includegraphics[draft=\draft,width=0.125\linewidth, height=0.125\linewidth]{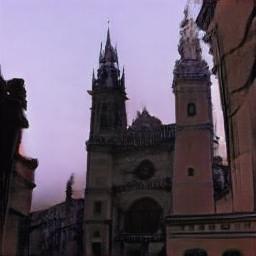}} &
\frame{\includegraphics[draft=\draft,width=0.125\linewidth, height=0.125\linewidth]{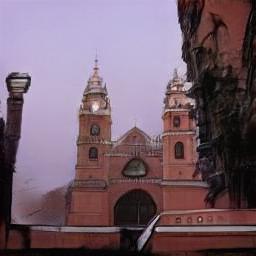}} &
\frame{\includegraphics[draft=\draft,width=0.125\linewidth, height=0.125\linewidth]{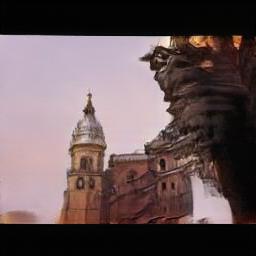}} &
\frame{\includegraphics[draft=\draft,width=0.125\linewidth, height=0.125\linewidth]{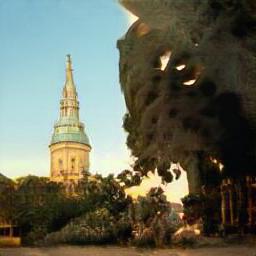}} &
\frame{\includegraphics[draft=\draft,width=0.125\linewidth, height=0.125\linewidth]{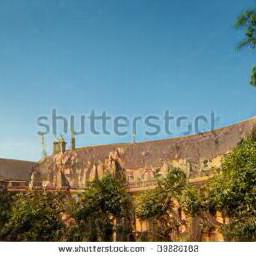}} &
\frame{\includegraphics[draft=\draft,width=0.125\linewidth, height=0.125\linewidth]{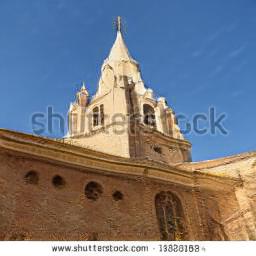}} &
~~\rotatebox{270}{{\hspace{-0.0em} \footnotesize \hspace{0.0em} }}
\tabularnewline[-3pt]

~~\rotatebox{90}{{\hspace{1.00em} \small -- \hspace{-3.0em} }} &
\frame{\includegraphics[draft=\draft,width=0.125\linewidth, height=0.125\linewidth]{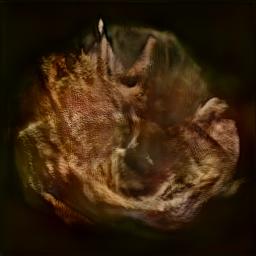}} & 
\frame{\includegraphics[draft=\draft,width=0.125\linewidth, height=0.125\linewidth]{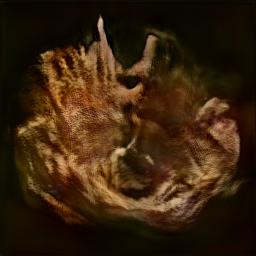}} &
\frame{\includegraphics[draft=\draft,width=0.125\linewidth, height=0.125\linewidth]{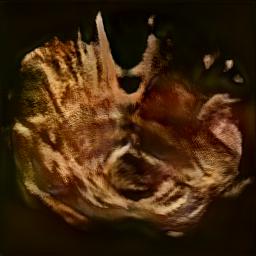}} &
\frame{\includegraphics[draft=\draft,width=0.125\linewidth, height=0.125\linewidth]{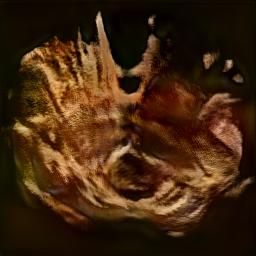}} &
\frame{\includegraphics[draft=\draft,width=0.125\linewidth, height=0.125\linewidth]{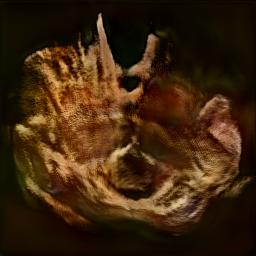}} &
\frame{\includegraphics[draft=\draft,width=0.125\linewidth, height=0.125\linewidth]{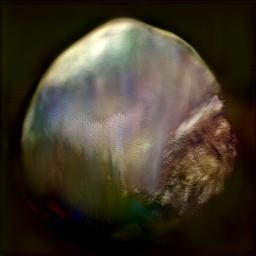}} &
\frame{\includegraphics[draft=\draft,width=0.125\linewidth, height=0.125\linewidth]{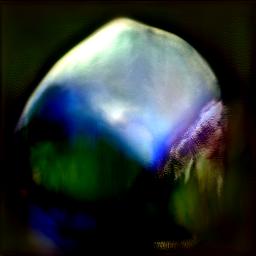}} &
~~\rotatebox{270}{{\hspace{-1.9em} \small -- \hspace{0.0em} }}
\tabularnewline[-3pt]

~~\rotatebox{90}{{\hspace{1.00em} \small -- \hspace{-3.0em} }} &
\frame{\includegraphics[draft=\draft,width=0.125\linewidth, height=0.125\linewidth]{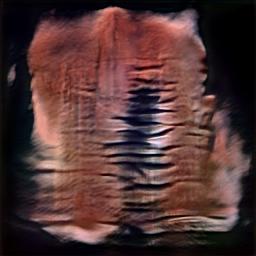}} & 
\frame{\includegraphics[draft=\draft,width=0.125\linewidth, height=0.125\linewidth]{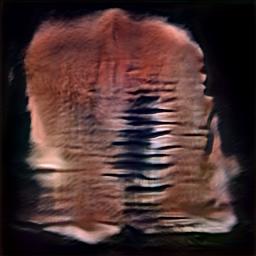}} &
\frame{\includegraphics[draft=\draft,width=0.125\linewidth, height=0.125\linewidth]{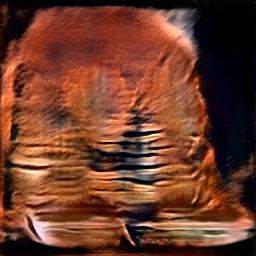}} &
\frame{\includegraphics[draft=\draft,width=0.125\linewidth, height=0.125\linewidth]{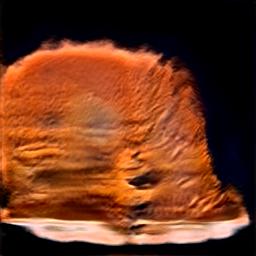}} &
\frame{\includegraphics[draft=\draft,width=0.125\linewidth, height=0.125\linewidth]{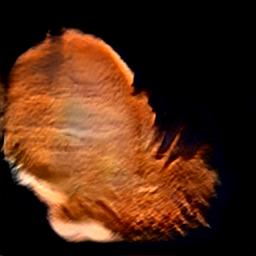}} &
\frame{\includegraphics[draft=\draft,width=0.125\linewidth, height=0.125\linewidth]{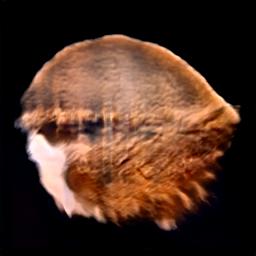}} &
\frame{\includegraphics[draft=\draft,width=0.125\linewidth, height=0.125\linewidth]{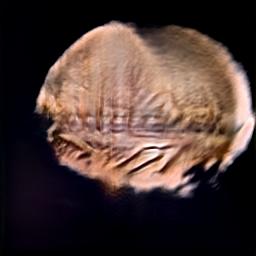}} &
~~\rotatebox{270}{{\hspace{-2.6em} \footnotesize $\mathcal{L}_{SS}$ \hspace{-3.0em} }}
\tabularnewline[-3pt]

~~\rotatebox{90}{{\hspace{0.60em} \footnotesize $\mathcal{L}_{all}$ \hspace{-3.0em} }} &
\frame{\includegraphics[draft=\draft,width=0.125\linewidth, height=0.125\linewidth]{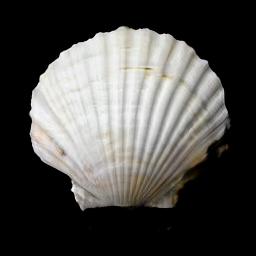}} & 
\frame{\includegraphics[draft=\draft,width=0.125\linewidth, height=0.125\linewidth]{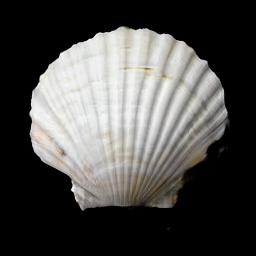}} &
\frame{\includegraphics[draft=\draft,width=0.125\linewidth, height=0.125\linewidth]{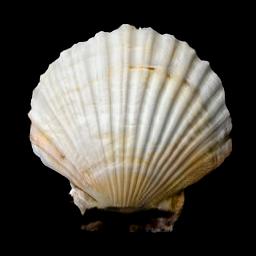}} &
\frame{\includegraphics[draft=\draft,width=0.125\linewidth, height=0.125\linewidth]{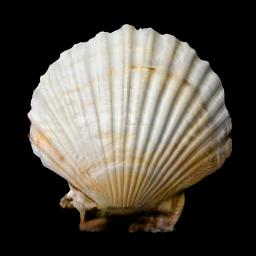}} &
\frame{\includegraphics[draft=\draft,width=0.125\linewidth, height=0.125\linewidth]{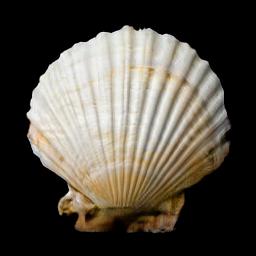}} &
\frame{\includegraphics[draft=\draft,width=0.125\linewidth, height=0.125\linewidth]{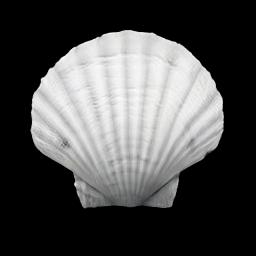}} &
\frame{\includegraphics[draft=\draft,width=0.125\linewidth, height=0.125\linewidth]{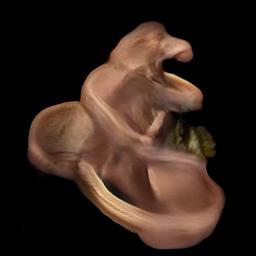}} &
~~\rotatebox{270}{{\hspace{-1.9em} \small -- \hspace{0.0em} }}
\tabularnewline[-3pt]

~~\rotatebox{90}{{\hspace{0.60em} \footnotesize $\mathcal{L}_{all}$ \hspace{-3.0em} }} &
\frame{\includegraphics[draft=\draft,width=0.125\linewidth, height=0.125\linewidth]{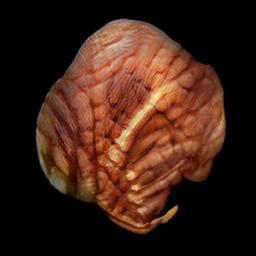}} & 
\frame{\includegraphics[draft=\draft,width=0.125\linewidth, height=0.125\linewidth]{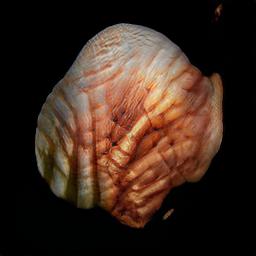}} &
\frame{\includegraphics[draft=\draft,width=0.125\linewidth, height=0.125\linewidth]{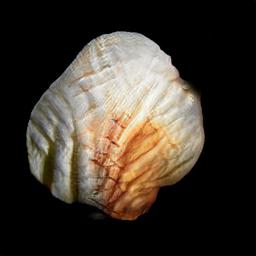}} &
\frame{\includegraphics[draft=\draft,width=0.125\linewidth, height=0.125\linewidth]{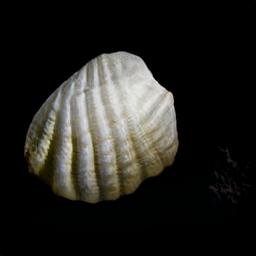}} &
\frame{\includegraphics[draft=\draft,width=0.125\linewidth, height=0.125\linewidth]{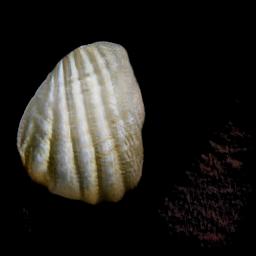}} &
\frame{\includegraphics[draft=\draft,width=0.125\linewidth, height=0.125\linewidth]{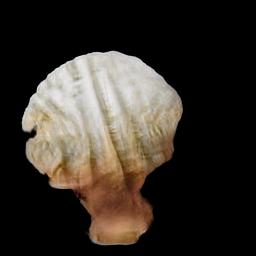}} &
\frame{\includegraphics[draft=\draft,width=0.125\linewidth, height=0.125\linewidth]{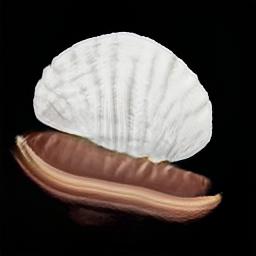}} &
~~\rotatebox{270}{{\hspace{-2.6em} \footnotesize $\mathcal{L}_{SS}$ \hspace{-3.0em} }}
\tabularnewline[-2pt]

~~\rotatebox{90}{{\hspace{-0.35em} \footnotesize Patch\cite{ojha2021few} \hspace{-4.0em} }} &
\frame{\includegraphics[draft=\draft,width=0.125\linewidth, height=0.125\linewidth]{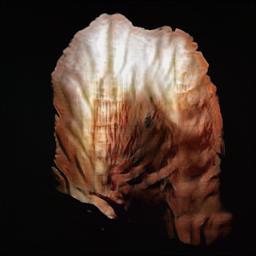}} & 
\frame{\includegraphics[draft=\draft,width=0.125\linewidth, height=0.125\linewidth]{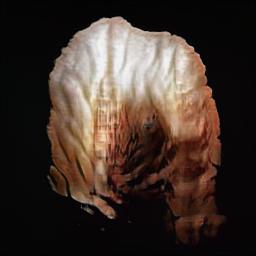}} &
\frame{\includegraphics[draft=\draft,width=0.125\linewidth, height=0.125\linewidth]{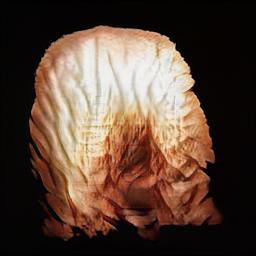}} &
\frame{\includegraphics[draft=\draft,width=0.125\linewidth, height=0.125\linewidth]{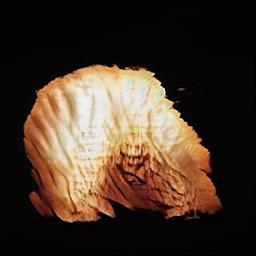}} &
\frame{\includegraphics[draft=\draft,width=0.125\linewidth, height=0.125\linewidth]{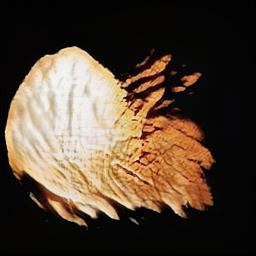}} &
\frame{\includegraphics[draft=\draft,width=0.125\linewidth, height=0.125\linewidth]{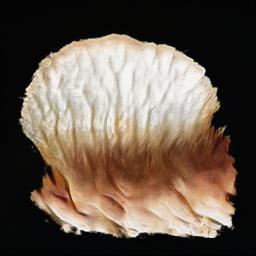}} &
\frame{\includegraphics[draft=\draft,width=0.125\linewidth, height=0.125\linewidth]{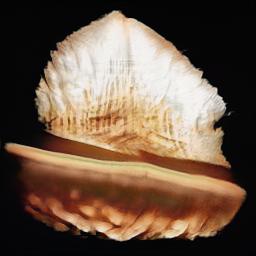}} &
~~\rotatebox{270}{{\hspace{-2.6em} \footnotesize $\mathcal{L}_{SS}$ \hspace{-3.0em} }}
\tabularnewline[-2pt]

~~\rotatebox{90}{{\hspace{0.60em} \footnotesize $\mathcal{L}_{all}$ \hspace{-3.0em} }} &
\frame{\includegraphics[draft=\draft,width=0.125\linewidth, height=0.125\linewidth]{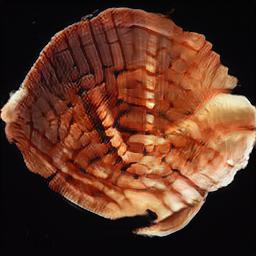}} & 
\frame{\includegraphics[draft=\draft,width=0.125\linewidth, height=0.125\linewidth]{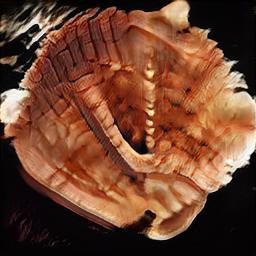}} &
\frame{\includegraphics[draft=\draft,width=0.125\linewidth, height=0.125\linewidth]{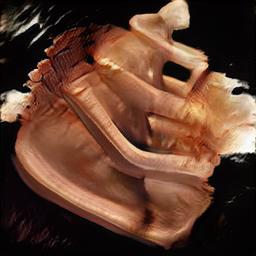}} &
\frame{\includegraphics[draft=\draft,width=0.125\linewidth, height=0.125\linewidth]{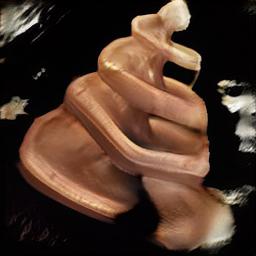}} &
\frame{\includegraphics[draft=\draft,width=0.125\linewidth, height=0.125\linewidth]{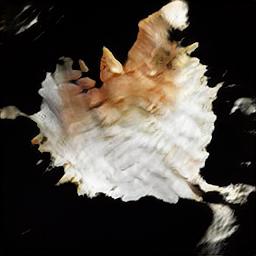}} &
\frame{\includegraphics[draft=\draft,width=0.125\linewidth, height=0.125\linewidth]{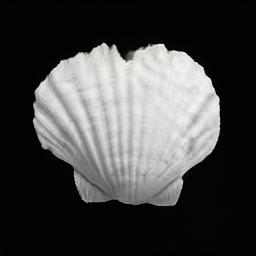}} &
\frame{\includegraphics[draft=\draft,width=0.125\linewidth, height=0.125\linewidth]{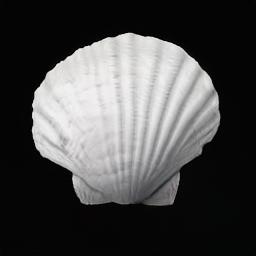}} &
~~\rotatebox{270}{{\hspace{-2.8em} \footnotesize PPL \hspace{-3.0em} }}
\tabularnewline[-3pt]

~~\rotatebox{90}{{\hspace{0.60em} \footnotesize $\mathcal{L}_{all}$ \hspace{-3.0em} }} &
\frame{\includegraphics[draft=\draft,width=0.125\linewidth, height=0.125\linewidth]{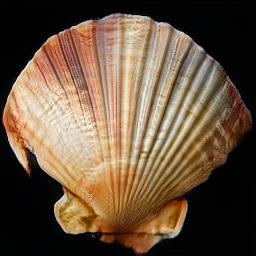}} & 
\frame{\includegraphics[draft=\draft,width=0.125\linewidth, height=0.125\linewidth]{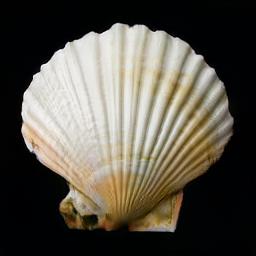}} &
\frame{\includegraphics[draft=\draft,width=0.125\linewidth, height=0.125\linewidth]{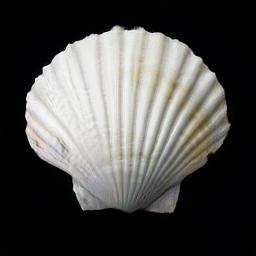}} &
\frame{\includegraphics[draft=\draft,width=0.125\linewidth, height=0.125\linewidth]{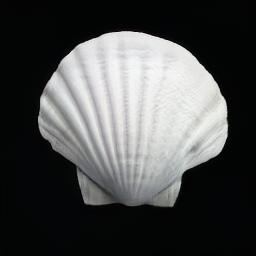}} &
\frame{\includegraphics[draft=\draft,width=0.125\linewidth, height=0.125\linewidth]{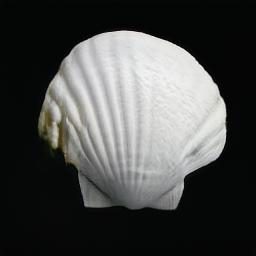}} &
\frame{\includegraphics[draft=\draft,width=0.125\linewidth, height=0.125\linewidth]{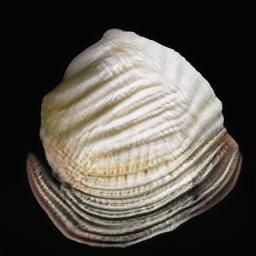}} &
\frame{\includegraphics[draft=\draft,width=0.125\linewidth, height=0.125\linewidth]{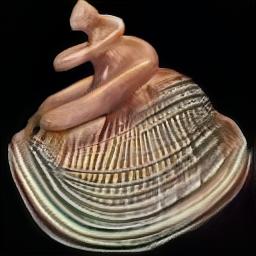}} &
~~\rotatebox{270}{{\hspace{-3.2em} \footnotesize MixDL \hspace{-3.0em} }} 
\tabularnewline[-3pt]

\end{tabular}
\par\end{centering}
\vspace{4pt}
\caption{Latent space interpolations of the source generator and the ablation models from Tables~\ref{table:ablation}-\ref{table:ablation_SS}. Leftmost and rightmost columns show the used $D$ loss and $G$ smoothness regularization.}
\label{fig:ablation_interp}
\vspace{-1.5ex}
\end{figure}

%% file: tables/ablation.tex
\begin{table}[t]

	\setlength{\tabcolsep}{0.55em}
	\renewcommand{\arraystretch}{0.95}
	\centering
		\begin{tabular}{@{\hskip -0.04in}c@{\hskip 0.02in}|@{\hskip 0.03in}c@{\hskip 0.03in}|c@{\hskip -0.01in}c@{\hskip 0.02in}|c@{\hskip -0.05in}c@{\hskip -0.06in}}
			\multirow{2}{*}{\footnotesize{} $D$ loss} & \multirow{1}{*}{\footnotesize{} Smooth} & \multicolumn{2}{c|}{\footnotesize{} \hspace{-0.7ex} \textbf{Face$\rightarrow$Anime}} & \multicolumn{2}{c}{\footnotesize{} \hspace{-1.5ex} \textbf{Church$\rightarrow$Shells}} 
            \tabularnewline 
            & \footnotesize{} reg. for $G$ & \footnotesize{} FID$\downarrow$  & \footnotesize{} LPIPS$\uparrow$  & \footnotesize{} FID$\downarrow$  & \footnotesize{} LPIPS$\uparrow$ 
            \tabularnewline
            
			\hline 	

			{\footnotesize{} StyleGANv2} &  - & \footnotesize{178.0} & \footnotesize{0.21}  & \footnotesize{243.8} & \footnotesize{0.17}  
            \tabularnewline

			{\footnotesize{} StyleGANv2} & \footnotesize{} SS (ours) & \footnotesize{180.7} & \textbf{\footnotesize{0.61}}  & \footnotesize{252.8} & \textbf{\footnotesize{0.62}}  
            \tabularnewline            

            \hdashline[0.1pt/0.5pt]

			{\footnotesize{} PatchGAN \cite{ojha2021few}} &  - & \footnotesize{145.2} & \footnotesize{0.37}  & \footnotesize{183.1} & \footnotesize{0.31}  
            \tabularnewline

			{\footnotesize{} PatchGAN \cite{ojha2021few}} & \footnotesize{} SS (ours) & \footnotesize{132.2} & \footnotesize{0.55}  & \footnotesize{184.2} & \footnotesize{0.56}  
            \tabularnewline

            \hdashline[0.1pt/0.5pt]

            \footnotesize{} $\mathcal{L}_{all}$ (ours) & - & \footnotesize{116.4} & \footnotesize{0.36}  & \footnotesize{175.4} & \footnotesize{0.43}  
            \tabularnewline

            \footnotesize{} $\mathcal{L}_{all}$ (ours) & \footnotesize{} SS (ours) & \textbf{\footnotesize{97.3}} & \footnotesize{0.57}   & \textbf{\footnotesize{140.5}} & \footnotesize{0.53} 

\end{tabular}
\vspace{0.5ex}
\caption{Impact of $\mathcal{L}_{all}$ and $\mathcal{L}_{SS}$. Bold denotes best performance.}
\label{table:ablation} %
\vspace{-1.5ex}
\end{table}

%% file: figures/plots_losses.tex
\begin{figure}[t]
\begin{centering}
\setlength{\tabcolsep}{0.53in}
\renewcommand{\arraystretch}{1}
\par\end{centering}
\begin{centering}
\vspace{-0.5ex}
\begin{tabular}{@{\hskip -0.05in}c@{\hskip 0.00in}c@{\hskip 0.00in}}

\small \textit{Face$\rightarrow$Anime} &
\small \textit{Church$\rightarrow$Shells} \tabularnewline[-2pt]

\includegraphics[width=0.50\linewidth]{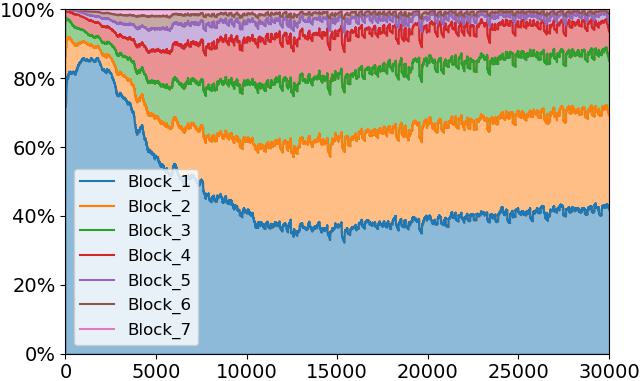} & 
\includegraphics[width=0.50\linewidth]{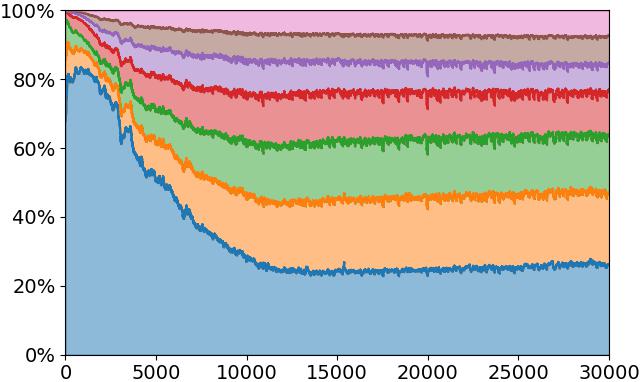} 

\end{tabular}

\par\end{centering}
\vspace{2pt}
\caption{The contribution of features at different $D$ blocks to the adversarial loss function $\mathcal{L}_{all}$. For two closer image domains (the left plot), the network concentrates mostly on earlier layers to compute the loss, while for less similar domains the network learns to use the later layers representing more high-level $D$ features.
}
\label{fig:plots_losses}
\vspace{-1.5ex}
\end{figure}

%% file: tables/ablation_SS.tex
\begin{table}[t]

	\setlength{\tabcolsep}{0.55em}
	\renewcommand{\arraystretch}{0.95}
	\centering
		\begin{tabular}{@{\hskip -0.04in}c@{\hskip 0.02in}|@{\hskip 0.03in}c@{\hskip 0.03in}|c@{\hskip -0.01in}c@{\hskip 0.02in}|c@{\hskip -0.05in}c@{\hskip -0.06in}}
			\multirow{2}{*}{\footnotesize{} $D$ loss} & \multirow{1}{*}{\footnotesize{} Smooth} & \multicolumn{2}{c|}{\footnotesize{} \hspace{-0.7ex} \textbf{Face$\rightarrow$Anime}} & \multicolumn{2}{c}{\footnotesize{} \hspace{-1.5ex} \textbf{Church$\rightarrow$Shells}} 
            \tabularnewline 
            & \footnotesize{} reg. for $G$ & \footnotesize{} FID$\downarrow$  & \footnotesize{} LPIPS$\uparrow$  & \footnotesize{} FID$\downarrow$  & \footnotesize{} LPIPS$\uparrow$ 
            \tabularnewline
            
			\hline

            \footnotesize{} $\mathcal{L}_{all}$ (ours) & - & \footnotesize{116.4} & \footnotesize{0.36}  & \footnotesize{175.4} & \footnotesize{0.43}  
            \tabularnewline

            \hdashline[0.1pt/0.5pt]

			{\footnotesize{} $\mathcal{L}_{all}$ (ours) } & \footnotesize{} PPL \cite{Karras2018ASG} & \footnotesize{107.8}  & \footnotesize{0.46} & \footnotesize{179.4}  & \footnotesize{0.44} 
            \tabularnewline

            \footnotesize{} $\mathcal{L}_{all}$ (ours) & \footnotesize{} MixDL \cite{kong2022few} & \footnotesize{105.9}  & \footnotesize{0.50} & \footnotesize{150.4}  & \footnotesize{0.51} 
            \tabularnewline

            \footnotesize{} $\mathcal{L}_{all}$ (ours) & \footnotesize{} SS (ours) & \textbf{\footnotesize{97.3}} & \textbf{\footnotesize{0.57}}   & \textbf{\footnotesize{140.5}} & \textbf{\footnotesize{0.53}} 

\end{tabular}
\vspace{0.5ex}
\caption{Comparison of smoothness similarity regularization $\mathcal{L}_{SS}$ with other regularizers. Bold denotes best performance.}
\label{table:ablation_SS} %
\vspace{-1.5ex}
\end{table}

%% file: figures/biggan.tex
\begin{figure*}[t]
\begin{centering}
\setlength{\tabcolsep}{0.01in}
\renewcommand{\arraystretch}{1}
\par\end{centering}
\begin{centering}

\begin{tabular}{@{\hskip -0.17in}c@{\hskip 0.01in}c@{\hskip 0.01in}c@{\hskip 0.01in}c@{\hskip 0.01in}c@{\hskip 0.01in}c@{\hskip 0.01in}c@{\hskip 0.02in}c@{\hskip 0.01in}c@{\hskip 0.01in}c@{\hskip 0.01in}c@{\hskip 0.01in}c@{\hskip 0.01in}c@{\hskip 0.01in}c}

\multicolumn{7}{c}{ Real samples -- 10-Shot \color{purple} Flowers \color{black} } &
\multicolumn{7}{c}{ Real samples -- 10-Shot \color{brown} Pokemons \color{black} \color{black} }
\tabularnewline[0pt]
\multicolumn{7}{@{\hskip -0.04in}c}{
\frame{\includegraphics[draft=\draft,width=0.0484\linewidth, height=0.0484\linewidth]{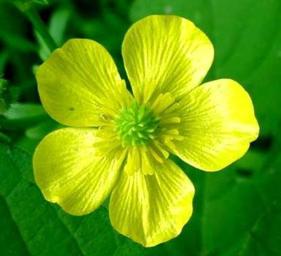}} \hspace{-1.0ex}
\frame{\includegraphics[draft=\draft,width=0.0484\linewidth, height=0.0484\linewidth]{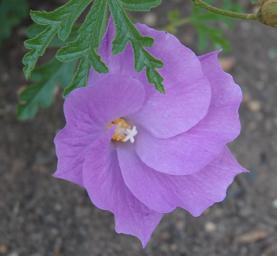}} \hspace{-1.0ex}
\frame{\includegraphics[draft=\draft,width=0.0484\linewidth, height=0.0484\linewidth]{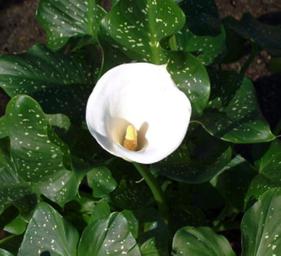}} \hspace{-1.0ex}
\frame{\includegraphics[draft=\draft,width=0.0484\linewidth, height=0.0484\linewidth]{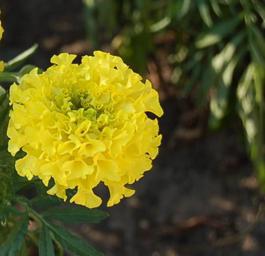}} \hspace{-1.0ex}
\frame{\includegraphics[draft=\draft,width=0.0484\linewidth, height=0.0484\linewidth]{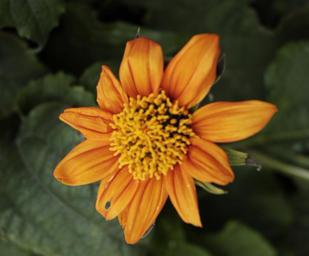}} \hspace{-1.0ex}
\frame{\includegraphics[draft=\draft,width=0.0484\linewidth, height=0.0484\linewidth]{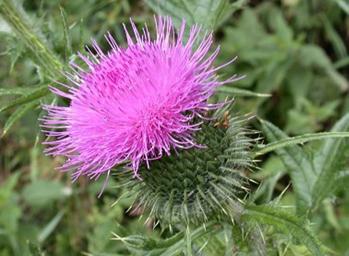}} \hspace{-1.0ex}
\frame{\includegraphics[draft=\draft,width=0.0484\linewidth, height=0.0484\linewidth]{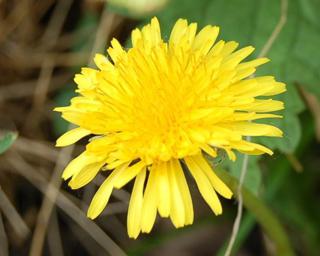}} \hspace{-1.0ex}
\frame{\includegraphics[draft=\draft,width=0.0484\linewidth, height=0.0484\linewidth]{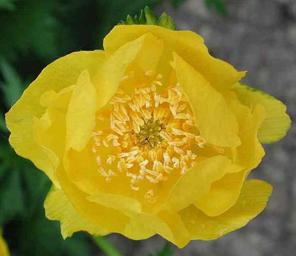}} \hspace{-1.0ex}
\frame{\includegraphics[draft=\draft,width=0.0484\linewidth, height=0.0484\linewidth]{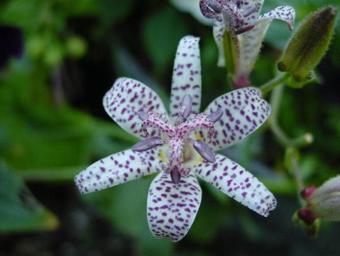}} \hspace{-1.0ex}
\frame{\includegraphics[draft=\draft,width=0.0484\linewidth, height=0.0484\linewidth]{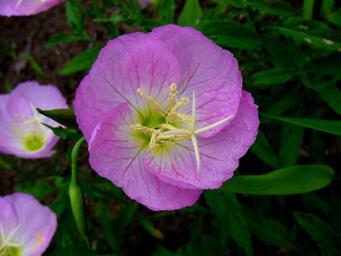}}  
} &
\multicolumn{7}{@{\hskip -0.04in}c}{
\frame{\includegraphics[draft=\draft,width=0.0484\linewidth, height=0.0484\linewidth]{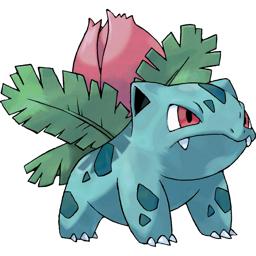}} \hspace{-1.0ex}
\frame{\includegraphics[draft=\draft,width=0.0484\linewidth, height=0.0484\linewidth]{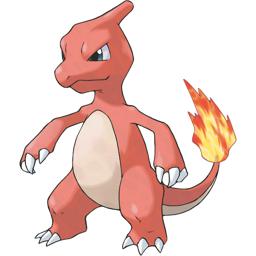}} \hspace{-1.0ex}
\frame{\includegraphics[draft=\draft,width=0.0484\linewidth, height=0.0484\linewidth]{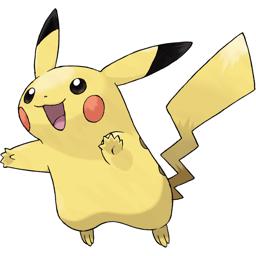}} \hspace{-1.0ex}
\frame{\includegraphics[draft=\draft,width=0.0484\linewidth, height=0.0484\linewidth]{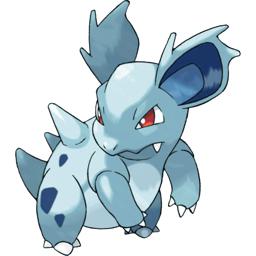}} \hspace{-1.0ex}
\frame{\includegraphics[draft=\draft,width=0.0484\linewidth, height=0.0484\linewidth]{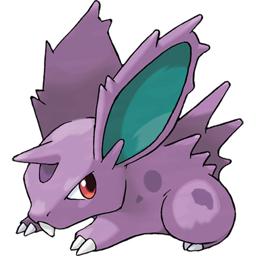}} \hspace{-1.0ex}
\frame{\includegraphics[draft=\draft,width=0.0484\linewidth, height=0.0484\linewidth]{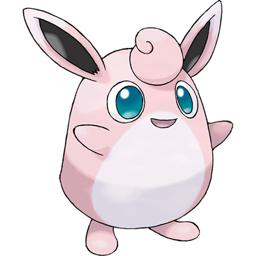}} \hspace{-1.0ex}
\frame{\includegraphics[draft=\draft,width=0.0484\linewidth, height=0.0484\linewidth]{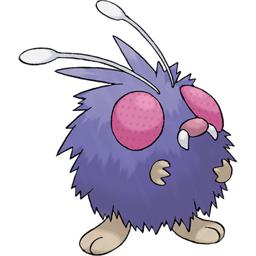}} \hspace{-1.0ex}
\frame{\includegraphics[draft=\draft,width=0.0484\linewidth, height=0.0484\linewidth]{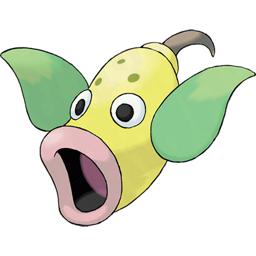}} \hspace{-1.0ex}
\frame{\includegraphics[draft=\draft,width=0.0484\linewidth, height=0.0484\linewidth]{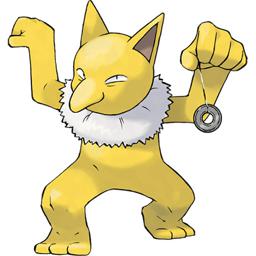}} \hspace{-1.0ex}
\frame{\includegraphics[draft=\draft,width=0.0484\linewidth, height=0.0484\linewidth]{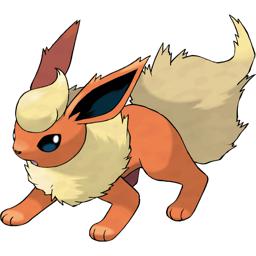}}  
}
\tabularnewline[0pt]
\end{tabular}

\begin{tabular}{ @{\hskip -0.10in} c @{\hskip 0.04in}
c @{\hskip 0.01in} c @{\hskip 0.01in} c @{\hskip 0.01in} c @{\hskip 0.01in} c @{\hskip 0.01in} c @{\hskip 0.01in} c @{\hskip 0.01in} c @{\hskip 0.01in} c@{\hskip 0.01in} c @{\hskip 0.01in} c @{\hskip 0.01in} c @{\hskip 0.01in} c @{\hskip 0.01in} c @{\hskip -0.01in} c @{\hskip 0.01in}}

& \multicolumn{15}{c}{10-Shot adaptation results of our method on class-conditional BigGAN \cite{Brock2019}, pre-trained on ImageNet}
\tabularnewline[-5pt]
~~\rotatebox{90}{{\hspace{-0.3em} \small ImageNet \hspace{0.0em} }} &
\frame{\includegraphics[draft=\draft,width=0.0677\linewidth, height=0.0677\linewidth]{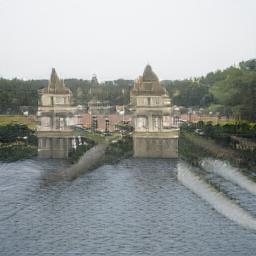}} & 
\frame{\includegraphics[draft=\draft,width=0.0677\linewidth, height=0.0677\linewidth]{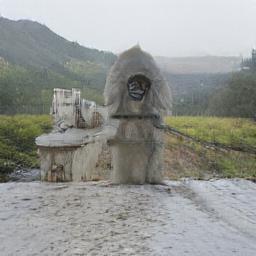}} & 
\frame{\includegraphics[draft=\draft,width=0.0677\linewidth, height=0.0677\linewidth]{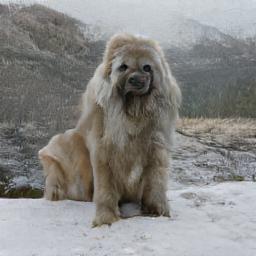}} & 
\frame{\includegraphics[draft=\draft,width=0.0677\linewidth, height=0.0677\linewidth]{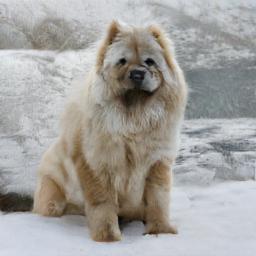}} & 
\frame{\includegraphics[draft=\draft,width=0.0677\linewidth, height=0.0677\linewidth]{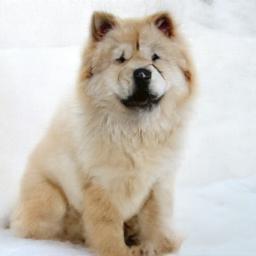}} & 
\frame{\includegraphics[draft=\draft,width=0.0677\linewidth, height=0.0677\linewidth]{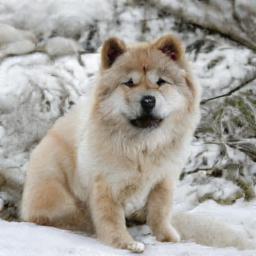}} & 
\frame{\includegraphics[draft=\draft,width=0.0677\linewidth, height=0.0677\linewidth]{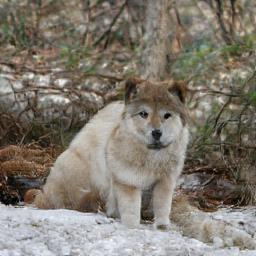}} & 
\frame{\includegraphics[draft=\draft,width=0.0677\linewidth, height=0.0677\linewidth]{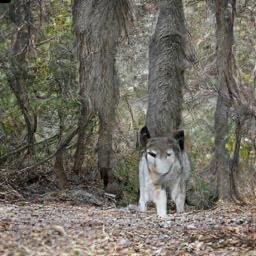}} & \frame{\includegraphics[draft=\draft,width=0.0677\linewidth, height=0.0677\linewidth]{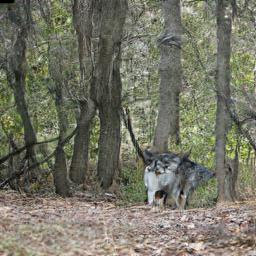}} & 
\frame{\includegraphics[draft=\draft,width=0.0677\linewidth, height=0.0677\linewidth]{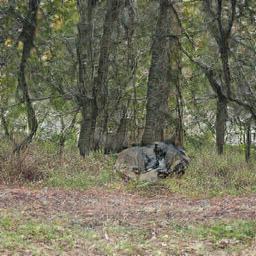}} & 
\frame{\includegraphics[draft=\draft,width=0.0677\linewidth, height=0.0677\linewidth]{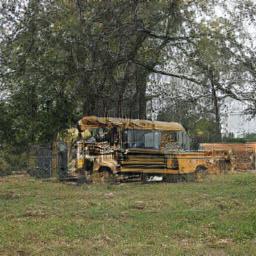}} & 
\frame{\includegraphics[draft=\draft,width=0.0677\linewidth, height=0.0677\linewidth]{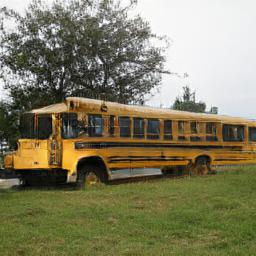}} & \frame{\includegraphics[draft=\draft,width=0.0677\linewidth, height=0.0677\linewidth]{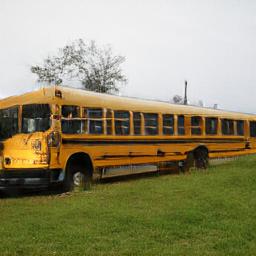}} & 
\frame{\includegraphics[draft=\draft,width=0.0677\linewidth, height=0.0677\linewidth]{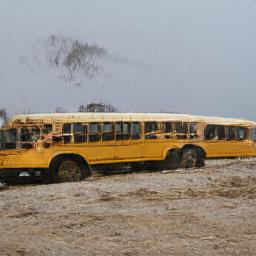}} & ~~\rotatebox{270}{{\hspace{-3.3em} \small Source \hspace{-5.0em} }}
\tabularnewline[1pt]

\multirow{2}{*}{~~\rotatebox{90}{{\hspace{0.2em} \color{purple} \small Flowers \color{black} \hspace{-1.3em} }}} &
\frame{\includegraphics[draft=\draft,width=0.0677\linewidth, height=0.0677\linewidth]{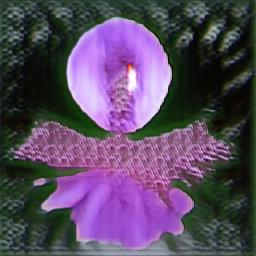}} & 
\frame{\includegraphics[draft=\draft,width=0.0677\linewidth, height=0.0677\linewidth]{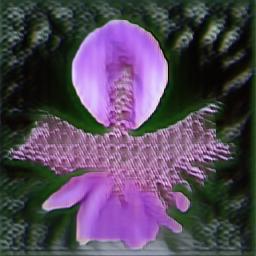}} & 
\frame{\includegraphics[draft=\draft,width=0.0677\linewidth, height=0.0677\linewidth]{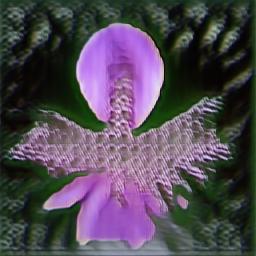}} & 
\frame{\includegraphics[draft=\draft,width=0.0677\linewidth, height=0.0677\linewidth]{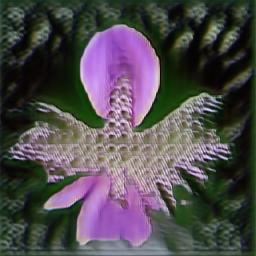}} & 
\frame{\includegraphics[draft=\draft,width=0.0677\linewidth, height=0.0677\linewidth]{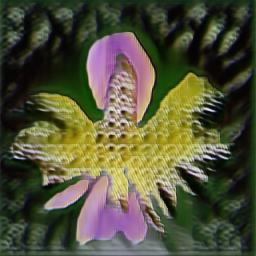}} & 
\frame{\includegraphics[draft=\draft,width=0.0677\linewidth, height=0.0677\linewidth]{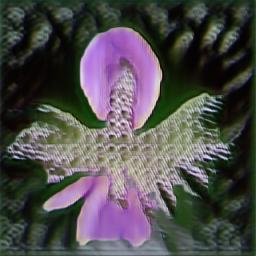}} & 
\frame{\includegraphics[draft=\draft,width=0.0677\linewidth, height=0.0677\linewidth]{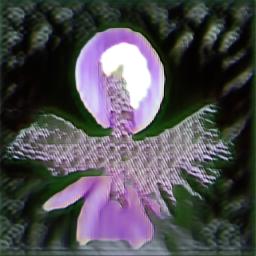}} & 
\frame{\includegraphics[draft=\draft,width=0.0677\linewidth, height=0.0677\linewidth]{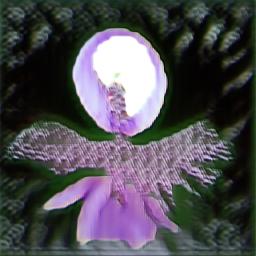}} & \frame{\includegraphics[draft=\draft,width=0.0677\linewidth, height=0.0677\linewidth]{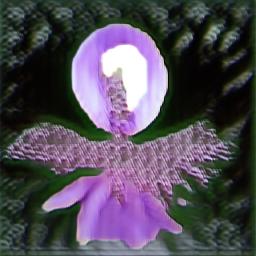}} & 
\frame{\includegraphics[draft=\draft,width=0.0677\linewidth, height=0.0677\linewidth]{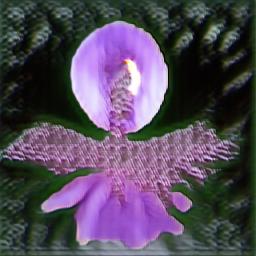}} & 
\frame{\includegraphics[draft=\draft,width=0.0677\linewidth, height=0.0677\linewidth]{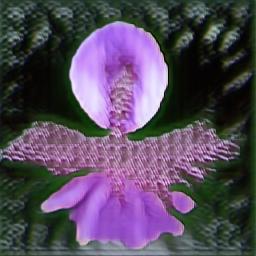}} & 
\frame{\includegraphics[draft=\draft,width=0.0677\linewidth, height=0.0677\linewidth]{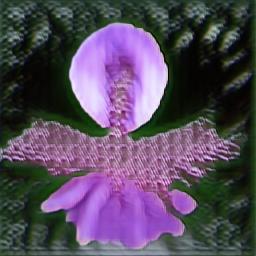}} & \frame{\includegraphics[draft=\draft,width=0.0677\linewidth, height=0.0677\linewidth]{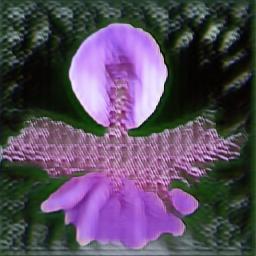}} & 
\frame{\includegraphics[draft=\draft,width=0.0677\linewidth, height=0.0677\linewidth]{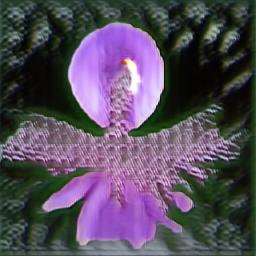}} & ~~\rotatebox{270}{{\hspace{-2.3em} \small FT \hspace{-5.0em} }}
\tabularnewline[-3pt]

&
\frame{\includegraphics[draft=\draft,width=0.0677\linewidth, height=0.0677\linewidth]{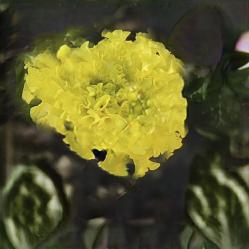}} & 
\frame{\includegraphics[draft=\draft,width=0.0677\linewidth, height=0.0677\linewidth]{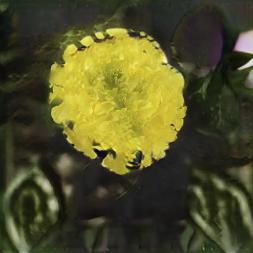}} & 
\frame{\includegraphics[draft=\draft,width=0.0677\linewidth, height=0.0677\linewidth]{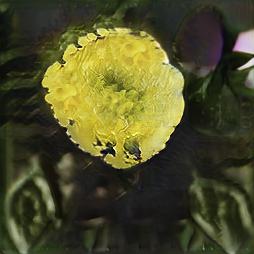}} & 
\frame{\includegraphics[draft=\draft,width=0.0677\linewidth, height=0.0677\linewidth]{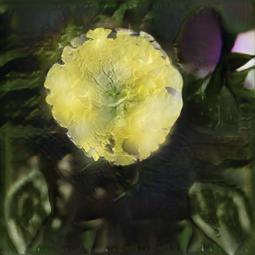}} & 
\frame{\includegraphics[draft=\draft,width=0.0677\linewidth, height=0.0677\linewidth]{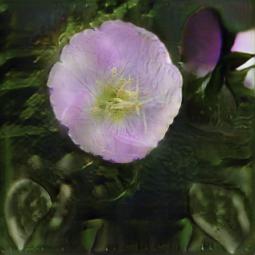}} & 
\frame{\includegraphics[draft=\draft,width=0.0677\linewidth, height=0.0677\linewidth]{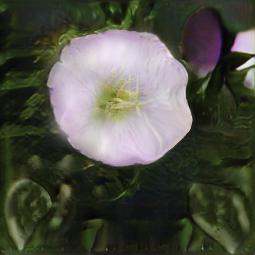}} & 
\frame{\includegraphics[draft=\draft,width=0.0677\linewidth, height=0.0677\linewidth]{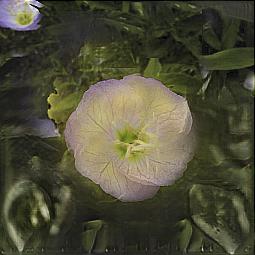}} & 
\frame{\includegraphics[draft=\draft,width=0.0677\linewidth, height=0.0677\linewidth]{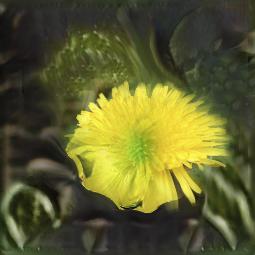}} & \frame{\includegraphics[draft=\draft,width=0.0677\linewidth, height=0.0677\linewidth]{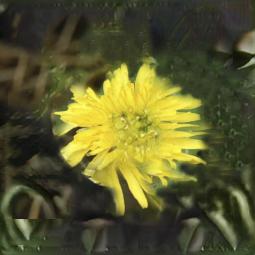}} & 
\frame{\includegraphics[draft=\draft,width=0.0677\linewidth, height=0.0677\linewidth]{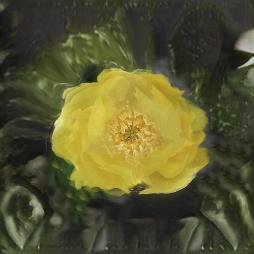}} & 
\frame{\includegraphics[draft=\draft,width=0.0677\linewidth, height=0.0677\linewidth]{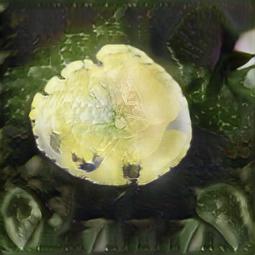}} & 
\frame{\includegraphics[draft=\draft,width=0.0677\linewidth, height=0.0677\linewidth]{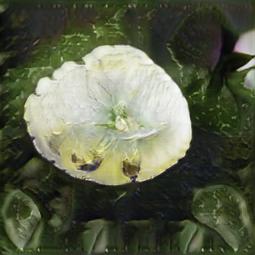}} & \frame{\includegraphics[draft=\draft,width=0.0677\linewidth, height=0.0677\linewidth]{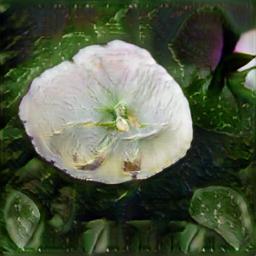}} & 
\frame{\includegraphics[draft=\draft,width=0.0677\linewidth, height=0.0677\linewidth]{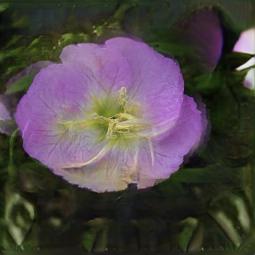}} & ~~\rotatebox{270}{{\hspace{-2.9em} \small \textbf{Ours} \hspace{-5.0em} }}
\tabularnewline[1pt]


\multirow{2}{*}{~~\rotatebox{90}{{\hspace{3.9em} \color{brown} \small Pokemons \color{black} \hspace{-1.5em} }}} &
\frame{\includegraphics[draft=\draft,width=0.0677\linewidth, height=0.0677\linewidth]{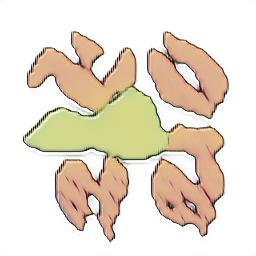}} & 
\frame{\includegraphics[draft=\draft,width=0.0677\linewidth, height=0.0677\linewidth]{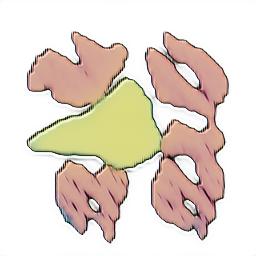}} & 
\frame{\includegraphics[draft=\draft,width=0.0677\linewidth, height=0.0677\linewidth]{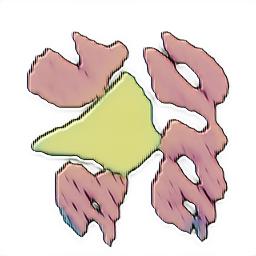}} & 
\frame{\includegraphics[draft=\draft,width=0.0677\linewidth, height=0.0677\linewidth]{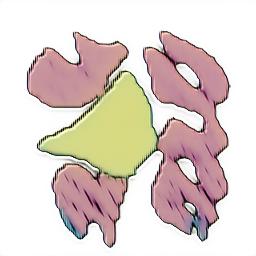}} & 
\frame{\includegraphics[draft=\draft,width=0.0677\linewidth, height=0.0677\linewidth]{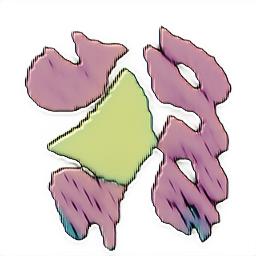}} & 
\frame{\includegraphics[draft=\draft,width=0.0677\linewidth, height=0.0677\linewidth]{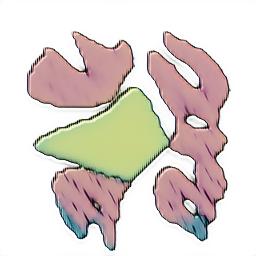}} & 
\frame{\includegraphics[draft=\draft,width=0.0677\linewidth, height=0.0677\linewidth]{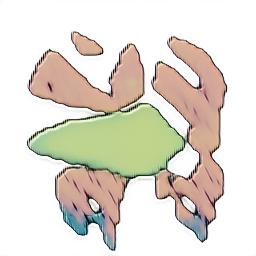}} & 
\frame{\includegraphics[draft=\draft,width=0.0677\linewidth, height=0.0677\linewidth]{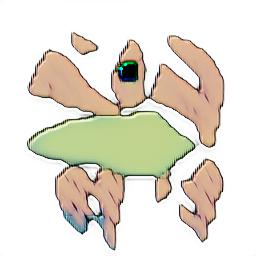}} & 
\frame{\includegraphics[draft=\draft,width=0.0677\linewidth, height=0.0677\linewidth]{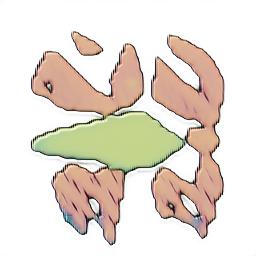}} & 
\frame{\includegraphics[draft=\draft,width=0.0677\linewidth, height=0.0677\linewidth]{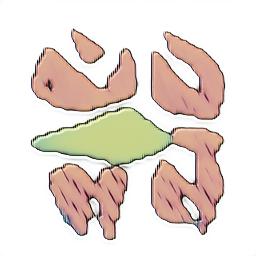}} & 
\frame{\includegraphics[draft=\draft,width=0.0677\linewidth, height=0.0677\linewidth]{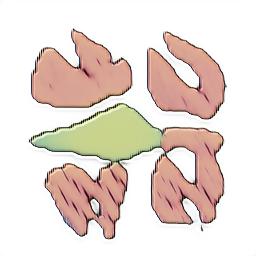}} & 
\frame{\includegraphics[draft=\draft,width=0.0677\linewidth, height=0.0677\linewidth]{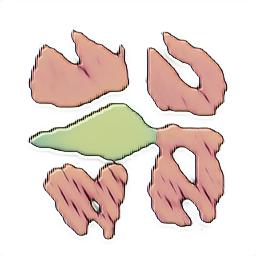}} & 
\frame{\includegraphics[draft=\draft,width=0.0677\linewidth, height=0.0677\linewidth]{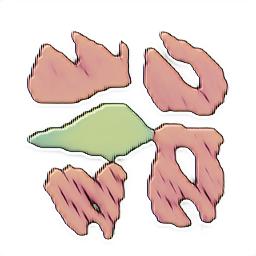}} & 
\frame{\includegraphics[draft=\draft,width=0.0677\linewidth, height=0.0677\linewidth]{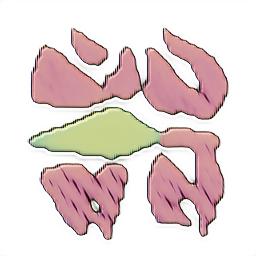}} & ~~\rotatebox{270}{{\hspace{-2.3em} \small FT \hspace{-5.0em} }}
\tabularnewline[-3pt]

&
\frame{\includegraphics[draft=\draft,width=0.0677\linewidth, height=0.0677\linewidth]{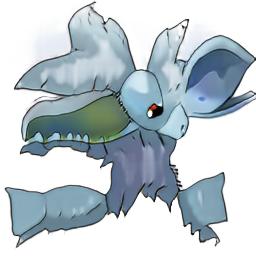}} & 
\frame{\includegraphics[draft=\draft,width=0.0677\linewidth, height=0.0677\linewidth]{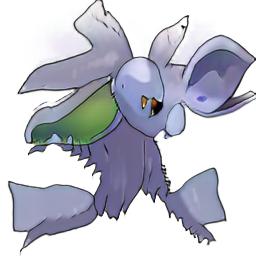}} & 
\frame{\includegraphics[draft=\draft,width=0.0677\linewidth, height=0.0677\linewidth]{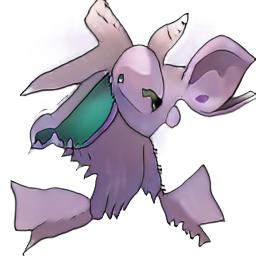}} & 
\frame{\includegraphics[draft=\draft,width=0.0677\linewidth, height=0.0677\linewidth]{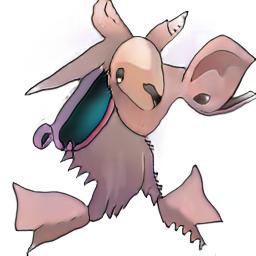}} & 
\frame{\includegraphics[draft=\draft,width=0.0677\linewidth, height=0.0677\linewidth]{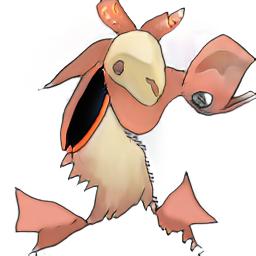}} & 
\frame{\includegraphics[draft=\draft,width=0.0677\linewidth, height=0.0677\linewidth]{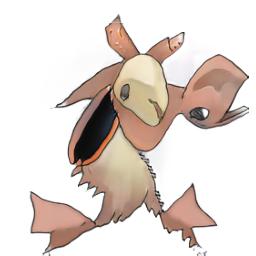}} & 
\frame{\includegraphics[draft=\draft,width=0.0677\linewidth, height=0.0677\linewidth]{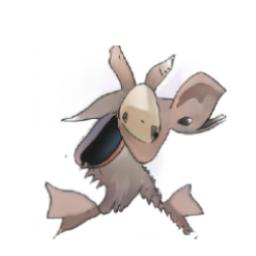}} & 
\frame{\includegraphics[draft=\draft,width=0.0677\linewidth, height=0.0677\linewidth]{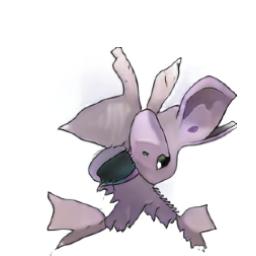}} & 
\frame{\includegraphics[draft=\draft,width=0.0677\linewidth, height=0.0677\linewidth]{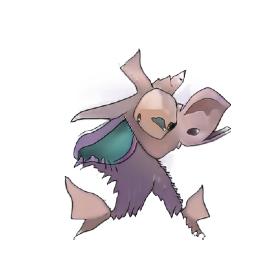}} & 
\frame{\includegraphics[draft=\draft,width=0.0677\linewidth, height=0.0677\linewidth]{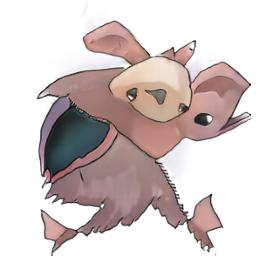}} & 
\frame{\includegraphics[draft=\draft,width=0.0677\linewidth, height=0.0677\linewidth]{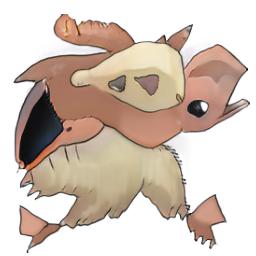}} & 
\frame{\includegraphics[draft=\draft,width=0.0677\linewidth, height=0.0677\linewidth]{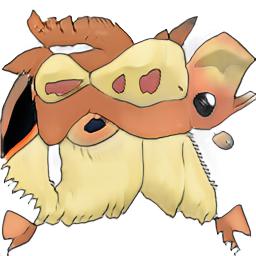}} & 
\frame{\includegraphics[draft=\draft,width=0.0677\linewidth, height=0.0677\linewidth]{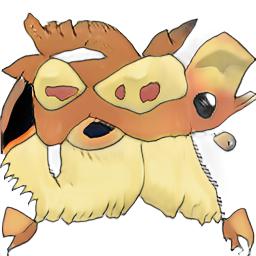}} & 
\frame{\includegraphics[draft=\draft,width=0.0677\linewidth, height=0.0677\linewidth]{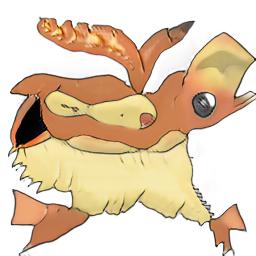}} & ~~\rotatebox{270}{{\hspace{-2.9em} \small \textbf{Ours} \hspace{-5.0em} }}
\tabularnewline[-3pt]
\end{tabular}

\par\end{centering}
\vspace{5pt}
\caption{10-shot adaptation results for the class-conditional BigGAN \cite{Brock2019} pre-trained on ImageNet. While simple fine-tuning (FT) suffers from training instabilities and mode collapse, our method helps to achieve much higher image quality and diversity, transferring smooth and realistic image transitions from the source domain, e.g., objects smoothly changing their locations, size, and shape.
}
\label{fig:qualitative_biggan}
\vspace{-3ex}
\end{figure*}

%% file: tables/biggan.tex
\begin{table}[t]

	\setlength{\tabcolsep}{0.55em}
	\renewcommand{\arraystretch}{0.95}
	\centering
		\begin{tabular}{@{\hskip -0.04in}c@{\hskip 0.02in}|@{\hskip 0.03in}c@{\hskip 0.03in}|c@{\hskip -0.15in}c@{\hskip -0.02in}|c@{\hskip -0.15in}c@{\hskip -0.00in}}
			\multirow{2}{*}{\footnotesize{} $D$ loss} & \multirow{1}{*}{\footnotesize{} Smooth} & \multicolumn{2}{c|}{\footnotesize{} \hspace{-1.5ex} \textbf{ImageNet$\rightarrow$Flowers}} & \multicolumn{2}{c}{\footnotesize{} \hspace{-1.5ex} \textbf{ImageNet$\rightarrow$Pokemons}} 
            \tabularnewline 
            & \footnotesize{} reg. for $G$ & \footnotesize{} ~~~FID$\downarrow$  & \footnotesize{} ~LPIPS$\uparrow$  & \footnotesize{} ~~~FID$\downarrow$  & \footnotesize{} LPIPS$\uparrow$ 
            \tabularnewline
            
			\hline 	

			{\footnotesize{} BigGAN} &  - & \footnotesize{~~~213.3} & \footnotesize{~0.29}  & \footnotesize{~~~226.8} & \footnotesize{~0.15}  
            \tabularnewline

			{\footnotesize{} BigGAN} & \footnotesize{} SS (ours) & \footnotesize{~~~225.6} & \footnotesize{~0.47}  & \footnotesize{~~208.3} & \footnotesize{~0.47}  
            \tabularnewline            

            \footnotesize{} $\mathcal{L}_{all}$ (ours) & - & \footnotesize{~~~123.9} & \footnotesize{~0.28}  & \footnotesize{~~~129.4} & \footnotesize{0.27}  
            \tabularnewline

            \footnotesize{} $\mathcal{L}_{all}$ (ours) & \footnotesize{} SS (ours) & \textbf{\footnotesize{~~~106.4}} & \textbf{\footnotesize{~0.55}}   & \textbf{\footnotesize{~~~89.6}} & \textbf{\footnotesize{0.56}} 
\end{tabular}
\vspace{0.5ex}
\caption{Ablation on the performance when adapting the class-conditional BigGAN model \cite{Brock2019} pre-trained on ImageNet.}
\label{table:biggan} %
\vspace{-2ex}
\end{table}

%% file: tex/5_conclusion.tex
\section{Conclusion}
\label{sec:conclusion}

In this work, we presented a new method for few-shot adaptation of GAN models.
It transfers the smooth latent space of a pre-trained GAN, which was trained on a large dataset, to a new domain with very few images. We addressed the case of few-shot GAN adaptation when the source and target domains are structurally dissimilar, which is a common issue in applications. Our extensive results demonstrate that in this setting our approach outperforms previous works in terms of image quality and diversity. 




\section*{Acknowledgement}

The work has been supported by the Deutsche Forschungsgemeinschaft (DFG, German Research Foundation) under Germany’s Excellence Strategy – EXC 2070 -390732324 and the ERC Consolidator Grant FORHUE (101044724). We thank Jinhui Yi for providing and analyzing the sugar beet data used in supplementary experiments.

%% file: tex/7_supplementary.tex
\section{Additional quantitative analysis}
\label{sec:supp:quantitative}

Fig.~\ref{fig:plots_metrics1} shows the progression of the image quality and diversity metrics, FID and LPIPS, for different methods during few-shot adaptation. For these visualizations, we pick a pair of structurally dissimilar source-target domain pairs (\textit{Horses$\rightarrow$Pokemons}). The curves in the figure correspond to the results in Table \ref{table:comparison_unrelated} in the main paper. 

\input{figures/plots_metrics1}

Our observation from Fig.~\ref{fig:plots_metrics1} is that in the challenging adaptation scenario (\textit{Horses$\rightarrow$Pokemons}) prior methods achieve the best performance in FID (left plot) very early during the training, and are then unable to improve the image quality at later stages. For example, the methods without any diversity preserving regularization (TGAN, FreezeD, AdAM) suffer from training instabilities, indicated by an early collapse in the FID curves. On the other hand, the FID of the models that regularize diversity degradation (CDC, RSSA) remains stable without improvements. We hypothesize that these methods can successfully adapt the colors of objects from the source domain to the style of the target domain quickly, but they are not able to learn more high-level properties like shape of objects at later stages. This is confirmed by the visual results in Fig.~\ref{fig:qualitative_comparison_distant3}, where the images generated by these methods strongly follow the structure of the source domain. In contrast, our method allows to improve FID throughout the whole training and thus achieves higher image quality (yellow curve in the left plot). Next, the diversity evaluation (right plot) demonstrates that the LPIPS of TGAN, FreezeD, and AdAM collapses to low values very quickly, indicating training instabilities. Similarly, CDC and RSSA also suffer from diversity degradation, but it is slowed down with the help of the diversity regularizations used in these methods. Finally, our method allows to maintain high diversity scores throughout the whole training process.


\section{The effect of the FID evaluation protocol.}
\label{sec:supp:fid}

Our FID evaluation protocol differs from prior works in two ways. Firstly, as in the regime of dissimilar source-target domains the best performance can be achieved at later training epochs (e.g., see Fig.~\ref{fig:plots_metrics1}), we extend the duration of the training procedure from 5k to 30k epochs. We additionally evaluate all methods at epochs 500 and 750 since we found this beneficial to achieve superior FID scores for some close source-target domain pairs.

\input{tables/suppl_fid}

\input{figures/ablation_suppl}

Secondly, we resort to evaluating FID between the sets of generated and real images of the same size, which is a standard practice in the community. In contrast, prior work \cite{ojha2021few} proposed to compute FID between 5000 generated images and the whole validation set, which often contains a significantly smaller amount of images. We note that computing FID between sets of different sizes is generally not advisable due to a mismatch in estimation of variances of the first two moments between real and generated distributions \cite{binkowskidemystifying}. The difference in evaluation results between our (FID$_{\text{val}}$) and prior protocols (FID$_{\text{\cite{ojha2021few}}}$) is demonstrated in Table \ref{table:fid_effect}, where the Sketch and Sunglasses domains have 290 and 2683 images, respectively. Due to a larger generated set, FID$_{\text{\cite{ojha2021few}}}$ tends to output consistently lower scores than our reported numbers, but it does not change the ranking of the models.

\section{Additional qualitative results}
\label{sec:supp:qualitative}

We provide additional visual results with StyleGANv2 for the dissimilar source-target domains \textit{Face$\rightarrow$Cats} \cite{yunqingfew} and \textit{Horses$\rightarrow$Pokemons} in Fig.~\ref{fig:qualitative_comparison_distant3}. In both cases, our method generates diverse images that inherit the variation of the source images and flexibly combine features of different target images. In contrast, in most cases for the prior methods we observe inferior performance due to either memorization issues, training instabilities, or inability to learn the shape of objects in the target domain. We note that while the generation results of our method in the \textit{Pokemons} domain exhibit the most realistic shapes and the largest variation in colors, it is still challenging to generate fully realistic new pokemons in the 10-shot regime. Further improvement of few-shot synthesis for such challenging datasets is an interesting direction for future work.

Fig.~\ref{fig:qualitative_comparison_close2} shows results for the more similar domain pairs \textit{Face$\rightarrow$Babies} and \textit{Churches$\rightarrow$Haunted Houses}. We find that the results are consistent with Fig.~\ref{fig:qualitative_comparison_close1}: our method successfully adapts images of churches to a new style or converts adult faces into babies, performing on par with previous state-of-the-art approaches.

\section{Ablation on the parameters of the smoothness similarity regularization}
\label{sec:supp:ablation_ss}

Our smoothness similarity regularization has two parameters: the regularization strength $\lambda_{SS}$ and the resolution of features $G^l$. All the experiments in the main paper were conducted with $\lambda = 5.0$ and $G^l$ at resolution (32$\times$32). In Fig.~\ref{fig:ablation_suppl} and Table~\ref{table:ablation_SS_suppl} we provide an ablation on both these parameters. Firstly, we observe the effect of $\lambda_{SS}$ (rows 3-6 in Fig.~\ref{fig:ablation_suppl} and rows 2-5 in Table~\ref{table:ablation_SS_suppl}). As seen from the ablation study, compared to the model without any regularization, our smoothness similarity regularization helps to overcome memorization and achieve diverse synthesis. The effect of $\mathcal{L}_{SS}$ is, as expected, higher when $\lambda_{SS}$ is increased, which is indicated by increasing LPIPS scores. Yet, we find that setting a high $\lambda_{SS}$ starts to compromise the image quality, as the loss starts to overtake the adversarial loss supervision. We found that $\lambda = 5.0$ consistently achieves a good trade-off between image quality and diversity across many source-target domains. 

\input{figures/1-5-shot}

\input{tables/ablation_SS_lambda_l.tex}

Furthermore, we observe the effect of using features at different resolutions, corresponding to different generator blocks (rows 7-10 in Fig.~\ref{fig:ablation_suppl} and rows 6-9 in Table~\ref{table:ablation_SS_suppl}). We find that using later generator blocks at higher resolution increases the impact of the regularization. However, we also observe that using a very high resolution leads to the transfer of image transitions from the source domain at more fine-grained level, which can compromise image quality, for example transferring minor details that do not look realistic in the target domain. Based on the results in Table~\ref{table:ablation_SS_suppl}), we concluded that the resolution (32$\times$32) provides a good quality-diversity trade-off as it transfers high-level, more interpretable image variations without compromising the high-level coherency of objects in the target domain.

\section{Additional analysis on $\mathcal{L}_{all}$}
\label{sec:supp:ablation_Lall}

The second component of our model is a new way to compute the $D$'s loss. As discussed in Sec.~\ref{sec:experiments:stylegan2}, allowing the discriminator to compute the loss at different layers is strongly beneficial for improving the quality of synthesized images. Interestingly, even though the formulation of $\mathcal{L}_{all}$ in Eq.~\ref{eq:d_loss} has equal weights for all layers, the final contributions can still be different because activations $s^i \circ D^i(x)$  can have different magnitudes for different layers. In effect, this leads to an automatic discovery of the correct loss contribution of each layer depending on the source-target domains, as shown in Fig.~\ref{fig:plots_losses}.

The fact that optimal contributions of different layers in Fig.~\ref{fig:plots_losses} are different suggests using alternative weighting schemes rather than using equal weights for all layers. For comparison, we consider two alternative strategies: assigning higher weights on earlier or later $D$ layers. For this, instead of the uniform weights [1.0, 1.0, 1.0, 1.0, 1.0, 1.0, 1.0]/7 in Eq.~\ref{eq:d_loss}, we use either the weighting [1.6, 1.4, 1.2, 1.0, 0.8, 0.6, 0.4]/7 or [0.4, 0.6, 0.8, 1.0, 1.2, 1.4, 1.6]/7, referred to as ``Earlier'' or ``Later'' in Table ~\ref{table:suppl_weigths_Lall}.

\input{tables/suppl_weights_Lall}

\input{figures/crops}

We note that there exists a trade-off. On one hand, while using higher weights for earlier layers is beneficial for the closer domains \textit{Face$\rightarrow$Anime} (improved FID and LPIPS), it also leads to degraded image quality for the more distant domains \textit{Church$\rightarrow$Shells} (higher FID). On the other hand, the ``later'' strategy universally improves the image quality, but leads to memorization of training images and thus lower LPIPS scores. For this reason, we select the uniform weighting as it is the simplest solution which already allows D to adjust the contributions of different layers, while providing a reasonable balance between image quality and diversity for diverse source-target domain pairs.

\section{1-shot and 5-shot adaptation performance}
\label{sec:supp:1-5-shot}

In the main paper, we mainly focus on the 10-shot target datasets. Following prior work, we extend our analysis to 5-shot and 1-shot setups. Consistent with Sec.~\ref{sec:experiments}, our main focus is on the challenging case of structurally dissimilar source and target domains. We thus construct 1-shot and 5-shot scenarios of the adaptation between \textit{Face}$\rightarrow$\textit{Cats} and \textit{Horses}$\rightarrow$\textit{Pokemons} (10-shot results for these datasets are shown in Fig.~\ref{fig:qualitative_comparison_distant3}). We compare our method to CDC \cite{ojha2021few}, which is a popular baseline from the literature. Our observations from Fig.~\ref{fig:1-5-shot} are consistent with the main paper: while the prior method CDC cannot learn the shapes of objects in the new domain, our method achieves more realistic synthesis, successfully transferring meaningful high-level image variations even from structurally dissimilar datasets.


\section{Application: detection of nutrient deficiencies of
crops}
\label{sec:crops}

We investigate the application of our model to the task of visual detection of nutrient deficiencies in crop science  \cite{yi2020deep}. In agriculture, this task is important to enable timely actions to prevent major losses of crops caused by lack of nutrients, such as nitrogen. From the data collection perspective, this task refers to restricted image domains, since it typically requires manual photographing of growing crops and expert knowledge for obtaining correct annotations. Therefore, we explore whether our model can be trained on a limited set of images depicting sugar beats (see Fig.~\ref{fig:crops}).

For our experiments, we pick two random 10-shot subsets of the DND-SB dataset \cite{yi2020deep}, consisting of images with healthy sugar beats and crops suffering from nitrogen nutrient deficiencies (see Fig.~\ref{fig:crops}). We use the StyleGANv2 checkpoint pre-trained on FFHQ \cite{Karras2018ASG}. Despite using such a dissimilar source domain, we observe that our model still achieves photorealistic synthesis of new crops, for example changing the shape or locations of leaves of the training examples.

To verify that our generated images preserve the characteristics of interest of the training images, we take a classification network, which was pre-trained to perform the \textit{healthy-deficient} binary classification on images of the same resolution (256$\times$256). We observe that generated images from the healthy subset were identified correctly in 98.9\% cases, while nitrogen deficiencies were detected correctly for 95.6\% of the images generated from the second subset. We consider this experiment as a promising example which suggests future utilization of our model for data augmentation in restricted image domains.




\input{tables/biggan_suppl_ablation}

\section{Additional details in the class-conditional GAN setting}
\label{sec:supp:biggan}

For our experiments in Sec.~\ref{sec:experiments:biggan}, we pre-train the class-conditional BigGAN model \cite{Brock2019} (without BigGAN-deep extensions) on ImageNet at the image resolution of (256$\times$256). The model achieves FID of 9.23 on the ImageNet validation set. We then fine-tune both the pre-trained generator and discriminator on the provided few-shot dataset using our proposed loss terms as presented in Sec.~\ref{Sec:method}. We use batch size of 32, decay of 0.999 for the generator's exponential moving averages, and learning rates of 2e-4 and 8e-4 for the generator and discriminator, respectively, while preserving all the other hyperparameters that were used for pre-training.

The generator of BigGAN takes two inputs, a noise vector and a class label. The input label is then projected into a continuous embedding space via a learnable linear mapping. To enable the adaptation of the generator to unconditional few-shot datasets, we do not inject class labels in our approach but directly operate with the pre-learnt continuous class embedding. At each fine-tuning epoch, we therefore sample a Gaussian vector in a joint noise-class space.

The discriminator of BigGAN takes a class label only at the final layer, where it is processed via a linear projection layer \cite{miyato2018cgans} and added to the output features of the last discriminator's block. In our experiments, we remove this conditioning mechanism and simply pass unmodified features after the last block to the final layer to compute the adversarial loss. This way, our whole model can be trained on the provided dataset in an unconditional fashion.

We apply our smoothness similarity regularization using the generator's features at resolution (32$\times$32). We explored two different ways for the implementation of the regularization, considering smoothness with respect to only the noise space or the joint noise-class space. We found that using the class embeddings for the regularization is important, as achieving a high synthesis diversity without it is difficult (see Table~\ref{table:biggan_supl_ablation}). We hypothesize that this happens because a large part of transferable image variations in the source domains is contained not only in the interpolations between different noise vectors, but also in the interpolations between different classes.

\input{figures/qual_comparison_distant3.tex}

\input{figures/qual_comparison_close2.tex}

%% file: figures/plots_metrics1.tex
\begin{figure*}[t]
\begin{centering}
\setlength{\tabcolsep}{0.53in}
\renewcommand{\arraystretch}{1}
\par\end{centering}
\begin{centering}
\begin{tabular}{@{\hskip -0.05in}c@{\hskip 0.00in}c@{\hskip 0.00in}c@{\hskip 0.00in}c@{\hskip 0.00in}}

\includegraphics[width=0.47\linewidth, height=0.35\linewidth]{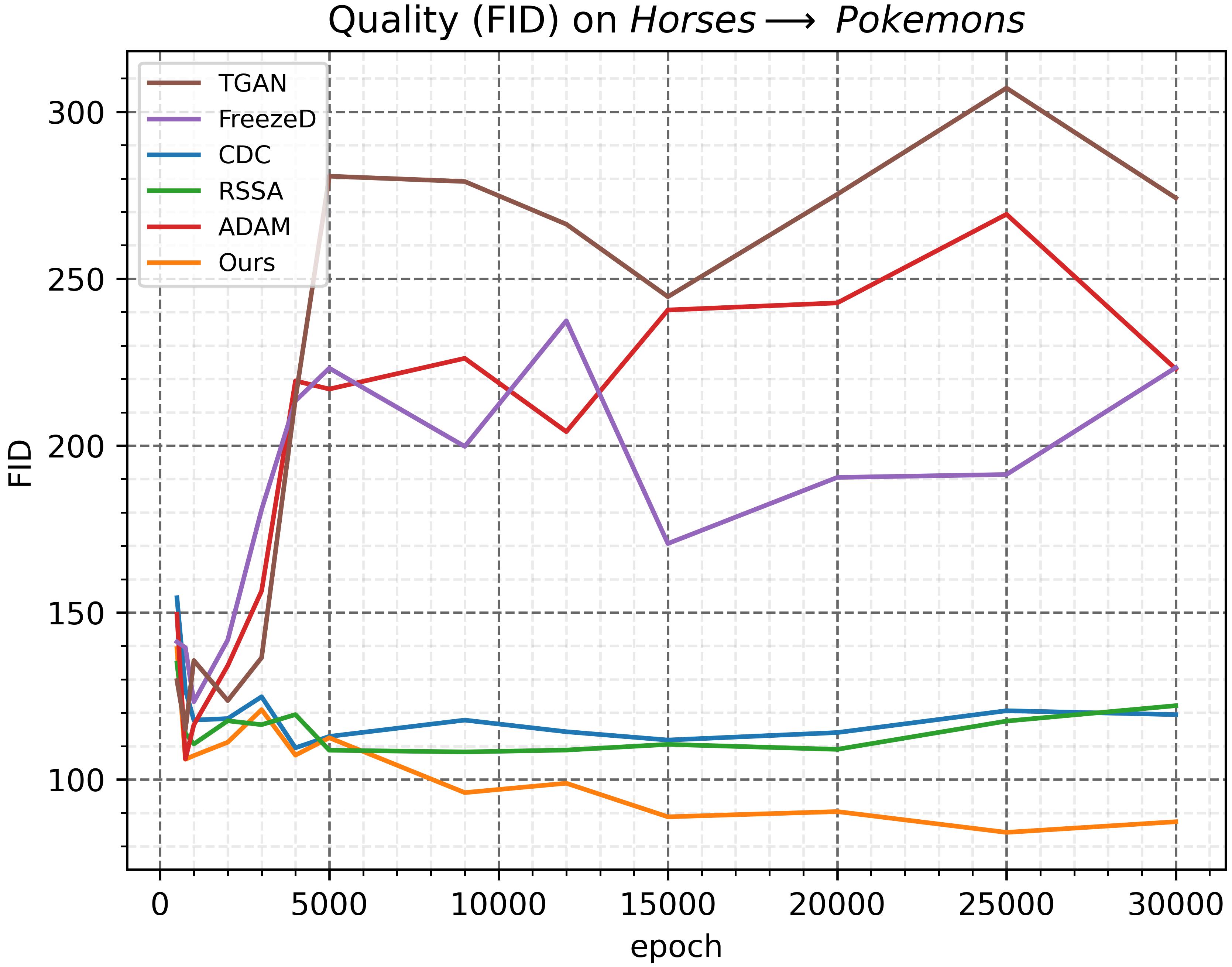} ~~~ 
\includegraphics[width=0.47\linewidth, height=0.35\linewidth]{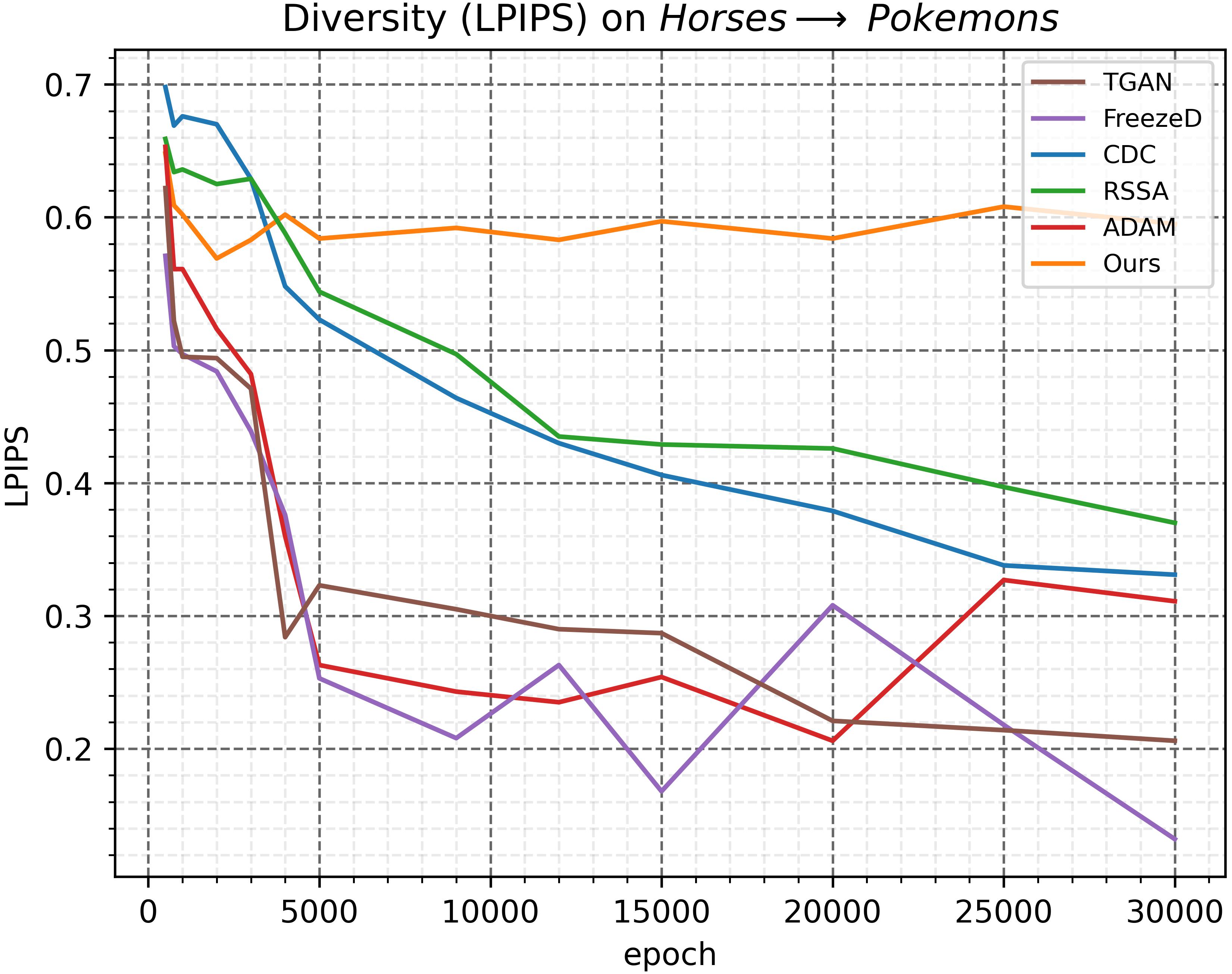} 

\end{tabular}

\par\end{centering}
\caption{The FID and LPIPS curves of different methods for the few-shot adaptation between dissimilar domains \textit{Horses$\rightarrow$Pokemons.}
}
\label{fig:plots_metrics1}
\end{figure*}

%% file: tables/suppl_fid.tex
\begin{table}[t]

	\setlength{\tabcolsep}{0.55em}
	\renewcommand{\arraystretch}{0.95}
	\centering
		\begin{tabular}{@{\hskip -0.04in}l@{\hskip 0.02in}|c@{\hskip 0.01in}c@{\hskip 0.02in}|c@{\hskip -0.05in}c@{\hskip -0.06in}}
			
   \multirow{2}{*}{\footnotesize{} Method} & \multicolumn{2}{c|}{\footnotesize{} \hspace{-0.7ex} \textbf{Face$\rightarrow$Sketch}} & \multicolumn{2}{c}{\footnotesize{} \hspace{-1.5ex} \textbf{Face$\rightarrow$Sunglasses}} 
            \tabularnewline 
            & \footnotesize{} FID$_{\text{val}}$$\downarrow$  & \footnotesize{} FID$_\text{\cite{ojha2021few}}$$\downarrow$ & \footnotesize{} FID$_{\text{val}}$$\downarrow$  & \footnotesize{} FID$_\text{\cite{ojha2021few}}$$\downarrow$
            \tabularnewline
            
			\hline

			{\footnotesize{} \textbf{TGAN} \cite{Wang2018TransferringGG} } & \footnotesize{54.2} & \footnotesize{47.3}  & \footnotesize{36.8} & \footnotesize{36.2}   \tabularnewline

			{\footnotesize{} \textbf{FreezeD} \cite{Mo2020FreezeDA} } & \footnotesize{48.8} & \footnotesize{40.8}  & \footnotesize{32.0} & \footnotesize{31.9}    \tabularnewline

			{\footnotesize{} \textbf{CDC} \cite{ojha2021few} } & \footnotesize{54.2} & \footnotesize{46.8}  & \footnotesize{30.5} & \footnotesize{31.2}   \tabularnewline

			{\footnotesize{} \textbf{RSSA} \cite{xiao2022few} } & \footnotesize{61.4} & \footnotesize{51.8}  & \footnotesize{36.3} & \footnotesize{35.9}     \tabularnewline

			{\footnotesize{} \textbf{AdAM} \cite{yunqingfew} } & \footnotesize{56.3} & \footnotesize{47.8}   & \footnotesize{31.1} & \footnotesize{29.7}   \tabularnewline

			{\footnotesize{} \textbf{Ours}} & \textbf{\footnotesize{45.2}} & \textbf{\footnotesize{39.9}}   & \textbf{\footnotesize{27.5}} & \textbf{\footnotesize{27.0}}   \tabularnewline

\end{tabular}
\vspace{0.5ex}
\caption{Effect of a different FID evaluation protocol. The differences between the protocols are described in Sec. \ref{sec:supp:fid}.}
\label{table:fid_effect} %
\end{table}

%% file: figures/ablation_suppl.tex
\begin{figure}[t]
\begin{centering}
\setlength{\tabcolsep}{0.01in}
\renewcommand{\arraystretch}{1}
\par\end{centering}
\begin{centering}

\vspace{-0.5ex}
\begin{tabular}{@{\hskip -0.08in}c@{\hskip 0.05in}c@{\hskip 0.01in}c@{\hskip 0.01in}c@{\hskip 0.01in}c@{\hskip 0.01in}c@{\hskip 0.01in}c@{\hskip 0.01in}c@{\hskip -0.01in}c}

\multicolumn{8}{c}{\small $\leftarrow$ Interpolation in the latent space $\rightarrow$}
\tabularnewline[0.3pt]

~~\rotatebox{90}{{\hspace{-0.2em} \small (Source) \hspace{-3.0em} }} &
\frame{\includegraphics[draft=\draft,width=0.125\linewidth, height=0.125\linewidth]{figures/ablation_interp/church-shells/source/z1_sample74.jpg}} & 
\frame{\includegraphics[draft=\draft,width=0.125\linewidth, height=0.125\linewidth]{figures/ablation_interp/church-shells/source/z1_sample77.jpg}} &
\frame{\includegraphics[draft=\draft,width=0.125\linewidth, height=0.125\linewidth]{figures/ablation_interp/church-shells/source/z1_sample97.jpg}} &
\frame{\includegraphics[draft=\draft,width=0.125\linewidth, height=0.125\linewidth]{figures/ablation_interp/church-shells/source/z1_sample102.jpg}} &
\frame{\includegraphics[draft=\draft,width=0.125\linewidth, height=0.125\linewidth]{figures/ablation_interp/church-shells/source/z1_sample109.jpg}} &
\frame{\includegraphics[draft=\draft,width=0.125\linewidth, height=0.125\linewidth]{figures/ablation_interp/church-shells/source/z1_sample131.jpg}} &
\frame{\includegraphics[draft=\draft,width=0.125\linewidth, height=0.125\linewidth]{figures/ablation_interp/church-shells/source/z1_sample140.jpg}} &
~~\rotatebox{270}{{\hspace{-0.0em} \footnotesize \hspace{0.0em} }}
\tabularnewline[-1pt]

~~\rotatebox{90}{{\hspace{1.00em} \small -- \hspace{-3.0em} }} &
\frame{\includegraphics[draft=\draft,width=0.125\linewidth, height=0.125\linewidth]{figures/ablation_interp/church-shells/baseline/z2_sample74.jpg}} & 
\frame{\includegraphics[draft=\draft,width=0.125\linewidth, height=0.125\linewidth]{figures/ablation_interp/church-shells/baseline/z2_sample77.jpg}} &
\frame{\includegraphics[draft=\draft,width=0.125\linewidth, height=0.125\linewidth]{figures/ablation_interp/church-shells/baseline/z2_sample97.jpg}} &
\frame{\includegraphics[draft=\draft,width=0.125\linewidth, height=0.125\linewidth]{figures/ablation_interp/church-shells/baseline/z2_sample102.jpg}} &
\frame{\includegraphics[draft=\draft,width=0.125\linewidth, height=0.125\linewidth]{figures/ablation_interp/church-shells/baseline/z2_sample109.jpg}} &
\frame{\includegraphics[draft=\draft,width=0.125\linewidth, height=0.125\linewidth]{figures/ablation_interp/church-shells/baseline/z2_sample131.jpg}} &
\frame{\includegraphics[draft=\draft,width=0.125\linewidth, height=0.125\linewidth]{figures/ablation_interp/church-shells/baseline/z2_sample140.jpg}} &
~~\rotatebox{270}{{\hspace{-1.9em} \small -- \hspace{0.0em} }}
\tabularnewline[-1pt]

~~\rotatebox{90}{{\hspace{1.00em} \small 0.2 \hspace{-3.0em} }} &
\frame{\includegraphics[draft=\draft,width=0.125\linewidth, height=0.125\linewidth]{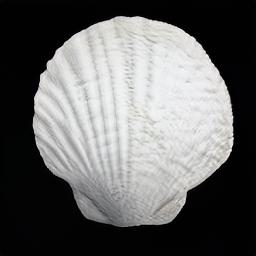}} & 
\frame{\includegraphics[draft=\draft,width=0.125\linewidth, height=0.125\linewidth]{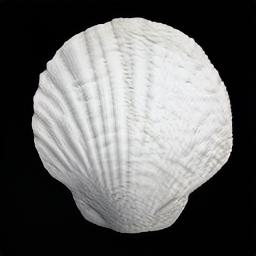}} & 
\frame{\includegraphics[draft=\draft,width=0.125\linewidth, height=0.125\linewidth]{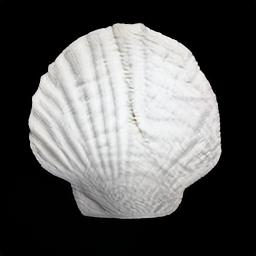}} & 
\frame{\includegraphics[draft=\draft,width=0.125\linewidth, height=0.125\linewidth]{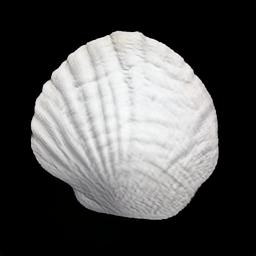}} & 
\frame{\includegraphics[draft=\draft,width=0.125\linewidth, height=0.125\linewidth]{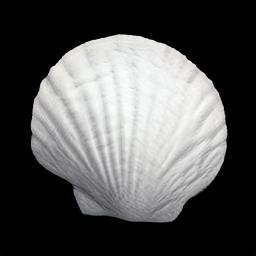}} & 
\frame{\includegraphics[draft=\draft,width=0.125\linewidth, height=0.125\linewidth]{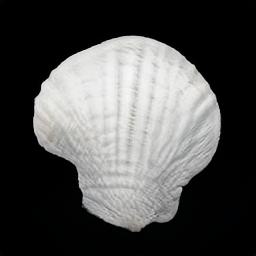}} & 
\frame{\includegraphics[draft=\draft,width=0.125\linewidth, height=0.125\linewidth]{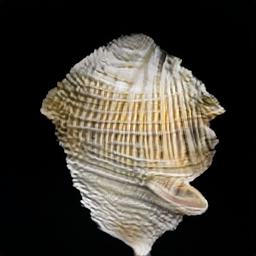}} & 
~~\rotatebox{270}{{\hspace{-2.9em} \footnotesize 32$\times$32 \hspace{-3.0em} }}
\tabularnewline[-3pt]

~~\rotatebox{90}{{\hspace{0.60em} \small 1.0 \hspace{-3.0em} }} &
\frame{\includegraphics[draft=\draft,width=0.125\linewidth, height=0.125\linewidth]{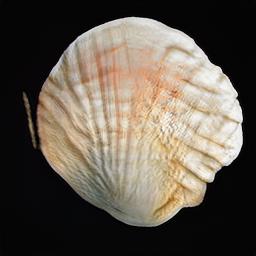}} & 
\frame{\includegraphics[draft=\draft,width=0.125\linewidth, height=0.125\linewidth]{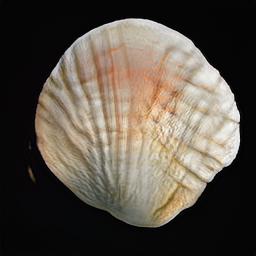}} & 
\frame{\includegraphics[draft=\draft,width=0.125\linewidth, height=0.125\linewidth]{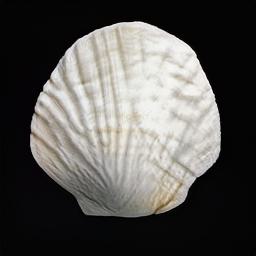}} & 
\frame{\includegraphics[draft=\draft,width=0.125\linewidth, height=0.125\linewidth]{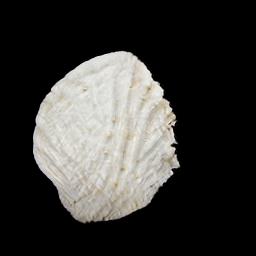}} & 
\frame{\includegraphics[draft=\draft,width=0.125\linewidth, height=0.125\linewidth]{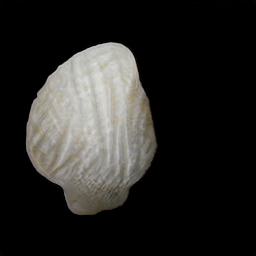}} & 
\frame{\includegraphics[draft=\draft,width=0.125\linewidth, height=0.125\linewidth]{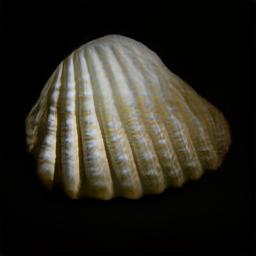}} & 
\frame{\includegraphics[draft=\draft,width=0.125\linewidth, height=0.125\linewidth]{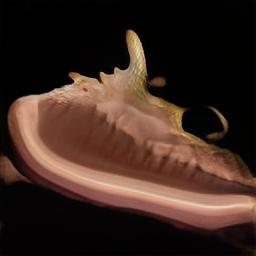}} & 
~~\rotatebox{270}{{\hspace{-2.9em} \footnotesize 32$\times$32 \hspace{-3.0em} }}
\tabularnewline[-3pt]

~~\rotatebox{90}{{\hspace{0.60em} \small 25.0 \hspace{-3.0em} }} &
\frame{\includegraphics[draft=\draft,width=0.125\linewidth, height=0.125\linewidth]{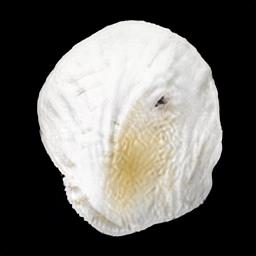}} & 
\frame{\includegraphics[draft=\draft,width=0.125\linewidth, height=0.125\linewidth]{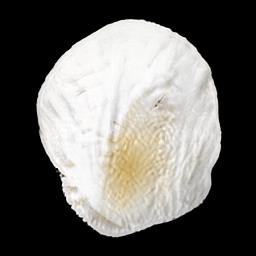}} & 
\frame{\includegraphics[draft=\draft,width=0.125\linewidth, height=0.125\linewidth]{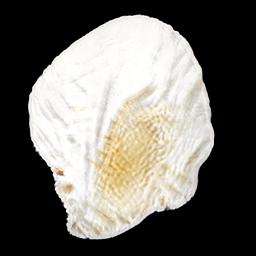}} & 
\frame{\includegraphics[draft=\draft,width=0.125\linewidth, height=0.125\linewidth]{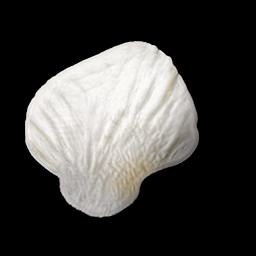}} & 
\frame{\includegraphics[draft=\draft,width=0.125\linewidth, height=0.125\linewidth]{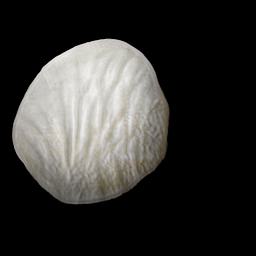}} & 
\frame{\includegraphics[draft=\draft,width=0.125\linewidth, height=0.125\linewidth]{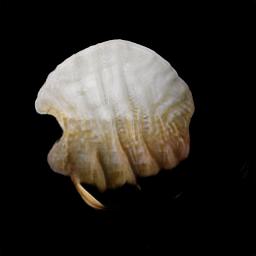}} & 
\frame{\includegraphics[draft=\draft,width=0.125\linewidth, height=0.125\linewidth]{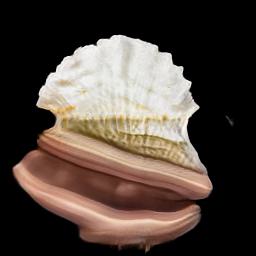}} & 
~~\rotatebox{270}{{\hspace{-2.9em} \footnotesize 32$\times$32 \hspace{-3.0em} }}
\tabularnewline[-3pt]

~~\rotatebox{90}{{\hspace{0.35em} \small 125.0 \hspace{-4.0em} }} &
\frame{\includegraphics[draft=\draft,width=0.125\linewidth, height=0.125\linewidth]{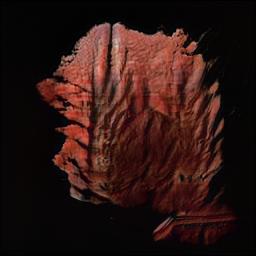}} & 
\frame{\includegraphics[draft=\draft,width=0.125\linewidth, height=0.125\linewidth]{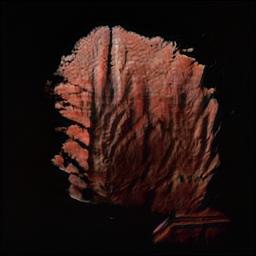}} & 
\frame{\includegraphics[draft=\draft,width=0.125\linewidth, height=0.125\linewidth]{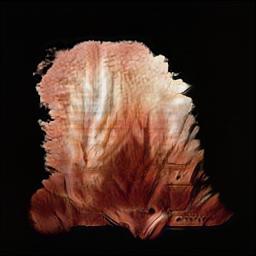}} & 
\frame{\includegraphics[draft=\draft,width=0.125\linewidth, height=0.125\linewidth]{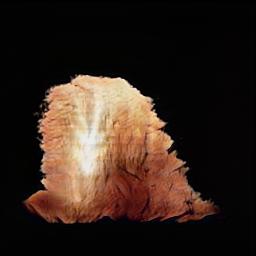}} & 
\frame{\includegraphics[draft=\draft,width=0.125\linewidth, height=0.125\linewidth]{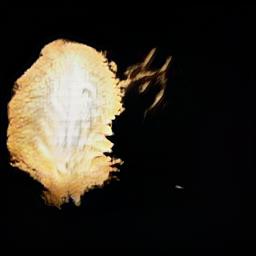}} & 
\frame{\includegraphics[draft=\draft,width=0.125\linewidth, height=0.125\linewidth]{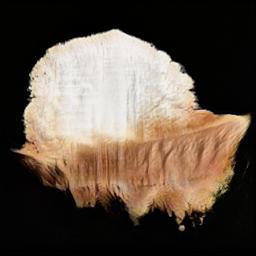}} & 
\frame{\includegraphics[draft=\draft,width=0.125\linewidth, height=0.125\linewidth]{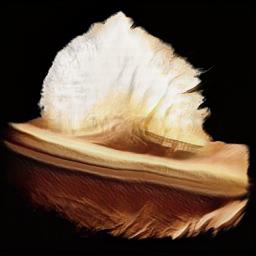}} & 
~~\rotatebox{270}{{\hspace{-2.9em} \footnotesize 32$\times$32 \hspace{-3.0em} }}
\tabularnewline[-1pt]

~~\rotatebox{90}{{\hspace{0.60em} \small 5.0 \hspace{-3.0em} }} &
\frame{\includegraphics[draft=\draft,width=0.125\linewidth, height=0.125\linewidth]{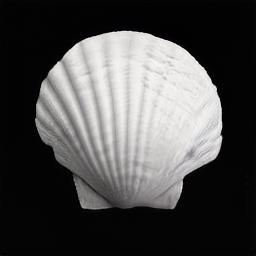}} & 
\frame{\includegraphics[draft=\draft,width=0.125\linewidth, height=0.125\linewidth]{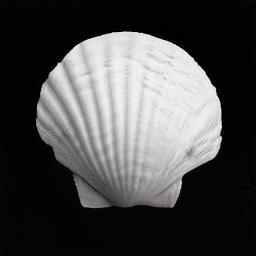}} & 
\frame{\includegraphics[draft=\draft,width=0.125\linewidth, height=0.125\linewidth]{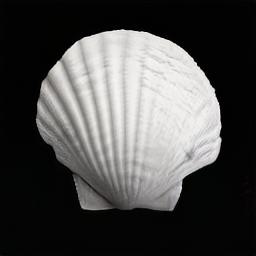}} & 
\frame{\includegraphics[draft=\draft,width=0.125\linewidth, height=0.125\linewidth]{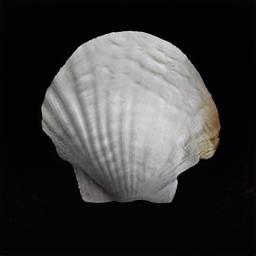}} & 
\frame{\includegraphics[draft=\draft,width=0.125\linewidth, height=0.125\linewidth]{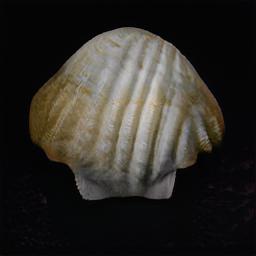}} & 
\frame{\includegraphics[draft=\draft,width=0.125\linewidth, height=0.125\linewidth]{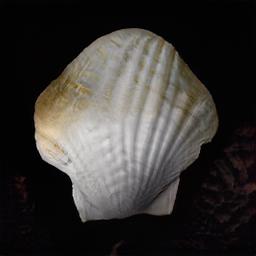}} & 
\frame{\includegraphics[draft=\draft,width=0.125\linewidth, height=0.125\linewidth]{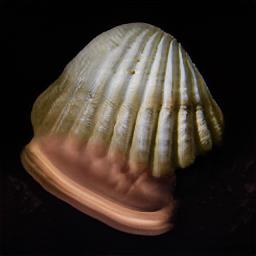}} & 
~~\rotatebox{270}{{\hspace{-2.9em} \footnotesize ~8$\times$8 \hspace{-3.0em} }}
\tabularnewline[-3pt]

~~\rotatebox{90}{{\hspace{0.60em} \small 5.0 \hspace{-3.0em} }} &
\frame{\includegraphics[draft=\draft,width=0.125\linewidth, height=0.125\linewidth]{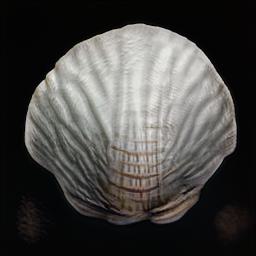}} & 
\frame{\includegraphics[draft=\draft,width=0.125\linewidth, height=0.125\linewidth]{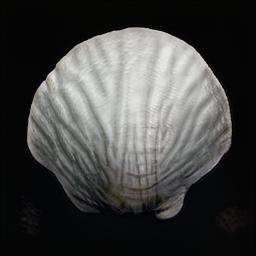}} & 
\frame{\includegraphics[draft=\draft,width=0.125\linewidth, height=0.125\linewidth]{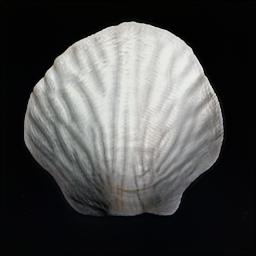}} & 
\frame{\includegraphics[draft=\draft,width=0.125\linewidth, height=0.125\linewidth]{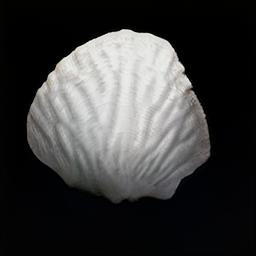}} & 
\frame{\includegraphics[draft=\draft,width=0.125\linewidth, height=0.125\linewidth]{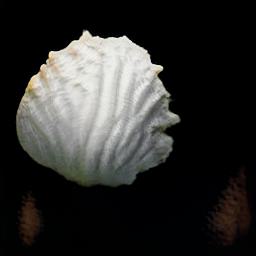}} & 
\frame{\includegraphics[draft=\draft,width=0.125\linewidth, height=0.125\linewidth]{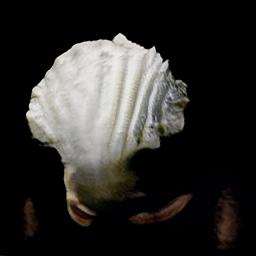}} & 
\frame{\includegraphics[draft=\draft,width=0.125\linewidth, height=0.125\linewidth]{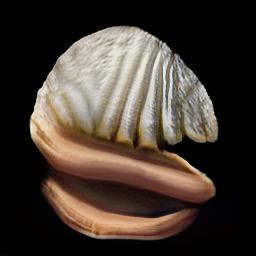}} & 
~~\rotatebox{270}{{\hspace{-2.9em} \footnotesize 16$\times$16 \hspace{-3.0em} }}
\tabularnewline[-3pt]

~~\rotatebox{90}{{\hspace{0.60em} \small 5.0 \hspace{-3.0em} }} &
\frame{\includegraphics[draft=\draft,width=0.125\linewidth, height=0.125\linewidth]{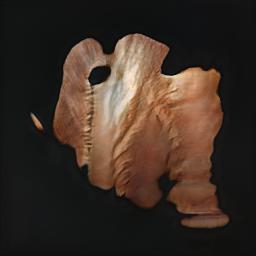}} & 
\frame{\includegraphics[draft=\draft,width=0.125\linewidth, height=0.125\linewidth]{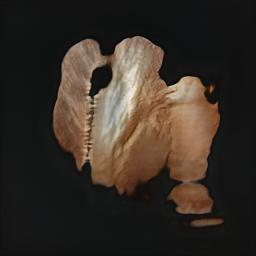}} & 
\frame{\includegraphics[draft=\draft,width=0.125\linewidth, height=0.125\linewidth]{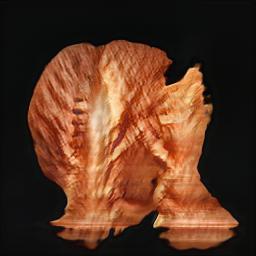}} & 
\frame{\includegraphics[draft=\draft,width=0.125\linewidth, height=0.125\linewidth]{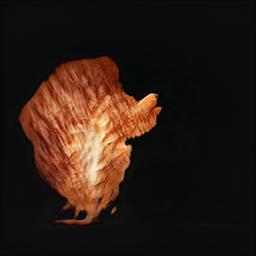}} & 
\frame{\includegraphics[draft=\draft,width=0.125\linewidth, height=0.125\linewidth]{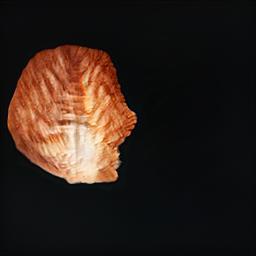}} & 
\frame{\includegraphics[draft=\draft,width=0.125\linewidth, height=0.125\linewidth]{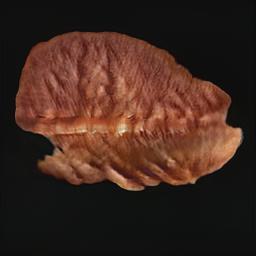}} & 
\frame{\includegraphics[draft=\draft,width=0.125\linewidth, height=0.125\linewidth]{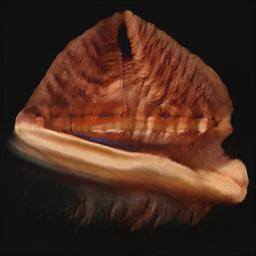}} & 
~~\rotatebox{270}{{\hspace{-2.9em} \footnotesize 64$\times$64 \hspace{-3.0em} }}
\tabularnewline[-3pt]

~~\rotatebox{90}{{\hspace{0.60em} \small 5.0 \hspace{-3.0em} }} &
\frame{\includegraphics[draft=\draft,width=0.125\linewidth, height=0.125\linewidth]{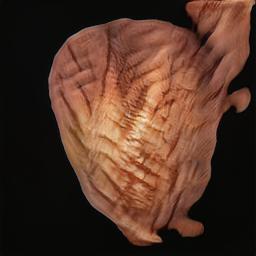}} & 
\frame{\includegraphics[draft=\draft,width=0.125\linewidth, height=0.125\linewidth]{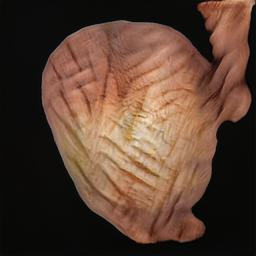}} & 
\frame{\includegraphics[draft=\draft,width=0.125\linewidth, height=0.125\linewidth]{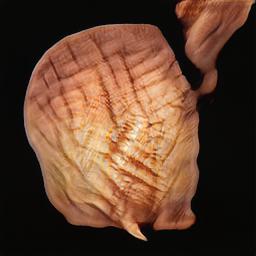}} & 
\frame{\includegraphics[draft=\draft,width=0.125\linewidth, height=0.125\linewidth]{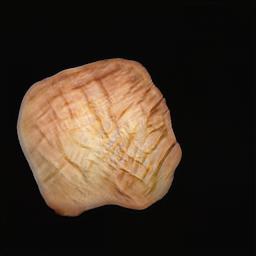}} & 
\frame{\includegraphics[draft=\draft,width=0.125\linewidth, height=0.125\linewidth]{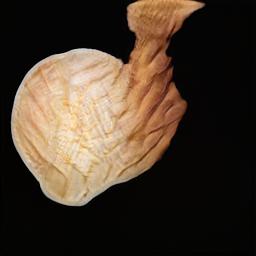}} & 
\frame{\includegraphics[draft=\draft,width=0.125\linewidth, height=0.125\linewidth]{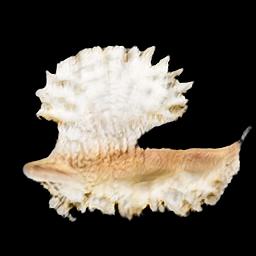}} & 
\frame{\includegraphics[draft=\draft,width=0.125\linewidth, height=0.125\linewidth]{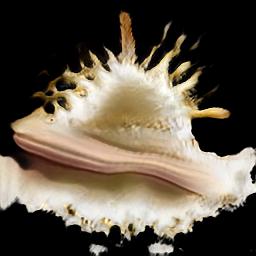}} & 
~~\rotatebox{270}{{\hspace{-3.4em} \footnotesize 128$\times$128 \hspace{-3.0em} }}
\tabularnewline[-1pt]

~~\rotatebox{90}{{\hspace{0.60em} \small 5.0 \hspace{-3.0em} }} &
\frame{\includegraphics[draft=\draft,width=0.125\linewidth, height=0.125\linewidth]{figures/ablation_interp/church-shells/ours/z2_sample74.jpg}} & 
\frame{\includegraphics[draft=\draft,width=0.125\linewidth, height=0.125\linewidth]{figures/ablation_interp/church-shells/ours/z2_sample77.jpg}} &
\frame{\includegraphics[draft=\draft,width=0.125\linewidth, height=0.125\linewidth]{figures/ablation_interp/church-shells/ours/z2_sample97.jpg}} &
\frame{\includegraphics[draft=\draft,width=0.125\linewidth, height=0.125\linewidth]{figures/ablation_interp/church-shells/ours/z2_sample102.jpg}} &
\frame{\includegraphics[draft=\draft,width=0.125\linewidth, height=0.125\linewidth]{figures/ablation_interp/church-shells/ours/z2_sample109.jpg}} &
\frame{\includegraphics[draft=\draft,width=0.125\linewidth, height=0.125\linewidth]{figures/ablation_interp/church-shells/ours/z2_sample131.jpg}} &
\frame{\includegraphics[draft=\draft,width=0.125\linewidth, height=0.125\linewidth]{figures/ablation_interp/church-shells/ours/z2_sample140.jpg}} &
~~\rotatebox{270}{{\hspace{-2.9em} \footnotesize 32$\times$32 \hspace{-3.0em} }}
\tabularnewline[-3pt]

\end{tabular}
\par\end{centering}
\vspace{4pt}
\caption{Latent space interpolations of the source generator and the ablation models from  Table \ref{table:ablation_SS_suppl}. Leftmost and rightmost columns show the used $\lambda_{SS}$ and the resolution of $G^l$.}
\label{fig:ablation_suppl}
\vspace{-1.5ex}
\end{figure}

%% file: figures/1-5-shot.tex
\begin{figure*}[t]
\begin{centering}
\setlength{\tabcolsep}{0.01in}
\renewcommand{\arraystretch}{1}
\par\end{centering}
\begin{centering}

\begin{tabular}{@{\hskip -0.13in}c@{\hskip 0.10in}c@{\hskip 0.01in}c@{\hskip 0.01in}c@{\hskip 0.01in}c@{\hskip 0.01in}c@{\hskip 0.01in}c@{\hskip 0.01in}c:c@{\hskip 0.01in}c@{\hskip 0.01in}c@{\hskip 0.01in}c@{\hskip 0.01in}c@{\hskip 0.01in}c@{\hskip 0.01in}c}

~~\rotatebox{90}{{\hspace{0.6em} Real \hspace{0.0em} }} &

\multicolumn{7}{@{\hskip -0.04in}c:}{
\frame{\includegraphics[width=0.07\linewidth, height=0.07\linewidth]{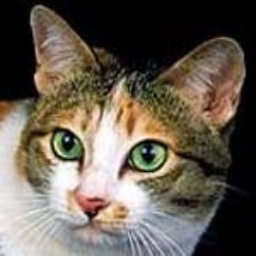}} 
} &
\multicolumn{7}{@{\hskip 0.04in}c}{
\frame{\includegraphics[width=0.07\linewidth, height=0.07\linewidth]{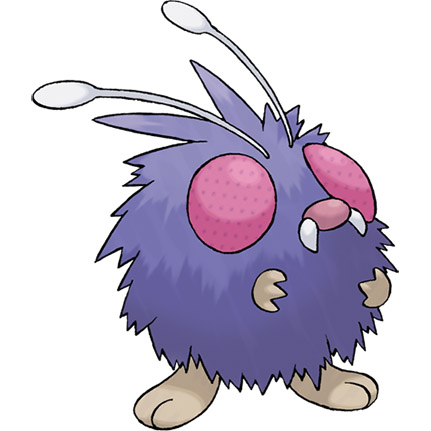}}
}

\tabularnewline[2pt]

~~\rotatebox{90}{{\hspace{0.0em} Source \hspace{-3.0em} }} &
\frame{\includegraphics[width=0.067\linewidth, height=0.067\linewidth]{figures/qualitative_results/ffhq-sketch/source/p_2_000000.jpg}} & 
\frame{\includegraphics[width=0.067\linewidth, height=0.067\linewidth]{figures/qualitative_results/ffhq-sketch/source/p_6_000000.jpg}} &
\frame{\includegraphics[width=0.067\linewidth, height=0.067\linewidth]{figures/qualitative_results/ffhq-sketch/source/p_7_000000.jpg}} &
\frame{\includegraphics[width=0.067\linewidth, height=0.067\linewidth]{figures/qualitative_results/ffhq-sketch/source/p_8_000000.jpg}} &
\frame{\includegraphics[width=0.067\linewidth, height=0.067\linewidth]{figures/qualitative_results/ffhq-sketch/source/p_17_000000.jpg}} &
\frame{\includegraphics[width=0.067\linewidth, height=0.067\linewidth]{figures/qualitative_results/ffhq-sketch/source/p_22_000000.jpg}} &
\frame{\includegraphics[width=0.067\linewidth, height=0.067\linewidth]{figures/qualitative_results/ffhq-sketch/source/p_23_000000.jpg}} &

\frame{\includegraphics[width=0.067\linewidth, height=0.067\linewidth]{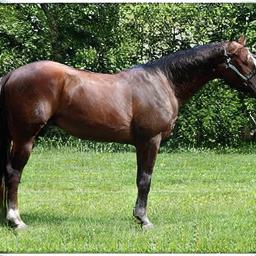}} & 
\frame{\includegraphics[width=0.067\linewidth, height=0.067\linewidth]{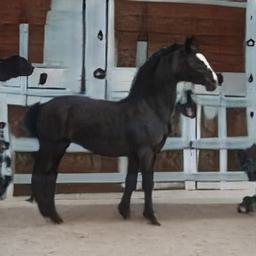}} & 
\frame{\includegraphics[width=0.067\linewidth, height=0.067\linewidth]{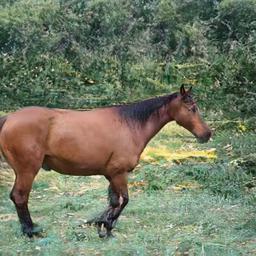}} & 
\frame{\includegraphics[width=0.067\linewidth, height=0.067\linewidth]{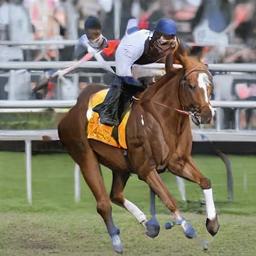}} & 
\frame{\includegraphics[width=0.067\linewidth, height=0.067\linewidth]{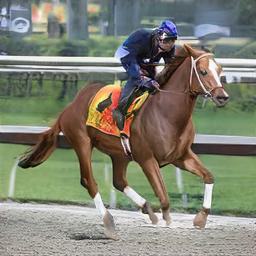}} & 
\frame{\includegraphics[width=0.067\linewidth, height=0.067\linewidth]{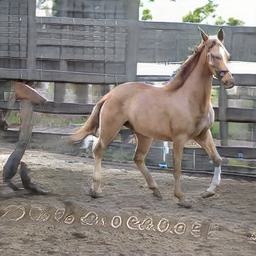}} & 
\frame{\includegraphics[width=0.067\linewidth, height=0.067\linewidth]{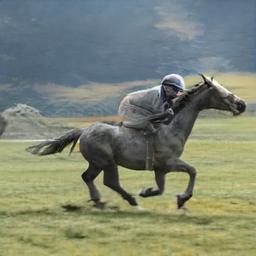}} 
\tabularnewline[-3pt]

~~\rotatebox{90}{{\hspace{0.4em} CDC \hspace{-3.0em} }} &
\frame{\includegraphics[width=0.067\linewidth, height=0.067\linewidth]{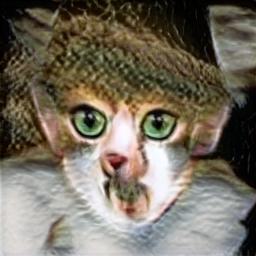}} & 
\frame{\includegraphics[width=0.067\linewidth, height=0.067\linewidth]{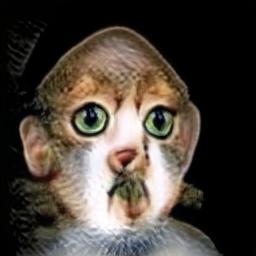}} &
\frame{\includegraphics[width=0.067\linewidth, height=0.067\linewidth]{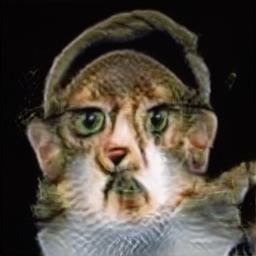}} &
\frame{\includegraphics[width=0.067\linewidth, height=0.067\linewidth]{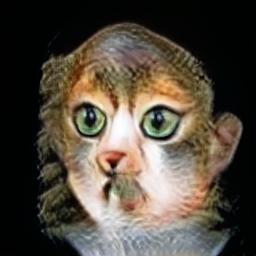}} &
\frame{\includegraphics[width=0.067\linewidth, height=0.067\linewidth]{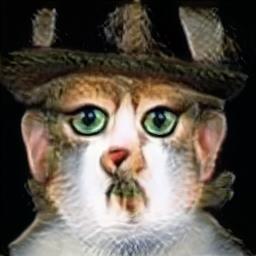}} &
\frame{\includegraphics[width=0.067\linewidth, height=0.067\linewidth]{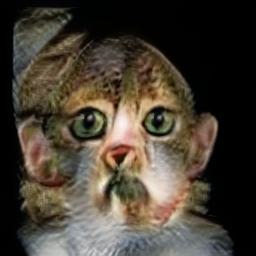}} &
\frame{\includegraphics[width=0.067\linewidth, height=0.067\linewidth]{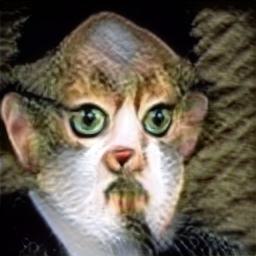}} &

\frame{\includegraphics[width=0.067\linewidth, height=0.067\linewidth]{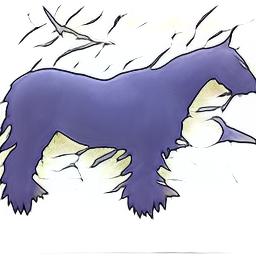}} & 
\frame{\includegraphics[width=0.067\linewidth, height=0.067\linewidth]{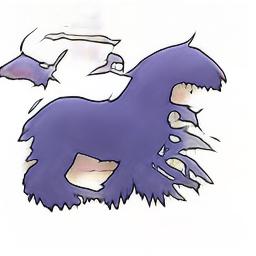}} & 
\frame{\includegraphics[width=0.067\linewidth, height=0.067\linewidth]{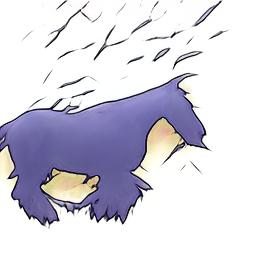}} & 
\frame{\includegraphics[width=0.067\linewidth, height=0.067\linewidth]{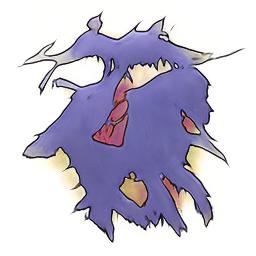}} & 
\frame{\includegraphics[width=0.067\linewidth, height=0.067\linewidth]{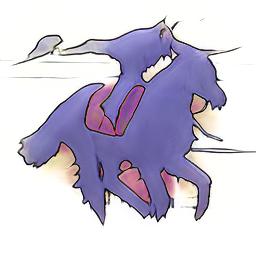}} & 
\frame{\includegraphics[width=0.067\linewidth, height=0.067\linewidth]{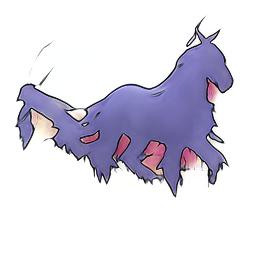}} & 
\frame{\includegraphics[width=0.067\linewidth, height=0.067\linewidth]{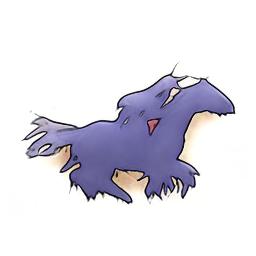}} 
\tabularnewline[-3pt]

~~\rotatebox{90}{{\hspace{0.4em} Ours \hspace{-3.0em} }} &
\frame{\includegraphics[width=0.067\linewidth, height=0.067\linewidth]{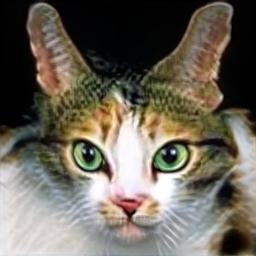}} & 
\frame{\includegraphics[width=0.067\linewidth, height=0.067\linewidth]{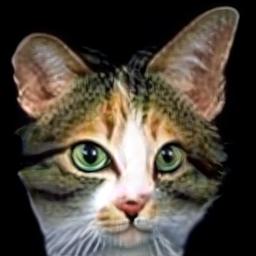}} &
\frame{\includegraphics[width=0.067\linewidth, height=0.067\linewidth]{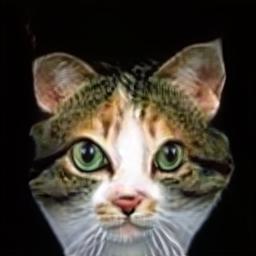}} &
\frame{\includegraphics[width=0.067\linewidth, height=0.067\linewidth]{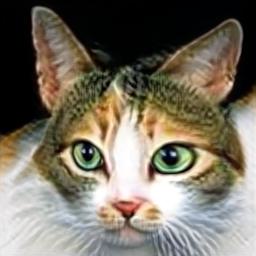}} &
\frame{\includegraphics[width=0.067\linewidth, height=0.067\linewidth]{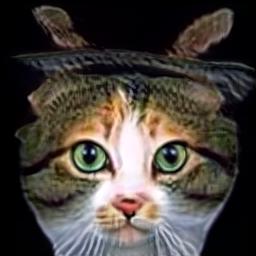}} &
\frame{\includegraphics[width=0.067\linewidth, height=0.067\linewidth]{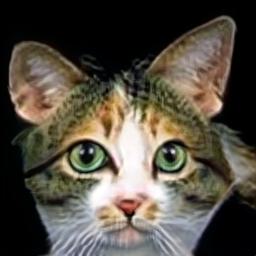}} &
\frame{\includegraphics[width=0.067\linewidth, height=0.067\linewidth]{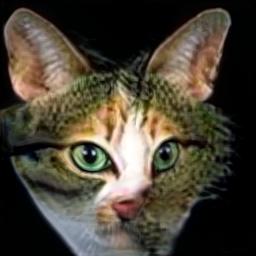}} &

\frame{\includegraphics[width=0.067\linewidth, height=0.067\linewidth]{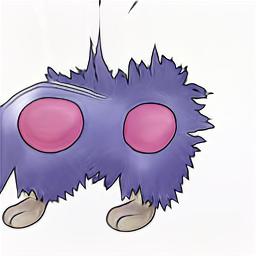}} & 
\frame{\includegraphics[width=0.067\linewidth, height=0.067\linewidth]{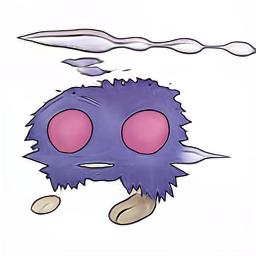}} & 
\frame{\includegraphics[width=0.067\linewidth, height=0.067\linewidth]{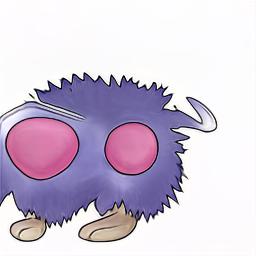}} & 
\frame{\includegraphics[width=0.067\linewidth, height=0.067\linewidth]{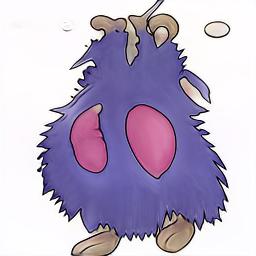}} & 
\frame{\includegraphics[width=0.067\linewidth, height=0.067\linewidth]{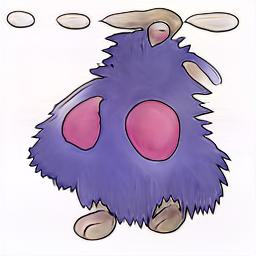}} & 
\frame{\includegraphics[width=0.067\linewidth, height=0.067\linewidth]{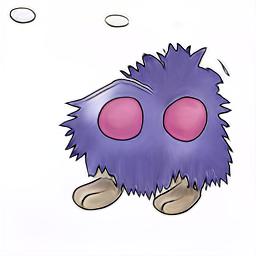}} & 
\frame{\includegraphics[width=0.067\linewidth, height=0.067\linewidth]{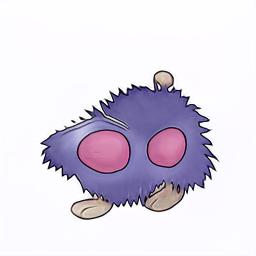}} \tabularnewline[5pt]
\hline & &  & &  & &  & &  & &  & & 
\tabularnewline[2pt]


~~\rotatebox{90}{{\hspace{0.6em} Real \hspace{0.0em} }} &

\multicolumn{7}{@{\hskip -0.04in}c:}{
\frame{\includegraphics[width=0.07\linewidth, height=0.07\linewidth]{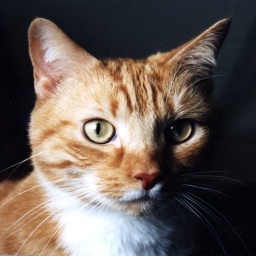}} 
\frame{\includegraphics[width=0.07\linewidth, height=0.07\linewidth]{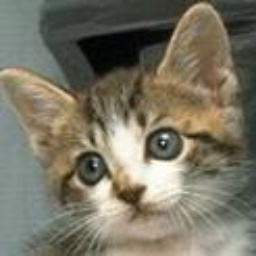}} 
\frame{\includegraphics[width=0.07\linewidth, height=0.07\linewidth]{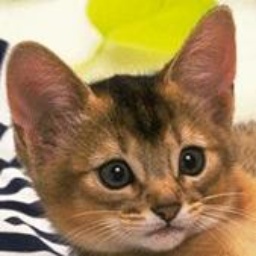}} 
\frame{\includegraphics[width=0.07\linewidth, height=0.07\linewidth]{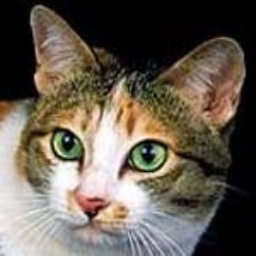}} 
\frame{\includegraphics[width=0.07\linewidth, height=0.07\linewidth]{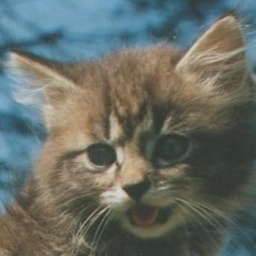}}  
} &
\multicolumn{7}{@{\hskip 0.04in}c}{
\frame{\includegraphics[width=0.07\linewidth, height=0.07\linewidth]{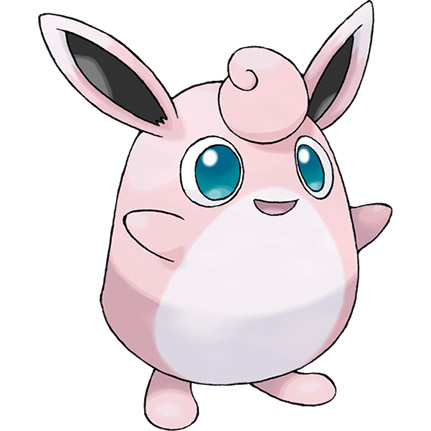}} 
\frame{\includegraphics[width=0.07\linewidth, height=0.07\linewidth]{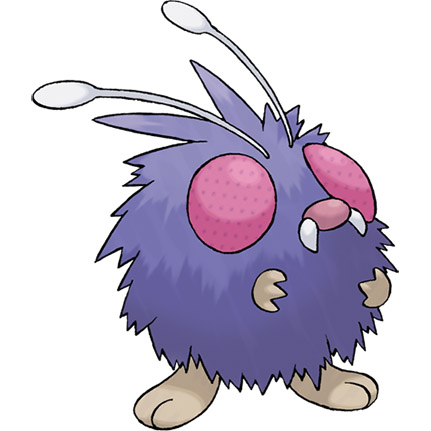}} 
\frame{\includegraphics[width=0.07\linewidth, height=0.07\linewidth]{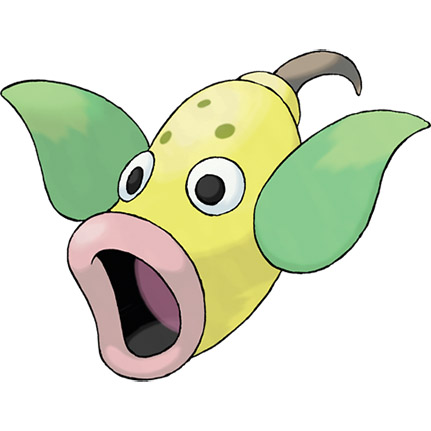}} 
\frame{\includegraphics[width=0.07\linewidth, height=0.07\linewidth]{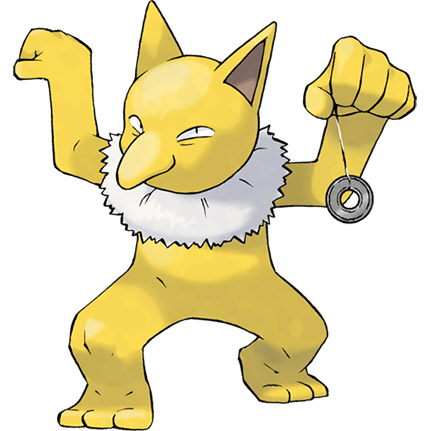}} 
\frame{\includegraphics[width=0.07\linewidth, height=0.07\linewidth]{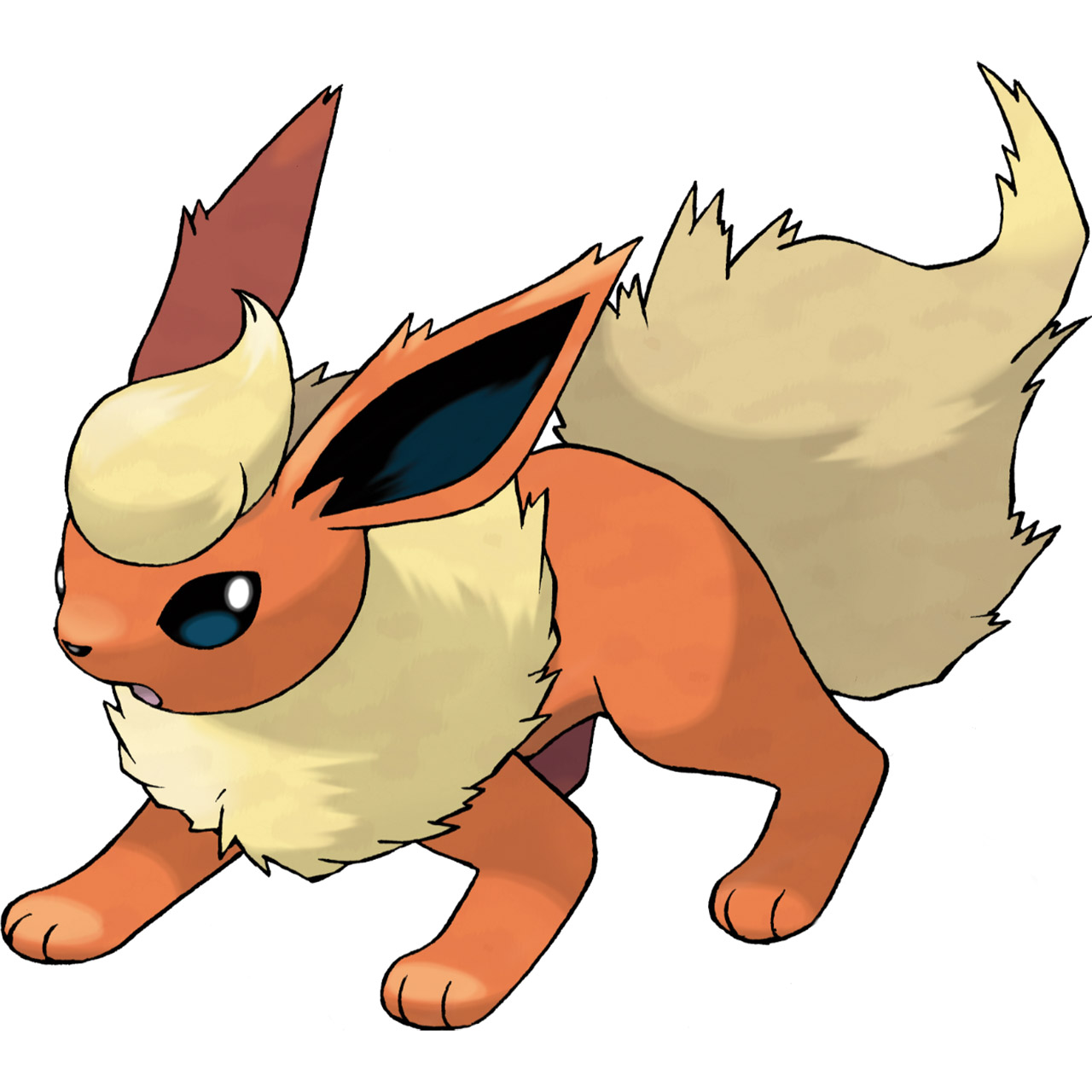}}  
}

\tabularnewline[2pt]

~~\rotatebox{90}{{\hspace{0.0em} Source \hspace{-3.0em} }} &
\frame{\includegraphics[width=0.067\linewidth, height=0.067\linewidth]{figures/qualitative_results/ffhq-sketch/source/p_2_000000.jpg}} & 
\frame{\includegraphics[width=0.067\linewidth, height=0.067\linewidth]{figures/qualitative_results/ffhq-sketch/source/p_6_000000.jpg}} &
\frame{\includegraphics[width=0.067\linewidth, height=0.067\linewidth]{figures/qualitative_results/ffhq-sketch/source/p_7_000000.jpg}} &
\frame{\includegraphics[width=0.067\linewidth, height=0.067\linewidth]{figures/qualitative_results/ffhq-sketch/source/p_8_000000.jpg}} &
\frame{\includegraphics[width=0.067\linewidth, height=0.067\linewidth]{figures/qualitative_results/ffhq-sketch/source/p_17_000000.jpg}} &
\frame{\includegraphics[width=0.067\linewidth, height=0.067\linewidth]{figures/qualitative_results/ffhq-sketch/source/p_22_000000.jpg}} &
\frame{\includegraphics[width=0.067\linewidth, height=0.067\linewidth]{figures/qualitative_results/ffhq-sketch/source/p_23_000000.jpg}} &

\frame{\includegraphics[width=0.067\linewidth, height=0.067\linewidth]{figures/qualitative_results/1-5-shot/pokemons-source/z1_sample14.jpg}} & 
\frame{\includegraphics[width=0.067\linewidth, height=0.067\linewidth]{figures/qualitative_results/1-5-shot/pokemons-source/z1_sample108.jpg}} & 
\frame{\includegraphics[width=0.067\linewidth, height=0.067\linewidth]{figures/qualitative_results/1-5-shot/pokemons-source/z1_sample262.jpg}} & 
\frame{\includegraphics[width=0.067\linewidth, height=0.067\linewidth]{figures/qualitative_results/1-5-shot/pokemons-source/z1_sample660.jpg}} & 
\frame{\includegraphics[width=0.067\linewidth, height=0.067\linewidth]{figures/qualitative_results/1-5-shot/pokemons-source/z1_sample668.jpg}} & 
\frame{\includegraphics[width=0.067\linewidth, height=0.067\linewidth]{figures/qualitative_results/1-5-shot/pokemons-source/z1_sample694.jpg}} & 
\frame{\includegraphics[width=0.067\linewidth, height=0.067\linewidth]{figures/qualitative_results/1-5-shot/pokemons-source/z1_sample950.jpg}} 
\tabularnewline[-3pt]

~~\rotatebox{90}{{\hspace{0.4em} CDC \hspace{-3.0em} }} &
\frame{\includegraphics[width=0.067\linewidth, height=0.067\linewidth]{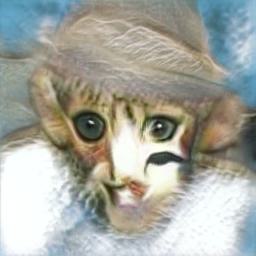}} & 
\frame{\includegraphics[width=0.067\linewidth, height=0.067\linewidth]{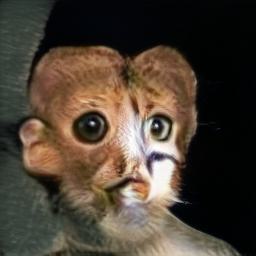}} &
\frame{\includegraphics[width=0.067\linewidth, height=0.067\linewidth]{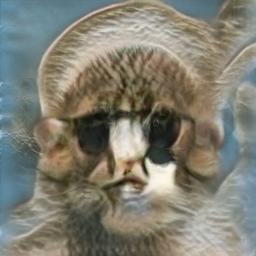}} &
\frame{\includegraphics[width=0.067\linewidth, height=0.067\linewidth]{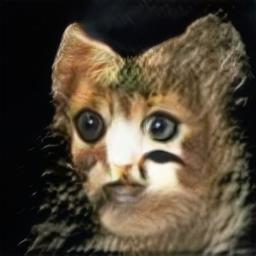}} &
\frame{\includegraphics[width=0.067\linewidth, height=0.067\linewidth]{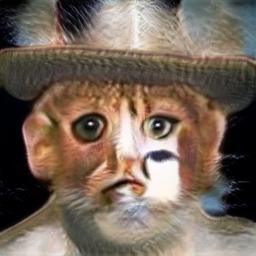}} &
\frame{\includegraphics[width=0.067\linewidth, height=0.067\linewidth]{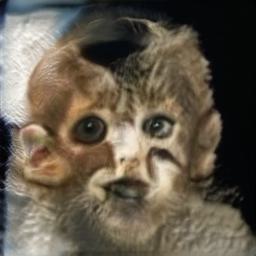}} &
\frame{\includegraphics[width=0.067\linewidth, height=0.067\linewidth]{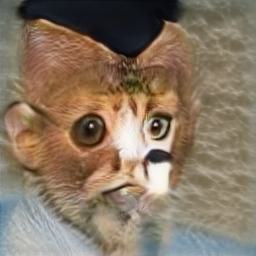}} &

\frame{\includegraphics[width=0.067\linewidth, height=0.067\linewidth]{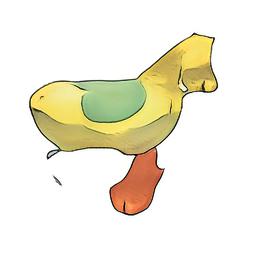}} & 
\frame{\includegraphics[width=0.067\linewidth, height=0.067\linewidth]{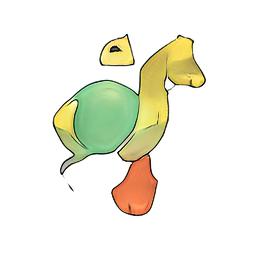}} & 
\frame{\includegraphics[width=0.067\linewidth, height=0.067\linewidth]{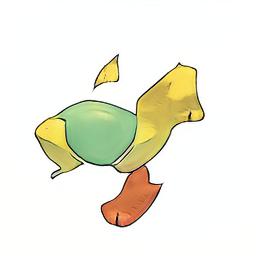}} & 
\frame{\includegraphics[width=0.067\linewidth, height=0.067\linewidth]{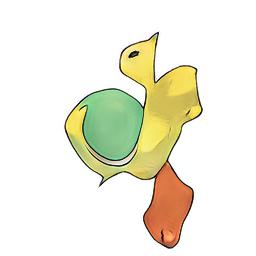}} & 
\frame{\includegraphics[width=0.067\linewidth, height=0.067\linewidth]{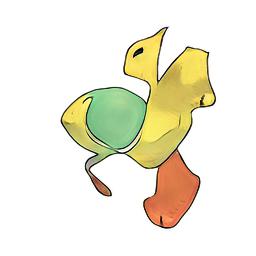}} & 
\frame{\includegraphics[width=0.067\linewidth, height=0.067\linewidth]{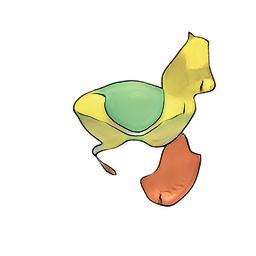}} & 
\frame{\includegraphics[width=0.067\linewidth, height=0.067\linewidth]{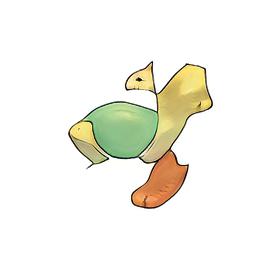}} 
\tabularnewline[-3pt]

~~\rotatebox{90}{{\hspace{0.4em} Ours \hspace{-3.0em} }} &
\frame{\includegraphics[width=0.067\linewidth, height=0.067\linewidth]{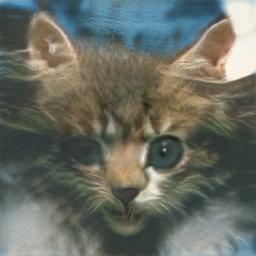}} & 
\frame{\includegraphics[width=0.067\linewidth, height=0.067\linewidth]{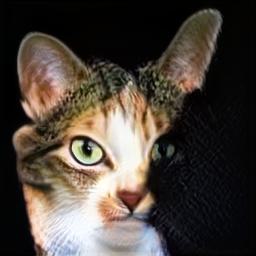}} &
\frame{\includegraphics[width=0.067\linewidth, height=0.067\linewidth]{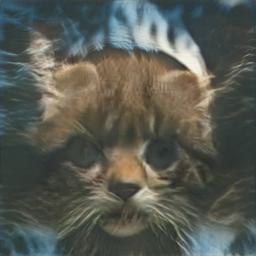}} &
\frame{\includegraphics[width=0.067\linewidth, height=0.067\linewidth]{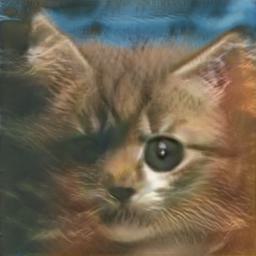}} &
\frame{\includegraphics[width=0.067\linewidth, height=0.067\linewidth]{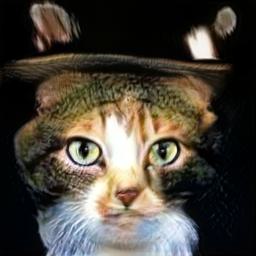}} &
\frame{\includegraphics[width=0.067\linewidth, height=0.067\linewidth]{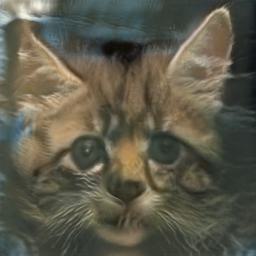}} &
\frame{\includegraphics[width=0.067\linewidth, height=0.067\linewidth]{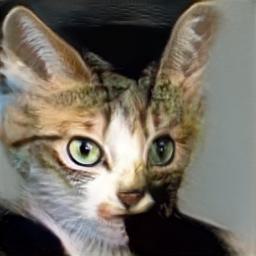}} &

\frame{\includegraphics[width=0.067\linewidth, height=0.067\linewidth]{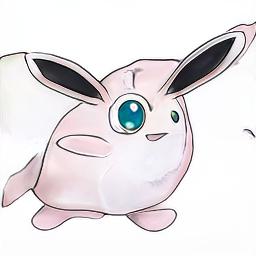}} & 
\frame{\includegraphics[width=0.067\linewidth, height=0.067\linewidth]{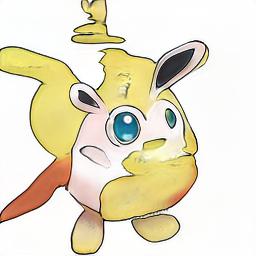}} & 
\frame{\includegraphics[width=0.067\linewidth, height=0.067\linewidth]{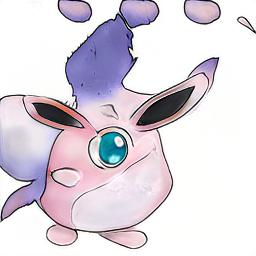}} & 
\frame{\includegraphics[width=0.067\linewidth, height=0.067\linewidth]{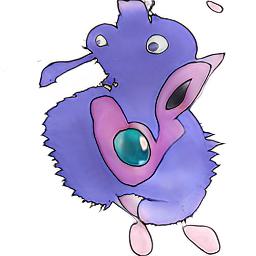}} & 
\frame{\includegraphics[width=0.067\linewidth, height=0.067\linewidth]{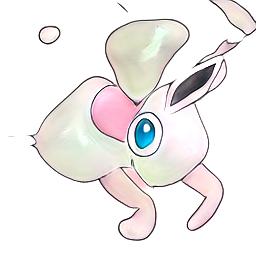}} & 
\frame{\includegraphics[width=0.067\linewidth, height=0.067\linewidth]{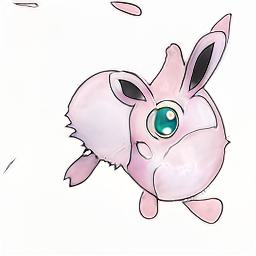}} & 
\frame{\includegraphics[width=0.067\linewidth, height=0.067\linewidth]{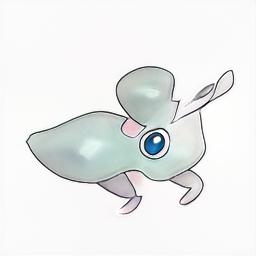}} 
\tabularnewline[-3pt]

\end{tabular}
\par\end{centering}
\vspace{5pt}
\caption{1-shot and 5-shot GAN adaptation results on the Cats and Pokemons datasets.}
\label{fig:1-5-shot}

\end{figure*}

%% file: tables/ablation_SS_lambda_l.tex
\begin{table}[t]

	\setlength{\tabcolsep}{0.55em}
	\renewcommand{\arraystretch}{0.95}
	\centering
		\begin{tabular}{@{\hskip -0.04in}c@{\hskip 0.02in}|@{\hskip 0.03in}c@{\hskip 0.03in}|c@{\hskip -0.01in}c@{\hskip 0.02in}|c@{\hskip -0.05in}c@{\hskip -0.06in}}
			\multirow{2}{*}{\footnotesize{} $\lambda_{SS}$} & \multirow{1}{*}{\footnotesize{} Res.} & \multicolumn{2}{c|}{\footnotesize{} \hspace{-0.7ex} \textbf{Face$\rightarrow$Anime}} & \multicolumn{2}{c}{\footnotesize{} \hspace{-1.5ex} \textbf{Church$\rightarrow$Shells}} 
            \tabularnewline 
            & \footnotesize{} of $G^l$ & \footnotesize{} FID$\downarrow$  & \footnotesize{} LPIPS$\uparrow$  & \footnotesize{} FID$\downarrow$  & \footnotesize{} LPIPS$\uparrow$ 
            \tabularnewline
            
			\hline

             - & - & \footnotesize{116.4} & \footnotesize{0.36}  & \footnotesize{175.4} & \footnotesize{0.43}  
            \tabularnewline

            \hdashline[0.1pt/0.5pt]

            \footnotesize{} 0.2 & \footnotesize{} ~32$\times$32 & \footnotesize{110.0} & \footnotesize{0.41} & \footnotesize{160.2} & \footnotesize{0.44} \tabularnewline
            \footnotesize{} 1.0 & \footnotesize{} ~32$\times$32 & \textbf{\footnotesize{96.4}} & \footnotesize{0.51} &\footnotesize{144.5} & \footnotesize{0.50} \tabularnewline
            \footnotesize{} 25.0 & \footnotesize{} ~32$\times$32 & \footnotesize{105.2} & \footnotesize{0.58} & \footnotesize{171.0} & \footnotesize{0.55} \tabularnewline
            \footnotesize{} 125.0 & \footnotesize{} ~32$\times$32 & \footnotesize{131.3} & \textbf{\footnotesize{0.64}} & \footnotesize{188.5} & \textbf{\footnotesize{0.57}} \tabularnewline

            \hdashline[0.1pt/0.5pt]
            
            \footnotesize{} 5.0 & \footnotesize{} ~8$\times$8 & \footnotesize{104.1} & \footnotesize{0.44} & \footnotesize{156.6} & \footnotesize{0.45} \tabularnewline
            \footnotesize{} 5.0 & \footnotesize{} ~16$\times$16 & \footnotesize{101.4} & \footnotesize{0.55} & \footnotesize{150.2} & \footnotesize{0.48}  \tabularnewline
            \footnotesize{} 5.0 & \footnotesize{} ~64$\times$64 & \footnotesize{114.7} & \footnotesize{0.59} & \footnotesize{165.5} & \footnotesize{0.54}  \tabularnewline
            \footnotesize{} 5.0 & \footnotesize{} ~128$\times$128 & \footnotesize{128.2} & \footnotesize{0.60} & \footnotesize{182.2} & \textbf{\footnotesize{0.57}} \tabularnewline            

            \hdashline[0.1pt/0.5pt]

            \footnotesize{} 5.0 & \footnotesize{} ~32$\times$32 & {\footnotesize{97.3}} & {\footnotesize{0.57}}   & \textbf{\footnotesize{140.5}} & {\footnotesize{0.53}} 

\end{tabular}
\vspace{0.5ex}
\caption{Ablation on $\lambda_{SS}$ and the resolution of $G^l$ used for the smoothness similarity regularization.}
\label{table:ablation_SS_suppl} %
\vspace{-1.5ex}
\end{table}

%% file: tables/suppl_weights_Lall.tex
\begin{table}[t]
\setlength{\tabcolsep}{0.55em}
\renewcommand{\arraystretch}{0.95}
\centering
		\begin{tabular}{@{\hskip -0.04in}c@{\hskip 0.04in}|c@{\hskip -0.01in}c@{\hskip 0.02in}|c@{\hskip -0.05in}c@{\hskip -0.06in}}
			\multirow{2}{*}{\footnotesize{} $\mathcal{L}_{all}$ weighting}  & \multicolumn{2}{c|}{\footnotesize{} \hspace{-0.7ex} \textbf{Face$\rightarrow$Anime}} & \multicolumn{2}{c}{\footnotesize{} \hspace{-1.5ex} \textbf{Church$\rightarrow$Shells}} 
            \tabularnewline 
            & \footnotesize{} FID$\downarrow$  & \footnotesize{} LPIPS$\uparrow$  & \footnotesize{} FID$\downarrow$  & \footnotesize{} LPIPS$\uparrow$ 
            \tabularnewline
			\hline 	
			{\footnotesize{} ``Earlier''} & \footnotesize{96.8} & \footnotesize{\textbf{0.59}}  & \footnotesize{157.4} & \footnotesize{0.52}  
            \tabularnewline
            \footnotesize{} Uniform (ours) & {\footnotesize{97.3}} & \footnotesize{0.57}   & {\footnotesize{140.5}} & \textbf{{\footnotesize{0.53}}} 
            \tabularnewline
            \footnotesize{} ``Later'' & \textbf{\footnotesize{93.2}} & \footnotesize{0.53}  & \textbf{\footnotesize{138.4}} & \footnotesize{0.48}  
\end{tabular}
\caption{Effect of using different weights for different layers in $L_{all}$. Bold denotes the best performance.}
\label{table:suppl_weigths_Lall} %
\end{table}

%% file: figures/crops.tex
\begin{figure*}[t]
\begin{centering}
\setlength{\tabcolsep}{0.01in}
\renewcommand{\arraystretch}{1}
\par\end{centering}
\begin{centering}

\begin{tabular}{@{\hskip -0.17in}c@{\hskip 0.01in}c@{\hskip 0.01in}c@{\hskip 0.01in}c@{\hskip 0.01in}c@{\hskip 0.13in}c@{\hskip 0.01in}c@{\hskip 0.02in}c@{\hskip 0.01in}c@{\hskip 0.01in}c@{\hskip 0.01in}c@{\hskip 0.01in}c@{\hskip 0.01in}c@{\hskip 0.01in}c}

\multicolumn{5}{c}{ \hspace{-5.0ex} Real samples -- 10-Shot healthy Crops } &
\multicolumn{5}{c}{ Real samples -- 10-Shot Crops with nitrogen deficiencies}
\tabularnewline[0pt]

\frame{\includegraphics[draft=\draft,width=0.0968\linewidth, height=0.0968\linewidth]{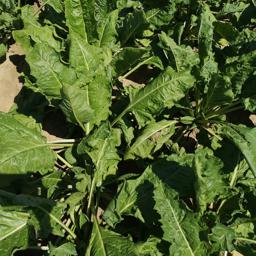}} \hspace{-1.0ex} &
\frame{\includegraphics[draft=\draft,width=0.0968\linewidth, height=0.0968\linewidth]{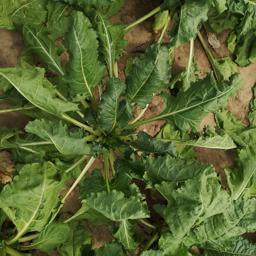}} \hspace{-1.0ex} &
\frame{\includegraphics[draft=\draft,width=0.0968\linewidth, height=0.0968\linewidth]{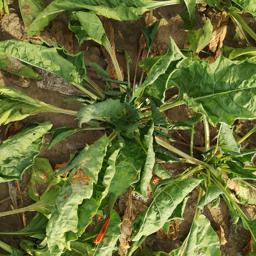}} \hspace{-1.0ex} &
\frame{\includegraphics[draft=\draft,width=0.0968\linewidth, height=0.0968\linewidth]{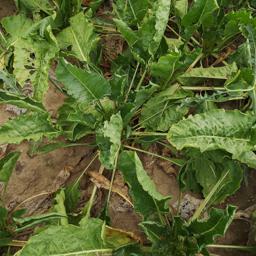}} \hspace{-1.0ex} &
\frame{\includegraphics[draft=\draft,width=0.0968\linewidth, height=0.0968\linewidth]{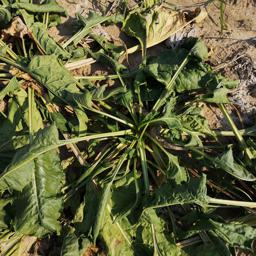}} \hspace{-1.0ex} &

\frame{\includegraphics[draft=\draft,width=0.0968\linewidth, height=0.0968\linewidth]{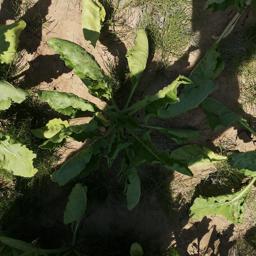}} \hspace{-1.0ex} &
\frame{\includegraphics[draft=\draft,width=0.0968\linewidth, height=0.0968\linewidth]{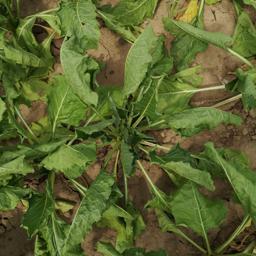}} \hspace{-1.0ex} &
\frame{\includegraphics[draft=\draft,width=0.0968\linewidth, height=0.0968\linewidth]{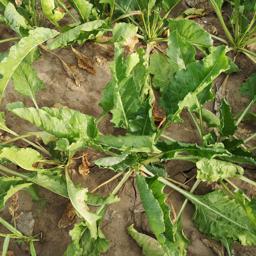}} \hspace{-1.0ex} &
\frame{\includegraphics[draft=\draft,width=0.0968\linewidth, height=0.0968\linewidth]{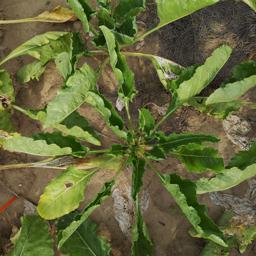}} \hspace{-1.0ex} &
\frame{\includegraphics[draft=\draft,width=0.0968\linewidth, height=0.0968\linewidth]{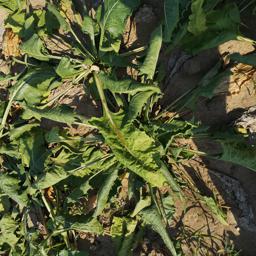}} 
\tabularnewline[-3pt]

\frame{\includegraphics[draft=\draft,width=0.0968\linewidth, height=0.0968\linewidth]{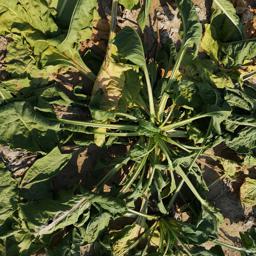}} \hspace{-1.0ex} & 
\frame{\includegraphics[draft=\draft,width=0.0968\linewidth, height=0.0968\linewidth]{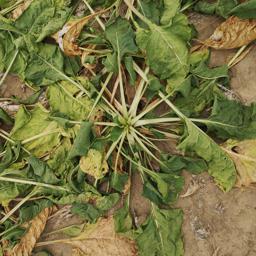}} \hspace{-1.0ex} &
\frame{\includegraphics[draft=\draft,width=0.0968\linewidth, height=0.0968\linewidth]{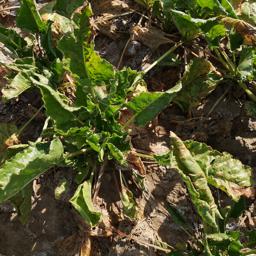}} \hspace{-1.0ex} &
\frame{\includegraphics[draft=\draft,width=0.0968\linewidth, height=0.0968\linewidth]{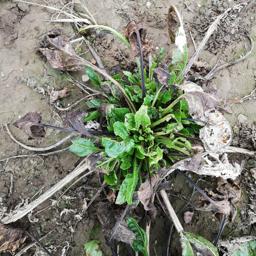}} \hspace{-1.0ex} &
\frame{\includegraphics[draft=\draft,width=0.0968\linewidth, height=0.0968\linewidth]{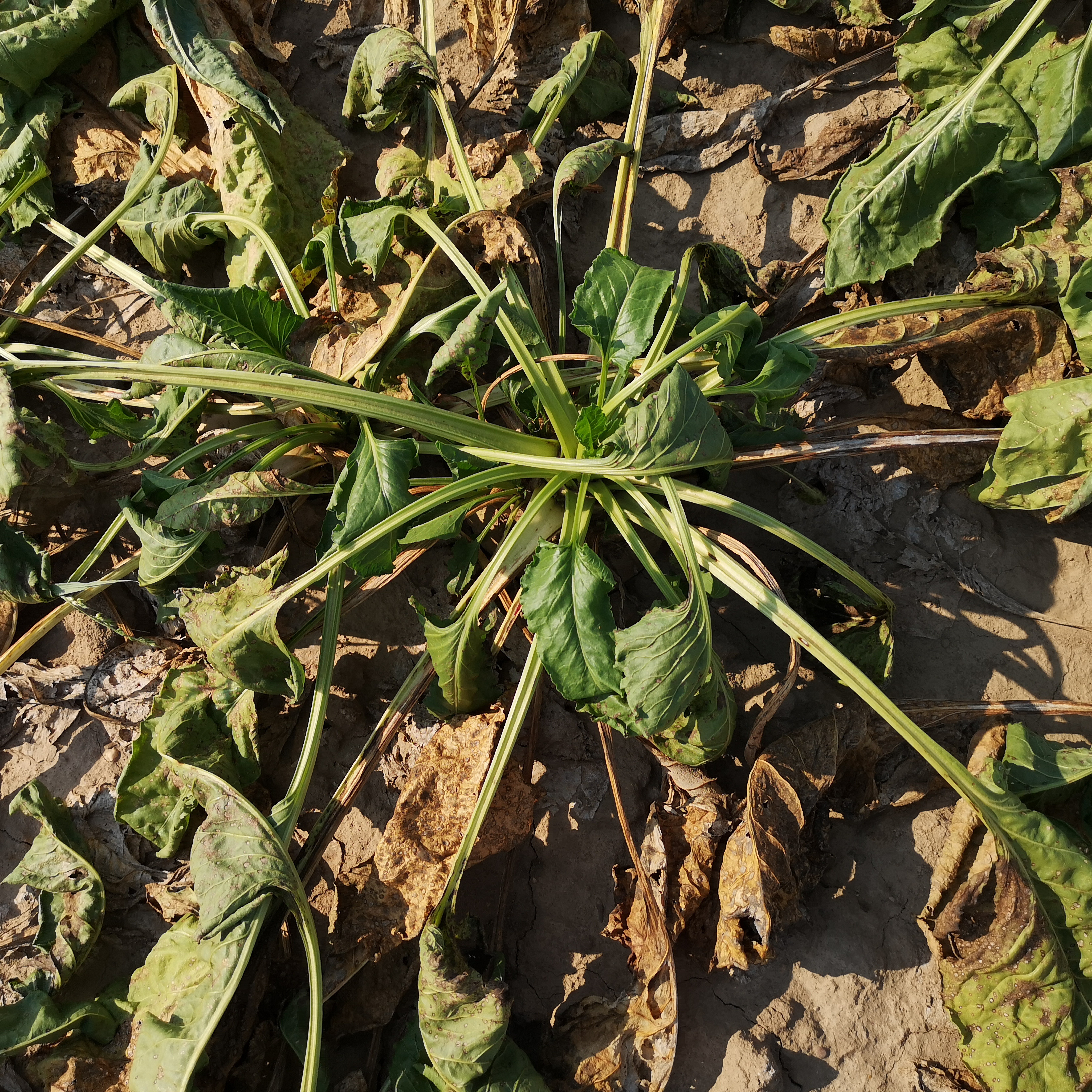}} \hspace{-1.0ex} &

\frame{\includegraphics[draft=\draft,width=0.0968\linewidth, height=0.0968\linewidth]{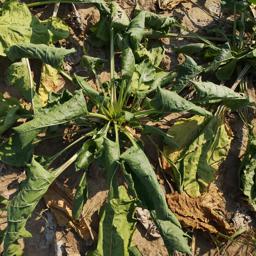}} \hspace{-1.0ex} &
\frame{\includegraphics[draft=\draft,width=0.0968\linewidth, height=0.0968\linewidth]{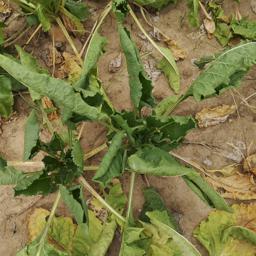}} \hspace{-1.0ex} &
\frame{\includegraphics[draft=\draft,width=0.0968\linewidth, height=0.0968\linewidth]{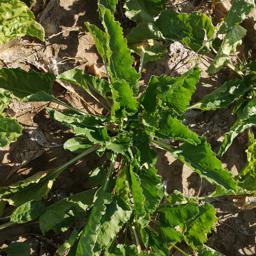}} \hspace{-1.0ex} &
\frame{\includegraphics[draft=\draft,width=0.0968\linewidth, height=0.0968\linewidth]{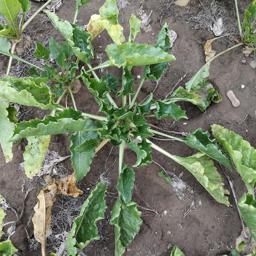}} \hspace{-1.0ex} &
\frame{\includegraphics[draft=\draft,width=0.0968\linewidth, height=0.0968\linewidth]{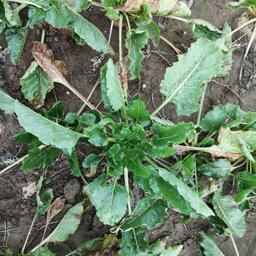}} 

\tabularnewline[3pt]

\multicolumn{5}{c}{ Generated samples: } &
\multicolumn{5}{c}{ Generated samples: }
\tabularnewline[0pt]

\frame{\includegraphics[draft=\draft,width=0.0968\linewidth, height=0.0968\linewidth]{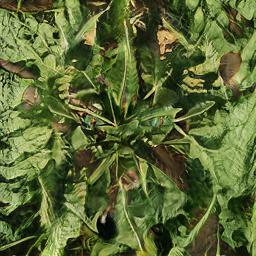}} \hspace{-1.0ex} &
\frame{\includegraphics[draft=\draft,width=0.0968\linewidth, height=0.0968\linewidth]{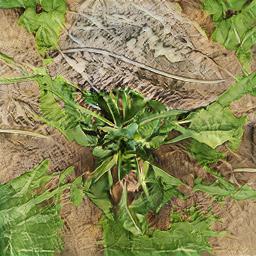}} \hspace{-1.0ex} &
\frame{\includegraphics[draft=\draft,width=0.0968\linewidth, height=0.0968\linewidth]{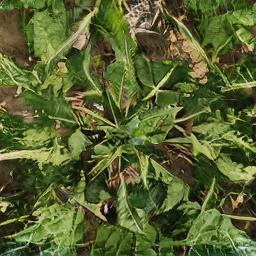}} \hspace{-1.0ex} &
\frame{\includegraphics[draft=\draft,width=0.0968\linewidth, height=0.0968\linewidth]{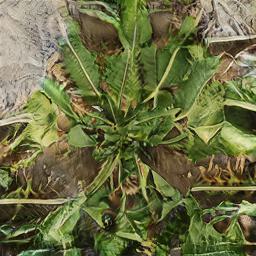}} \hspace{-1.0ex} &
\frame{\includegraphics[draft=\draft,width=0.0968\linewidth, height=0.0968\linewidth]{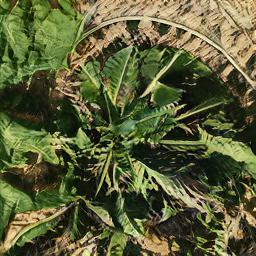}} \hspace{-1.0ex} &

\frame{\includegraphics[draft=\draft,width=0.0968\linewidth, height=0.0968\linewidth]{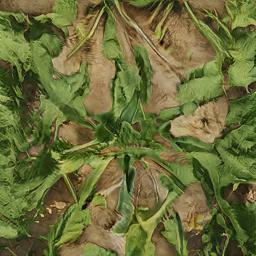}} \hspace{-1.0ex} &
\frame{\includegraphics[draft=\draft,width=0.0968\linewidth, height=0.0968\linewidth]{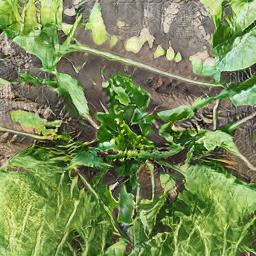}} \hspace{-1.0ex} &
\frame{\includegraphics[draft=\draft,width=0.0968\linewidth, height=0.0968\linewidth]{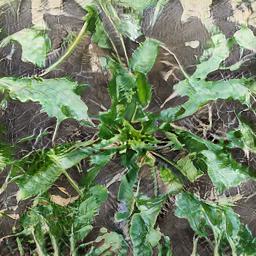}} \hspace{-1.0ex} &
\frame{\includegraphics[draft=\draft,width=0.0968\linewidth, height=0.0968\linewidth]{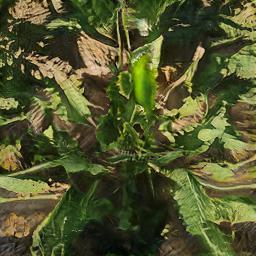}} \hspace{-1.0ex} &
\frame{\includegraphics[draft=\draft,width=0.0968\linewidth, height=0.0968\linewidth]{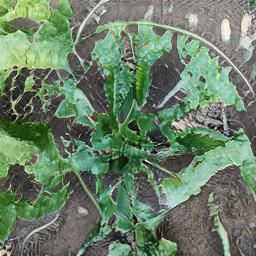}}
\tabularnewline[-3pt]

\frame{\includegraphics[draft=\draft,width=0.0968\linewidth, height=0.0968\linewidth]{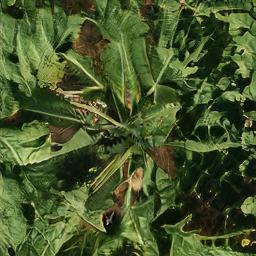}} \hspace{-1.0ex} &
\frame{\includegraphics[draft=\draft,width=0.0968\linewidth, height=0.0968\linewidth]{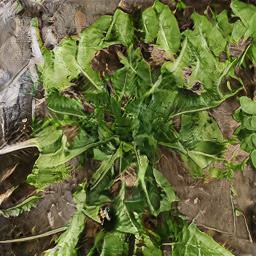}} \hspace{-1.0ex} &
\frame{\includegraphics[draft=\draft,width=0.0968\linewidth, height=0.0968\linewidth]{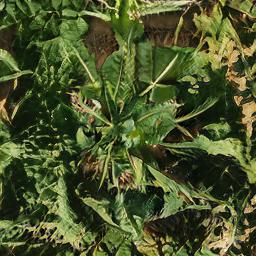}} \hspace{-1.0ex} &
\frame{\includegraphics[draft=\draft,width=0.0968\linewidth, height=0.0968\linewidth]{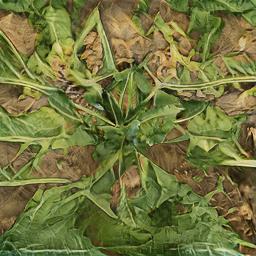}} \hspace{-1.0ex} &
\frame{\includegraphics[draft=\draft,width=0.0968\linewidth, height=0.0968\linewidth]{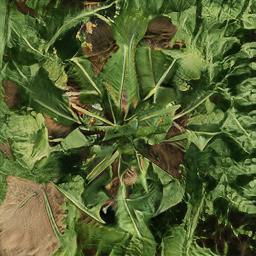}} \hspace{-1.0ex} &

\frame{\includegraphics[draft=\draft,width=0.0968\linewidth, height=0.0968\linewidth]{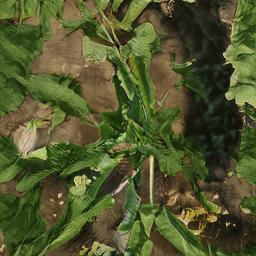}} \hspace{-1.0ex} &
\frame{\includegraphics[draft=\draft,width=0.0968\linewidth, height=0.0968\linewidth]{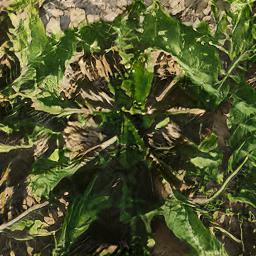}} \hspace{-1.0ex} &
\frame{\includegraphics[draft=\draft,width=0.0968\linewidth, height=0.0968\linewidth]{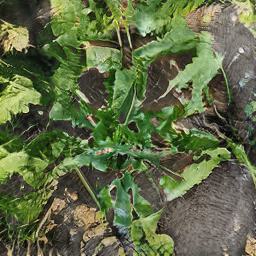}} \hspace{-1.0ex} &
\frame{\includegraphics[draft=\draft,width=0.0968\linewidth, height=0.0968\linewidth]{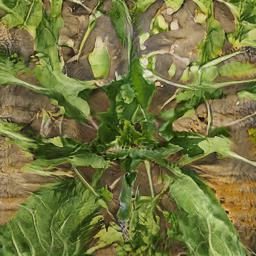}} \hspace{-1.0ex} &
\frame{\includegraphics[draft=\draft,width=0.0968\linewidth, height=0.0968\linewidth]{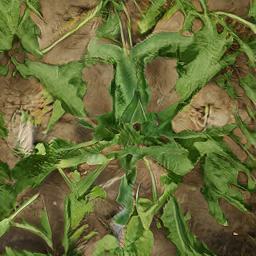}}

\tabularnewline[0pt]
\end{tabular}

\par\end{centering}
\vspace{2pt}
\caption{Above: 10-shot subsets of the DND-SB dataset \cite{yi2020deep}, depicting healthy sugar beat crops or crops with nitrogen nutrition deficiency. Below: few-shot adaptation results with StyleGAN pre-trained on FFHQ. 
}
\label{fig:crops}
\vspace{-2ex}
\end{figure*}

%% file: tables/biggan_suppl_ablation.tex
\begin{table}[t]

	\setlength{\tabcolsep}{0.55em}
	\renewcommand{\arraystretch}{0.95}
	\centering
		\begin{tabular}{@{\hskip 0.03in}c@{\hskip 0.03in}|@{\hskip 0.11in}c@{\hskip -0.02in}|c@{\hskip -0.15in}c@{\hskip -0.02in}|c@{\hskip -0.15in}c@{\hskip -0.00in}}
			\multicolumn{2}{c|}{\footnotesize{} Smooth. reg. w.r.t.:} & \multicolumn{2}{c|}{\footnotesize{} \hspace{-1.5ex} \textbf{ImageNet$\rightarrow$Flowers}} & \multicolumn{2}{c}{\footnotesize{} \hspace{-1.5ex} \textbf{ImageNet$\rightarrow$Pokemons}} 
            \tabularnewline 
            \footnotesize{}  Noise & \footnotesize{} Class & \footnotesize{} ~~~FID$\downarrow$  & \footnotesize{} ~LPIPS$\uparrow$  & \footnotesize{} ~~~FID$\downarrow$  & \footnotesize{} LPIPS$\uparrow$ 
            \tabularnewline
            
			\hline 	

			\xmark \color{black} &  \xmark \color{black} & \footnotesize{~~~123.9} & \footnotesize{~0.28}  & \footnotesize{~~~129.4} & \footnotesize{0.27}  
            \tabularnewline

			\cmark \color{black} &  \xmark \color{black} & \footnotesize{~~~114.0} & \footnotesize{~0.39}  & \footnotesize{~~104.6} & \footnotesize{~0.41}  
            \tabularnewline            

            \cmark \color{black} &  \cmark \color{black} & \textbf{\footnotesize{~~~106.4}} & \textbf{\footnotesize{~0.55}}   & \textbf{\footnotesize{~~~89.6}} & \textbf{\footnotesize{0.56}} 
\end{tabular}
\vspace{0.5ex}
\caption{Ablation on the performance when adapting the class-conditional BigGAN model \cite{Brock2019} pre-trained on ImageNet.}
\label{table:biggan_supl_ablation} %
\end{table}

%% file: figures/qual_comparison_distant3.tex
\begin{figure*}[t]
\begin{centering}
\setlength{\tabcolsep}{0.53in}
\renewcommand{\arraystretch}{1}
\par\end{centering}
\begin{centering}

\begin{tabular}{@{\hskip -0.02in}c@{\hskip 0.04in}:c@{\hskip 0.01in}c@{\hskip 0.01in}c@{\hskip 0.01in}c@{\hskip 0.01in}c@{\hskip 0.01in}c@{\hskip 0.01in}c@{\hskip 0.01in}c@{\hskip 0.01in}c@{\hskip 0.01in}c@{\hskip 0.01in}c@{\hskip 0.01in}c@{\hskip 0.01in}c}

  \multirow{-1}{*}{\begin{tabular}{c@{\hskip 0.01in}c@{\hskip 0.03in}} 
  \frame{\includegraphics[draft=\draft,width=0.075\linewidth, height=0.070\linewidth]{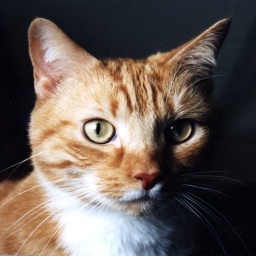}} &
  \frame{\includegraphics[draft=\draft,width=0.075\linewidth, height=0.070\linewidth]{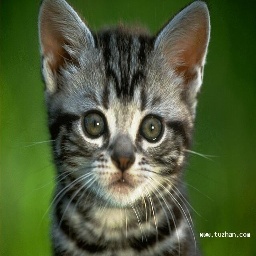}} \tabularnewline[-3pt]	
  \frame{\includegraphics[draft=\draft,width=0.075\linewidth, height=0.070\linewidth]{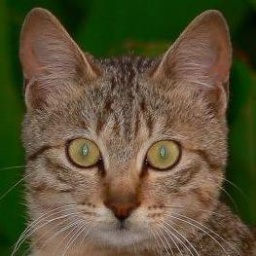}} &
  \frame{\includegraphics[draft=\draft,width=0.075\linewidth, height=0.070\linewidth]{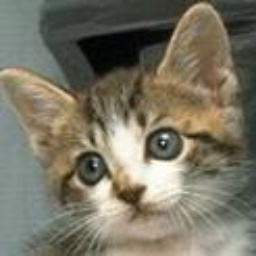}} \tabularnewline[-3pt]
  \frame{\includegraphics[draft=\draft,width=0.075\linewidth, height=0.070\linewidth]{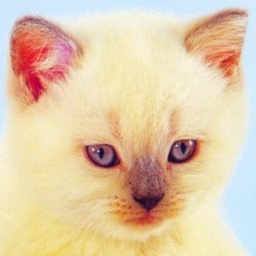}} &
  \frame{\includegraphics[draft=\draft,width=0.075\linewidth, height=0.070\linewidth]{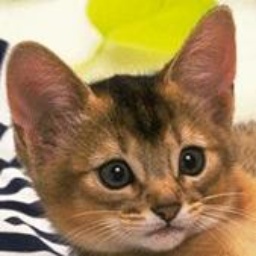}} \tabularnewline[-3pt] 	
  \frame{\includegraphics[draft=\draft,width=0.075\linewidth, height=0.070\linewidth]{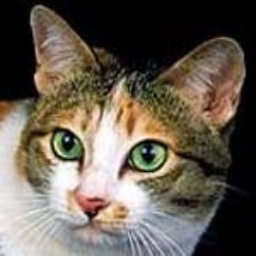}} &
  \frame{\includegraphics[draft=\draft,width=0.075\linewidth, height=0.070\linewidth]{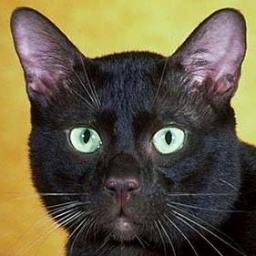}} \tabularnewline[-3pt]
  \frame{\includegraphics[draft=\draft,width=0.075\linewidth, height=0.070\linewidth]{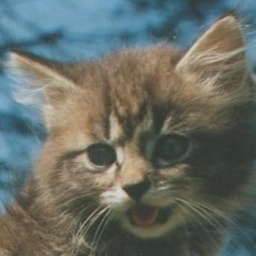}} &
  \frame{\includegraphics[draft=\draft,width=0.075\linewidth, height=0.070\linewidth]{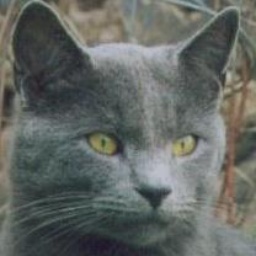}} \tabularnewline[+1pt] 
  \multicolumn{2}{@{\hskip -0.02in}c@{\hskip 0.03in}}{\small 10-shot} \tabularnewline[-2pt] 
  \multicolumn{2}{@{\hskip -0.02in}c@{\hskip 0.03in}}{\small Cats} \tabularnewline[-3pt]
  \end{tabular}} &
  
  \frame{\includegraphics[draft=\draft,width=0.075\linewidth, height=0.070\linewidth]{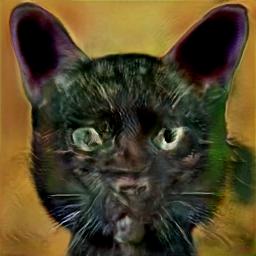}} & 
  \frame{\includegraphics[draft=\draft,width=0.075\linewidth, height=0.070\linewidth]{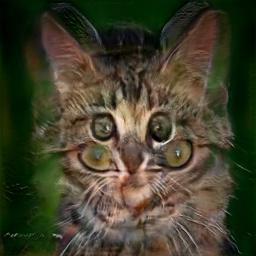}} & 
  \frame{\includegraphics[draft=\draft,width=0.075\linewidth, height=0.070\linewidth]{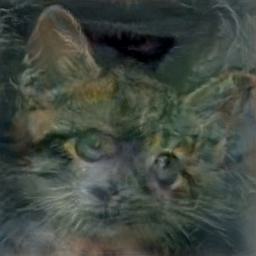}} & 
  \frame{\includegraphics[draft=\draft,width=0.075\linewidth, height=0.070\linewidth]{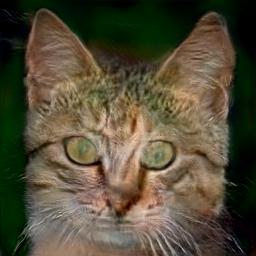}} & 
  \frame{\includegraphics[draft=\draft,width=0.075\linewidth, height=0.070\linewidth]{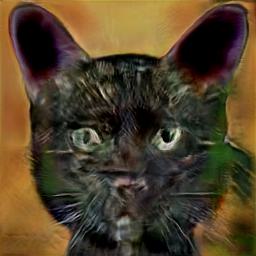}} & 
  \frame{\includegraphics[draft=\draft,width=0.075\linewidth, height=0.070\linewidth]{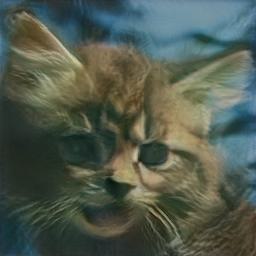}} & 
  \frame{\includegraphics[draft=\draft,width=0.075\linewidth, height=0.070\linewidth]{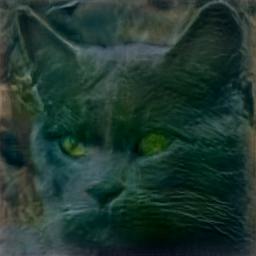}} & 
  \frame{\includegraphics[draft=\draft,width=0.075\linewidth, height=0.070\linewidth]{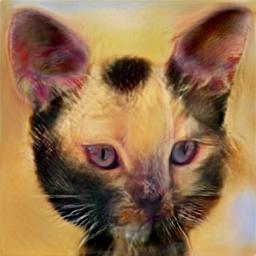}} & 
  \frame{\includegraphics[draft=\draft,width=0.075\linewidth, height=0.070\linewidth]{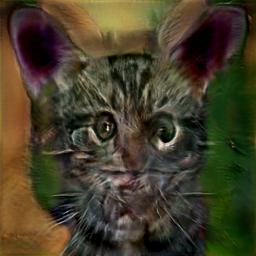}} & 
  \frame{\includegraphics[draft=\draft,width=0.075\linewidth, height=0.070\linewidth]{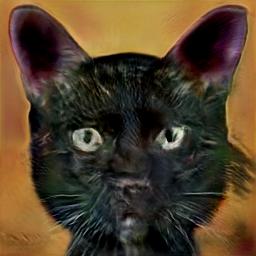}} &~~\rotatebox{270}{{\hspace{-3.6em} TGAN \hspace{-3.0em} } }
  \tabularnewline[-3pt]
  &
  \frame{\includegraphics[draft=\draft,width=0.075\linewidth, height=0.070\linewidth]{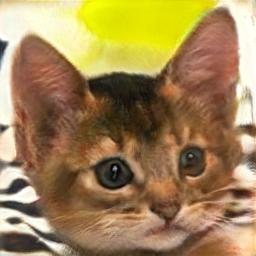}} & 
  \frame{\includegraphics[draft=\draft,width=0.075\linewidth, height=0.070\linewidth]{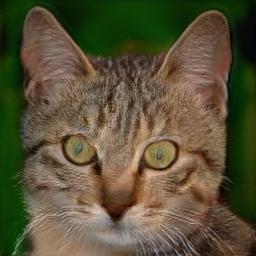}} & 
  \frame{\includegraphics[draft=\draft,width=0.075\linewidth, height=0.070\linewidth]{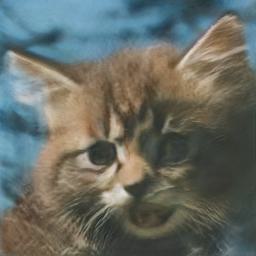}} & 
  \frame{\includegraphics[draft=\draft,width=0.075\linewidth, height=0.070\linewidth]{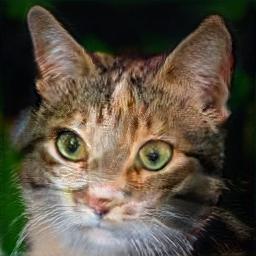}} & 
  \frame{\includegraphics[draft=\draft,width=0.075\linewidth, height=0.070\linewidth]{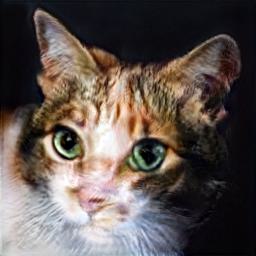}} & 
  \frame{\includegraphics[draft=\draft,width=0.075\linewidth, height=0.070\linewidth]{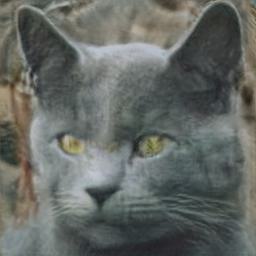}} & 
  \frame{\includegraphics[draft=\draft,width=0.075\linewidth, height=0.070\linewidth]{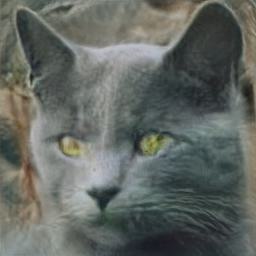}} & 
  \frame{\includegraphics[draft=\draft,width=0.075\linewidth, height=0.070\linewidth]{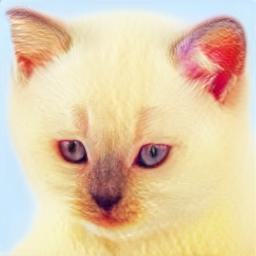}} & 
  \frame{\includegraphics[draft=\draft,width=0.075\linewidth, height=0.070\linewidth]{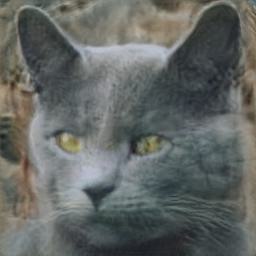}} & 
  \frame{\includegraphics[draft=\draft,width=0.075\linewidth, height=0.070\linewidth]{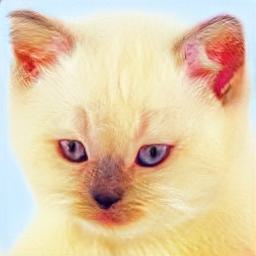}} &~~\rotatebox{270}{{\hspace{-3.7em} FreezeD \hspace{-3.0em} } }
  \tabularnewline[-3pt]
  &
  \frame{\includegraphics[draft=\draft,width=0.075\linewidth, height=0.070\linewidth]{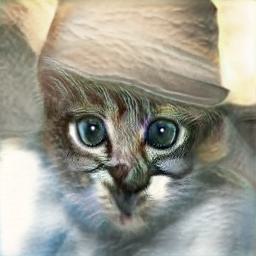}} & 
  \frame{\includegraphics[draft=\draft,width=0.075\linewidth, height=0.070\linewidth]{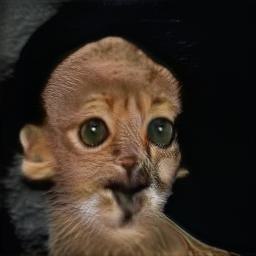}} & 
  \frame{\includegraphics[draft=\draft,width=0.075\linewidth, height=0.070\linewidth]{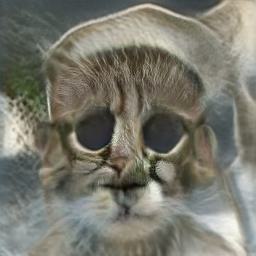}} & 
  \frame{\includegraphics[draft=\draft,width=0.075\linewidth, height=0.070\linewidth]{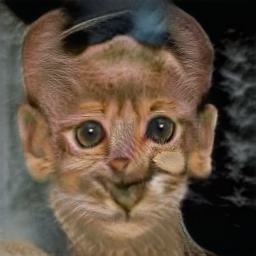}} & 
  \frame{\includegraphics[draft=\draft,width=0.075\linewidth, height=0.070\linewidth]{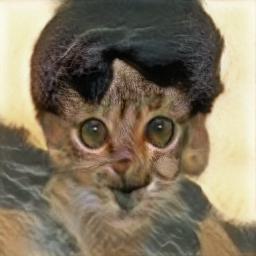}} & 
  \frame{\includegraphics[draft=\draft,width=0.075\linewidth, height=0.070\linewidth]{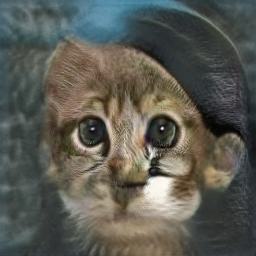}} & 
  \frame{\includegraphics[draft=\draft,width=0.075\linewidth, height=0.070\linewidth]{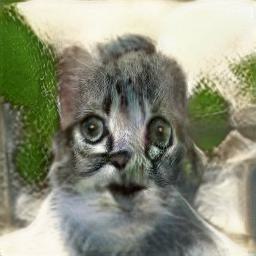}} & 
  \frame{\includegraphics[draft=\draft,width=0.075\linewidth, height=0.070\linewidth]{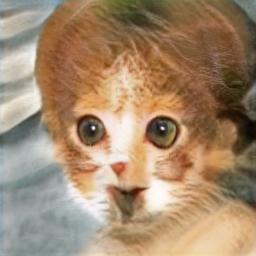}} & 
  \frame{\includegraphics[draft=\draft,width=0.075\linewidth, height=0.070\linewidth]{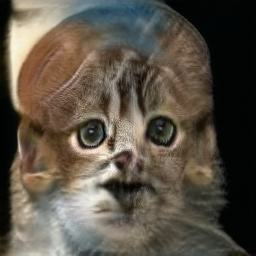}} & 
  \frame{\includegraphics[draft=\draft,width=0.075\linewidth, height=0.070\linewidth]{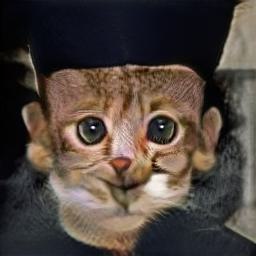}} &  ~~\rotatebox{270}{{\hspace{-3.0em} CDC \hspace{-3.0em} } }
  \tabularnewline[-3pt]
  &
  \frame{\includegraphics[draft=\draft,width=0.075\linewidth, height=0.070\linewidth]{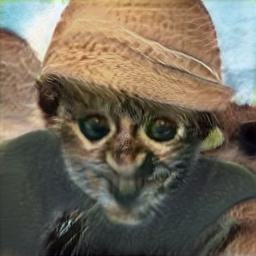}} & 
  \frame{\includegraphics[draft=\draft,width=0.075\linewidth, height=0.070\linewidth]{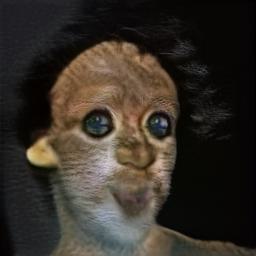}} & 
  \frame{\includegraphics[draft=\draft,width=0.075\linewidth, height=0.070\linewidth]{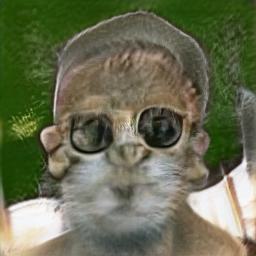}} & 
  \frame{\includegraphics[draft=\draft,width=0.075\linewidth, height=0.070\linewidth]{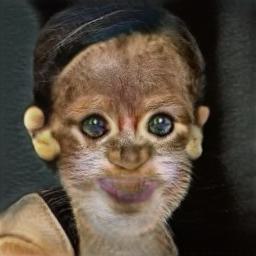}} & 
  \frame{\includegraphics[draft=\draft,width=0.075\linewidth, height=0.070\linewidth]{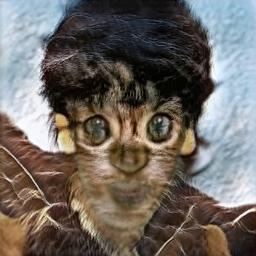}} & 
  \frame{\includegraphics[draft=\draft,width=0.075\linewidth, height=0.070\linewidth]{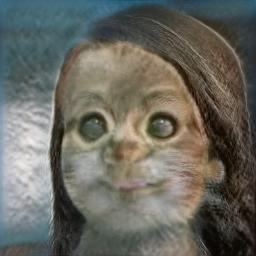}} & 
  \frame{\includegraphics[draft=\draft,width=0.075\linewidth, height=0.070\linewidth]{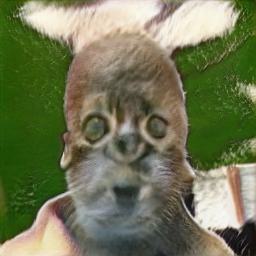}} & 
  \frame{\includegraphics[draft=\draft,width=0.075\linewidth, height=0.070\linewidth]{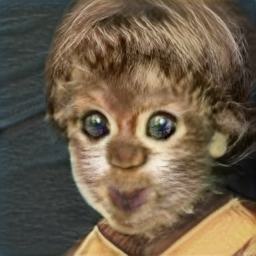}} & 
  \frame{\includegraphics[draft=\draft,width=0.075\linewidth, height=0.070\linewidth]{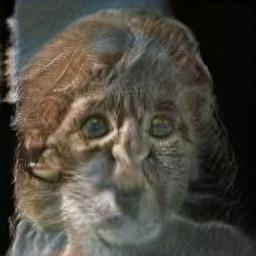}} & 
  \frame{\includegraphics[draft=\draft,width=0.075\linewidth, height=0.070\linewidth]{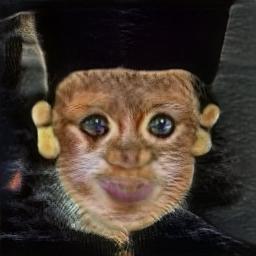}} &  ~~\rotatebox{270}{{\hspace{-3.2em} RSSA \hspace{-3.0em} } }
  \tabularnewline[-3pt]
  &
  \frame{\includegraphics[draft=\draft,width=0.075\linewidth, height=0.070\linewidth]{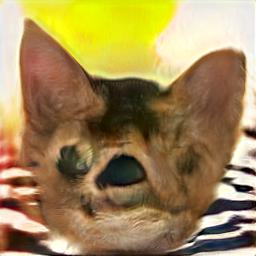}} & 
  \frame{\includegraphics[draft=\draft,width=0.075\linewidth, height=0.070\linewidth]{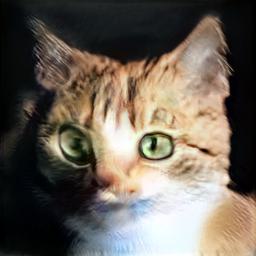}} & 
  \frame{\includegraphics[draft=\draft,width=0.075\linewidth, height=0.070\linewidth]{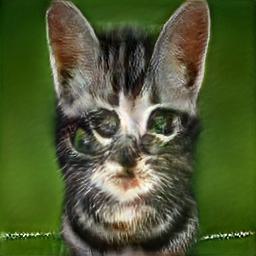}} & 
  \frame{\includegraphics[draft=\draft,width=0.075\linewidth, height=0.070\linewidth]{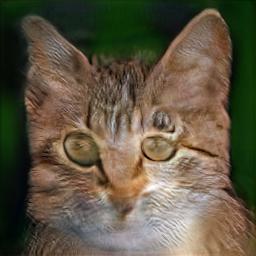}} & 
  \frame{\includegraphics[draft=\draft,width=0.075\linewidth, height=0.070\linewidth]{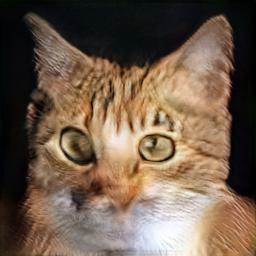}} & 
  \frame{\includegraphics[draft=\draft,width=0.075\linewidth, height=0.070\linewidth]{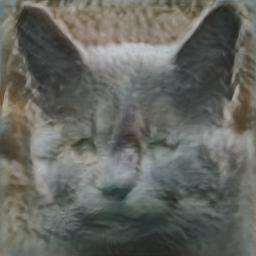}} & 
  \frame{\includegraphics[draft=\draft,width=0.075\linewidth, height=0.070\linewidth]{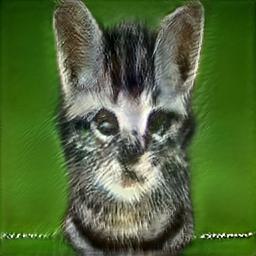}} & 
  \frame{\includegraphics[draft=\draft,width=0.075\linewidth, height=0.070\linewidth]{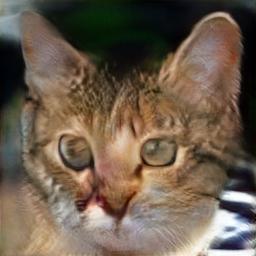}} & 
  \frame{\includegraphics[draft=\draft,width=0.075\linewidth, height=0.070\linewidth]{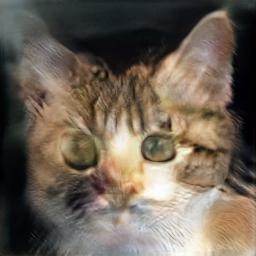}} & 
  \frame{\includegraphics[draft=\draft,width=0.075\linewidth, height=0.070\linewidth]{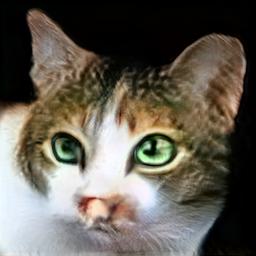}} &  ~~\rotatebox{270}{{\hspace{-3.4em} AdAM \hspace{-3.0em} } }
  \tabularnewline[-3pt]
  &
  \frame{\includegraphics[draft=\draft,width=0.075\linewidth, height=0.070\linewidth]{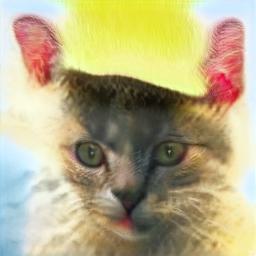}} & 
  \frame{\includegraphics[draft=\draft,width=0.075\linewidth, height=0.070\linewidth]{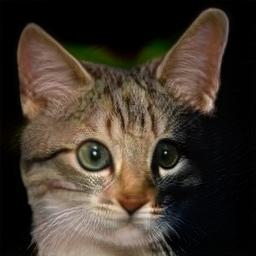}} & 
  \frame{\includegraphics[draft=\draft,width=0.075\linewidth, height=0.070\linewidth]{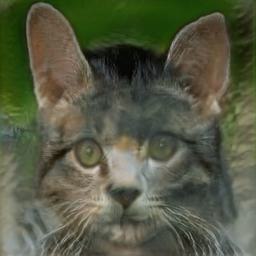}} & 
  \frame{\includegraphics[draft=\draft,width=0.075\linewidth, height=0.070\linewidth]{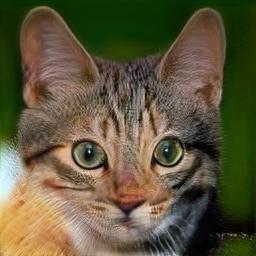}} & 
  \frame{\includegraphics[draft=\draft,width=0.075\linewidth, height=0.070\linewidth]{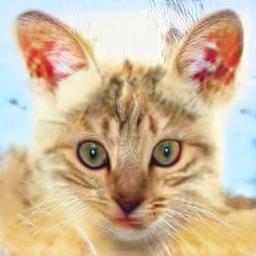}} & 
  \frame{\includegraphics[draft=\draft,width=0.075\linewidth, height=0.070\linewidth]{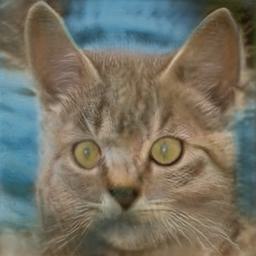}} & 
  \frame{\includegraphics[draft=\draft,width=0.075\linewidth, height=0.070\linewidth]{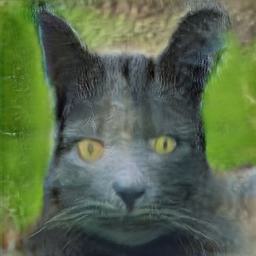}} & 
  \frame{\includegraphics[draft=\draft,width=0.075\linewidth, height=0.070\linewidth]{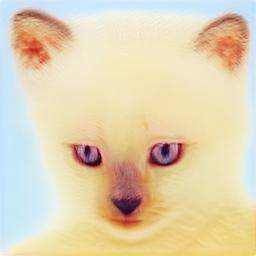}} & 
  \frame{\includegraphics[draft=\draft,width=0.075\linewidth, height=0.070\linewidth]{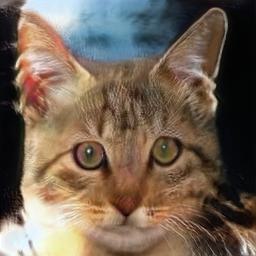}} & 
  \frame{\includegraphics[draft=\draft,width=0.075\linewidth, height=0.070\linewidth]{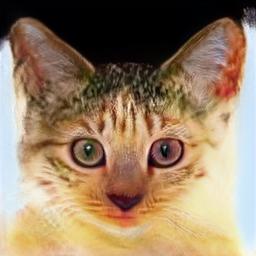}} & ~~\rotatebox{270}{{\hspace{-3.1em} \textbf{Ours} \hspace{-3.0em} } }
  \tabularnewline[-3pt]
  &
  \frame{\includegraphics[draft=\draft,width=0.075\linewidth, height=0.070\linewidth]{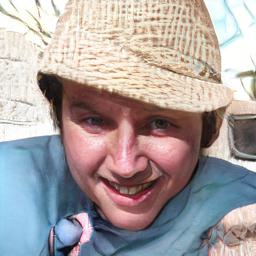}} & 
  \frame{\includegraphics[draft=\draft,width=0.075\linewidth, height=0.070\linewidth]{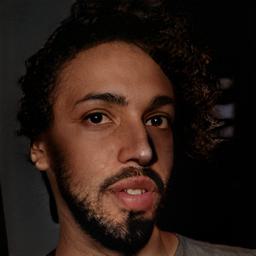}} & 
  \frame{\includegraphics[draft=\draft,width=0.075\linewidth, height=0.070\linewidth]{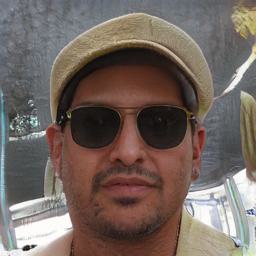}} & 
  \frame{\includegraphics[draft=\draft,width=0.075\linewidth, height=0.070\linewidth]{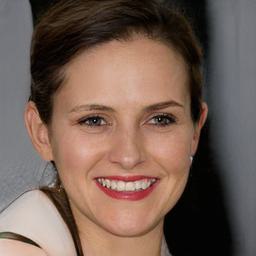}} & 
  \frame{\includegraphics[draft=\draft,width=0.075\linewidth, height=0.070\linewidth]{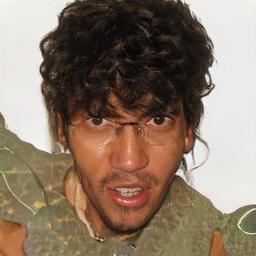}} & 
  \frame{\includegraphics[draft=\draft,width=0.075\linewidth, height=0.070\linewidth]{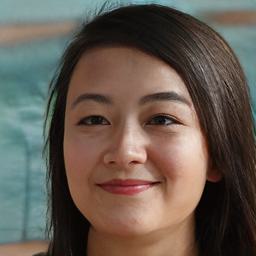}} & 
  \frame{\includegraphics[draft=\draft,width=0.075\linewidth, height=0.070\linewidth]{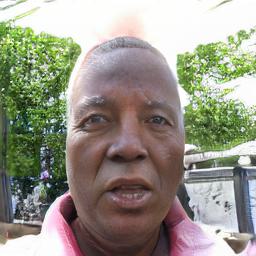}} & 
  \frame{\includegraphics[draft=\draft,width=0.075\linewidth, height=0.070\linewidth]{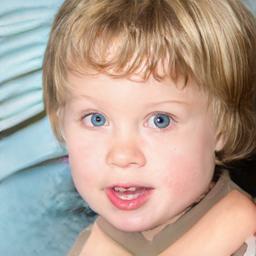}} & 
  \frame{\includegraphics[draft=\draft,width=0.075\linewidth, height=0.070\linewidth]{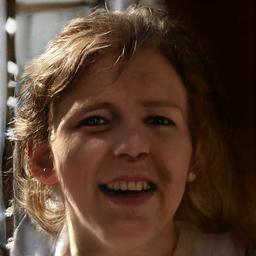}} & 
  \frame{\includegraphics[draft=\draft,width=0.075\linewidth, height=0.070\linewidth]{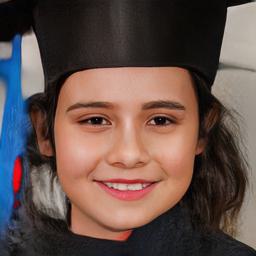}} & ~~\rotatebox{270}{{\hspace{-3.5em} Source 
  \hspace{-3.0em} } }
  \tabularnewline[-4pt]
\end{tabular}

\vspace{1.5ex}

\begin{tabular}{@{\hskip -0.02in}c@{\hskip 0.04in}:c@{\hskip 0.01in}c@{\hskip 0.01in}c@{\hskip 0.01in}c@{\hskip 0.01in}c@{\hskip 0.01in}c@{\hskip 0.01in}c@{\hskip 0.01in}c@{\hskip 0.01in}c@{\hskip 0.01in}c@{\hskip 0.01in}c@{\hskip 0.01in}c@{\hskip 0.01in}c}

  \multirow{-1}{*}{\begin{tabular}{c@{\hskip 0.01in}c@{\hskip 0.03in}} 
  \frame{\includegraphics[width=0.075\linewidth, height=0.070\linewidth]{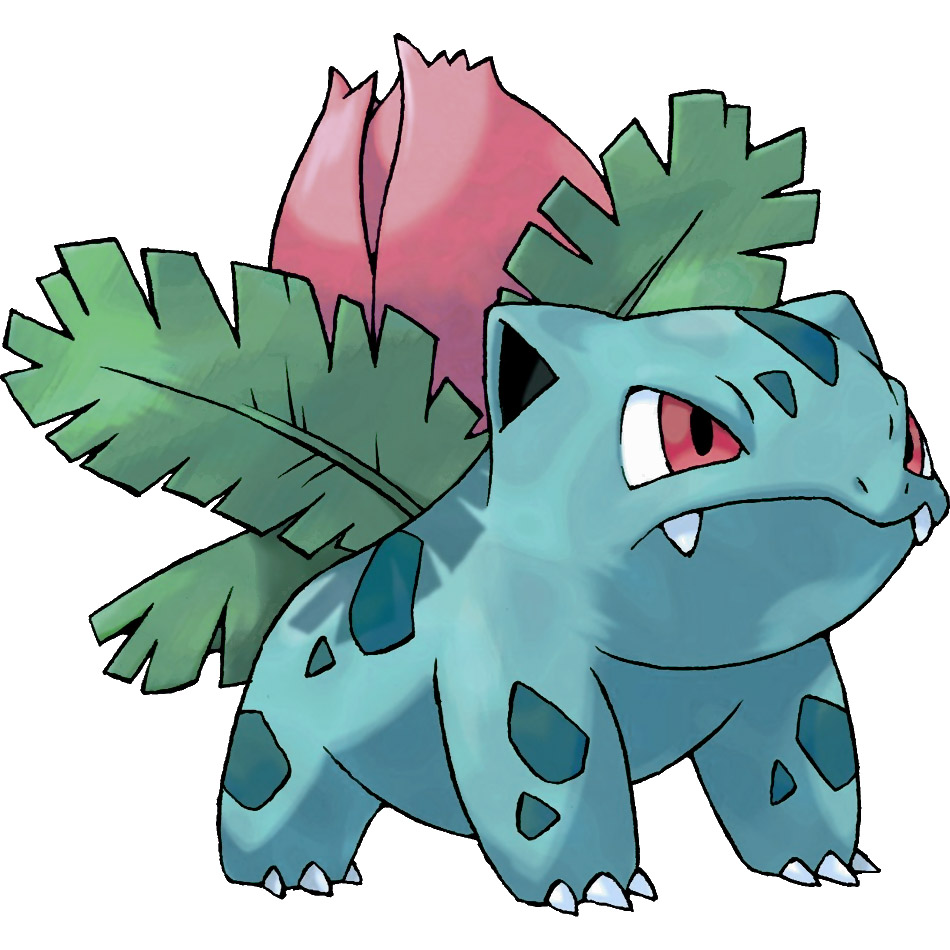}} &
  \frame{\includegraphics[width=0.075\linewidth, height=0.070\linewidth]{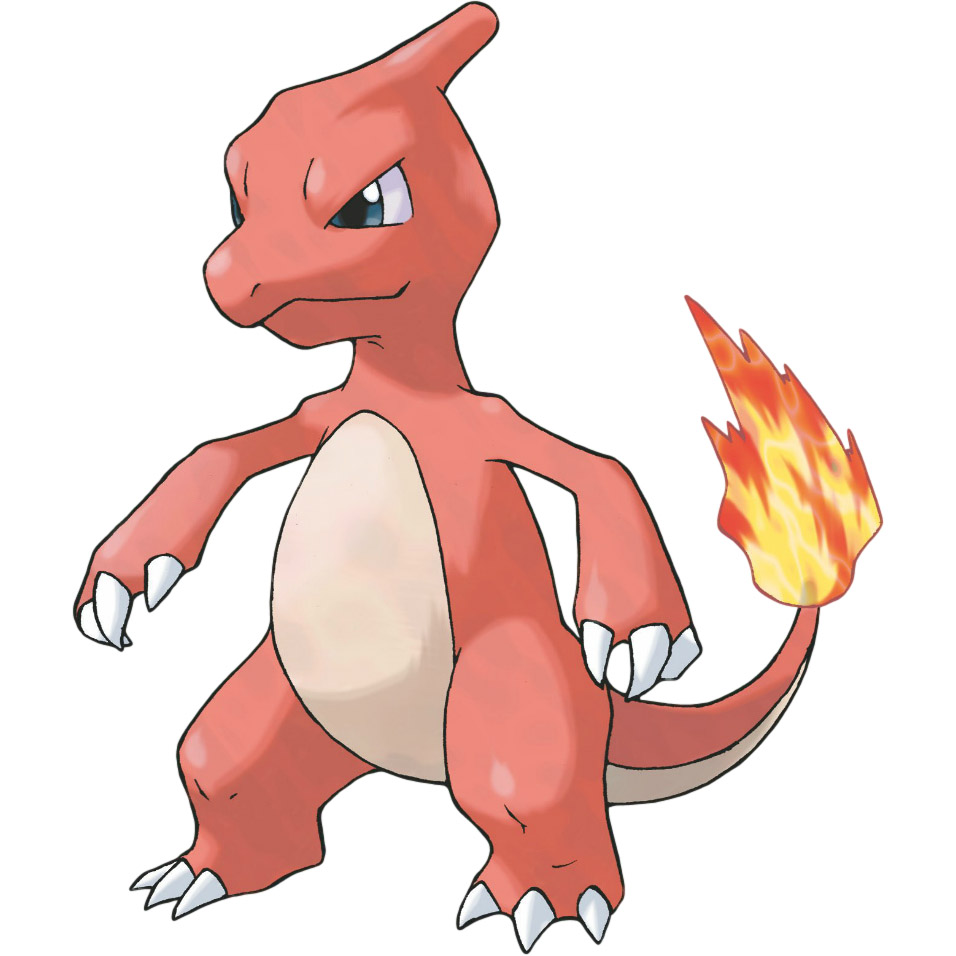}} \tabularnewline[-3pt]	
  \frame{\includegraphics[width=0.075\linewidth, height=0.070\linewidth]{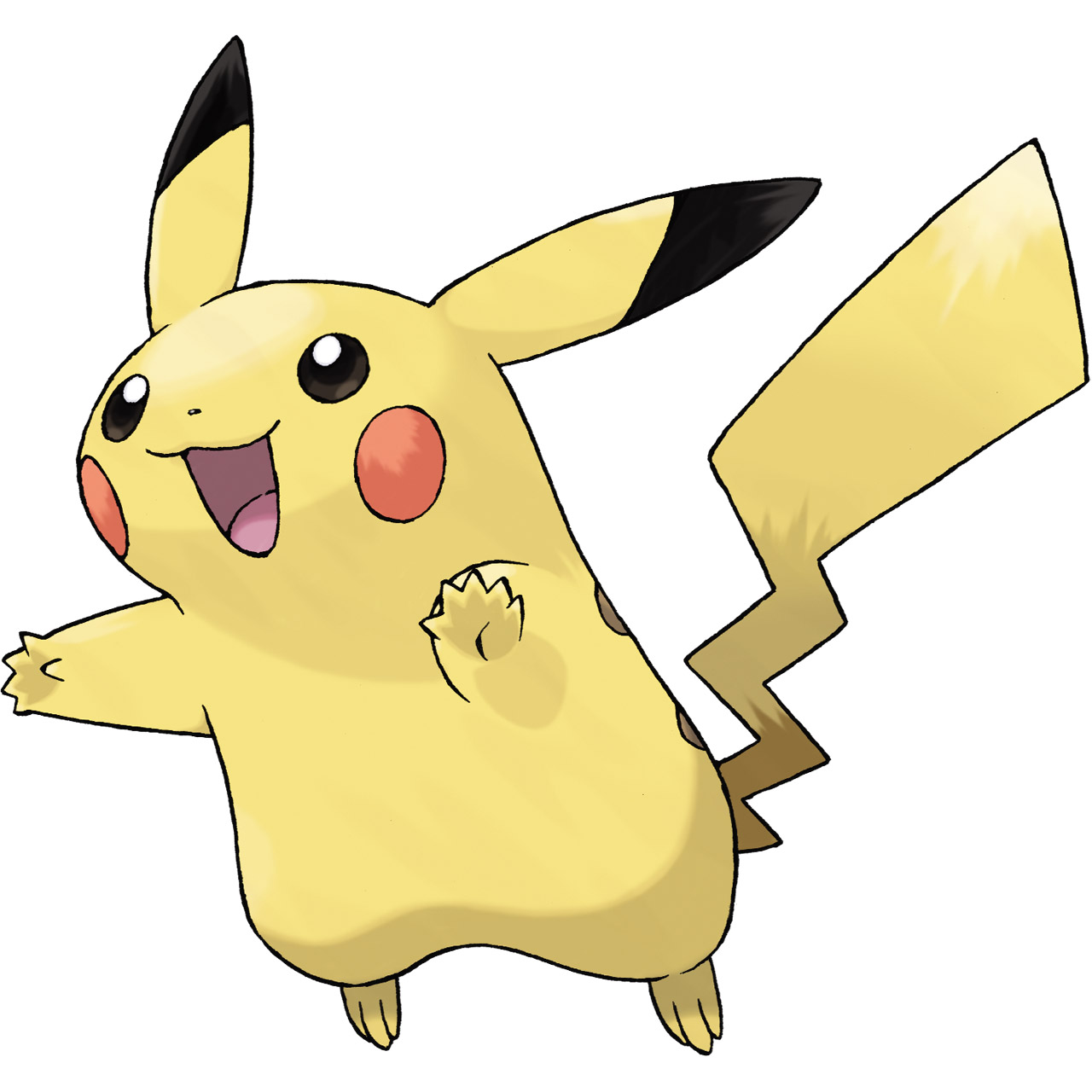}} &
  \frame{\includegraphics[width=0.075\linewidth, height=0.070\linewidth]{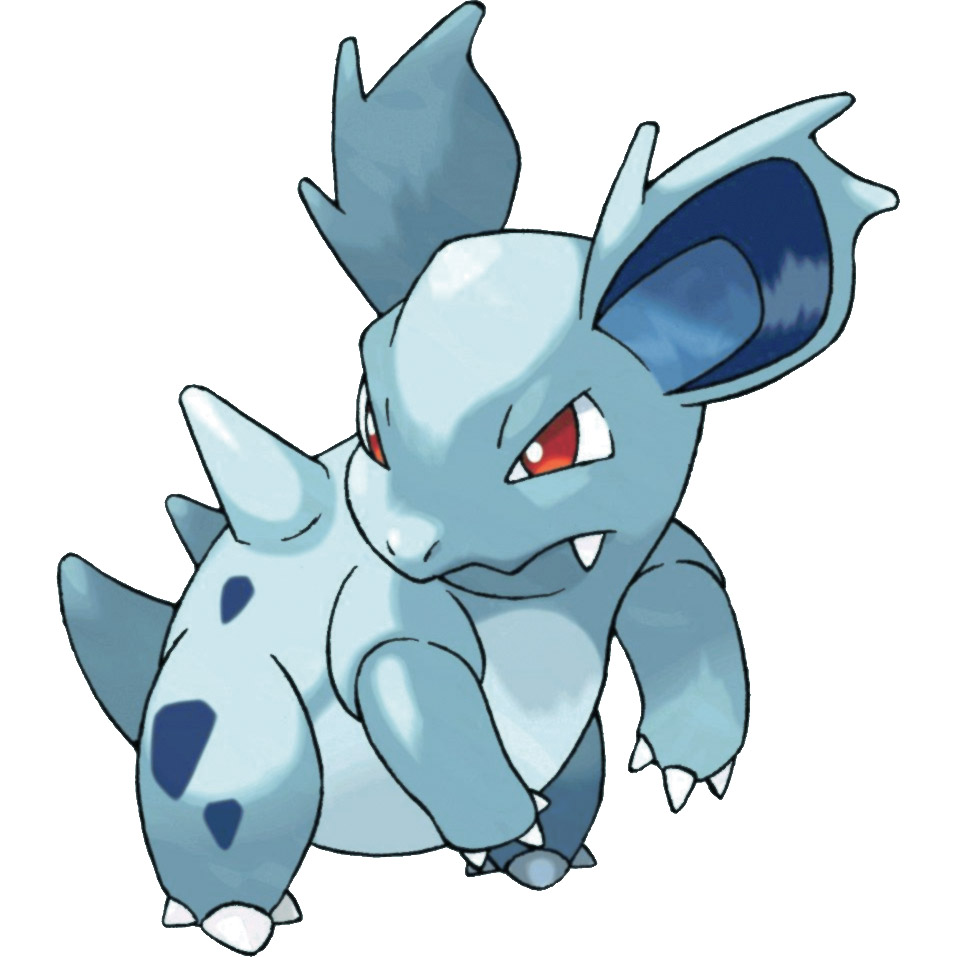}} \tabularnewline[-3pt]	
  \frame{\includegraphics[width=0.075\linewidth, height=0.070\linewidth]{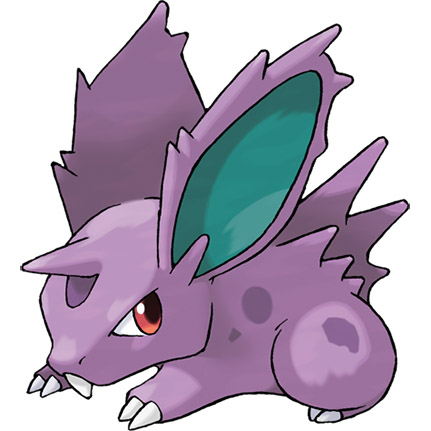}} &
  \frame{\includegraphics[width=0.075\linewidth, height=0.070\linewidth]{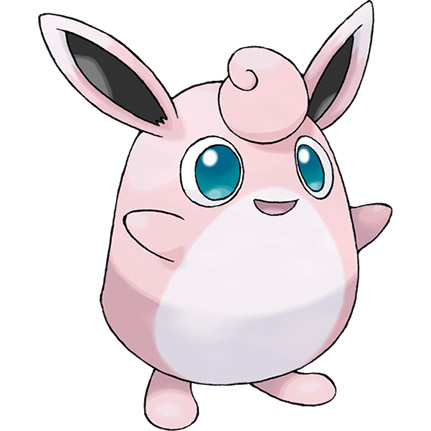}} \tabularnewline[-3pt]	
  \frame{\includegraphics[width=0.075\linewidth, height=0.070\linewidth]{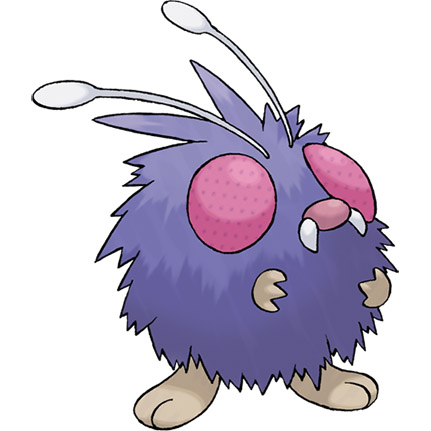}} &
  \frame{\includegraphics[width=0.075\linewidth, height=0.070\linewidth]{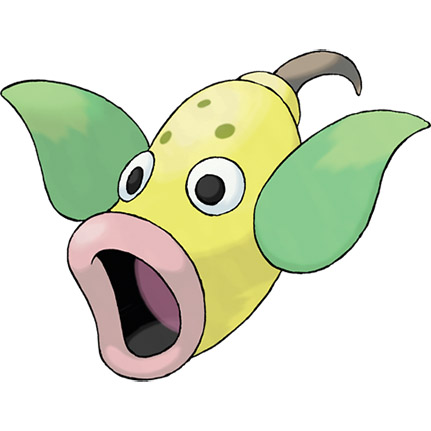}} \tabularnewline[-3pt]	
  \frame{\includegraphics[width=0.075\linewidth, height=0.070\linewidth]{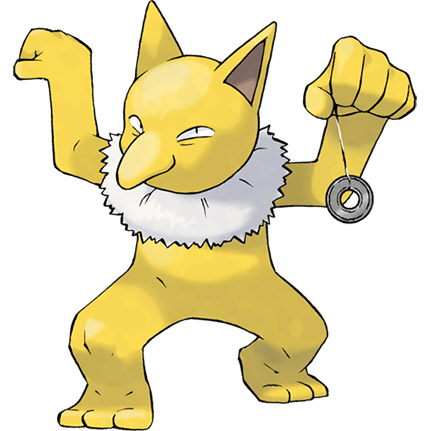}} &
  \frame{\includegraphics[width=0.075\linewidth, height=0.070\linewidth]{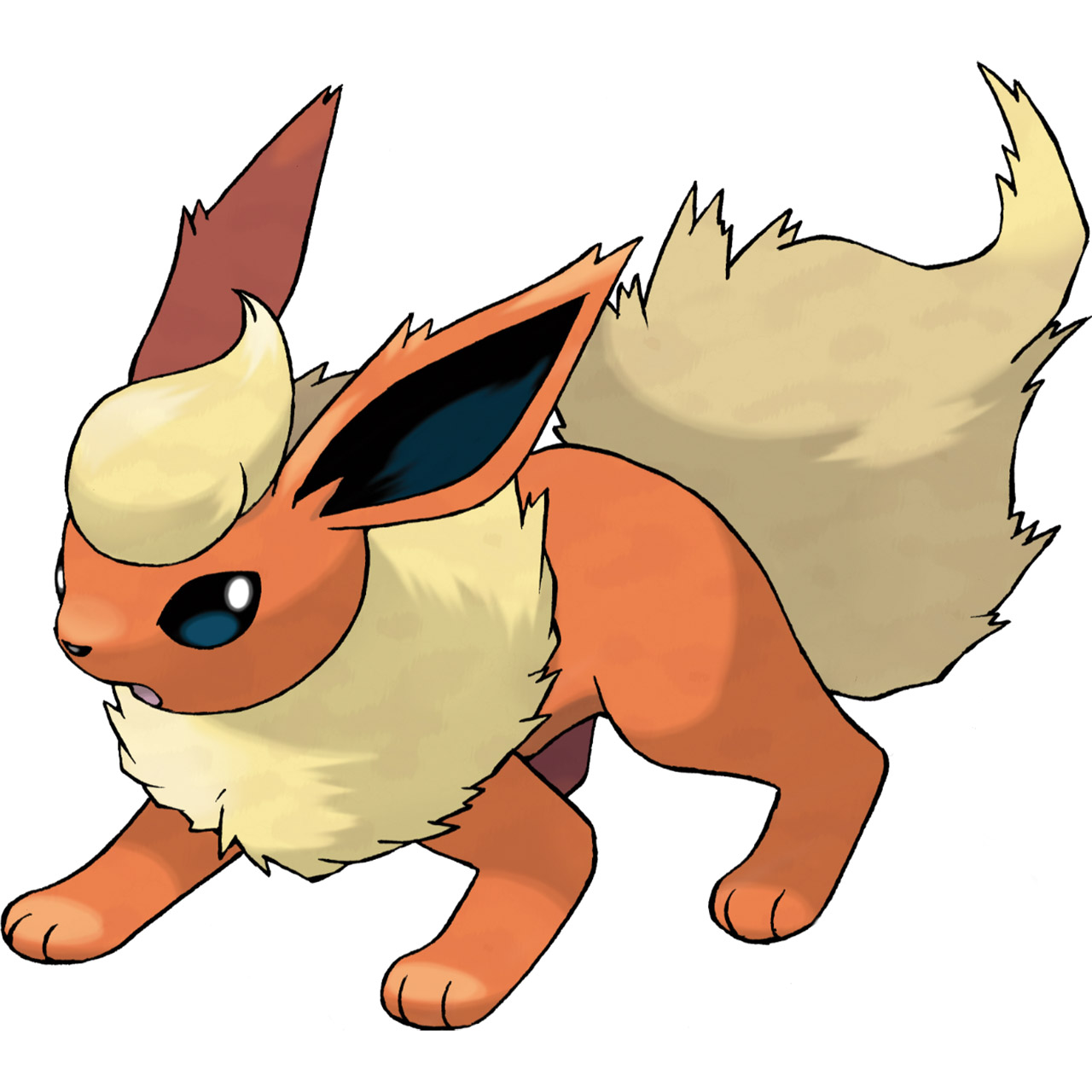}} \tabularnewline[+1pt]	
  \multicolumn{2}{@{\hskip -0.02in}c@{\hskip 0.03in}}{\small 10-shot} \tabularnewline[-2pt] 
  \multicolumn{2}{@{\hskip -0.02in}c@{\hskip 0.03in}}{\small Pokemons} \tabularnewline[-3pt]
  \end{tabular}} &
  
  \frame{\includegraphics[width=0.075\linewidth, height=0.070\linewidth]{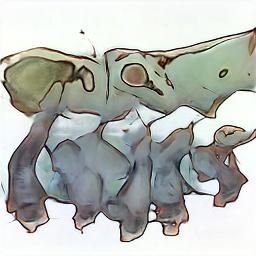}} & 
  \frame{\includegraphics[width=0.075\linewidth, height=0.070\linewidth]{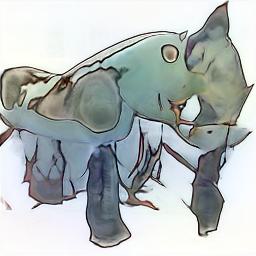}} & 
  \frame{\includegraphics[width=0.075\linewidth, height=0.070\linewidth]{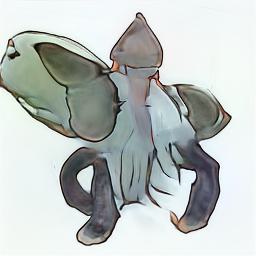}} & 
  \frame{\includegraphics[width=0.075\linewidth, height=0.070\linewidth]{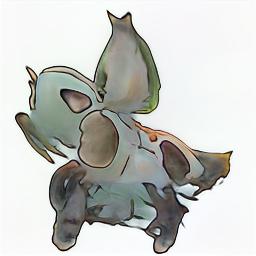}} & 
  \frame{\includegraphics[width=0.075\linewidth, height=0.070\linewidth]{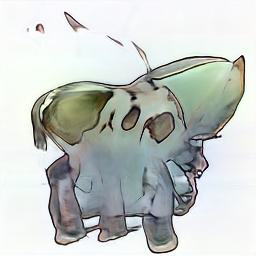}} & 
  \frame{\includegraphics[width=0.075\linewidth, height=0.070\linewidth]{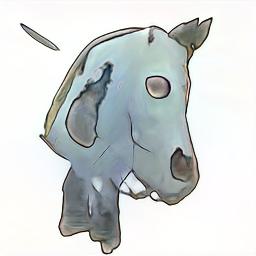}} & 
  \frame{\includegraphics[width=0.075\linewidth, height=0.070\linewidth]{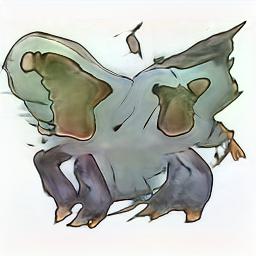}} & 
  \frame{\includegraphics[width=0.075\linewidth, height=0.070\linewidth]{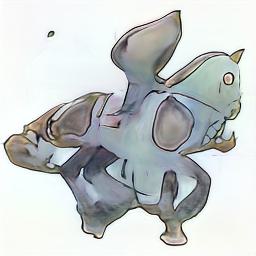}} & 
  \frame{\includegraphics[width=0.075\linewidth, height=0.070\linewidth]{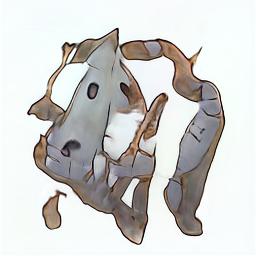}} & 
  \frame{\includegraphics[width=0.075\linewidth, height=0.070\linewidth]{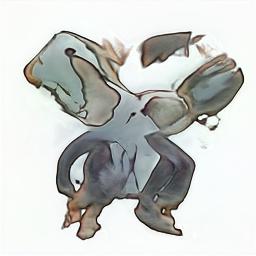}} & ~~\rotatebox{270}{{\hspace{-3.6em} TGAN \hspace{-3.0em} } }
  \tabularnewline[-3pt]
  &
  \frame{\includegraphics[width=0.075\linewidth, height=0.070\linewidth]{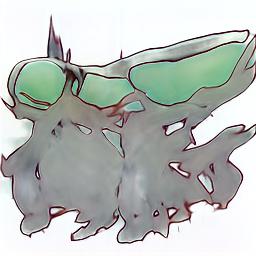}} & 
  \frame{\includegraphics[width=0.075\linewidth, height=0.070\linewidth]{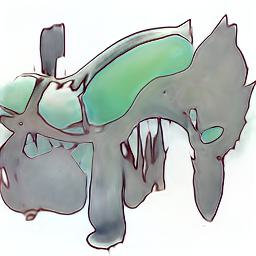}} & 
  \frame{\includegraphics[width=0.075\linewidth, height=0.070\linewidth]{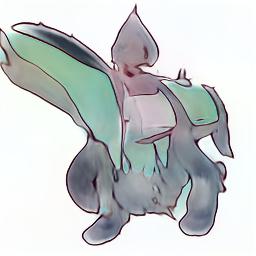}} & 
  \frame{\includegraphics[width=0.075\linewidth, height=0.070\linewidth]{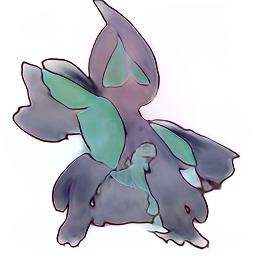}} & 
  \frame{\includegraphics[width=0.075\linewidth, height=0.070\linewidth]{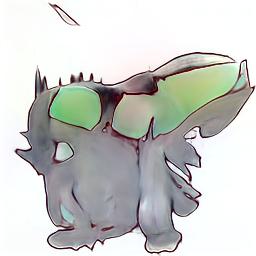}} & 
  \frame{\includegraphics[width=0.075\linewidth, height=0.070\linewidth]{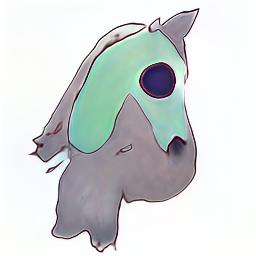}} & 
  \frame{\includegraphics[width=0.075\linewidth, height=0.070\linewidth]{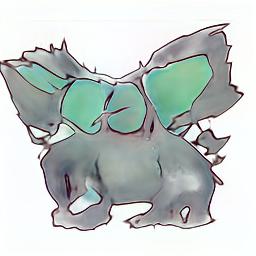}} & 
  \frame{\includegraphics[width=0.075\linewidth, height=0.070\linewidth]{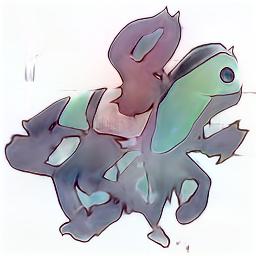}} & 
  \frame{\includegraphics[width=0.075\linewidth, height=0.070\linewidth]{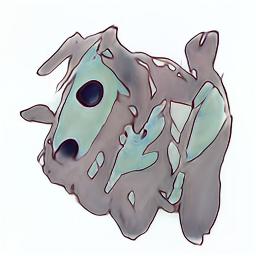}} & 
  \frame{\includegraphics[width=0.075\linewidth, height=0.070\linewidth]{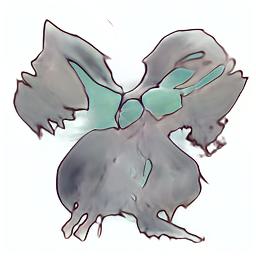}} & ~~\rotatebox{270}{{\hspace{-3.7em} FreezeD \hspace{-3.0em} } }
  \tabularnewline[-3pt]
  &
  \frame{\includegraphics[width=0.075\linewidth, height=0.070\linewidth]{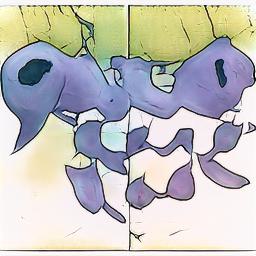}} & 
  \frame{\includegraphics[width=0.075\linewidth, height=0.070\linewidth]{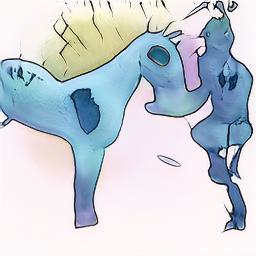}} & 
  \frame{\includegraphics[width=0.075\linewidth, height=0.070\linewidth]{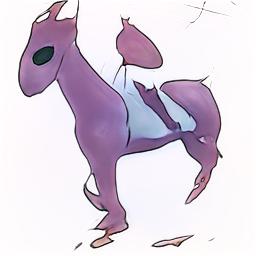}} & 
  \frame{\includegraphics[width=0.075\linewidth, height=0.070\linewidth]{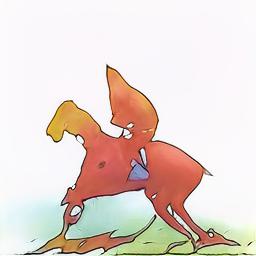}} & 
  \frame{\includegraphics[width=0.075\linewidth, height=0.070\linewidth]{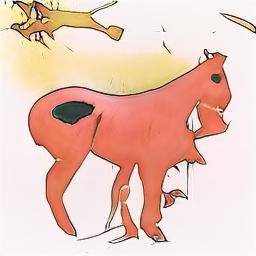}} & 
  \frame{\includegraphics[width=0.075\linewidth, height=0.070\linewidth]{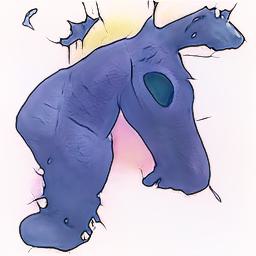}} & 
  \frame{\includegraphics[width=0.075\linewidth, height=0.070\linewidth]{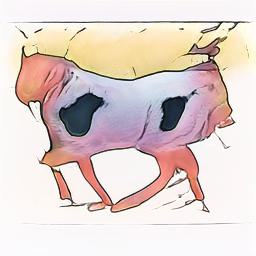}} & 
  \frame{\includegraphics[width=0.075\linewidth, height=0.070\linewidth]{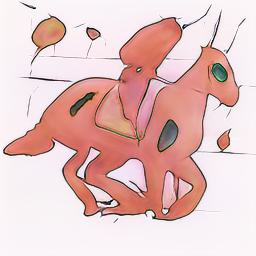}} & 
  \frame{\includegraphics[width=0.075\linewidth, height=0.070\linewidth]{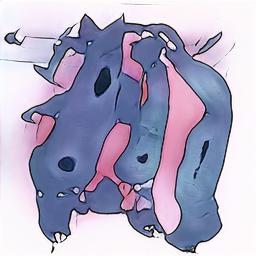}} & 
  \frame{\includegraphics[width=0.075\linewidth, height=0.070\linewidth]{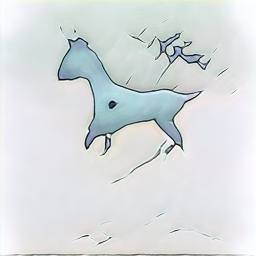}} & ~~\rotatebox{270}{{\hspace{-3.0em} CDC \hspace{-3.0em} } }
  \tabularnewline[-3pt]
  &
  \frame{\includegraphics[width=0.075\linewidth, height=0.070\linewidth]{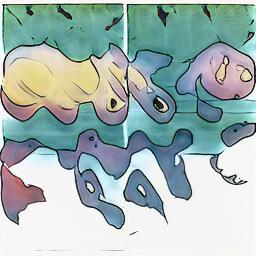}} & 
  \frame{\includegraphics[width=0.075\linewidth, height=0.070\linewidth]{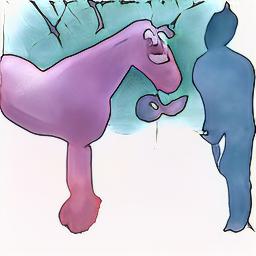}} & 
  \frame{\includegraphics[width=0.075\linewidth, height=0.070\linewidth]{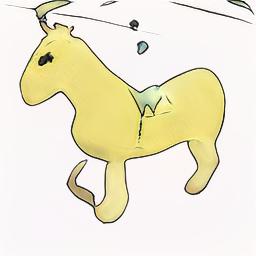}} & 
  \frame{\includegraphics[width=0.075\linewidth, height=0.070\linewidth]{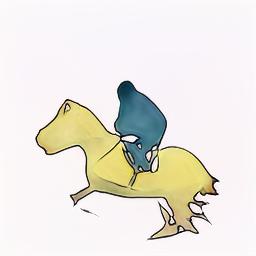}} & 
  \frame{\includegraphics[width=0.075\linewidth, height=0.070\linewidth]{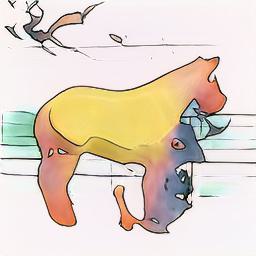}} & 
  \frame{\includegraphics[width=0.075\linewidth, height=0.070\linewidth]{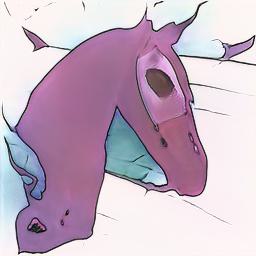}} & 
  \frame{\includegraphics[width=0.075\linewidth, height=0.070\linewidth]{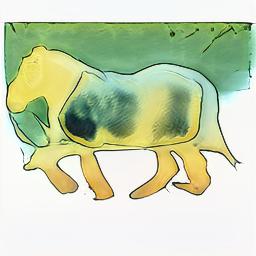}} & 
  \frame{\includegraphics[width=0.075\linewidth, height=0.070\linewidth]{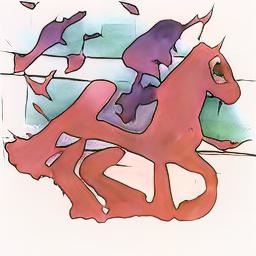}} & 
  \frame{\includegraphics[width=0.075\linewidth, height=0.070\linewidth]{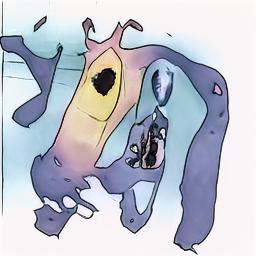}} & 
  \frame{\includegraphics[width=0.075\linewidth, height=0.070\linewidth]{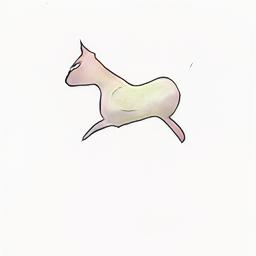}} & ~~\rotatebox{270}{{\hspace{-3.2em} RSSA \hspace{-3.0em} } }
  \tabularnewline[-3pt]
  &
  \frame{\includegraphics[width=0.075\linewidth, height=0.070\linewidth]{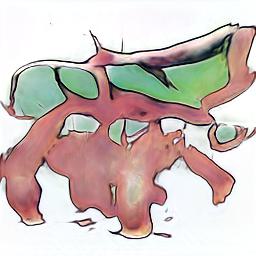}} & 
  \frame{\includegraphics[width=0.075\linewidth, height=0.070\linewidth]{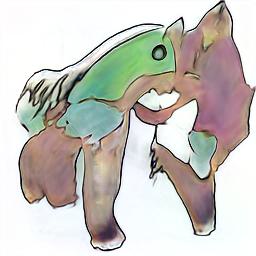}} & 
  \frame{\includegraphics[width=0.075\linewidth, height=0.070\linewidth]{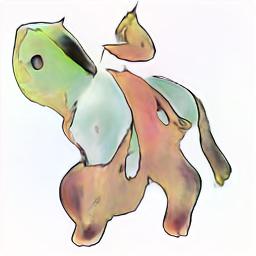}} & 
  \frame{\includegraphics[width=0.075\linewidth, height=0.070\linewidth]{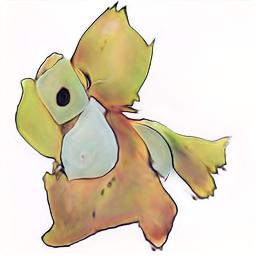}} & 
  \frame{\includegraphics[width=0.075\linewidth, height=0.070\linewidth]{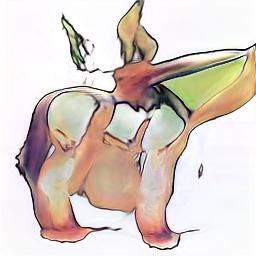}} & 
  \frame{\includegraphics[width=0.075\linewidth, height=0.070\linewidth]{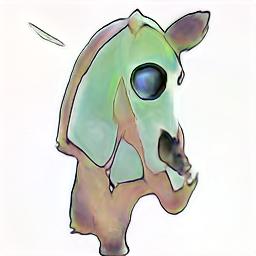}} & 
  \frame{\includegraphics[width=0.075\linewidth, height=0.070\linewidth]{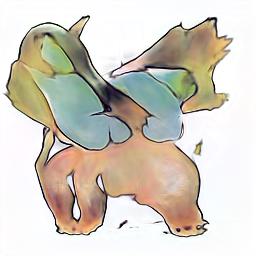}} & 
  \frame{\includegraphics[width=0.075\linewidth, height=0.070\linewidth]{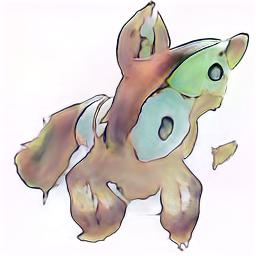}} & 
  \frame{\includegraphics[width=0.075\linewidth, height=0.070\linewidth]{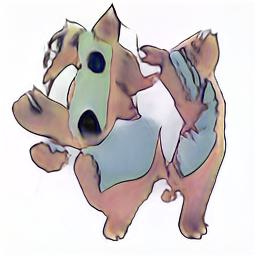}} & 
  \frame{\includegraphics[width=0.075\linewidth, height=0.070\linewidth]{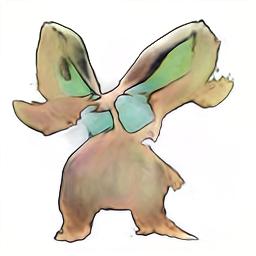}} & ~~\rotatebox{270}{{\hspace{-3.4em} AdAM \hspace{-3.0em} } }
  \tabularnewline[-3pt]
  &
  \frame{\includegraphics[width=0.075\linewidth, height=0.070\linewidth]{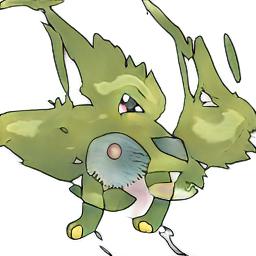}} & 
  \frame{\includegraphics[width=0.075\linewidth, height=0.070\linewidth]{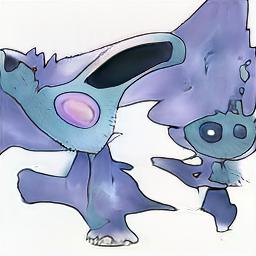}} & 
  \frame{\includegraphics[width=0.075\linewidth, height=0.070\linewidth]{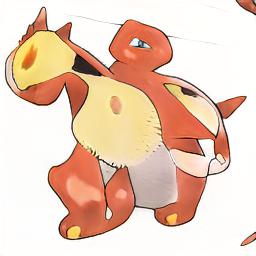}} & 
  \frame{\includegraphics[width=0.075\linewidth, height=0.070\linewidth]{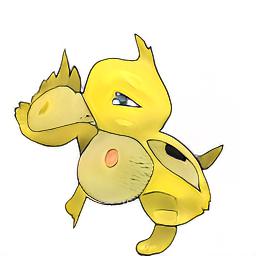}} & 
  \frame{\includegraphics[width=0.075\linewidth, height=0.070\linewidth]{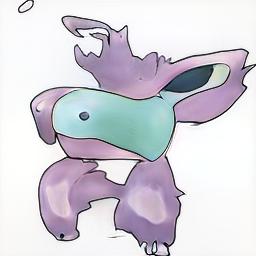}} & 
  \frame{\includegraphics[width=0.075\linewidth, height=0.070\linewidth]{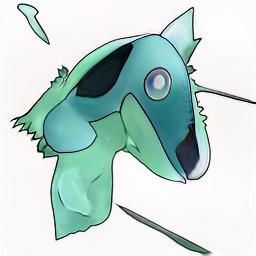}} & 
  \frame{\includegraphics[width=0.075\linewidth, height=0.070\linewidth]{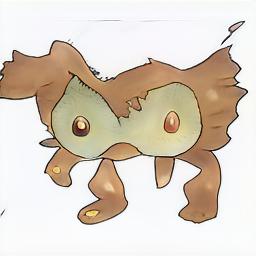}} & 
  \frame{\includegraphics[width=0.075\linewidth, height=0.070\linewidth]{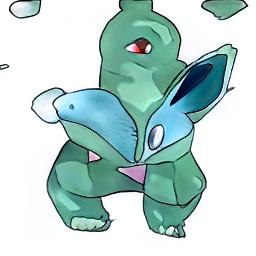}} & 
  \frame{\includegraphics[width=0.075\linewidth, height=0.070\linewidth]{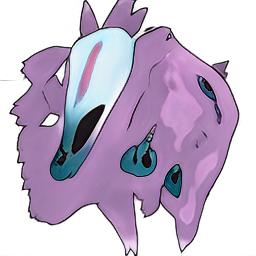}} & 
  \frame{\includegraphics[width=0.075\linewidth, height=0.070\linewidth]{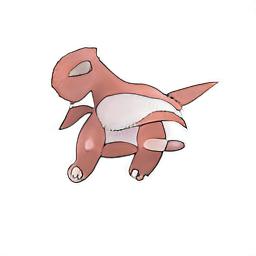}} & ~~\rotatebox{270}{{\hspace{-3.1em} \textbf{Ours} \hspace{-3.0em} } }
  \tabularnewline[-3pt]
  &
  \frame{\includegraphics[width=0.075\linewidth, height=0.070\linewidth]{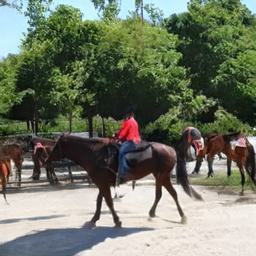}} & 
  \frame{\includegraphics[width=0.075\linewidth, height=0.070\linewidth]{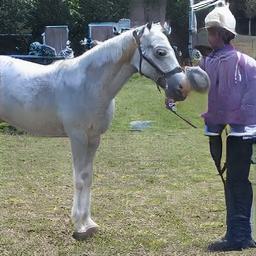}} & 
  \frame{\includegraphics[width=0.075\linewidth, height=0.070\linewidth]{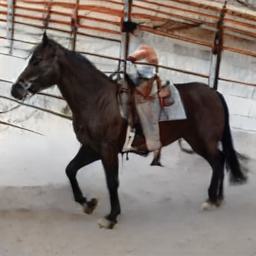}} & 
  \frame{\includegraphics[width=0.075\linewidth, height=0.070\linewidth]{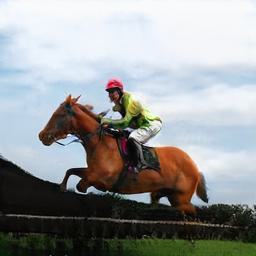}} & 
  \frame{\includegraphics[width=0.075\linewidth, height=0.070\linewidth]{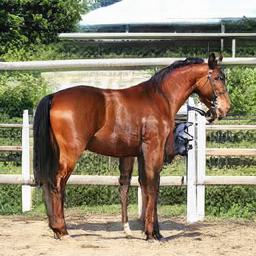}} & 
  \frame{\includegraphics[width=0.075\linewidth, height=0.070\linewidth]{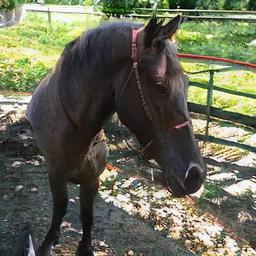}} & 
  \frame{\includegraphics[width=0.075\linewidth, height=0.070\linewidth]{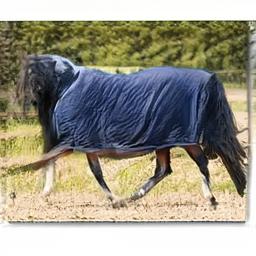}} & 
  \frame{\includegraphics[width=0.075\linewidth, height=0.070\linewidth]{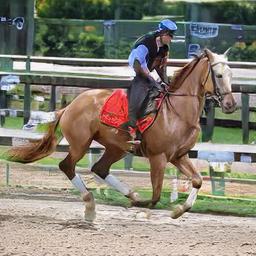}} & 
  \frame{\includegraphics[width=0.075\linewidth, height=0.070\linewidth]{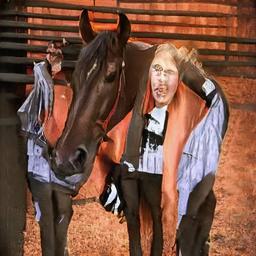}} & 
  \frame{\includegraphics[width=0.075\linewidth, height=0.070\linewidth]{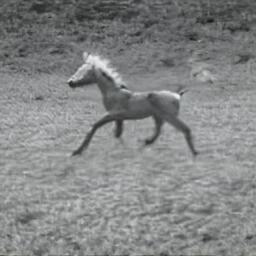}} & ~~\rotatebox{270}{{\hspace{-3.5em} Source 
  \hspace{-3.0em} } }
  \tabularnewline[-4pt]
 
\end{tabular}

\par\end{centering}
\vspace{5pt}
\caption{Additional visual comparison to prior methods on \textit{Face$\rightarrow$Cats} and \textit{Horses$\rightarrow$Pokemons}, the source-target dataset pairs with a dissimilar structure (e.g., shapes of objects). 
}
\label{fig:qualitative_comparison_distant3}
\end{figure*}

%% file: figures/qual_comparison_close2.tex
\begin{figure*}[t]
\begin{centering}
\setlength{\tabcolsep}{0.01in}
\renewcommand{\arraystretch}{1}
\par\end{centering}
\begin{centering}

\begin{tabular}{@{\hskip -0.13in}c@{\hskip 0.10in}c@{\hskip 0.01in}c@{\hskip 0.01in}c@{\hskip 0.01in}c@{\hskip 0.01in}c@{\hskip 0.01in}c@{\hskip 0.01in}c:c@{\hskip 0.01in}c@{\hskip 0.01in}c@{\hskip 0.01in}c@{\hskip 0.01in}c@{\hskip 0.01in}c@{\hskip 0.01in}c}

~~\rotatebox{90}{{\hspace{0.0em} Real \hspace{0.0em} }} &

\multicolumn{7}{@{\hskip -0.04in}c:}{
\frame{\includegraphics[width=0.0465\linewidth, height=0.0465\linewidth]{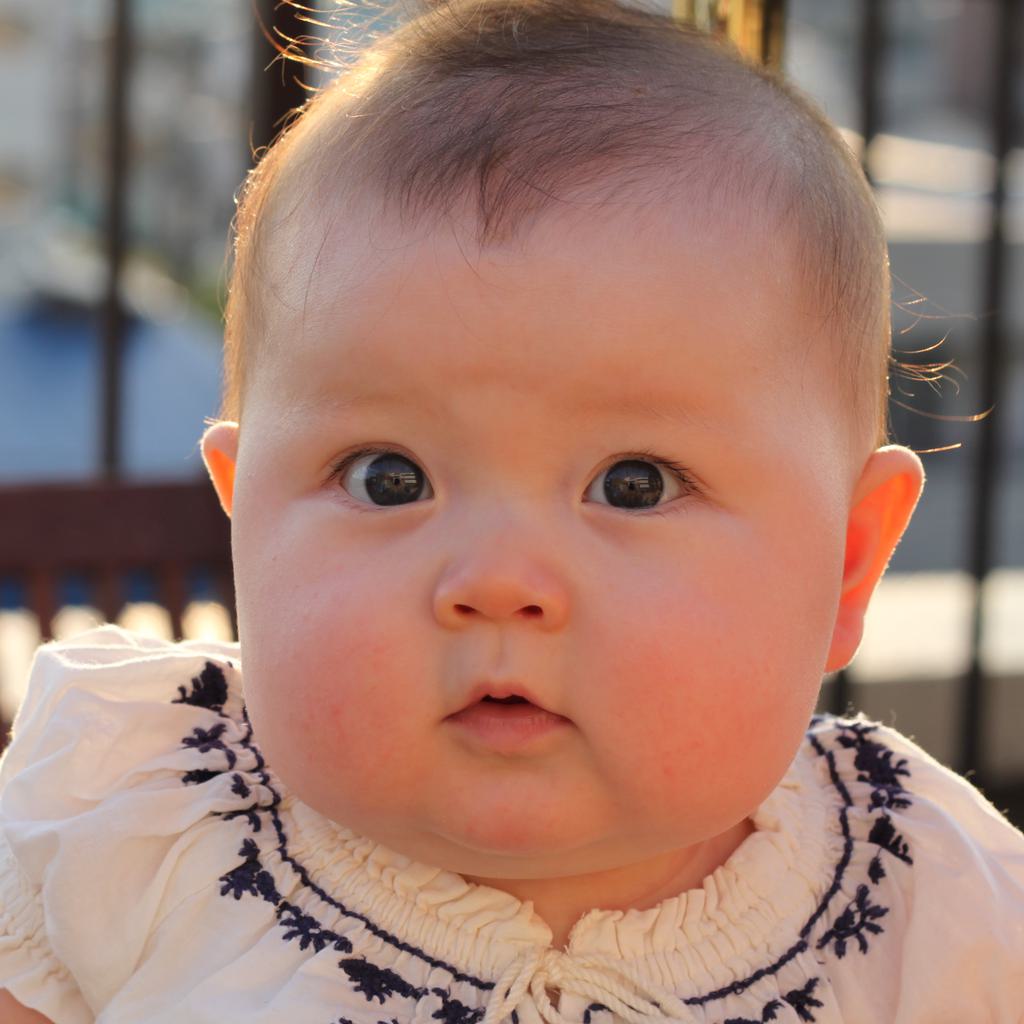}} \hspace{-1.0ex}
\frame{\includegraphics[width=0.0465\linewidth, height=0.0465\linewidth]{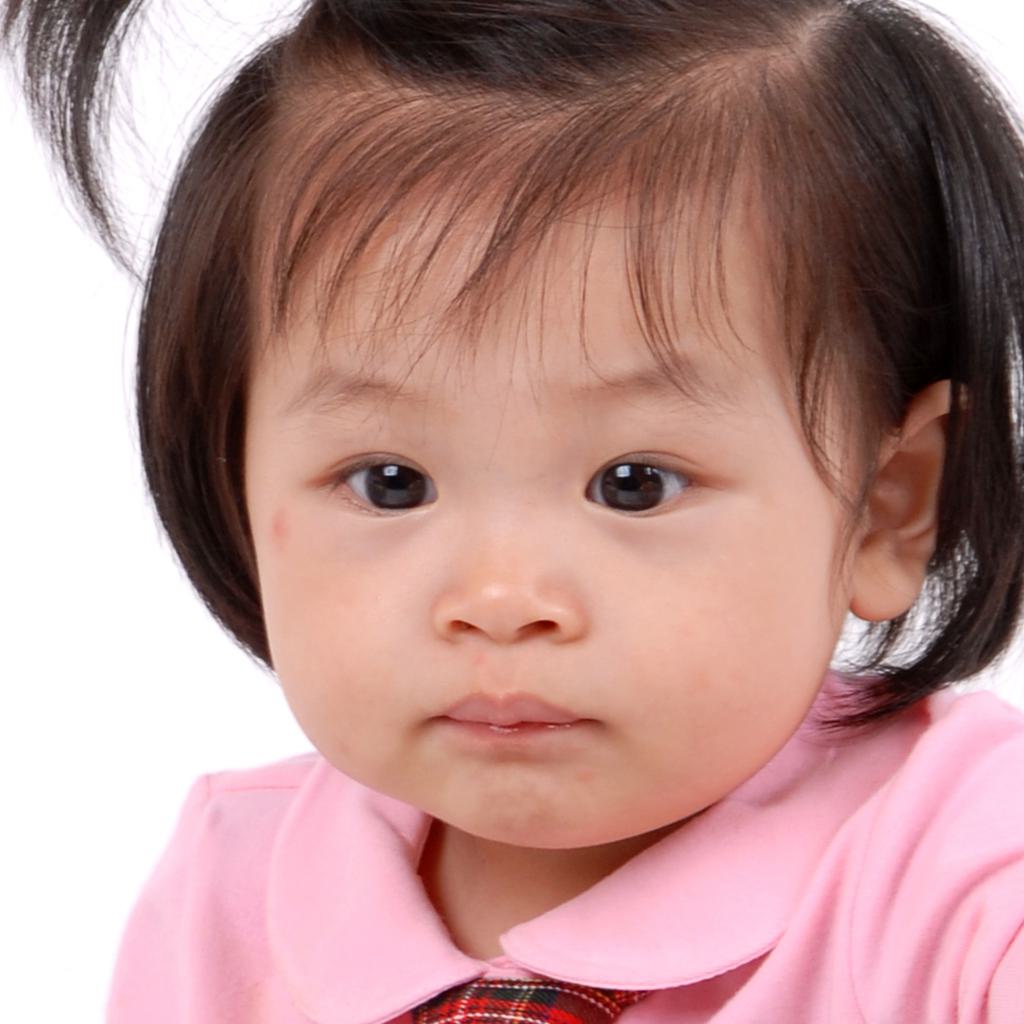}} \hspace{-1.0ex}
\frame{\includegraphics[width=0.0465\linewidth, height=0.0465\linewidth]{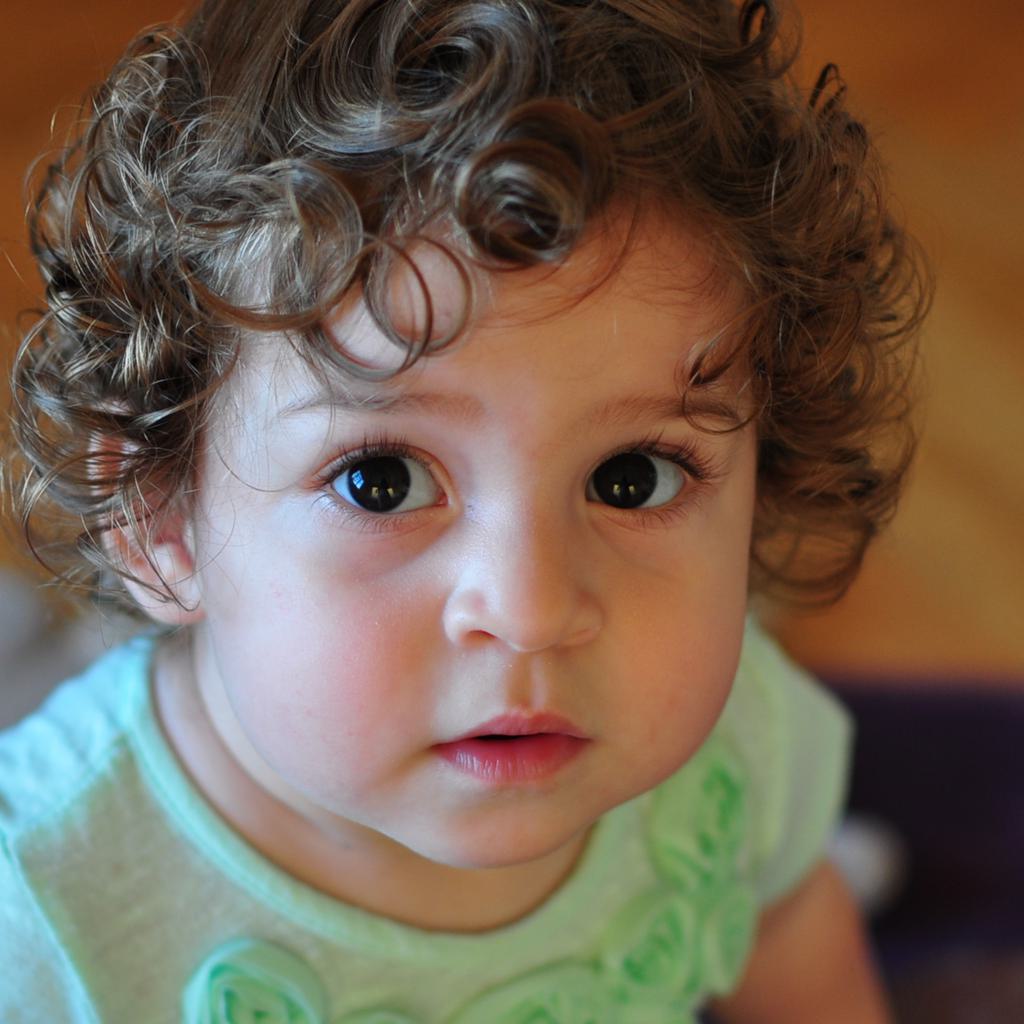}} \hspace{-1.0ex}
\frame{\includegraphics[width=0.0465\linewidth, height=0.0465\linewidth]{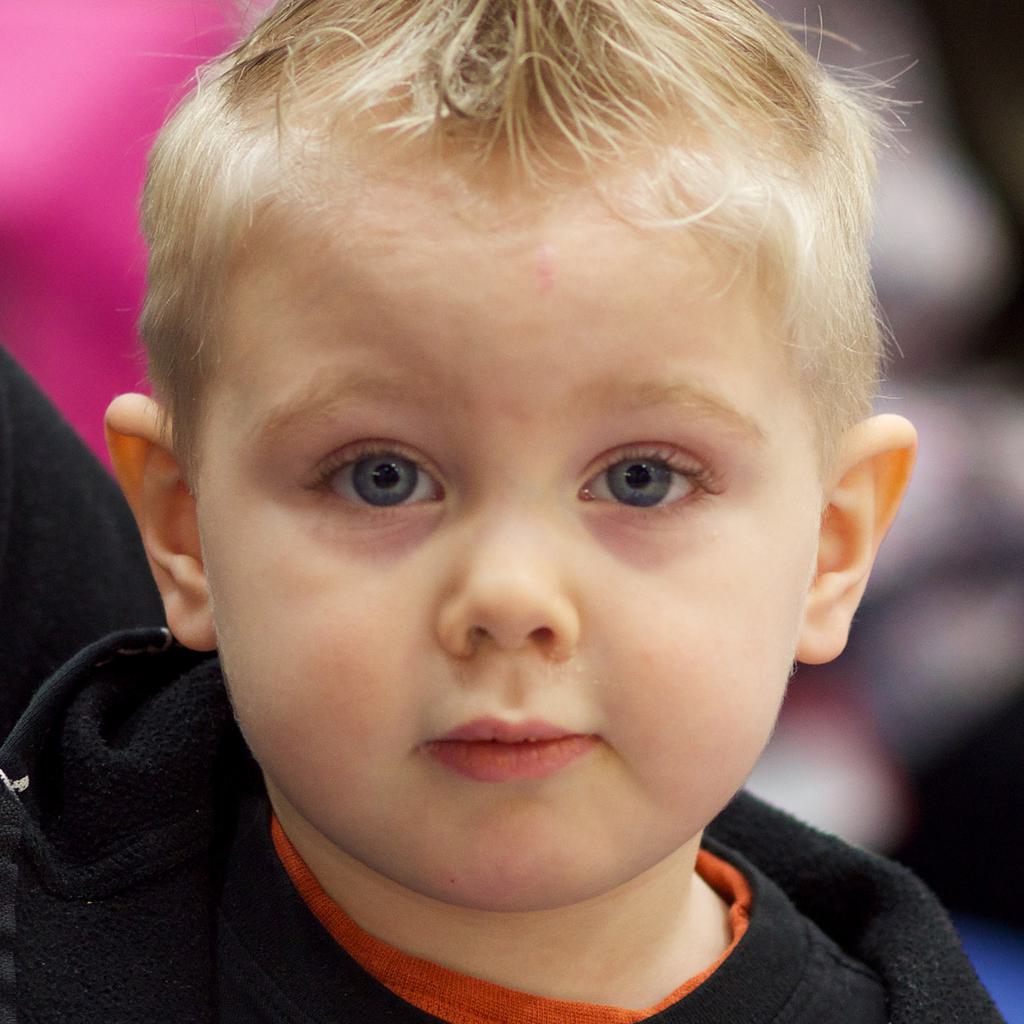}} \hspace{-1.0ex}
\frame{\includegraphics[width=0.0465\linewidth, height=0.0465\linewidth]{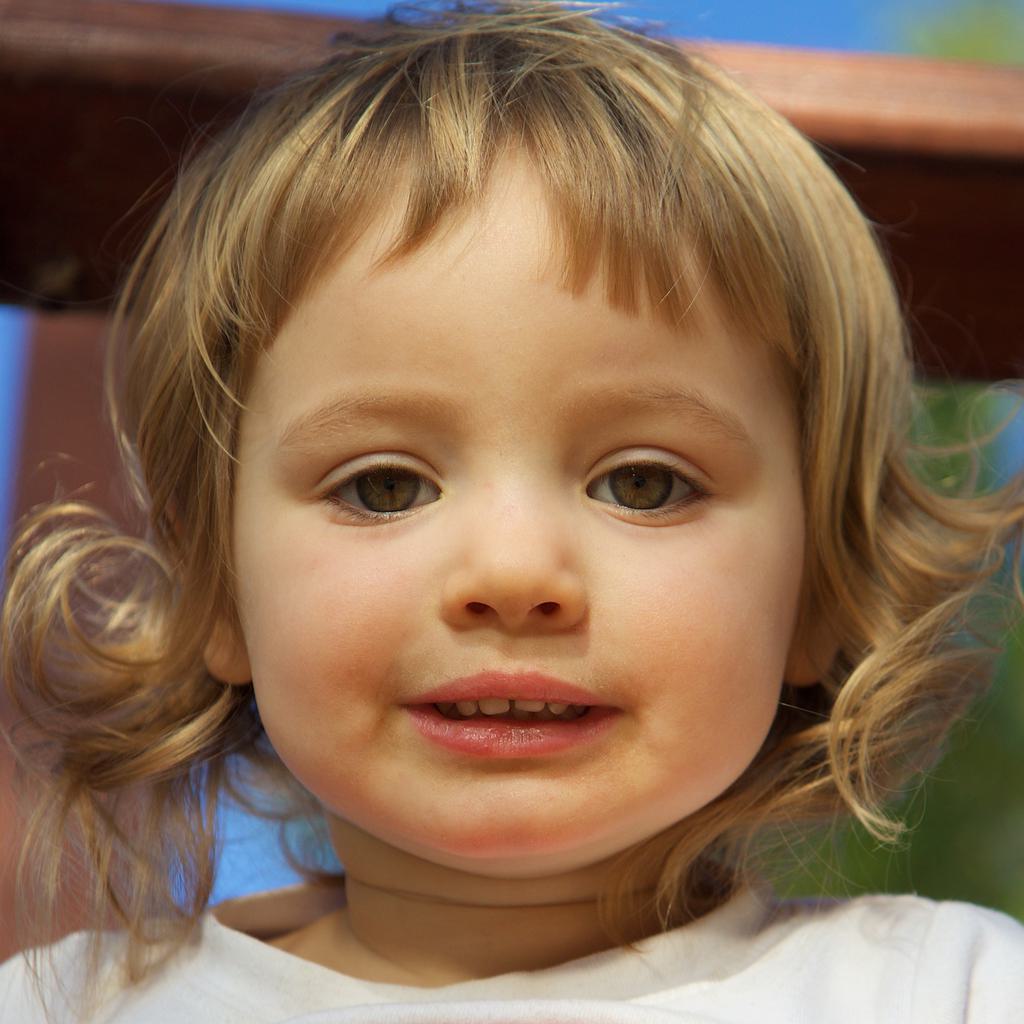}} \hspace{-1.0ex}
\frame{\includegraphics[width=0.0465\linewidth, height=0.0465\linewidth]{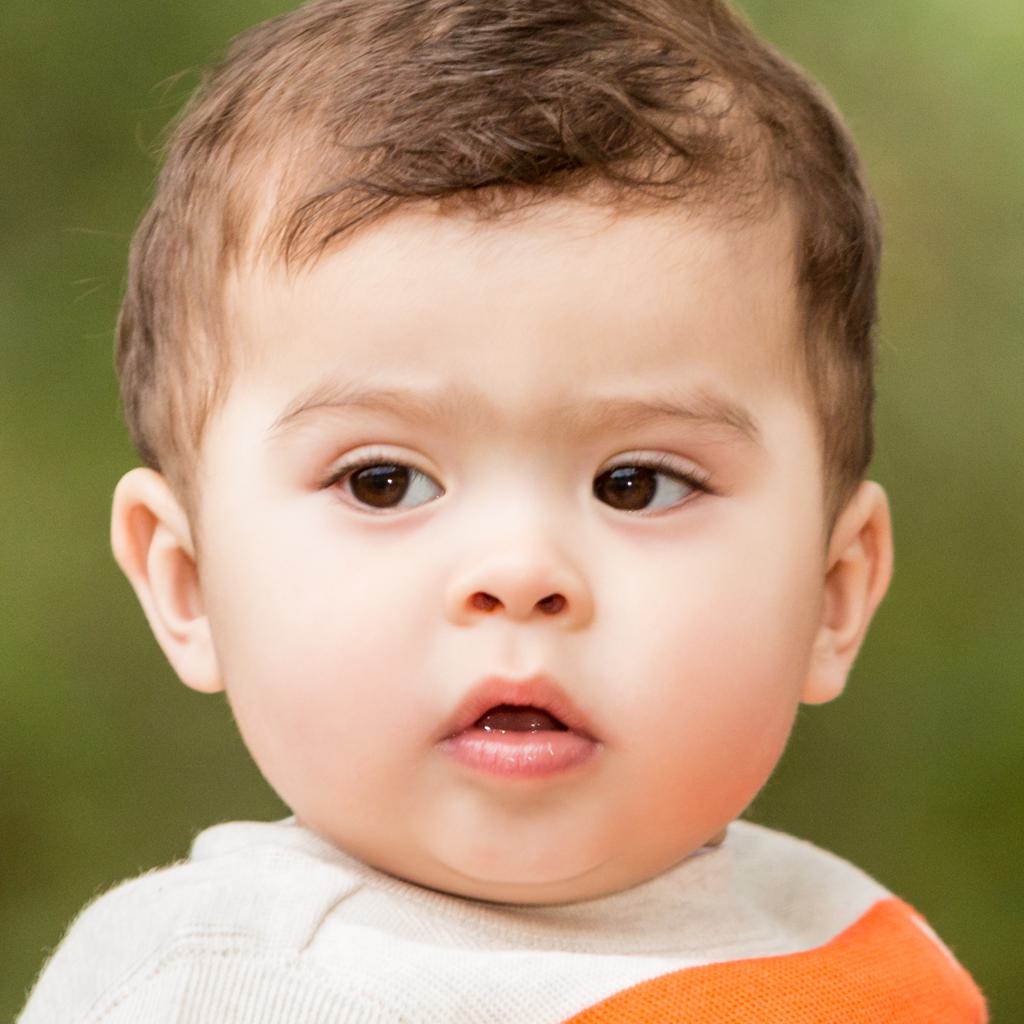}} \hspace{-1.0ex}
\frame{\includegraphics[width=0.0465\linewidth, height=0.0465\linewidth]{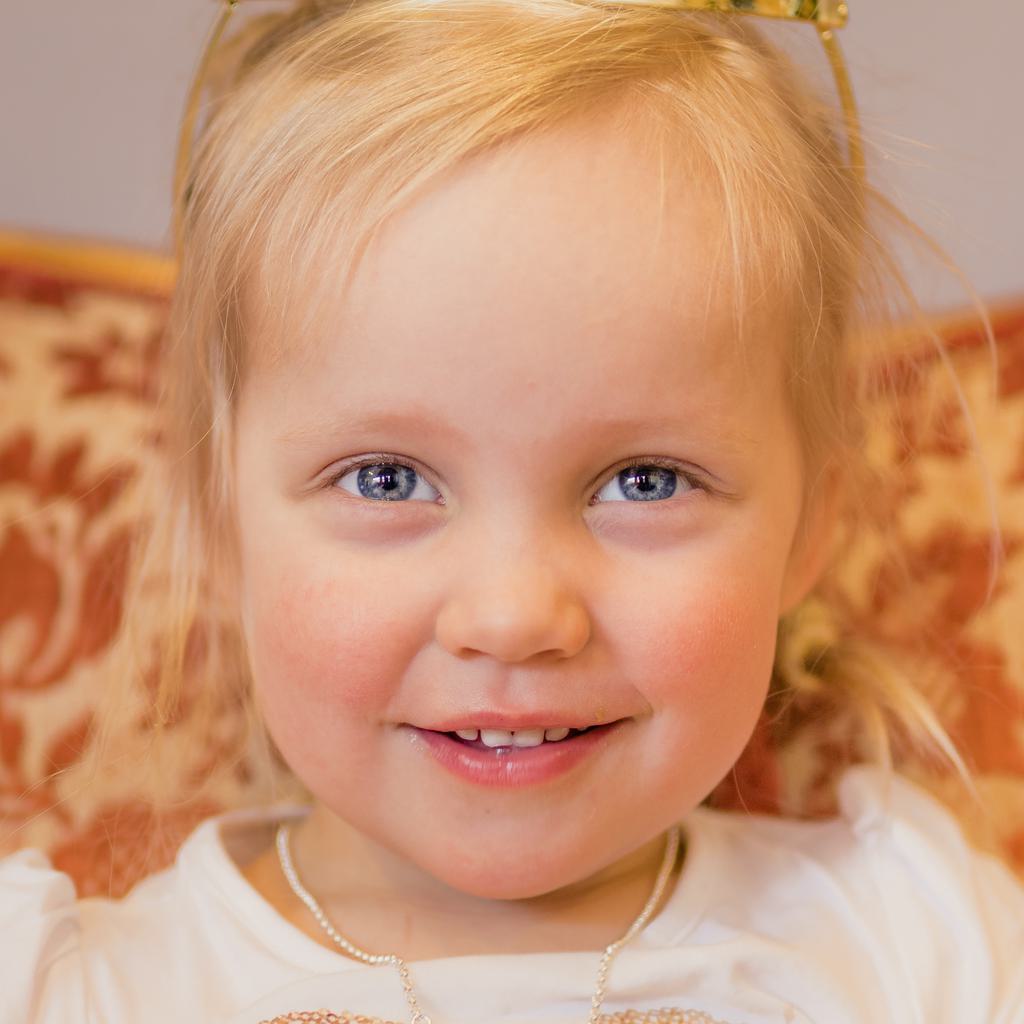}} \hspace{-1.0ex}
\frame{\includegraphics[width=0.0465\linewidth, height=0.0465\linewidth]{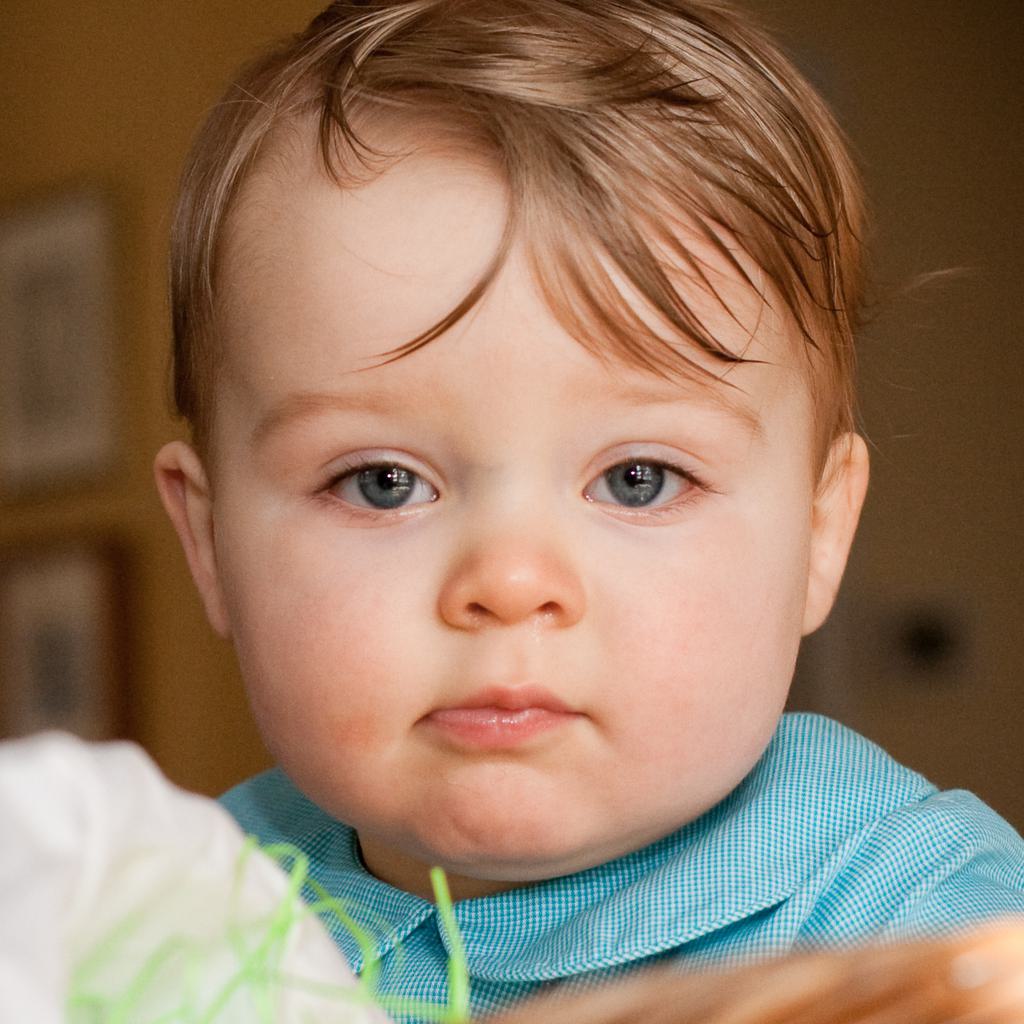}} \hspace{-1.0ex}
\frame{\includegraphics[width=0.0465\linewidth, height=0.0465\linewidth]{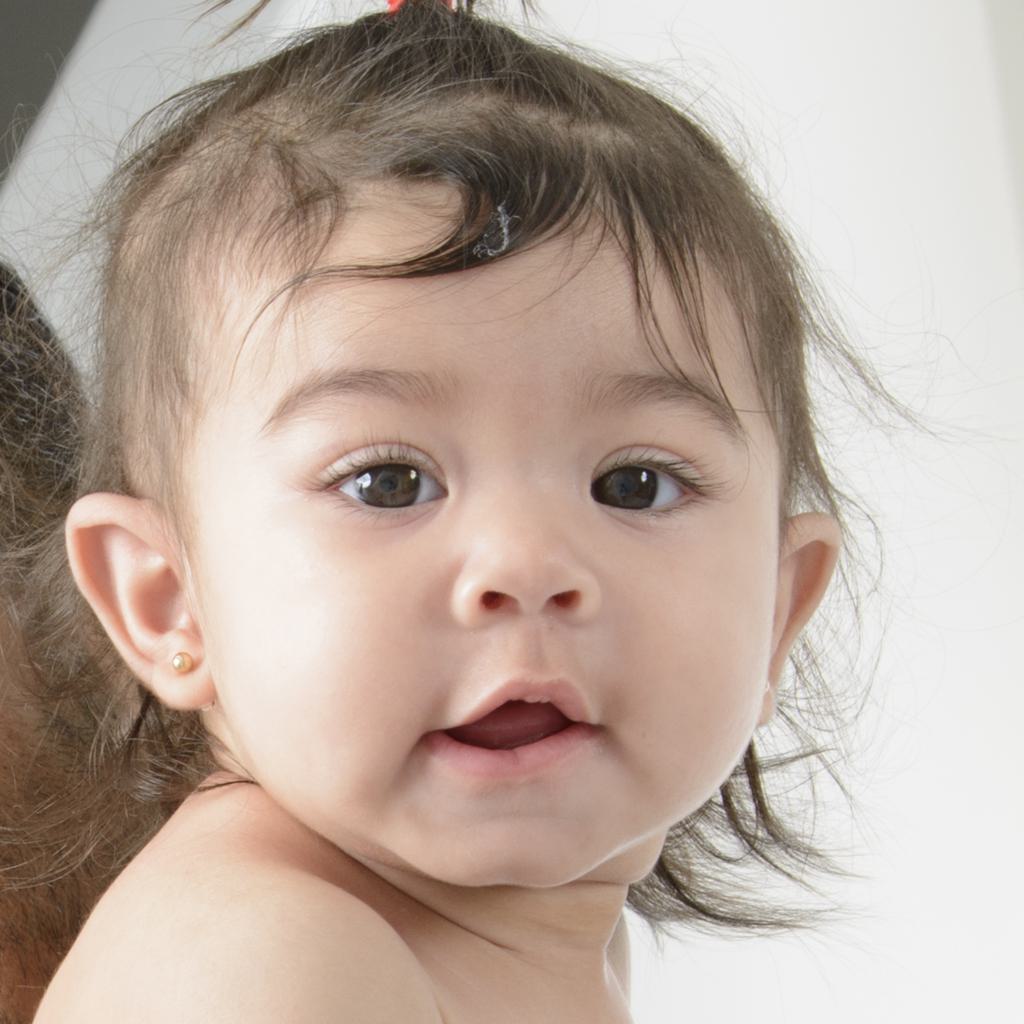}} \hspace{-1.0ex}
\frame{\includegraphics[width=0.0465\linewidth, height=0.0465\linewidth]{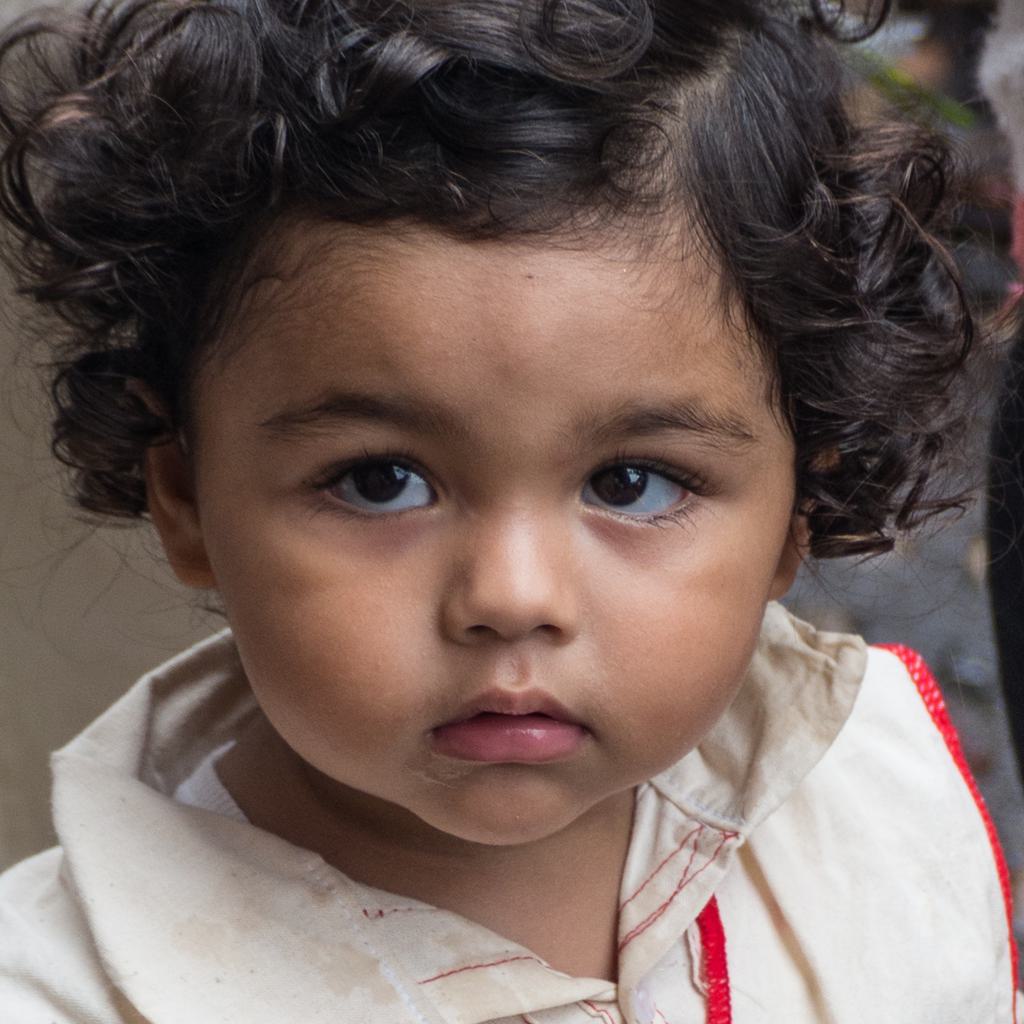}}  
} &
\multicolumn{7}{@{\hskip 0.04in}c}{
\frame{\includegraphics[width=0.0465\linewidth, height=0.0465\linewidth]{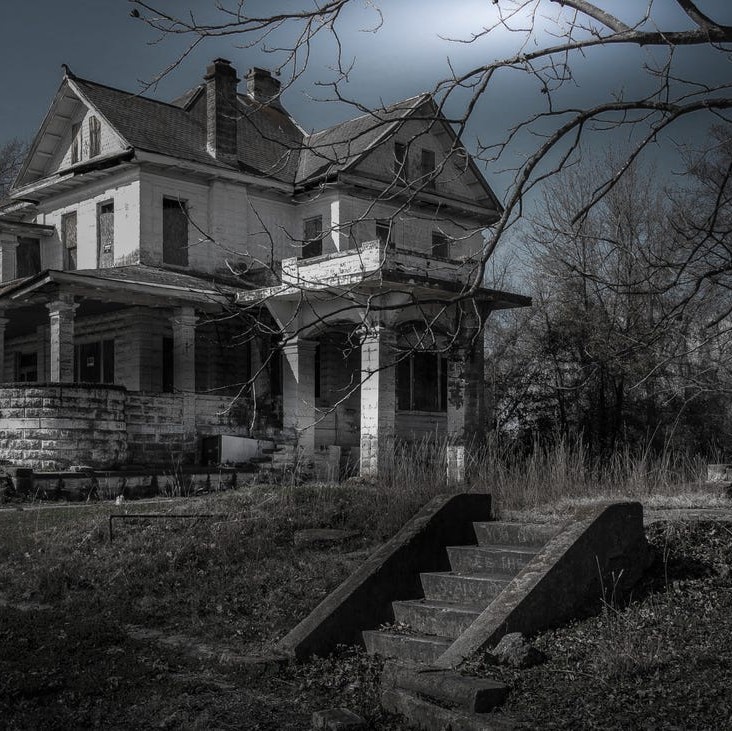}} \hspace{-1.0ex}
\frame{\includegraphics[width=0.0465\linewidth, height=0.0465\linewidth]{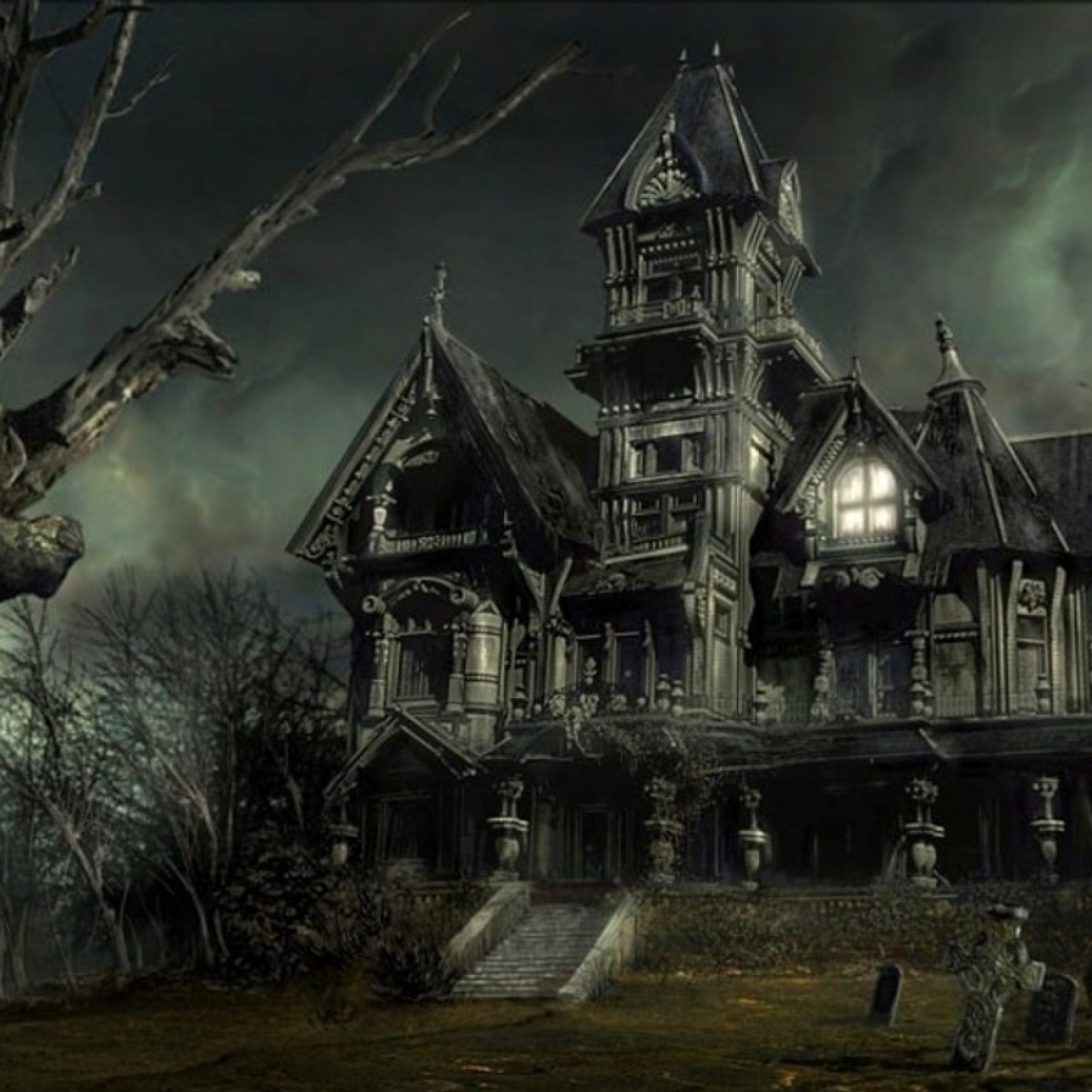}} \hspace{-1.0ex}
\frame{\includegraphics[width=0.0465\linewidth, height=0.0465\linewidth]{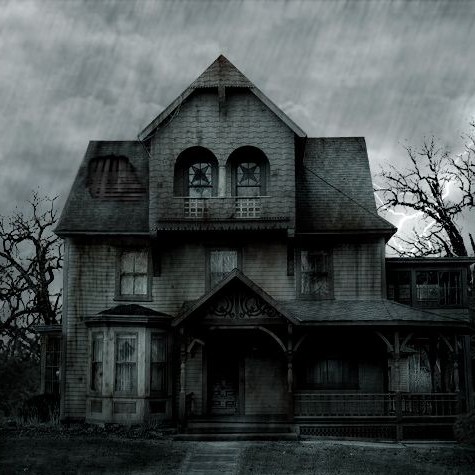}} \hspace{-1.0ex}
\frame{\includegraphics[width=0.0465\linewidth, height=0.0465\linewidth]{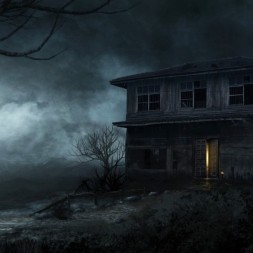}} \hspace{-1.0ex}
\frame{\includegraphics[width=0.0465\linewidth, height=0.0465\linewidth]{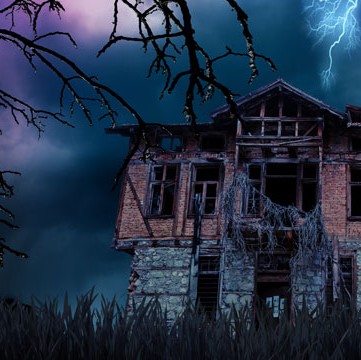}} \hspace{-1.0ex}
\frame{\includegraphics[width=0.0465\linewidth, height=0.0465\linewidth]{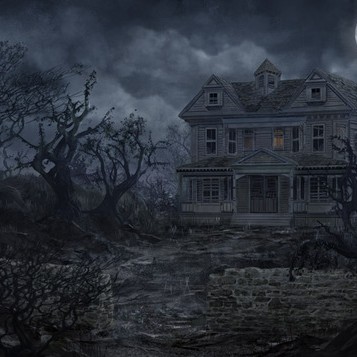}} \hspace{-1.0ex}
\frame{\includegraphics[width=0.0465\linewidth, height=0.0465\linewidth]{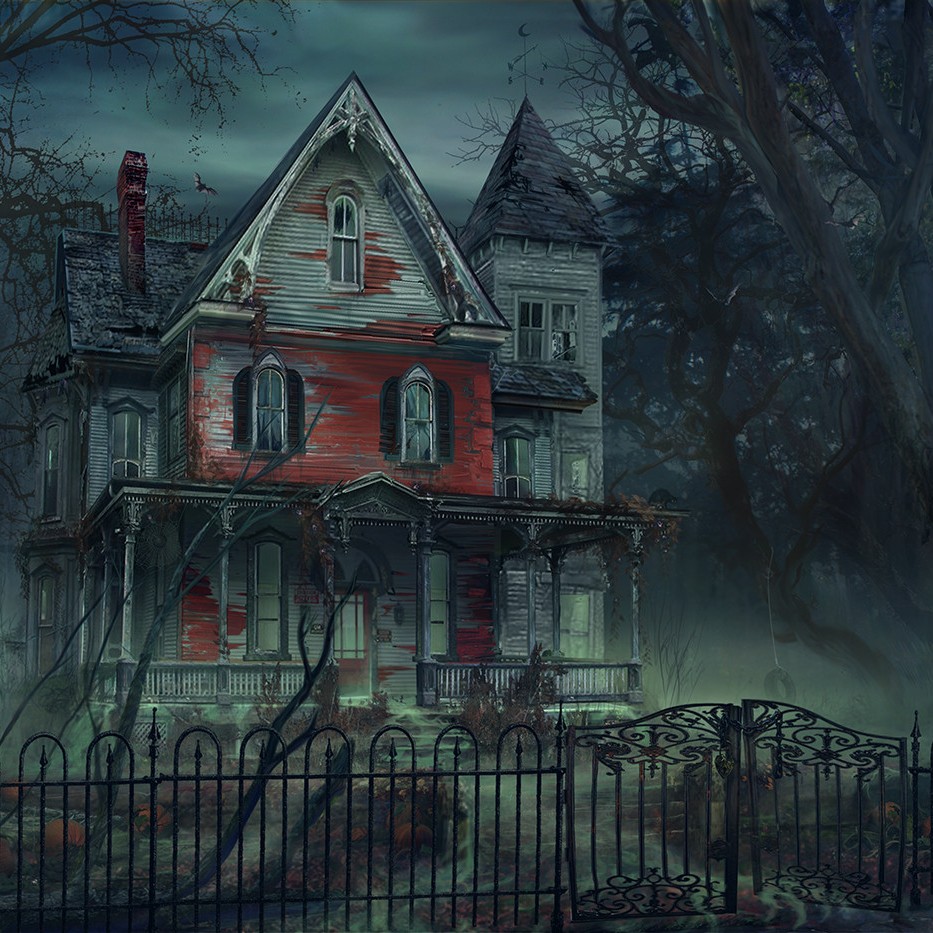}} \hspace{-1.0ex}
\frame{\includegraphics[width=0.0465\linewidth, height=0.0465\linewidth]{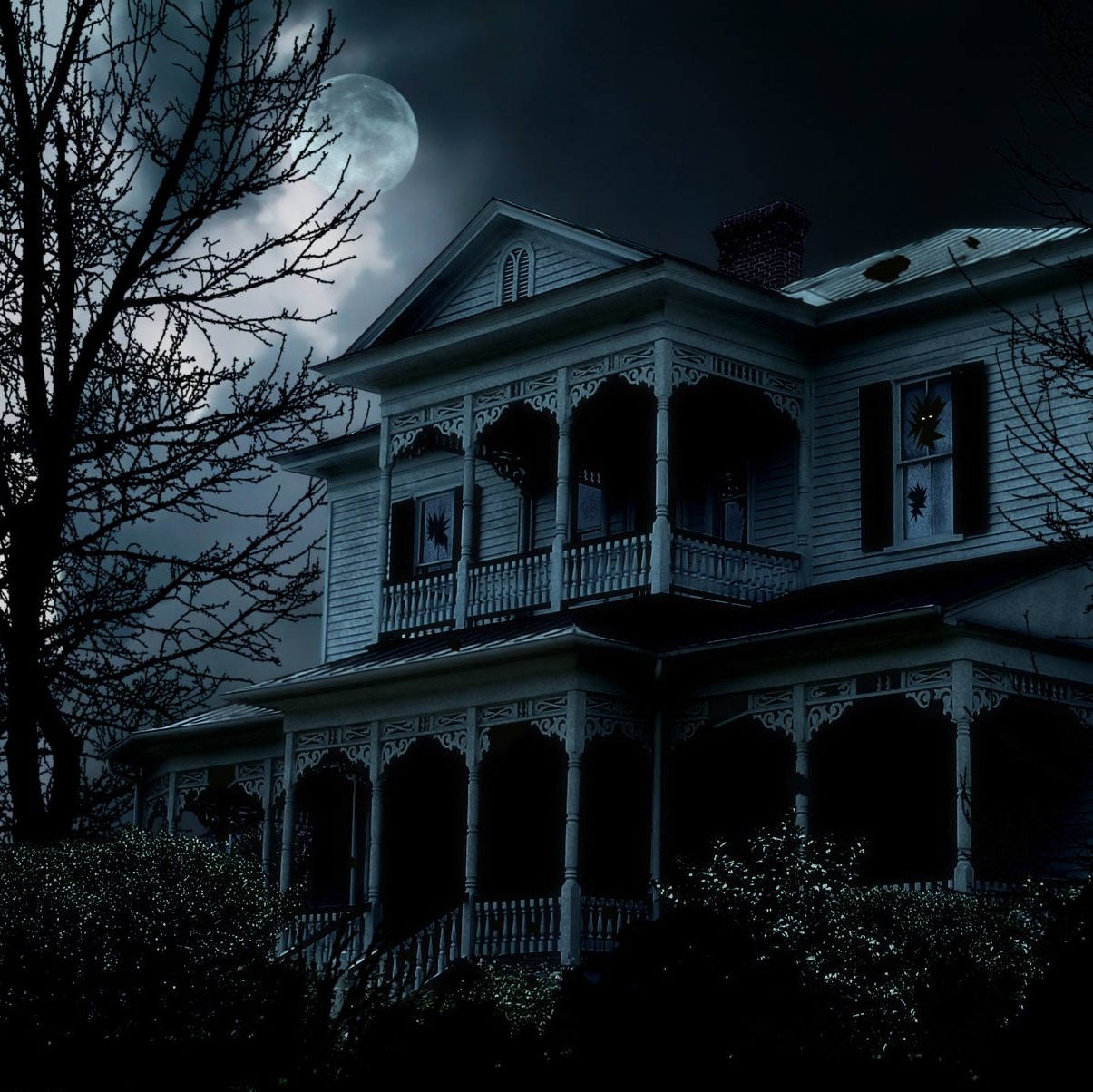}} \hspace{-1.0ex}
\frame{\includegraphics[width=0.0465\linewidth, height=0.0465\linewidth]{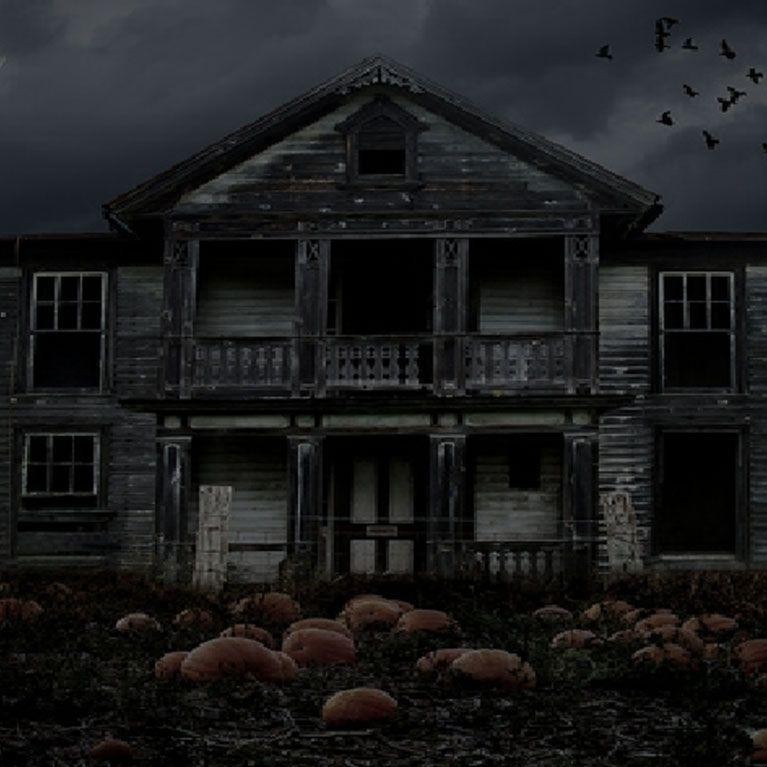}} \hspace{-1.0ex}
\frame{\includegraphics[width=0.0465\linewidth, height=0.0465\linewidth]{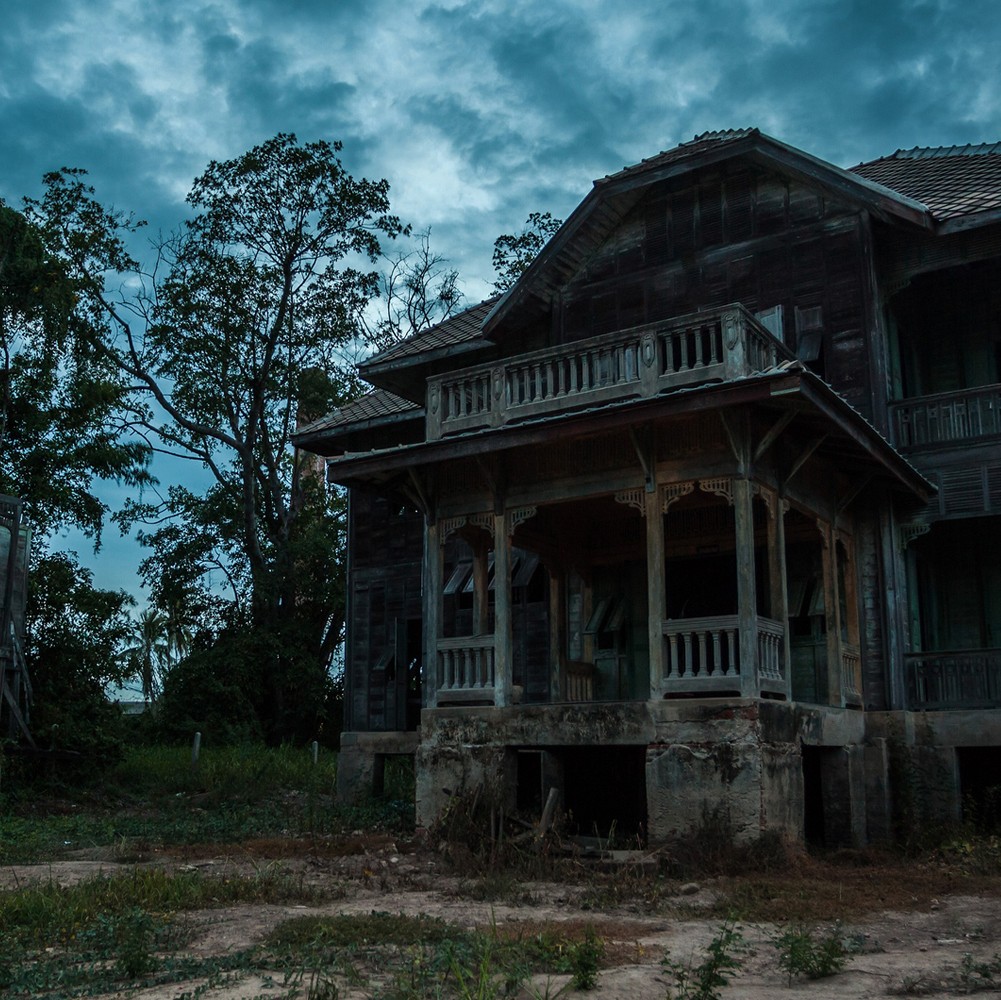}}  
}

\tabularnewline[2pt]

~~\rotatebox{90}{{\hspace{0.0em} Source \hspace{-3.0em} }} &
\frame{\includegraphics[width=0.067\linewidth, height=0.067\linewidth]{figures/qualitative_results/ffhq-sketch/source/p_2_000000.jpg}} & 
\frame{\includegraphics[width=0.067\linewidth, height=0.067\linewidth]{figures/qualitative_results/ffhq-sketch/source/p_6_000000.jpg}} &
\frame{\includegraphics[width=0.067\linewidth, height=0.067\linewidth]{figures/qualitative_results/ffhq-sketch/source/p_7_000000.jpg}} &
\frame{\includegraphics[width=0.067\linewidth, height=0.067\linewidth]{figures/qualitative_results/ffhq-sketch/source/p_8_000000.jpg}} &
\frame{\includegraphics[width=0.067\linewidth, height=0.067\linewidth]{figures/qualitative_results/ffhq-sketch/source/p_17_000000.jpg}} &
\frame{\includegraphics[width=0.067\linewidth, height=0.067\linewidth]{figures/qualitative_results/ffhq-sketch/source/p_22_000000.jpg}} &
\frame{\includegraphics[width=0.067\linewidth, height=0.067\linewidth]{figures/qualitative_results/ffhq-sketch/source/p_23_000000.jpg}} &

\frame{\includegraphics[width=0.067\linewidth, height=0.067\linewidth]{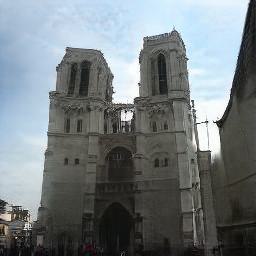}} & 
\frame{\includegraphics[width=0.067\linewidth, height=0.067\linewidth]{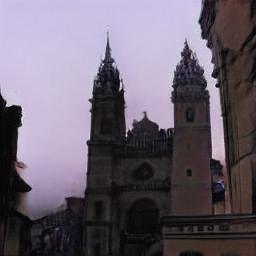}} &
\frame{\includegraphics[width=0.067\linewidth, height=0.067\linewidth]{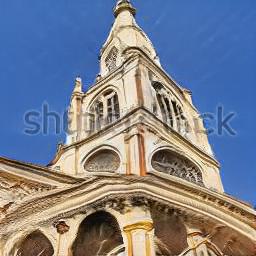}} &
\frame{\includegraphics[width=0.067\linewidth, height=0.067\linewidth]{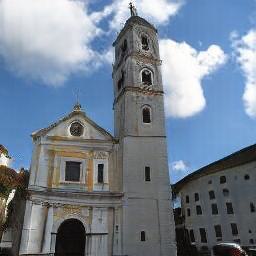}} &
\frame{\includegraphics[width=0.067\linewidth, height=0.067\linewidth]{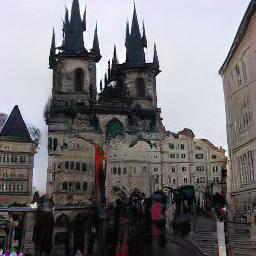}} &
\frame{\includegraphics[width=0.067\linewidth, height=0.067\linewidth]{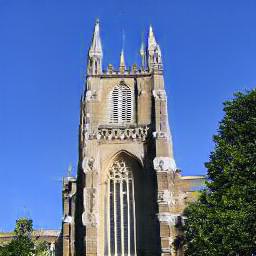}} &
\frame{\includegraphics[width=0.067\linewidth, height=0.067\linewidth]{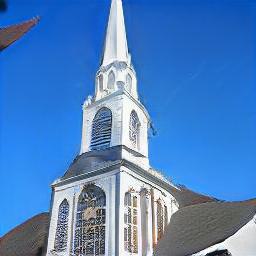}}
\tabularnewline[-3pt]

~~\rotatebox{90}{{\hspace{0.4em} CDC \hspace{-3.0em} }} &
\frame{\includegraphics[width=0.067\linewidth, height=0.067\linewidth]{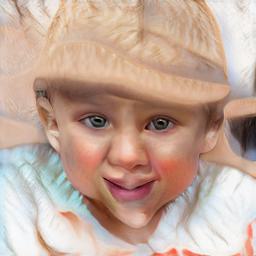}} & 
\frame{\includegraphics[width=0.067\linewidth, height=0.067\linewidth]{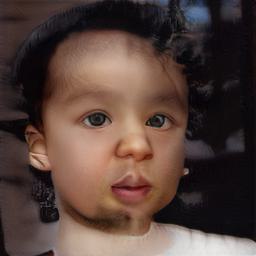}} &
\frame{\includegraphics[width=0.067\linewidth, height=0.067\linewidth]{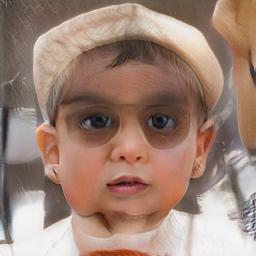}} &
\frame{\includegraphics[width=0.067\linewidth, height=0.067\linewidth]{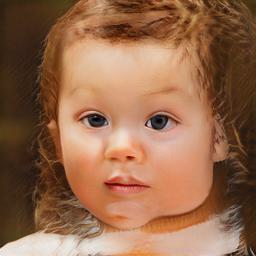}} &
\frame{\includegraphics[width=0.067\linewidth, height=0.067\linewidth]{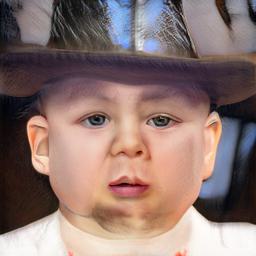}} &
\frame{\includegraphics[width=0.067\linewidth, height=0.067\linewidth]{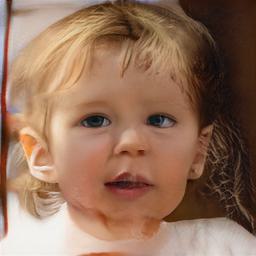}} &
\frame{\includegraphics[width=0.067\linewidth, height=0.067\linewidth]{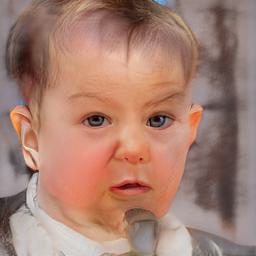}} &

\frame{\includegraphics[width=0.067\linewidth, height=0.067\linewidth]{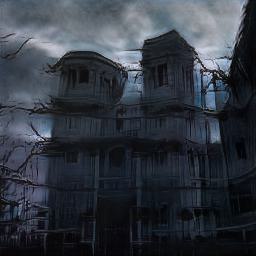}} & 
\frame{\includegraphics[width=0.067\linewidth, height=0.067\linewidth]{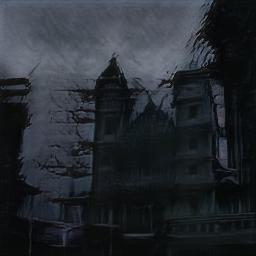}} &
\frame{\includegraphics[width=0.067\linewidth, height=0.067\linewidth]{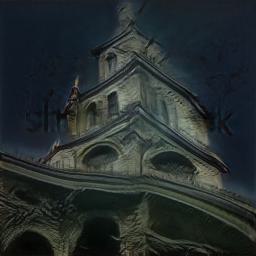}} &
\frame{\includegraphics[width=0.067\linewidth, height=0.067\linewidth]{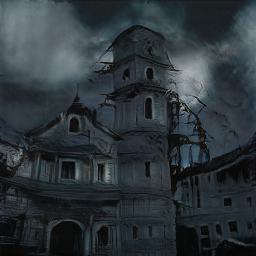}} &
\frame{\includegraphics[width=0.067\linewidth, height=0.067\linewidth]{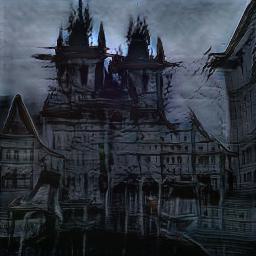}} &
\frame{\includegraphics[width=0.067\linewidth, height=0.067\linewidth]{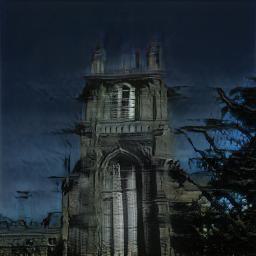}} &
\frame{\includegraphics[width=0.067\linewidth, height=0.067\linewidth]{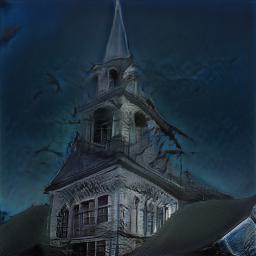}}
\tabularnewline[-3pt]

~~\rotatebox{90}{{\hspace{0.3em} RSSA \hspace{-3.0em} }} &
\frame{\includegraphics[width=0.067\linewidth, height=0.067\linewidth]{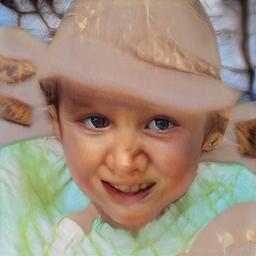}} & 
\frame{\includegraphics[width=0.067\linewidth, height=0.067\linewidth]{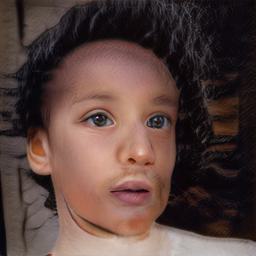}} &
\frame{\includegraphics[width=0.067\linewidth, height=0.067\linewidth]{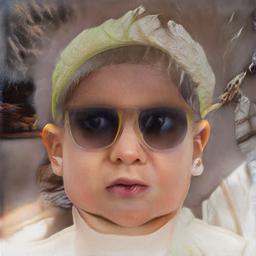}} &
\frame{\includegraphics[width=0.067\linewidth, height=0.067\linewidth]{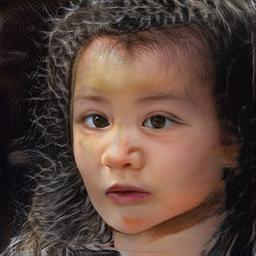}} &
\frame{\includegraphics[width=0.067\linewidth, height=0.067\linewidth]{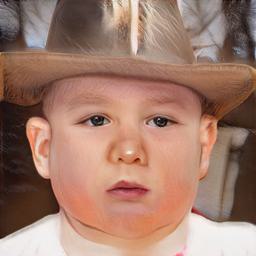}} &
\frame{\includegraphics[width=0.067\linewidth, height=0.067\linewidth]{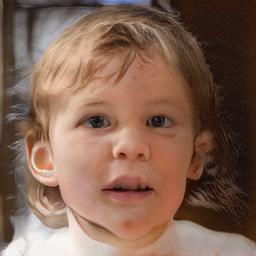}} &
\frame{\includegraphics[width=0.067\linewidth, height=0.067\linewidth]{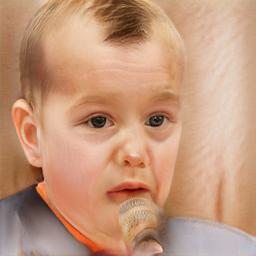}} &

\frame{\includegraphics[width=0.067\linewidth, height=0.067\linewidth]{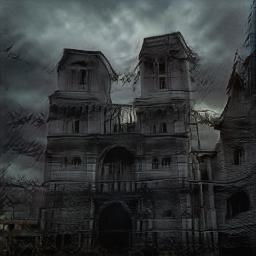}} & 
\frame{\includegraphics[width=0.067\linewidth, height=0.067\linewidth]{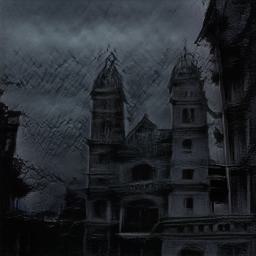}} &
\frame{\includegraphics[width=0.067\linewidth, height=0.067\linewidth]{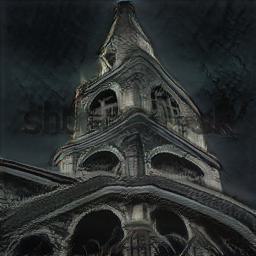}} &
\frame{\includegraphics[width=0.067\linewidth, height=0.067\linewidth]{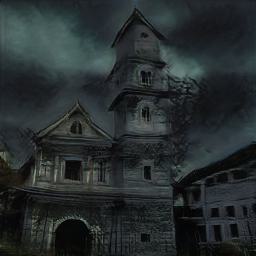}} &
\frame{\includegraphics[width=0.067\linewidth, height=0.067\linewidth]{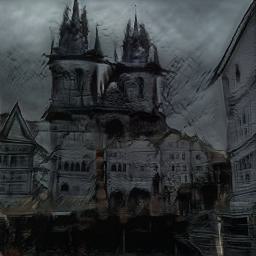}} &
\frame{\includegraphics[width=0.067\linewidth, height=0.067\linewidth]{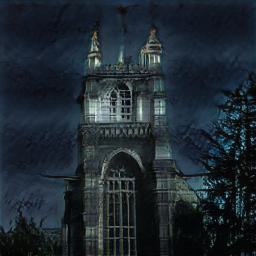}} &
\frame{\includegraphics[width=0.067\linewidth, height=0.067\linewidth]{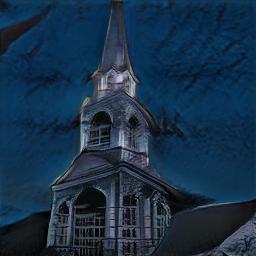}}
\tabularnewline[-3pt]

~~\rotatebox{90}{{\hspace{0.1em} AdAM \hspace{-3.0em} }} &
\frame{\includegraphics[width=0.067\linewidth, height=0.067\linewidth]{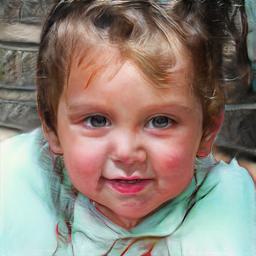}} & 
\frame{\includegraphics[width=0.067\linewidth, height=0.067\linewidth]{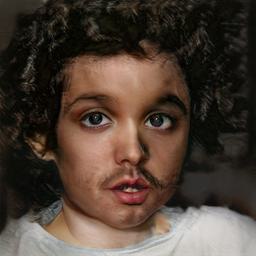}} &
\frame{\includegraphics[width=0.067\linewidth, height=0.067\linewidth]{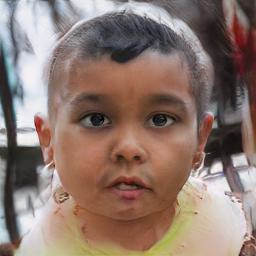}} &
\frame{\includegraphics[width=0.067\linewidth, height=0.067\linewidth]{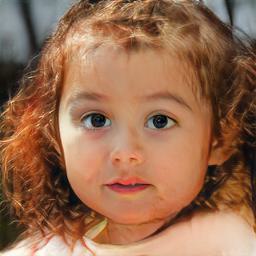}} &
\frame{\includegraphics[width=0.067\linewidth, height=0.067\linewidth]{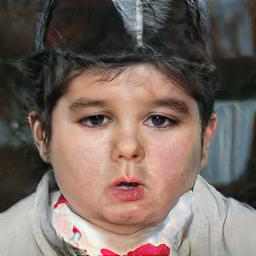}} &
\frame{\includegraphics[width=0.067\linewidth, height=0.067\linewidth]{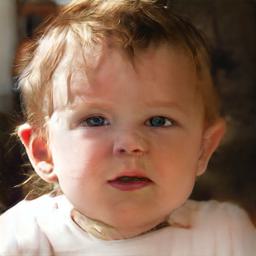}} &
\frame{\includegraphics[width=0.067\linewidth, height=0.067\linewidth]{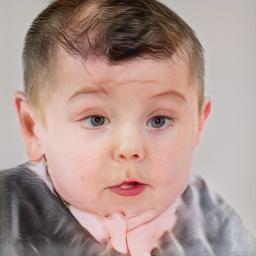}} &

\frame{\includegraphics[width=0.067\linewidth, height=0.067\linewidth]{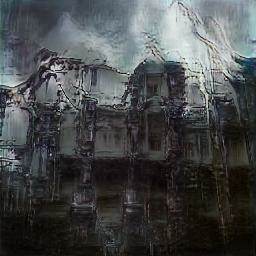}} & 
\frame{\includegraphics[width=0.067\linewidth, height=0.067\linewidth]{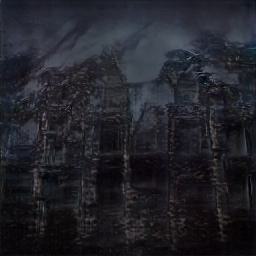}} &
\frame{\includegraphics[width=0.067\linewidth, height=0.067\linewidth]{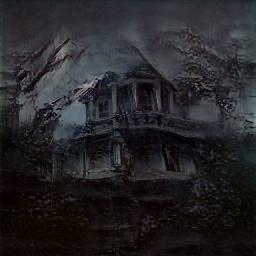}} &
\frame{\includegraphics[width=0.067\linewidth, height=0.067\linewidth]{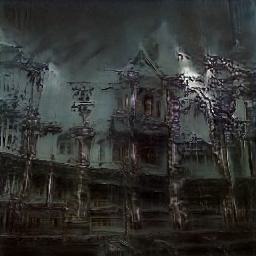}} &
\frame{\includegraphics[width=0.067\linewidth, height=0.067\linewidth]{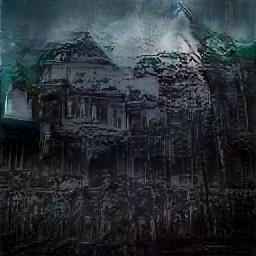}} &
\frame{\includegraphics[width=0.067\linewidth, height=0.067\linewidth]{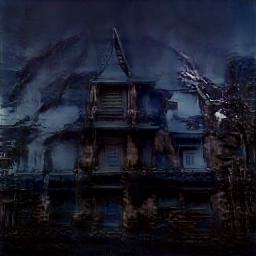}} &
\frame{\includegraphics[width=0.067\linewidth, height=0.067\linewidth]{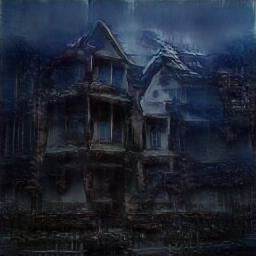}}
\tabularnewline[-3pt]

~~\rotatebox{90}{{\hspace{0.4em} Ours \hspace{-3.0em} }} &
\frame{\includegraphics[width=0.067\linewidth, height=0.067\linewidth]{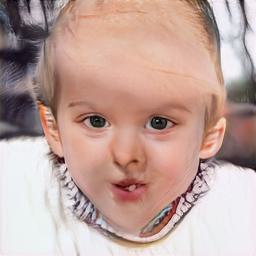}} & 
\frame{\includegraphics[width=0.067\linewidth, height=0.067\linewidth]{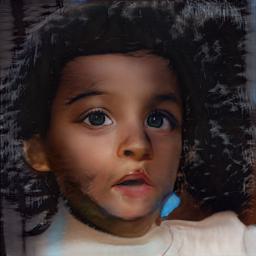}} &
\frame{\includegraphics[width=0.067\linewidth, height=0.067\linewidth]{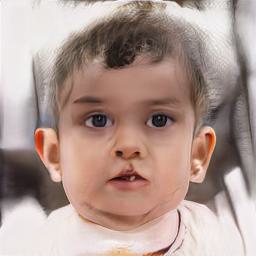}} &
\frame{\includegraphics[width=0.067\linewidth, height=0.067\linewidth]{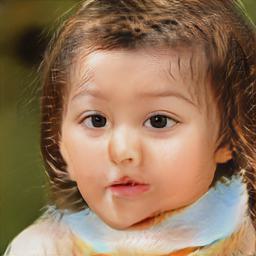}} &
\frame{\includegraphics[width=0.067\linewidth, height=0.067\linewidth]{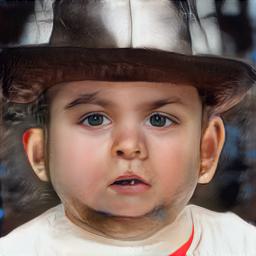}} &
\frame{\includegraphics[width=0.067\linewidth, height=0.067\linewidth]{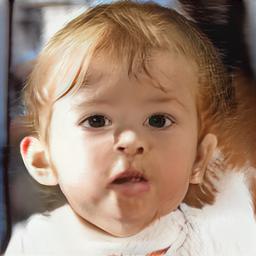}} &
\frame{\includegraphics[width=0.067\linewidth, height=0.067\linewidth]{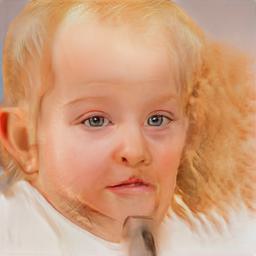}} &

\frame{\includegraphics[width=0.067\linewidth, height=0.067\linewidth]{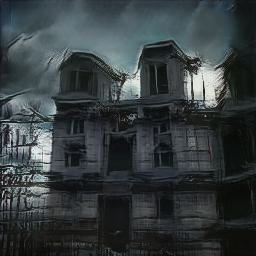}} & 
\frame{\includegraphics[width=0.067\linewidth, height=0.067\linewidth]{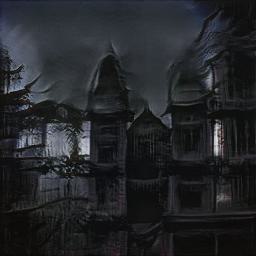}} &
\frame{\includegraphics[width=0.067\linewidth, height=0.067\linewidth]{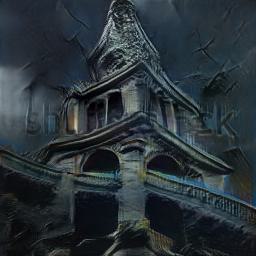}} &
\frame{\includegraphics[width=0.067\linewidth, height=0.067\linewidth]{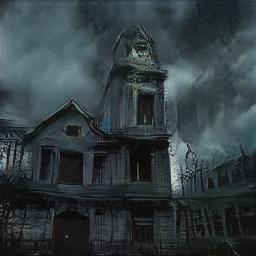}} &
\frame{\includegraphics[width=0.067\linewidth, height=0.067\linewidth]{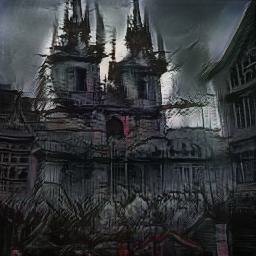}} &
\frame{\includegraphics[width=0.067\linewidth, height=0.067\linewidth]{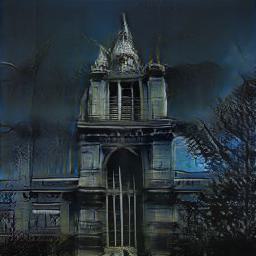}} &
\frame{\includegraphics[width=0.067\linewidth, height=0.067\linewidth]{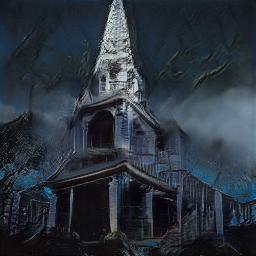}}
\tabularnewline[-3pt]

\end{tabular}
\par\end{centering}
\vspace{5pt}
\caption{Additional visual comparison to most recent prior methods on related domains.}
\label{fig:qualitative_comparison_close2}

\end{figure*}